\documentclass[manuscript]{acmart}
\usepackage{subcaption}
\usepackage{pifont}       
\usepackage[normalem]{ulem}
\newcommand{\cmark}{\ding{51}}
\DeclareMathOperator{\sgn}{sgn}
\definecolor{highlightCol}{rgb}{0,0,0}
\definecolor{deleteCol}{rgb}{0,0,0}

\DeclareMathOperator{\support}{support}

\AtBeginDocument{%
  }
\setcopyright{acmlicensed}
\copyrightyear{2025}
\acmYear{2025}
\acmDOI{XXXXXXX.XXXXXXX}

\acmJournal{JACM}
\acmVolume{37}
\acmNumber{4}
\acmArticle{111}
\acmMonth{8}

\begin{document}

\title{Pattern-Based Graph Classification: Comparison of Quality Measures and Importance of Preprocessing}


\author{Lucas Potin}
\email{lucas.potin@univ-avignon.fr}
\affiliation{%
  \institution{Laboratoire Informatique d'Avignon -- UPR 4128, F-84911}
  \city{Avignon}
  \country{France}
}

\author{Rosa Figueiredo}
\email{rosa.figueiredo@univ-avignon.fr}
\affiliation{%
  \institution{Laboratoire Informatique d'Avignon -- UPR 4128, F-84911}
  \city{Avignon}
  \country{France}
}

\author{Vincent Labatut}
\email{vincent.labatut@univ-avignon.fr}
\affiliation{%
  \institution{Laboratoire Informatique d'Avignon -- UPR 4128, F-84911}
  \city{Avignon}
  \country{France}
}

\author{Christine Largeron}
\email{christine.largeron@univ-st-etienne.fr}
\affiliation{%
  \institution{Laboratoire Hubert Curien -- UMR 5516, F-42023}
  \city{Saint-Etienne}
  \country{France}
}

\renewcommand{\shortauthors}{*}

\begin{abstract}
Graph classification aims to categorize graphs based on their structural and attribute features, with applications in diverse fields such as social network analysis and bioinformatics. Among the methods proposed to solve this task, those relying on patterns (i.e. subgraphs) provide good explainability, as the patterns used for classification can be directly interpreted. To identify meaningful patterns, a standard approach is to use a quality measure, i.e. a function that evaluates the discriminative power of each pattern. However, the literature provides tens of such measures, making it difficult to select the most appropriate for a given application. Only a handful of surveys try to provide some insight by comparing these measures, and none of them specifically focuses on graphs. This typically results in the systematic use of the most widespread measures, without thorough evaluation. 
To address this issue, we present a comparative analysis of \textcolor{highlightCol}{38 quality measures} from the literature. We characterize them theoretically, based on four mathematical properties. We leverage publicly available datasets to constitute a benchmark, and propose a method to elaborate a gold standard ranking of the patterns. We exploit these resources to perform an empirical comparison of the measures, both in terms of pattern ranking and classification performance. Moreover, we propose a clustering-based preprocessing step, which groups patterns appearing in the same graphs to enhance classification performance. Our experimental results demonstrate the effectiveness of this step, reducing the number of patterns to be processed while achieving comparable performance. Additionally, we show that some popular measures widely used in the literature are not associated with the best results.
\end{abstract}



\maketitle


\section{Introduction}
Graph classification is a fundamental task in graph theory and machine learning, aiming at partitioning a graph collection into different categories, based on their structural and attribute features~\cite{Tsuda2010}. It finds application in diverse fields such as chemistry~\cite{Jippo2019}, \textcolor{highlightCol}{API misuse
detection}~\cite{Kang2022}, and supply chain optimization~\cite{Wu2023}. There are three main approaches to tackle this task: graph kernels, graph neural networks and subgraph mining.

Graph kernels~\cite{Kriege2016, Kriege2020} allow measuring the similarity between all pairs of graphs based on predefined features. The resulting graph kernel matrix can then be used as a representation of the graph collection. Graph kernels have been applied in various domains. For instance, in bioinformatics for classifying chemical compounds and proteins by comparing their molecular structures~\cite{Kashima2003} or in image classification~\cite{Harchaoui2007}. This type of approach is highly flexible and can be adapted to different types of graphs~\cite{Kriege2020}. In particular, it can handle graphs with varying sizes and structures, as well as attributed graphs. 

Graph Neural Networks (GNN)~\cite{Wu2021, Zhou2020} leverage deep learning techniques to automatically learn graph representations. Typically, a GNN  first initializes vertex embeddings based on their features. Through iterative message passing, the embedding of each vertex is updated by aggregating information from its neighbors. After several iterations, a readout function aggregates these final vertex embeddings into a single vector that represents the entire graph. The addition of a layer able to take advantage of these representations allows performing the graph classification task. Several algorithms have been proposed following this principle, notably DGCNN~\cite{Wu2018} and U2GNN~\cite{Nguyen2022}, applied both to chemical compound and citation network classification.

Subgraph mining~\cite{Guvenoglu2018, Jiang2012} identifies specific subgraphs, called patterns, that frequently occur within the considered collection of graphs. Methods using subgraph mining convert the mined subgraphs into features and represent each graph as a binary vector, whose components indicate the presence or absence of these patterns~\cite{AcostaMendoza2016}. These vector representations are then used to train a standard classifier. For instance, \citet{Potin2023d} use a Support Vector Machine to distinguish between fraudulent and lawful public procurement contracts, and \citet{Karbalaie2012} use a Random Forest to distinguish malware from benign software.

It is worth stressing that all three approaches rely on specific fixed-size vector representations of the graphs, which are fetched to different types of classifiers. These representations exhibit different levels of interpretability. In this work, we focus on subgraph mining methods, because they provide better explainability than both other approaches. Indeed, patterns can be directly interpreted, providing insights into structural features that differentiate graph classes. However, the choice of the subgraphs selected as features should be made carefully: one wants to focus on patterns that have a high discriminative power with regard to the classification task.


To obtain patterns particularly related to a class, a common strategy is to mine all \textit{frequent} patterns, and keep only the most discriminative~\cite{AcostaMendoza2016}. This approach requires using a function, called a \textit{quality measure}~\cite{LoyolaGonzalez2014, Ventura2016}, to assess how well each pattern distinguishes between classes. \textcolor{highlightCol}{As we show later (Section~\ref{sec:RelatedWork}), although quality measures have been extensively studied in the context of \textit{tabular} data mining, their assessment in the context of graph pattern mining remains underexplored. Thus, existing surveys~\cite{LoyolaGonzalez2014, Ventura2016, Chen2022} provide only partial guidance, as, among other limitations, they primarily focus on patterns based on items, by opposition to subgraphs.}

In this paper, we focus on evaluating the effectiveness of quality measures able to assess the discriminative power of subgraphs within the context of pattern-based graph classification. For the sake of concision and clarity, we make two important assumptions. First, we work within the context of \textit{binary} classification. Indeed, many graph classification applications are based on two classes, often interpreted as the presence or absence of a particular characteristic, for instance, in chemistry with the absence or presence of chemical compounds~\cite{Kashima2003}, or in economics with the absence or presence of fraud~\cite{Li2022, Potin2023d}. As a result, the majority of available benchmark datasets for graph classification contain two classes. In addition, most studies that deal with the multi-class situation transform it into a two-class case, using the one-vs-rest approach~\cite{GarcaBorroto2013}. This may be due to most quality measures being designed to handle only two classes~\cite{LoyolaGonzalez2014, Chen2022, Ventura2016, Dong2012}. Second, we assume that the classes are balanced, i.e. contain the same number of graphs. This is a common methodological choice in the literature, justified by the fact that it allows a fair comparison of the quality measures, and avoids hiding correlations between equivalent measures~\cite{LoyolaGonzalez2014}.

\textcolor{highlightCol}{We aim to address three research questions: 
\begin{itemize}
    \item[RQ1] Is it possible to achieve more compact graph representation without compromising classification performance?
    \item[RQ2] Do quality measures behave consistently across graph datasets?
    \item[RQ3] Can we identify quality measures that tend to perform better than others? 
\end{itemize}
To answer these questions, we propose a two-step methodology: first, by comparing the way quality measures rank patterns, and second, by assessing how these rankings affect classification performance. 
This paper makes five main contributions. 
\begin{enumerate}
    \item We review a comprehensive set of \textcolor{highlightCol}{38} quality measures, and propose a typology based on three mathematical properties from the literature, as well as an additional original one that we introduce to better characterize the measures.
    \item We constitute a benchmark of graph datasets, and propose a method based on the Shapley Value~\cite{Shapley1953} to produce pattern rankings on these datasets that can be used as gold standard for classification evaluation. 
    \item We design a preprocessing step relying on cluster analysis to improve the pattern-based representation of graphs, and assess its effect on graph classification and runtime. 
    \item  We evaluate the effectiveness of the selected measures experimentally, and compare them in terms of classification performance using our benchmark and gold standard.
    \item We release all datasets, code, and experimental results in an open-source repository\footnote{\url{https://github.com/CompNet/gpQualMeasComp}} to support reproducibility and further research.
\end{enumerate}}

The rest of the paper is organized as follows. \textcolor{highlightCol}{Section~\ref{sec:RelatedWork} provides an overview of existing studies on quality measures in pattern mining, highlighting their current relevance, but also the limitations of these studies, particularly regarding graph data.} Section~\ref{sec:Prob} defines the problem, terminology and notations used throughout this article. Section~\ref{sec:QualityMeasures} lists the quality measures considered in our experiments, and discusses their properties. Section~\ref{sec:Meth} describes our methods and framework. In Section~\ref{sec:Exp}, we present our experimental protocol and discuss our results. Finally, we summarize our main findings in Section~\ref{sec:Conclusion}, and propose some perspectives.

\color{highlightCol}
\section{Related Work}
\label{sec:RelatedWork}

\paragraph{Relevance of Quality Measures}
Quality measures play a fundamental role in pattern mining, and this is especially true for \textit{graph} pattern mining. In this context, they have been widely adopted for various tasks, such as graph classification, subgraph selection, and explainability. For instance, \citet{Kang2022} use the $\chi^2$ statistic~\cite{Bay1999} to identify discriminative subgraphs for API misuse detection, while \citet{Alam2021} employ Information Gain~\cite{Church1990} to select subgraphs that maximize the reduction of entropy, enhancing supervised graph embeddings.

Beyond subgraph selection, quality measures also play an essential role in graph mining algorithms. \citet{He2024} introduce All-Confidence as an alternative to Lift~\cite{PiatetskyShapiro2000} for mining credible attribute rules in dynamic attributed graphs. They argue that Lift is unreliable due to its lack of anti-monotonicity, which makes it less suitable for hierarchical rule evaluation. Similarly, \citet{Chowdhury2021} analyze correlation in frequent subgraph mining, highlighting the limitations of the Confidence metric~\cite{Agrawal1993} in this context. 

In the field of GNNs, quality measures contribute both to performance enhancement and explainability. For instance, \citet{Kikaj2024} improve message-passing neural networks by selecting informative subgraphs using $\chi^2$ and Mutual Information~\cite{Fang2011}. Meanwhile, \citet{Veyrin-Forrer2022} focus on GNN explainability, identifying characteristic subgraphs through a modified version of WRACC~\cite{Lavrac2004} within the gSpan framework, improving model interpretability. 

But the impact of quality measures extends beyond graph pattern mining. Over the past five years, articles citing at least one of the quality measures assessed in this study have accumulated more than 15,000 citations per year according to GoogleScholar. These articles cover a wide range of applications, including analyzing cab customer behavior~\cite{GarciaVico2020}, predicting grocery stocking needs~\cite{Bandyopadhyaya2021} or discovering play patterns in video games~\cite{Mathonat2020}. This widespread impact highlights the central role of quality measures in various domains. But whatever the context, a recurrent methodological challenge persists: selecting the most appropriate quality measure for the application at hand. This choice can only be made based on a thorough comparison between the many existing measures. 

\paragraph{Tabular Quality Measure Surveys}
A handful of studies have investigated quality measures but specifically in the field of \textit{tabular} (or item-based) pattern mining~\cite{LoyolaGonzalez2014, Chen2022, Ventura2016, Dong2012}. Among them, \citet{Chen2022} cite 13 different measures, while \citet{Dong2012} list 10. Both surveys essentially provide a description without pushing their analysis deeper, though. \citet{Ventura2016} go further by evaluating three mathematical properties for nine selected measures. Additionally, they analyze the relationships between each pair of measures, providing a structured comparison that highlights their dependencies and redundancies. \citet{LoyolaGonzalez2014} take the most experimental approach when studying 33 different measures. Their work includes a correlation analysis between pairs of metrics with the goal of forming clusters of similar measures, providing insights into which measures capture redundant or complementary aspects of pattern quality. 

Although two of these surveys~\cite{Ventura2016, Chen2022} briefly touch upon the topic of \textit{graph} pattern mining, their focus remains limited to \textit{contrast} subgraphs~\cite{Ting2006}. This concept refers to a subgraph that appears \textit{only} in one class, and \textit{never} in another. This makes it a distinguishing feature that helps to differentiate between classes. The main advantage of these subgraphs is their ability to provide clear and interpretable features that highlight the differences between classes. However, their very strict definition (exclusive presence in a single class) can sometimes result in overlooking more subtle and potentially useful patterns that show a strong but not absolute association with one class. Most quality measures are able to handle non-contrast subgraphs, so by focusing only on this specific type of graph patterns, these two articles provide a very incomplete comparison regarding graph pattern mining. 

Because of the lack of dedicated studies, the selection of a quality measure for graph mining remains largely \textit{ad hoc}. Researchers often adopt measures based on convention rather than a principled evaluation of their effectiveness for specific tasks. This issue is compounded by the fact that existing surveys provide only partial guidance on the choice of the most appropriate quality measure. 

\paragraph{Limitations of Existing Work}
First, existing surveys remain incomplete in their coverage of quality measures. Although \citet{LoyolaGonzalez2014} present the widest range of quality measures among existing empirical studies, their analysis omits several essential measures, particularly those used in feature selection and classification, such as \textit{FPR}~\cite{GarciaVico2020} or \textit{AbsSupDif}~\cite{Thoma2010}. Another limitation of these surveys is that they do not cover more recent measures coming from the feature selection literature~\cite{Dhal2021, Theng2023, Barbieri2024}, which can also be used to assess the quality of patterns. Measures such as the \textit{Gini Index}, \textit{Entropy}, and \textit{Fisher Score}, rely on statistical techniques that follow the same principle as quality measures: each pattern is assigned a score that indicates its discriminative power. In our work, we extend this scope by incorporating such feature selection measures into our evaluation, ensuring a more complete benchmark. 

Second, there is a divide between theoretical and empirical approaches in existing surveys. Some studies provide a formal theoretical analysis based on mathematical properties of quality measures but lack empirical validation~\cite{Dong2012, Ventura2016}. Conversely, the most comprehensive study~\cite{LoyolaGonzalez2014} proposes an empirical comparison but lacks a theoretical foundation to select the quality measure and discuss the experimental results. Without theoretical guarantees, the results are highly sensitive to dataset biases, limiting their generalizability. To address this issue, our study adopts a dual approach that combines theoretical and empirical evaluation. We use four fundamental mathematical properties to characterize the 38 quality measures considered in this article, while simultaneously validating their effectiveness on eight datasets from diverse domains of application. 

Third, current evaluations lack a proper comparison against a ground truth. While \citet{LoyolaGonzalez2014} propose an experimental setup to assess quality measures, their approach is based on a classification method that is a deliberate oversimplification compared to standard approaches. Specifically, this method assigns a class to each instance by applying a voting mechanism among the highest-scoring patterns. This strategy creates a strong dependency on individual patterns, which is likely to limit the overall classification performance, possibly making measure comparison less reliable. To overcome this limitation, we construct a gold standard based on the Shapley Value~\cite{Shapley1953} to assess the quality measures, reducing the dependency on the classifier.

Finally, in addition to these limitations, an even more fundamental issue remains: existing studies do not specifically address the challenges posed by \textit{graphs}, and this leads to several limitations. First, these experiments leverage some pattern mining algorithms that directly mine patterns closely linked to a specific class~\cite{GarcaBorroto2010}. These methods are specific to \textit{tabular} data; there is no counterpart able to handle graphs. Second, dealing with tabular data means ignoring certain types of patterns specific to graphs, such as \textit{induced} subgraphs~\cite{Juttner2018}. In this work, we adopt an approach that is exclusively focused on graph data, which allows us to consider the effect of mining induced graph patterns.

\color{black}
\section{Definitions and Notations}
\label{sec:Prob}
Our work relies on the most general definition of an attributed graph, i.e. with attributes on vertices as well as on edges.
\begin{definition}[Attributed Graph]
    An attributed graph is defined as a tuple $G = (V,E,\mathbf{X},\mathbf{Y})$ in which $V$ is the set of $n$ vertices, $E$  the set of $m$ edges of $G$, $\mathbf{X}$ the $n \times d_V$ matrix whose row $\mathbf{x}_i$ is the $d_V$-dimensional attribute vector associated with vertex $v_i \in V$, and $\mathbf{Y}$ the $m \times d_E$ matrix whose row $\mathbf{y}_i$ is the $d_E$-dimensional attribute vector associated with edge $e_i \in E$.
\end{definition}
A non-attributed graph can be considered as a specific case of attributed graph, in which a single attribute is used to describe the vertices as well as the edges, with only one possible value, for example a weight equal to one on all edges. Consequently, the concepts and methods described in the following also apply to non-attributed graphs.

Let us consider a set $\mathcal{G} = \{ G_1,...,G_N \}$ of attributed graphs. Let us assume that each graph $G_i$ ($1 \leq i \leq N$) is associated with a label noted $\ell_i$ and defined in $\mathcal{L} =\{+, -\}$. The set $\mathcal{G}$ can therefore be split into two disjoint subsets, or classes: $\mathcal{G} = \mathcal{G}^+ \cup \mathcal{G}^-$ ($\mathcal{G}^+ \cap \mathcal{G}^- = \emptyset$), where $\mathcal{G}^+$ is the subset of graphs in the positive class and $\mathcal{G}^-$ is the subset of graphs in the negative class.

In this paper, we consider the problem that consists in classifying a graph from $\mathcal{G}$ as either positive or negative, based on its structure and attributes. We specifically focus on the case where these classes are balanced, i.e. $|\mathcal{G}^+| = |\mathcal{G}^-|$.

More particularly, we are interested in methods that rely on subgraphs to represent a graph and determine its class. These subgraphs, called \textit{patterns}, are defined as follows:
\begin{definition}[Pattern]
    Let $G = (V,E,\mathbf{X},\mathbf{Y})$ be an attributed graph. A graph $P$ is a pattern of $G$ if it is isomorphic to a subgraph $H$ of $G$, i.e. $\exists H \subseteq G: P \cong H$.
    \label{def:Pattern}
\end{definition}

We consider that $P$ is a pattern of a set of graphs $\mathcal{G}$ when $P$ is a pattern of at least one of its graphs. In order to mine the frequent patterns of a set of graphs, several algorithms have been proposed such as gSpan~\cite{Yan2002}, FFSM~\cite{Huan2003}, or more recently TKG~\cite{FournierViger2019}. We do not go into detail about these algorithms, as they are outside the scope of this paper. For more information, we refer the interested reader to the survey of~\citet{Guvenoglu2018}.

Applying one of these algorithms results in a set of patterns of $\mathcal{G}$, noted $\mathcal{P}$. Graphs can then be described in terms of whether or not they contain these patterns. However, not every pattern holds the same level of relevance to the classification task. In particular, some patterns are evenly distributed over $\mathcal{G}^+$ and $\mathcal{G}^-$, and therefore provide no information allowing to discriminate between the classes.

To discriminate between the patterns they detect in the dataset, these algorithms typically leverage the notion of \textit{graph support}.
\begin{definition}[Graph support]
    The graph support of a pattern $P$ in a set $\mathcal{G}$, noted $\support(P,\mathcal{G})$, is the number of graphs in $\mathcal{G}$ that contain $P$ as a pattern: $\support(P,\mathcal{G}) = \big| \{ G \in \mathcal{G} : \exists H \subseteq G \text{ s.t. } P \cong H \} \big|$.
\end{definition}

This support ranges from $0$ to $|\mathcal{G}|$, as it simply indicates the presence or absence of a pattern in a graph, without considering how many times it appears in the graph. It is worth stressing that the notion of support is ambiguously defined in the Pattern Mining literature: it is sometimes the \textit{number} of items containing the pattern~\cite{Aggarwal2014}, and sometimes the \textit{proportion} of such items~\cite{LoyolaGonzlez2020}. In the specific case of \textit{graph} pattern mining, the former version appears to be the most consensual~\cite{Yan2002, FournierViger2019}, thus we use it in this paper.

In order to select only the most interesting patterns, a common method~\cite{Guo2013} consists in ranking them using a so-called \textit{quality measure}.
\begin{definition}[Quality Measure]
    Let $\mathcal{P}$ be the set of patterns occurring in a collection of graphs $\mathcal{G}$. A quality measure $q(P,\mathcal{G}^+,\mathcal{G}^-)$ associates a numeric value to each pattern $P \in \mathcal{P}$, indicating its power to discriminate between classes $\mathcal{G}^+$ and $\mathcal{G}^-$.
    \label{def:Measure}
\end{definition}
In this article, we use the notation adopted by \citet{LoyolaGonzlez2020} to define a quality measure as a function of the considered pattern $P$ as well as both classes $\mathcal{G}^+$ and $\mathcal{G}^-$. In general, a high quality measure for a pattern indicates that it has a strong discriminative power. There exist many quality measures, which we review in Section~\ref{sec:QualityMeasures}. The main goal of this paper is to compare them.

Whatever the selected measure, it is possible to rank the patterns of $\mathcal{P}$ in order to obtain an ordered set noted $\mathcal{P}_r$.

From set $\mathcal{P}_r$, the $s$ most discriminative patterns ($s \leq |\mathcal{P}|$) are selected to define $\mathcal{P}_s$, the subset of patterns used to represent each graph. It is necessary to choose $s$ carefully, in order to retain the patterns that are necessary and useful for the classification. 

Although it is possible to directly use $\mathcal{P}_r$ (i.e. all available patterns) to represent the graphs, there are two major advantages with using a restricted subset. First, using fewer patterns reduces computation time. Second, previous work~\cite{Yang1997} has shown that reducing the size of the feature set used for classification may increase final performance. In order to select $\mathcal{P}_s$, it is therefore necessary to have an efficient way of distinguishing its patterns, using an adapted quality measure.

All the patterns in $\mathcal{P}_s$ are then used to build a matrix $\mathbf{H} \in \mathbb{R}^{|\mathcal{G}| \times s}$, where each row $\mathbf{h}_{i:}$ is the vector-based representation of graph $G_i$ using the patterns in $\mathcal{P}_s$. 

There are several possible approaches to create this representation. In this work, we focus on the most common one~\cite{AcostaMendoza2016}, which is called \textit{binary representation}. According to this approach, for each graph $G_i \in \mathcal{G}$ and each pattern $P_j \in \mathcal{P}_s$, $h_{ij} = 1$ if this pattern $P_j$ is present in $G_i$ and $h_{ij} = 0$ otherwise. This vector representation can be used as input by common classifiers~\cite{Chang2011} to predict the class of each graph.

Matrix $\mathbf{H}$ can also be viewed as a concatenation of columns instead of rows. Each column $\mathbf{h}_{:j}$ represents a pattern $P_j$, as each value $h_{ij}$ indicates whether $P_j$ is present or absent from graph $G_i$. We call such vector the \textit{footprint} of the pattern in $\mathcal{G}$.

\begin{figure}[htbp!]
    \centering 
    \includegraphics[width=\textwidth]{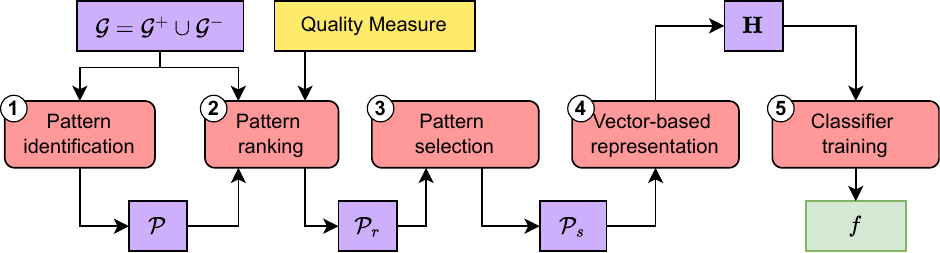}
    \caption{Processing steps of a standard pattern-based graph classification framework.}
    \label{fig:PANG_Framework}
    \Description{Description} 
\end{figure}

Figure~\ref{fig:PANG_Framework} summarizes each step of the classification process described in this section. Function $f$ denotes the model obtained by training the classifier.


\textcolor{highlightCol}{In this work, we compare quality measures to determine the best one for selecting the most discriminative set $\mathcal{P}_s$ from $\mathcal{P}$, as their performance is known to vary with both the task and the data \cite{LoyolaGonzalez2014}.} The following section provides an overview of the different quality measures considered in the rest of this article.

\section{Quality Measures}
\label{sec:QualityMeasures}
In this section, we first describe the measures selected to conduct our experiments (Section~\ref{sec:QualityMeasuresDef}), before discussing some of their properties (Section~\ref{sec:QualityMeasuresProp}).

\subsection{Definitions}
\label{sec:QualityMeasuresDef}
As mentioned in the introduction, the literature contains several papers that survey and compare quality measures designed for pattern-based classification~\cite{LoyolaGonzalez2014, Chen2022, Ventura2016, Dong2012}. \textcolor{highlightCol}{Similarly, in feature selection, several works have explored ranking methods for identifying discriminative features}~\cite{Dhal2021, Theng2023, Barbieri2024}. As is often the case, some measures appear in several reviews, such as \textsc{GR} in~\cite{LoyolaGonzalez2014, Chen2022, Dong2012}, and some measures appear under different names, like \textsc{Brins}~\cite{LoyolaGonzalez2014} \& \textsc{Conv}~\cite{Ventura2016}, or \textsc{Conf}~\cite{LoyolaGonzalez2014} \& \textsc{TPR}~\cite{Chen2022}. After resolving these differences, we identify a total of \textcolor{highlightCol}{41} distinct measures among these \textcolor{highlightCol}{seven} surveys.
It is worth stressing that all these measures were originally defined to characterize \textit{general} patterns, identified in \textit{tabular} (or item-based) data. However, their application to collections of graphs is straightforward for all but one, discussed below, by using the notion of graph support defined in Section~\ref{sec:Prob} instead of the standard support~\cite{Dong2012}.

Out of these \textcolor{highlightCol}{41} measures, we exclude three, which leaves us with \textcolor{highlightCol}{38} for our experiments. For the sake of comprehensiveness, the definition of the three discarded measures is provided in Appendix~\ref{sec:AppQM}. First, we discard \textsc{GenQuotient}~\cite{Chen2022}, because it requires the user to set a specific parameter. Moreover, using standard parameter values makes \textsc{GenQuotient} equivalent to other measures already present in our selection, such as \textsc{WRACC}. The second discarded measure is \textsc{SupMaxK}~\cite{Chen2022}, because it cannot handle graphs. Indeed, it requires considering each pattern as a set of separate items. Graph patterns can be seen as sets of vertices and edges, but the adaptation is not trivial. Third and finally, we exclude \textsc{PValue}~\cite{Chen2022}, because its computation is unsuitable to large datasets. Moreover, it should be noted that, unlike all the other measures, \textsc{FPR}, \textcolor{highlightCol}{\textsc{Gini}, and \textsc{Entropy}} assign lower values to stronger discriminative power~\cite{Chen2022, Theng2023}:  we therefore reverse their rankings for the sake of consistency.

\begin{figure}[htbp!]
    \centering
    \includegraphics[width=0.55\textwidth]{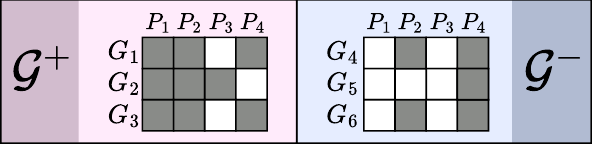} 
    \caption{Simplified representation of a collection $\mathcal{G}$ constituted of six graphs, distributed over two classes $\mathcal{G}^+$ and $\mathcal{G}^-$, and described according to four patterns.}
    \label{fig:Dataset}
    \Description{Description} 
\end{figure}

Figure~\ref{fig:Dataset} provides a visual representation of an example used throughout this section to illustrate the concepts and quality measures described here. It shows a set of six graphs $\mathcal{G} = \{ G_1,...,G_6 \}$, split in two classes $\mathcal{G}^+ = \{ G_1,..., G_3 \}$ and $\mathcal{G}^- = \{ G_4,...,G_6 \}$. This collection contains four distinct patterns $\mathcal{P} = \{ P_1,...,P_4 \}$, symbolized by squares. The graphs are only represented in terms of which patterns they contain (solid squares) or not (empty squares). For example, $G_4$ contains only $P_2$ and $P_4$. In this example, $\support(P_1, \mathcal{G}^+) = 3$, since $P_1$ is present in $G_1$, $G_2$, and $G_3$.

In the context of this paper, graph support can be tied to certain concepts from the classification field. Let us assume that, for a certain pattern $P$, graphs possessing $P$ tend to belong to class $\mathcal{G}^+$, whereas graphs not possessing $P$ tend to belong to class $\mathcal{G}^-$. According to the terminology of the Classification domain, such graphs possessing $P$ and belonging to $\mathcal{G}^+$ are called \textit{True Positives} (TP) and those not possessing $P$ and belonging to $\mathcal{G}^-$ are called \textit{True Negatives} (TN). On the contrary, a graph possessing $P$ but belonging to $\mathcal{G}^-$ is a \textit{False Positive} (FP), and a graph not possessing $P$ but belonging to $\mathcal{G}^+$ is a \textit{False Negative} (FN). Notation $\overline{P}$ expresses the \textit{absence} of pattern $P$, therefore the following support values can be used to count these cases: $\support(P, \mathcal{G}^+)$ (TP), $\support(\overline{P}, \mathcal{G}^-)$ (TN), $\support(P, \mathcal{G}^-)$ (FP), and $\support(\overline{P}, \mathcal{G}^+)$ (FN).

The concept of graph support allows us to derive several probabilities that constitute the building blocks of the quality measures. For convenience, they are all gathered in Table~\ref{tab:Probabilities}. They can be interpreted as follows. 
Probability $p(P)$ is the proportion of graphs in $\mathcal{G}$ that contain $P$ at least once, i.e. the probability of drawing a graph containing $P$ in $\mathcal{G}$. It is simply a normalization of the support: in our example, we have $p(P_1) = 3/6$. 
Probability $p(\mathcal{G}^+)$ is the proportion of graphs in $\mathcal{G}$ that belong to the positive class, i.e. the probability of drawing a graph with label $+$ in $\mathcal{G}$. As explained in Section~\ref{sec:Prob}, in the context of this paper, the classes are assumed balanced. Therefore, $p(\mathcal{G}^+) = p(\mathcal{G}^-) = 0.5$. 
Probability $p(P,\mathcal{G}^+)$ is the proportion of graphs in $\mathcal{G}$ which simultaneously belong to the positive class and contain pattern $P$. In our example, we have $p(P_4,\mathcal{G}^+) = 2/6$. 
Probability $p(\mathcal{G}^+ \mid P)$ is the proportion of positive graphs among the set of graphs containing pattern $P$. It can be seen as the ratio of the support of $P$ in $\mathcal{G}^+$ to the support of $P$ in $\mathcal{G}$. In our example, $p(\mathcal{G}^+ \mid P_4) = 2/5$. 
Probability $p(P \mid \mathcal{G}^+)$ is the proportion of graphs containing pattern $P$ among the set of positive graphs. Note that, because our classes are balanced, $|\mathcal{G}^+| = |\mathcal{G}|/2$, and thus $p(P \mid \mathcal{G}^+) = 2 p(P,\mathcal{G}^+)$. In our example, $p(P_4 \mid \mathcal{G}^+) = 2/3$. 
Probability $p(\overline{P},\mathcal{G}^+)$ is the proportion of graphs in $\mathcal{G}$ that simultaneously belong to the positive class and do not contain $P$. In our example, $p(\overline{P_4},\mathcal{G}^+) = 1/6$, since only graph $G_2$ is in the positive class while not containing $P_4$. 
All these interpretations are also valid for $\mathcal{G}^-$, this time considering graphs of the negative class. 

\begin{table}[htbp!]
    \centering
    \begin{tabular}{ll l ll}
        \toprule
        Probability & Formula & & Probability & Formula \\
        \cmidrule{1-2}\cmidrule{4-5}
        $p(P)$ & $\displaystyle\frac{\support(P,\mathcal{G})}{|\mathcal{G}|}$ & & $p(\overline{P})$ & $1 - p(P)$ \\[3mm]
        $p(\mathcal{G}^+)$ & $\displaystyle\frac{|\mathcal{G}^+|}{|\mathcal{G}|}$ & & $p(\mathcal{G}^-)$ & $1 - p(\mathcal{G}^+)$ \\[3mm]
        $p(P,\mathcal{G}^+)$ & $\displaystyle\frac{\support(P,\mathcal{G}^+)}{|\mathcal{G}|}$ & & $p(P,\mathcal{G}^-)$ & $\displaystyle\frac{\support(P,\mathcal{G}^-)}{|\mathcal{G}|}$ \\[3mm]
        $p(\overline{P},\mathcal{G}^+)$ & $\displaystyle\frac{|\mathcal{G}^+| - \support(P,\mathcal{G}^+)}{|\mathcal{G}|}$ & & $p(\overline{P},\mathcal{G}^-)$ & $\displaystyle\frac{|\mathcal{G}^-| - \support(P,\mathcal{G}^-)}{|\mathcal{G}|}$ \\[3mm]
        $p(\mathcal{G}^+ \mid P)$ & $\displaystyle\frac{\support(P,\mathcal{G}^+)}{\support(P,\mathcal{G})}$ & & $p(\mathcal{G}^+ \mid \overline{P})$ & $\displaystyle\frac{|\mathcal{G}^+| - \support(P,\mathcal{G}^+)}{|\mathcal{G}| - \support(P,\mathcal{G})}$ \\[3mm]
        $p(\mathcal{G}^- \mid P)$ & $\displaystyle\frac{\support(P,\mathcal{G}^-)}{\support(P,\mathcal{G})}$ & & $p(\mathcal{G}^- \mid \overline{P})$ & $\displaystyle\frac{|\mathcal{G}^-| - \support(P,\mathcal{G}^-)}{|\mathcal{G}| - \support(P,\mathcal{G})}$ \\[3mm]
        $p(P \mid \mathcal{G}^+)$ & $\displaystyle\frac{\support(P,\mathcal{G}^+)}{|\mathcal{G}^+|}$ & & $p(P \mid \mathcal{G}^-)$ & $\displaystyle\frac{\support(P,\mathcal{G}^-)}{|\mathcal{G}^-|}$ \\[3mm]
        $p(\overline{P} \mid \mathcal{G}^+)$ & $\displaystyle\frac{|\mathcal{G}^+| - \support(P,\mathcal{G}^+)}{|\mathcal{G}^+|}$ & & $p(\overline{P} \mid \mathcal{G}^-)$ & $\displaystyle\frac{|\mathcal{G}^-| - \support(P,\mathcal{G}^-)}{|\mathcal{G}|}$ \\[3mm]
        \bottomrule
    \end{tabular}
    \caption{Probabilities used to define the quality measures defined in Tables~\ref{tab:SurveyQualityMeasure1} and~\ref{tab:SurveyQualityMeasure2}.}
    \label{tab:Probabilities}
\end{table}

Tables~\ref{tab:SurveyQualityMeasure1} and~\ref{tab:SurveyQualityMeasure2} describe the \textcolor{highlightCol}{38} measures selected for this work. In order to ease the comparison, the formulas are all expressed using the concepts introduced before: graph support, and probabilities from Table~\ref{tab:Probabilities}. For each measure, column \textit{Bounds} shows its lower and upper bounds, and column \textit{Ref}. indicates the paper that originally introduced it. The four remaining columns state whether the measure possesses certain traits, which are discussed later.

Some of these measures take into account the total number of graphs, denoted $|\mathcal{G}|$, in their formula. This is notably the case for \textsc{Pearson} and $\chi^{2}$. Moreover, some measures are defined in relation to other measures: \textsc{Strength} and \textsc{ColStr} use \textsc{GR} and \textsc{Acc} respectively. 

\begin{table}[htbp!]
    \begin{center}
    \begin{tabular}{l l l l c c c c}
        \toprule
         Measure & Definition & Bounds & Ref. & Co. & Ju. & Cs. & Ps.  \\
        \midrule
        \textsc{AbsSupDif} & $\displaystyle |p(P \mid \mathcal{G}^+)-p(P \mid \mathcal{G}^-)|$ & $[0,1]$  & \cite{Thoma2010} & -- & \cmark & \cmark & \cmark \\[+4pt]
        \textsc{Acc} & $\displaystyle p(P, \mathcal{G}^+)+p(\overline{P} ,\mathcal{G}^-)$ & $[0,1]$ & \cite{Kodratoff2001} & \cmark & \cmark & -- & -- \\[+4pt] 
        \textsc{Brins} & $\displaystyle \frac{p(P) p(\mathcal{G}^-)}{p(P ,\mathcal{G}^-)}$ & $[0,+\infty[$ & \cite{Brin1997} & \cmark & -- & -- & -- \\[+8pt] 
        \textsc{CConf} & $\displaystyle p(\mathcal{G}^+ \mid P)-p(\mathcal{G}^+)$ & $[-0.5,0.5]$ & \cite{Lenca2007} & \cmark & -- & -- & -- \\[+4pt] 
        \textsc{CFactor} & $\displaystyle \frac{(p(P,\mathcal{G}^+)/p(P)) - p(\mathcal{G}^+) }{1 - p(\mathcal{G}^+)}$ & $[-1,1]$ & \cite{Ventura2016} & \cmark & -- & -- & -- \\[+8pt] 
        \textsc{Cole} & $\displaystyle \frac{p(\mathcal{G}^+ \mid P)-p(\mathcal{G}^+)}{1-p(\mathcal{G}^+)}$ & $[-1,1]$ & \cite{An1999} & \cmark & -- & -- & -- \\[+8pt] 
        \textsc{ColStr} & $\displaystyle \frac{p(P,\mathcal{G}^+)+p(\overline{P},\mathcal{G}^-)}{p(P)p(\mathcal{G}^+)+p(\overline{P})p(\mathcal{G}^-)} \frac{1-p(P) p(\mathcal{G}^+)- p(\overline{P})p(\mathcal{G}^-)}{1-p(P ,\mathcal{G}^+)-p(\mathcal{G}^- \mid \overline{P})}$ & $[-10,+\infty[$ & \cite{Tan2004} & -- & \cmark & -- & -- \\[+12pt]
        \textsc{Conf} & $\displaystyle p(\mathcal{G}^+ \mid P)$ & $[0,1]$ & \cite{Agrawal1993} & \cmark & -- & -- & -- \\[+6pt]
        \textsc{Cos} & $\displaystyle \sqrt{p(\mathcal{G}^+ \mid P) p(P \mid \mathcal{G}^+)}$ & $[0,1]$ & \cite{Tan2004} & \cmark & \cmark & -- & -- \\[+4pt]
        \textsc{Cover} & $\displaystyle p(P \mid \mathcal{G}^+)$ & $[0,1]$ & \cite{An2001} & -- & \cmark & -- & -- \\[+4pt]
        \textsc{Dep} & $\displaystyle |p(\mathcal{G}^-)-p(\mathcal{G}^- \mid P)|$ & $[0,0.5]$ & \cite{Kodratoff2001} & -- & -- & \cmark & -- \\[+4pt]
        \textcolor{highlightCol}{\textsc{Entropy}} & \textcolor{highlightCol}{$\displaystyle - p(\mathcal{G}^+ \mid P) \log p(\mathcal{G}^+ \mid P) - p(\mathcal{G}^- \mid P) \log p(\mathcal{G}^- \mid P)$} & \textcolor{highlightCol}{$[0,1]$} & \cite{Shannon1948} & \textcolor{highlightCol}{--} & \textcolor{highlightCol}{--} & \textcolor{highlightCol}{\cmark} & \textcolor{highlightCol}{--} \\[+4pt]
        \textsc{Excex} & $\displaystyle 1-\frac{p(\mathcal{G}^- \mid P)}{p(\mathcal{G}^+ \mid \overline{P})}$ & $]-\infty,1]$ & \cite{Gras1996} & \cmark & -- & -- & -- \\[+8pt]
        \textcolor{highlightCol}{\textsc{Fisher}} & \textcolor{highlightCol}{$\displaystyle \frac{(p(\mathcal{G}^+ \mid P) - p(\mathcal{G}^- \mid P))^2}{p(\mathcal{G}^+ \mid P)(1 - p(\mathcal{G}^+ \mid P)) + p(\mathcal{G}^- \mid P)(1 - p(\mathcal{G}^- \mid P))}$} & \textcolor{highlightCol}{$[0,+\infty[$} & \cite{Fisher1936} & \textcolor{highlightCol}{--} & \textcolor{highlightCol}{--} & \textcolor{highlightCol}{\cmark} & \textcolor{highlightCol}{--} \\[+8pt]
        \textsc{FPR} & $\displaystyle p(\mathcal{G}^+ \mid \overline{P})$ & $[0,1]$ & \cite{GarciaVico2020} & \cmark & \cmark & -- & -- \\[+4pt]
        \textsc{Gain} & $\displaystyle p(P,\mathcal{G}^+)(\log p(\mathcal{G}^+ \mid P)-\log p(\mathcal{G}^+))$ & $[-1,1]$ & \cite{Yin2003} & \cmark & \cmark & -- & -- \\[+4pt]
        \textcolor{highlightCol}{\textsc{Gini}} & \textcolor{highlightCol}{$\displaystyle 1 - p(\mathcal{G}^+ \mid P)^2 - p(\mathcal{G}^- \mid P)^2$} & \textcolor{highlightCol}{$[0,0.5]$} & \cite{Breiman1984} & \textcolor{highlightCol}{--} & \textcolor{highlightCol}{--} & \textcolor{highlightCol}{\cmark} & \textcolor{highlightCol}{--} \\[+4pt]  
        \textsc{GR} & $\displaystyle \frac{p(P \mid \mathcal{G}^+)}{p(P \mid \mathcal{G}^-)}$ & $[0,+\infty[$ & \cite{Dong1999} & \cmark & -- & -- & -- \\[+8pt]
        \textsc{InfGain} & $\displaystyle -\log p(\mathcal{G}^+)+\log p(\mathcal{G}^+ \mid P)$ & $]-\infty,0]$ & \cite{Church1990} & \cmark & -- & -- & -- \\[+4pt]
        \bottomrule
\end{tabular}
    \end{center}
    \caption{Name and formula of the first \textcolor{highlightCol}{19} quality measures considered in this article, out of \textcolor{highlightCol}{38} in total. The \textcolor{highlightCol}{19} remaining measures are listed in Table~\ref{tab:SurveyQualityMeasure2}. The four rightmost columns indicate the measure properties as discussed in the main text: Contrastivity (\textit{Co.}), Jumpiness (\textit{Ju.}), Class Symmetry (\textit{Cs.}) and Pattern Symmetry (Ps.).}
    \label{tab:SurveyQualityMeasure1}
\end{table}

\begin{table}[htbp!]
    \begin{center}
    \begin{tabular}{l l l l c c c c}
        \toprule
        Measure & Definition & Bounds & Ref. & Co. & Ju. & Cs. & Ps. \\
        \midrule        
        \textsc{Jacc} & $\displaystyle \frac{p(P,\mathcal{G}^+)}{p(P)+p(\mathcal{G}^+)-p(P,\mathcal{G}^+)}$ & $[0,1]$ & \cite{Tan2004} & \cmark & \cmark & -- & -- \\
        \textsc{Klos} & $\displaystyle \sqrt{p(P,\mathcal{G}^+)}(p(\mathcal{G}^+ \mid P)-p(\mathcal{G}^+))$ & [0,1] & \cite{Klosgen1996} & \cmark & \cmark & -- & -- \\[+4pt]
        \textsc{Lap} & $\displaystyle \frac{p(P, \mathcal{G}^+)+1 / |\mathcal{G}|}{p(P)+2 / |\mathcal{G}|}$ & $[0,1]$ & \cite{Harris1966} & \cmark & \cmark & -- & -- \\[+8pt]
        \textsc{Lever} & $\displaystyle p(P, \mathcal{G}^+)-p(P) p(\mathcal{G}^+)$ & $[-0.25,0.25]$ & \cite{Webb2005} & \cmark & \cmark & -- & -- \\[+4pt]
        \textsc{Lift} & $\displaystyle \frac{p(P, \mathcal{G}^+)}{p(P) p(\mathcal{G}^+)}$ 
        & $[0,2]$ & \cite{PiatetskyShapiro2000} & \cmark & -- & -- & -- \\[+8pt]
        \textsc{MDisc} & $\displaystyle \log \left(\frac{p(P ,\mathcal{G}^+) p(\overline{P}, \mathcal{G}^-)}{p(P ,\mathcal{G}^-) p(\overline{P},\mathcal{G}^+)}\right)$ & $]-\infty,+\infty[$ & \cite{An1998} & \cmark & -- & -- & -- \\[+8pt]
        \textsc{MutInf} & $\displaystyle \sum_{P_i \in \{P, \overline{P}\}} \sum_{\mathcal{G}_i \in \{\mathcal{G}^+, \mathcal{G}^-\}} \ p(P_i,\mathcal{G}_i) \log \frac{p(P_i,\mathcal{G}_i)}{p(P_i) p(\mathcal{G}_i)}$ & $[0,1]$ & \cite{Fang2011} & -- & \cmark & \cmark & \cmark \\[+8pt]
        \textsc{NetConf} & $\displaystyle \frac{p(\mathcal{G}^+ \mid P)-p(\mathcal{G}^+)}{1-p(P)}$ 
        & $[-1,1]$ & \cite{Ahn2004} & \cmark & \cmark & -- & -- \\[+8pt]
        \textsc{OddsR} & $\displaystyle \frac{p(P, \mathcal{G}^+) /(1-p(P, \mathcal{G}^+))}{p(P, \mathcal{G}^-) /(1-p(P, \mathcal{G}^-))}$ 
        & $[0,+\infty[$ & \cite{Tan2004} & \cmark & -- & -- & -- \\[+8pt]
        \textsc{Pearson} & $\displaystyle \frac{p(P, \mathcal{G}^+)-p(P) p(\mathcal{G}^+)}{\sqrt{N p(P) p(\mathcal{G}^+) p(\overline{P}) p(\mathcal{G}^-)}}$ & $[-1,1]$ & \cite{Pearson1896} & \cmark & \cmark & -- & -- \\[+12pt]
        \textsc{RelRisk} & $\displaystyle \frac{p(\mathcal{G}^+ \mid P)}{p(\mathcal{G}^+ \mid \overline{P})}$ 
        & $[0,+\infty[$ & \cite{Ali1997} & \cmark & \cmark & -- & -- \\[+8pt]
        \textsc{Sebag} & $\displaystyle \frac{p(P, \mathcal{G}^+)}{p(P, \mathcal{G}^-)}$ & $[0,+\infty[$ & \cite{Sebag1988} & \cmark & -- & -- & -- \\[+8pt]
        \textsc{Spec} & $\displaystyle p(\mathcal{G}^- \mid \overline{P})$ & $[0,1]$ & \cite{Lavrac1999} & \cmark & \cmark & -- & -- \\[+4pt]
        \textsc{Strength} & $\displaystyle \frac{\textsc{GR}(P,\mathcal{G}^+,\mathcal{G}^-)}{\textsc{GR}(P,\mathcal{G}^+,\mathcal{G}^-)+1}p(P, \mathcal{G}^+)$ & $[0,1]$ & \cite{Ramamohanarao2007,GarciaVico2020} & \cmark & \cmark & -- & -- \\[+8pt]
        \textsc{Sup} & $\displaystyle p(P,\mathcal{G}^+)$ & $[0,1]$ & \cite{Agrawal1993} & -- & \cmark & -- & -- \\[+4pt]
        \textsc{SupDif} & $\displaystyle p(P \mid \mathcal{G}^+)-p(P \mid \mathcal{G}^-)$ & $[-1,1]$ & \cite{LoyolaGonzalez2014} & \cmark & \cmark & -- & -- \\[+4pt]
        \textsc{WRACC} & $\displaystyle p(P)(p(\mathcal{G}^+ \mid P) -p(\mathcal{G}^+))$ & $[-1,1]$ & \cite{Lavrac2004,GarciaVico2020} & \cmark & \cmark & -- & -- \\[+4pt]
        \textsc{Zhang} & $\displaystyle \frac{p(P,\mathcal{G}^+)-p(P)p(\mathcal{G}^+)}{\max\{p(P, \mathcal{G}^+)p(\mathcal{G}^-),p(\mathcal{G}^+)p(P, \mathcal{G}^-)\}}$ & $[-1,1]$ & \cite{Zhang2011} & \cmark & -- & -- & -- \\[+8pt]
        $\chi^{2}$ & $\displaystyle N\frac{(p(P,\mathcal{G}^+)p(\overline{P},\mathcal{G}^-)-p(P,\mathcal{G}^-)p(\overline{P},\mathcal{G}^+))^{2}}{p(P)p(\mathcal{G}^+)p(\overline{P})p(\mathcal{G}^-)}$ & $[0,+\infty[$ & \cite{Bay1999} & -- & \cmark & \cmark & \cmark \\
        \bottomrule
        \end{tabular}
    \end{center}
    \caption{Name and formula of the last \textcolor{highlightCol}{19} quality measures considered in this article, out of \textcolor{highlightCol}{38} in total. The \textcolor{highlightCol}{19} other measures are listed in Table~\ref{tab:SurveyQualityMeasure1}.}
    \label{tab:SurveyQualityMeasure2}
\end{table}

Most measures have fixed bounds: $[0;1]$ \textcolor{highlightCol}{(14 measures)}, $[-1;1]$ (8), $[0;0.5]$ \textcolor{highlightCol}{(2)}, $[0;2]$ (\textsc{Lift}), $[-0.25;0.25]$ (\textsc{Lever}), $[-0.5;0.5]$ (\textsc{CConf}). A few have a fixed lower bound, but do not have any upper bound, i.e. $[0;+\infty[$ \textcolor{highlightCol}{(7)} or $[-10;+\infty[$ (\textsc{ColStr}). A few have a fixed upper bound, but do not have any lower bound, i.e. $]-\infty;0]$ (\textsc{InfGain}) or $]-\infty;1]$  (\textsc{Excex}). Only one has no bound at all, i.e. $]-\infty;+\infty[$ (\textsc{MDisc}). These differences have no effect on the experimental comparison that we perform later, as we consider how the measures \textit{rank} the patterns, by opposition to comparing directly the scores obtained with these measures.

\subsection{Properties}
\label{sec:QualityMeasuresProp}
In their survey, \citet{LoyolaGonzalez2014} propose a classification of quality measures into two categories: 
\begin{itemize}
    \item Those based on the notion of \textit{Independence}. A pattern $P$ is considered independent of the classes if $p(P,\mathcal{G}^+) = p(P) p(\mathcal{G}^+)$. The patterns considered as discriminative will then be those that deviate from such independence.
    \item Those based on the notion of \textit{Equilibrium}. A pattern $P$ is considered in equilibrium over the classes if $p(\mathcal{G}^+ \mid P) = p(\mathcal{G}^- \mid P)$. The patterns considered as discriminative will then be those that deviate from such equilibrium.
\end{itemize}

Under the assumption of balanced classes, however, the notions of \textit{Independence} and \textit{Equilibrium} are equivalent (cf. Appendix~\ref{sec:Proofs}), and this distinction is therefore irrelevant. \citet{Ventura2016} surveys seven other alternative mathematical properties proposed in the literature. However, they are designed in the context of association rule mining~\cite{Zhang2002}. As a consequence, some of them are irrelevant, or do not apply to subgraph pattern mining. We describe all these properties and discuss them in Appendix~\ref{sec:AddProps}. Our analysis reveals that three of these properties (\textit{Contrastivity}, \textit{Class Symmetry} and \textit{Pattern Symmetry}) are suitable to our case. In addition, we define another property (\textit{Jumpiness}) in order to fully describe the quality measures. In the end, we propose to characterize each measure using these four properties, which we present below. The last four columns in Tables~\ref{tab:SurveyQualityMeasure1} and~\ref{tab:SurveyQualityMeasure2} indicate whether a measure possesses these properties (\cmark) or not (--).

\paragraph{Contrastivity}
The concept of contrast has several slightly different meanings in the field of pattern mining~\cite{Chen2022, Dong2012}. In the specific context of graph pattern mining, \textit{contrast patterns} are subgraphs that belong specifically to the positive class, and do not appear at all in the negative class~\cite{Dong2012, Hellal2016, Chen2022}. Based on this notion designed to characterize patterns, we derive the property of \textit{Contrastivity}, which aims at describing quality measures. Our idea is to distinguish between measures that assess the discriminative power of patterns depending \textit{only} on their abundance in the positive class, and measures that \textit{also} consider their scarcity in the negative class. Formally, this translates as follows:
\begin{definition}[Contrastivity]
    \label{def:Contrastivity}
    A quality measure $q$ respects the \textit{Contrastivity} property iff 
    \begin{align*}
    \forall P_i,P_j,\; \big[ p(P_i, \mathcal{G}^+) = p(P_j, \mathcal{G}^+) \wedge p(P_i, \mathcal{G}^-) < p(P_j, \mathcal{G}^-) \big] \Rightarrow \big[ q(P_i,\mathcal{G}^+, \mathcal{G}^-) > q(P_j,\mathcal{G}^+, \mathcal{G}^-)  \big].
    \end{align*}
\end{definition}
This property is equivalent to PS3, the third property of \citet{PiatetskyShapiro1991a}, which we describe in Appendix~\ref{sec:AddPropsPSF}. As shown in Table~\ref{tab:Probabilities}, all the probabilities in the left term of Definition~\ref{def:Contrastivity} rely on the same denominator, therefore this term can be simplified, and expressed using only support: $[ \support(P_i, \mathcal{G}^+) = \support(P_j, \mathcal{G}^+) \wedge \support(P_i,\mathcal{G}^-) < \support(P_j,\mathcal{G}^-) ]$. More intuitively, when two patterns have the same support in $\mathcal{G}^+$, a contrastive quality measure will favor the one possessing the smallest support in $\mathcal{G}^-$. In other words, measures that possess the \textit{Contrastivity} property take into account \textit{False Positives} as defined in Section~\ref{sec:QualityMeasuresDef}.

Consider, for example, measure \textsc{Cover}, defined as $p(P \mid \mathcal{G}^+)$. By construction, if two patterns such as $P_1$ and $P_2$ in Figure~\ref{fig:Dataset} have the same support in $\mathcal{G}^+$ (here: 3), they have the same \textsc{Cover} score (here: 1), independently of their support in $\mathcal{G}^-$ (here: 0 and 2, respectively). Therefore, this measure does not possess the property of \textit{Contrastivity}. Let us now consider measure \textsc{GR}, defined as $p(P \mid \mathcal{G}^+)/p(P \mid \mathcal{G}^-)$. If both patterns have the same support in $\mathcal{G}^+$, then their numerators are equal, and the pattern with the largest support in $\mathcal{G}^-$ gets the smaller ratio. Thus, \textsc{GR} respects the \textit{Contrastivity} property.

It is worth stressing that this property treats both classes differently, and is not necessarily desirable, depending on the considered application. It is particularly the case if one gives as much importance to $\mathcal{G}^-$ as to $\mathcal{G}^+$. Assume that two patterns $P_i$ and $P_j$ are scarce in $\mathcal{G}^+$, but with the same support, and that they are common in $\mathcal{G}^-$, with different supports. Then, if the support of $P_i$ in $\mathcal{G}^-$ is higher than that of $P_j$, a contrastive quality measure will output a lower score for this pattern, when it is in fact more relevant than $P_j$ to distinguish both classes. This justifies additionally considering symmetry-related properties to fully characterize the measures: class symmetry, in particular, allows checking whether \textit{Contrastivity} also applies from the perspective of the negative class.

\paragraph{Jumpiness}
Previous works~\cite{Li2000} have introduced the notion of \textit{jumping emerging} pattern, which are patterns present in only one class. All such patterns are not necessarily interesting for classification: for example, a pattern appearing in a single graph is not very discriminative, overall. Nevertheless, some measures assign the same value to all emerging patterns, which makes it impossible to distinguish one jumping emerging pattern from another. As mentioned by \citet{LoyolaGonzalez2014}, this can lead to retaining a set of poorly discriminative patterns. Based on this observation, we define the property of \textit{Jumpiness}, which concerns quality measures able to distinguish between jumping emerging patterns.
\begin{definition}[Jumpiness]
    \label{def:Jumpiness}
    A quality measure $q$ respects the \textit{Jumpiness} property iff 
    \begin{align*}
    \forall P_i,P_j,\; \big[ p(\mathcal{G}^+ \mid P_i)=p(\mathcal{G}^+ \mid P_j)=1 \wedge p(P_i,\mathcal{G}^+) > p(P_j,\mathcal{G}^+) \big] \Rightarrow \big[ q(P_i,\mathcal{G}^+,\mathcal{G}^-) > q(P_j,\mathcal{G}^+,\mathcal{G}^-) \big].
    \end{align*}
\end{definition}
In other words, if all the occurrences of $P_i$ and $P_j$ belong to $\mathcal{G}^+$, then the most frequent of the two patterns gets a higher quality value. Consequently, measures possessing the \textit{Jumpiness} property take into account \textit{False Negatives} as defined in Section~\ref{sec:QualityMeasuresDef}. In addition, a measure that does not respect this property ranks contrast subgraphs of the positive class better than every other pattern, since contrast subgraphs are jumping emerging patterns. This property is original, it is not equivalent to any other from the literature (see Appendix~\ref{sec:AddProps}).

Consider $P_1$ and $P_3$ in our example from Figure~\ref{fig:Dataset}, and quality measure \textsc{Conf}, i.e. $p(\mathcal{G}^+ \mid P)$. Both patterns occur only in $\mathcal{G}^+$, and $P_1$ is more frequent than $P_3$ (3 vs. 1). Yet, \textsc{Conf} is 1 for both patterns, and consequently does not possess the \textit{Jumpiness} property. Consider now measure \textsc{Supdif}, i.e. $p(P \mid \mathcal{G}^+) - p(P \mid \mathcal{G}^-)$. The antecedent of  Definition~\ref{def:Contrastivity} yields $p(P \mid \mathcal{G}^-) = 0$, as both patterns are only present in $\mathcal{G}^+$. Therefore, in this case the measure only depends on $p(P \mid \mathcal{G}^+)$ and respects the Jumpiness property.

\paragraph{Class Symmetry}
In both previous properties, the positive and negative classes do not hold the same role. This is because in certain applications, users consider the positive class as the class of interest, and handle it differently from the negative class. Therefore, it can be interesting to distinguish these measures from those that make no difference between the classes. For this purpose, we define the \textit{Class Symmetry} property, which concerns measures for which both classes are interchangeable.
\begin{definition}[Class Symmetry]
    \label{def:ClassSym}
    A quality measure $q$ respects the Class Symmetry property iff
    \begin{align*}
    \forall P,\; q(P,\mathcal{G}^+,\mathcal{G}^-) = q(P,\mathcal{G}^-,\mathcal{G}^+).
    \end{align*}
\end{definition}
Our Class Symmetry is similar to T2b (Column Antisymmetry), the second variant of the second property defined by \citet{Tan2004}, except for the sign of the right-hand term (cf. Appendix~\ref{sec:AddPropsT}).

Consider $P_2$ in the example from Figure~\ref{fig:Dataset}, and measure \textsc{Conf}, defined as $p(\mathcal{G}^+ \mid P)$. We have $\textsc{Conf}(P_2, \mathcal{G}^+, \mathcal{G}^-) = 0.6$ and $\textsc{Conf}(P_2, \mathcal{G}^-, \mathcal{G}^+) = 0.4$, therefore this measure is not class symmetric. Consider measure \textsc{Dep} instead, which is defined as $|p(\mathcal{G}^-) - p(\mathcal{G}^- \mid P)|$. Using $p(\mathcal{G}^- \mid P) + p(\mathcal{G}^+ \mid P) = 1$, we can replace $p(\mathcal{G}^- \mid P)$ by $1 - p(\mathcal{G}^+ \mid P)$ in $\textsc{Dep}$, and get $|p(\mathcal{G}^-) - 1 + p(\mathcal{G}^+ \mid P)|$. 
Probabilities $p(\mathcal{G}^-)$ and $p(\mathcal{G}^+)$ also sum to one, 
which yields $|-p(\mathcal{G}^+) + p(\mathcal{G}^+ \mid P)| = |p(\mathcal{G}^+) - p(\mathcal{G}^+ \mid P)|$. In the end, $\textsc{Dep}(P, \mathcal{G}^+, \mathcal{G}^-) = \textsc{Dep}(P, \mathcal{G}^-, \mathcal{G}^+)$, and the property follows for this measure. 

\paragraph{Pattern Symmetry}
Classes can be characterized in terms of the presence of certain patterns, but also in terms of their absence. The \textit{Pattern Symmetry} property concerns measures for which being absent from a class is as important as being present, when ranking the patterns.
\begin{definition}[Pattern Symmetry]
    \label{def:PatternSym}
    A quality measure $q$ respects the Pattern Symmetry property iff
    \begin{align*}
    \forall P,\; q(P,\mathcal{G}^+,\mathcal{G}^-) = q(\overline{P},\mathcal{G}^+,\mathcal{G}^-).
    \end{align*}
\end{definition}
Pattern Symmetry is similar to T2a (Row Antisymmetry), the second variant of the second property defined by \citet{Tan2004}, except for the sign of the right-hand term (cf. Appendix~\ref{sec:AddPropsT}). When performing frequent pattern mining, one identifies the patterns with the highest overall support. In this context, it may be interesting to distinguish between measures that treat similarly the presence and absence of patterns, from those that favor their presence.

Consider $P_2$ in the example from Figure~\ref{fig:Dataset}, and measure \textsc{Conf}, defined as $p(\mathcal{G}^+ \mid P)$. We have $\textsc{Conf}(P_2, \mathcal{G}^+, \mathcal{G}^-) = 0.6$ and $\textsc{Conf}(\overline{P_2}, \mathcal{G}^+, \mathcal{G}^-) = 0$, therefore this measure is not pattern symmetric. Consider now measure \textsc{AbsSupDif}, which is defined as $|p(P \mid \mathcal{G}^+) - p(P \mid \mathcal{G}^-)|$. By definition, $p(P \mid \mathcal{G}^i) + p(\overline{P} \mid \mathcal{G}^i) = 1$ for $G^i\in \{G^+,G^-\}$, therefore we get $|1 - p(\overline{P} \mid \mathcal{G}^+) - 1 + p(\overline{P} \mid \mathcal{G}^-)| = |p(\overline{P} \mid \mathcal{G}^+) - p(\overline{P} \mid \mathcal{G}^-)|$. As a consequence, $\textsc{AbsSupDif}(P,\mathcal{G}^+,\mathcal{G}^-) = \textsc{AbsSupDif}(\overline{P},\mathcal{G}^+,\mathcal{G}^-)$, and this measure is pattern symmetric.

\section{Methodological Framework}
\label{sec:Meth}
\begin{figure}[hbtp!]
    \centering \includegraphics[width=\textwidth]{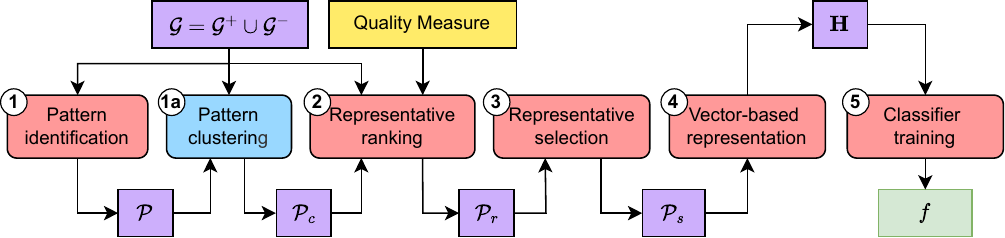}
    \caption{Processing steps of the extended framework, including the additional clustering step (1a), in blue.}
    \label{fig:UpdateFlowchart}
    \Description{Description} 
\end{figure}

The methods that we propose to assess the quality measures listed in Section~\ref{sec:QualityMeasures} rely on two distinct comparisons, each one corresponding to a specific step of the graph classification pipeline described in Section~\ref{sec:Prob} (see Figure~\ref{fig:PANG_Framework}). The first comparison is \textit{direct}, as it focuses on the way the quality measures rank the patterns, i.e. Step~2. It entails a major methodological difficulty: distinct patterns may result in similar vector representation, and this should be accounted for when comparing rankings. For this purpose, we propose an additional pattern clustering step, denoted Step~1a, and described in Section~\ref{sec:MethClust}. Its main effect is that the rest of the pipeline deals with only a subset of the original patterns, called \textit{representatives}. Figure~\ref{fig:UpdateFlowchart} shows the pipeline resulting from this modification, with the additional step in blue. We then discuss the appropriate correlation coefficients to compare pattern rankings in Section~\ref{sec:MethRanks}.

The second comparison is \textit{indirect}, as it relies on a \textit{task-driven evaluation} guided by the classification performance obtained after training (Step~5). The classifier training and its subsequent performance depend on the vector representation of the graphs, which is based itself on how the patterns are ranked using the quality measure, hence the indirect nature of this comparison. It is methodologically much simpler than ranking comparison, though, as it only requires standard classification performance measures, which we discuss in Section~\ref{sec:MethPerf}.

\subsection{Clustering Step}
\label{sec:MethClust}
In Section~\ref{sec:Prob}, we introduce the notion of footprint of a pattern: it corresponds to $\mathbf{h}_{:j}$, the binary vector in which each value $h_{ij}$ indicates whether pattern $P_j$ is present or absent from graph $G_i$. At this stage, we consider $\mathcal{P}$, the full set of patterns detected at Step~1 (Pattern identification). Let us assume that two patterns $P_i$ and $P_j$ have the same footprint, i.e. $\mathbf{h}_{:i} = \mathbf{h}_{:j}$. In other words, these patterns appear in exactly the same graphs of $\mathcal{P}$, and are absent from exactly the same graphs, too. Consequently, they have the same discriminative power, independently of the considered measure. In summary, they are two different subgraphs (i.e. they may contain different vertices and edges) that are identical from the classification perspective. Thus, they are interchangeable in the pattern ranking produced at Step~2.

\begin{figure}[hbtp!]
    \centering \includegraphics[width=0.75\textwidth]{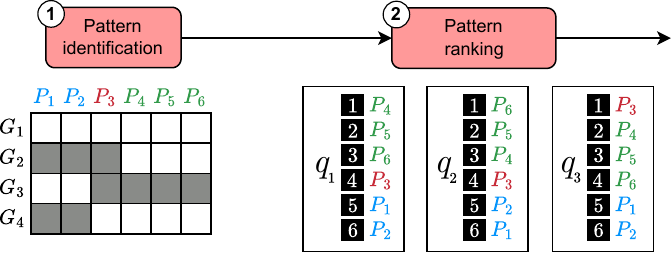}
    \caption{Illustration of the issue occurring when comparing rankings of patterns possessing similar footprints in the original framework.}
    \label{fig:RankingIssue}
    \Description{Description} 
\end{figure}

This behavior is an issue when comparing pattern rankings, as illustrated by Figure~\ref{fig:RankingIssue}. Its left part shows a dataset constituted of four graphs. Step~1 results in the identification of six patterns $P_1,...,P_6$. Some of them share the same footprint, as indicated by their colors. The right part of the figure shows the rankings obtained by three different quality measures ($q_1$, $q_2$, $q_3$). Patterns having the same footprint are likely to be placed consecutively in each ranking, but not necessarily in the same exact order. Measures $q_1$ and $q_2$ agree that the green footprint is the most discriminative, followed by the red and blue ones, but they differ in the way they rank patterns within a color group. As a consequence, using the top $k$ patterns of each ranking to train the classifier results in the exact same classification performance, as same-color patterns are interchangeable from the classification perspective. However, comparing the rankings with a measure such as Kendall's Tau (see Section~\ref{sec:MethRanksComp}) does not lead to a maximal similarity (here: $\tau = 0.47$). By comparison, the ranking obtained with the third quality measure $q_3$ places the red footprint before the green one, which is likely to affect the classification performance. However, it is more similar to $q_1$ in terms of rank correlation ($\tau = 0.60$), which constitutes an undesirable behavior.

\begin{figure}[hbtp!]
    \centering \includegraphics[width=\textwidth]{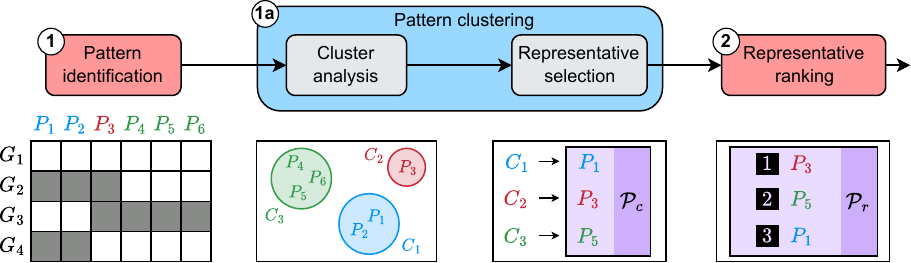}
    \caption{Detail of the pattern clustering process in our extended framework (Step 1a in Figure~\ref{fig:UpdateFlowchart}), in blue.}
    \label{fig:ClusteringFlowchart}
    \Description{Description} 
\end{figure}

In order to tackle this issue, we propose an additional Step~1a in the graph classification pipeline, which is detailed in Figure~\ref{fig:ClusteringFlowchart}. The top part of the figure positions this step in the general pipeline, whereas the bottom part provides an example that we discuss throughout this section. Step~1a takes place before Step~2 (\textit{Pattern Ranking}), and consists in performing a cluster analysis of the patterns identified at Step~1, in order to constitute groups of patterns with similar footprints. Grouping patterns with \textit{identical} footprints is necessary to conduct proper ranking comparison, but we hypothesize that more relaxed groups could also be a way to reduce noise in $\mathcal{P}_s$, the subset of patterns selected at Step~3 and used to build the vectors representing the graphs. For this reason, we do not focus only on strictly equal footprints, but also experiment with clusters that include patterns whose footprints are \textit{similar} (and not necessarily identical).

We do not have any \textit{a priori} idea of the number of clusters to expect, which rules out methods requiring the user to specify this as a parameter, such as $k$-means. Instead, we want to control how strict the method is when grouping patterns. Put differently, we want to specify how similar two patterns should be to be placed in the same cluster. For this reason, we adopt a standard hierarchical agglomerative method~\cite{Gentle1991}. In such ascending methods, all clusters are initially singletons, which are iteratively merged until only one cluster containing all the elements remains. The merged clusters are decided depending on the so-called \textit{linkage criterion}, a function that computes the dissimilarity between two clusters based on the dissimilarity between their constituting elements. We use the \textit{complete-linkage} criterion: the dissimilarity between two clusters corresponds to that of their most distant elements. In order to compare the elements themselves (i.e. the footprints), we select the Manhattan distance. Its advantage here is that its interpretation is straightforward: it corresponds to the number of graphs for which the compared patterns differ (i.e. one is present and the other absent).  Combining the Manhattan distance and the complete-linkage in such a way is particularly suitable to our case. Indeed, it means that, for a given iteration, all the patterns belonging to the same cluster are \textit{at most} as dissimilar as the last two merged clusters.

A hierarchical clustering method produces a \textit{dendrogram}, i.e. a tree showing the successive partitions: from singleton clusters at the first level, to a single all-encompassing cluster at the top level. In our case, the second level corresponds to clusters containing patterns with strictly \textit{identical} footprints, whereas higher levels gather \textit{similar} footprints together. The higher the level, the larger the clusters, and the less similar their constituting patterns' footprints. Thus, selecting a level in the dendrogram amounts to controlling how similar the footprints need to be for the patterns to be considered as interchangeable. We find it convenient to characterize each level by its so-called \textit{distance threshold}: the maximal distance allowed between two patterns belonging to the same cluster. In level 1, each pattern forms a cluster on its own; whereas in level 2, each cluster gathers patterns whose footprints are identical. In level 3, the distance between all patterns constituting a cluster is at most 1, meaning the patterns are present/absent from the same graphs but one. In level 4, this distance is at most 2, and so on.

The selection of the most appropriate level of the dendrogram is conducted empirically, which is why we discuss it later, in Section~\ref{sec:Exp}. Let us assume for now that we are able to pick the best level: at this stage, we have a partition of the pattern set $\mathcal{P}$, and we consider each of its clusters as a group of interchangeable patterns. Next, we compute the \textit{medoid} of each cluster, i.e. the pattern that minimizes the total distance to the rest of its cluster~\cite{Sokal1958}. We call this pattern the \textit{representative} of the cluster. All the other patterns from the same cluster are considered interchangeable with their representative, therefore we can ignore them in the rest of the process. This makes it possible to reduce the dimension of the representation space, and consequently the processing costs. We build $\mathcal{P}_c$, the subset of $\mathcal{P}$ that only contains the representatives. This set is then used in the next step, which consists in ranking the patterns (Step~2). 
It is worth stressing that the obtained clusters are also interesting from the perspective of interpretation: they allow identifying patterns that are possibly very different structurally, but exhibit similar footprints, and are therefore characteristic of the same class.

In the example represented in the bottom half of Figure~\ref{fig:ClusteringFlowchart}, the diagram associated to Step~1 represents a set of four graphs described by six patterns (the same as in Figure~\ref{fig:RankingIssue}). The result of the cluster analysis is shown next to it: three groups of patterns noted $C_1$, $C_2$ and $C_3$. The next part of the example shows the selection of a representative (medoid) for each cluster: $P_1$ for $C_1$; $P_3$ for $C_2$; and $P_5$ for $C_3$. These three patterns constitute $\mathcal{P}_c$, which is fetched to Step~2 (ranking step) in order to produce $\mathcal{P}_r$ and go on with the standard process.

\subsection{Ranking Evaluation}
\label{sec:MethRanks}
We now focus on our first method aiming at comparing the measures, which relies on the rankings they produce. The purpose of each quality measure is to rank the patterns depending on their hypothesized discriminative power. In order to assess the quality of such rankings, we need two resources: first, a gold standard, i.e. the ranking which is optimal for our classification task, and second, a coefficient able to assess the similarity between this gold standard and the ranking estimated by a given quality measure. Moreover, this coefficient can be used to compare pairs of measures, through their rankings. To elaborate the ground truth, we propose an approach based on the Shapley Value, which we introduce in Section~\ref{sec:MethRanksShap}. To compare rankings, we select two tools, described in Section~\ref{sec:MethRanksComp}: Kendall's Tau, a very widespread rank correlation coefficient~\cite{Kendall1938}, and Rank-Biased Overlap~\cite{Webber2010}, which is able to handle cases where the ranked objects are not exactly the same on both ranked lists.

\subsubsection{Shapley Value}
\label{sec:MethRanksShap}
The Shapley Value~\cite{Shapley1953} (SV) originates from Game Theory. It was designed in the context of cooperative games, to assess the contribution of individual players among coalitions of players. 
\begin{equation}
    SV(p) = \sum_{S \subset \mathcal{S} \setminus\{p\}} \frac{|S|! \, (|\mathcal{S}| - |S| - 1)!}{|\mathcal{S}|!} \big( f(S \cup \{p\}) - f(S) \big),
\end{equation}
\noindent where $p$ is a player, $\mathcal{S}$ is the set of all players, $S$ is used to consider all possible coalitions that exclude $p$, i.e. subsets of $\mathcal{S} \setminus\{p\}$, and $f(X)$ is the so-called \textit{characteristic function}, which assesses the contribution of coalition $X$. The right-hand factor is the difference between the contribution of a coalition $S$ with an additional player $p$ and the same coalition without $p$. One averages this quantity over all possible coalitions to compute the SV of $p$.

The SV was later used in machine learning, to perform feature selection, as it allows measuring how much individual features contribute to solving the task at hand~\cite{Fryer2021}. For instance, suppose that one has a collection of data points, described by certain features, and that they leverage them to train a classifier. In this analogy, the features correspond to the players of the original definition, and $f$ is the measure used to assess classification performance. In this context, the SV is interpreted as a score that represents the impact of a feature on the final performance.

The main drawback of the SV is its algorithmic complexity, which requires considering all possible coalitions of players (or combinations of features): this prevents from computing it in many real-world situations~\cite{Kumar2021}. For this reason, in practice, one generally computes an approximation. Several scores were designed specifically for machine learning applications, based on such approximations~\cite{Lundberg2017, Covert2020, Lundberg2020}. These approaches differ according to two aspects: 
\begin{itemize}
    \item Explanation \textit{scope}: some methods such as SHAP~\cite{Lundberg2017} provide a \textit{local} explanation, in the sense that each data point is treated separately. As a result, they assess the contribution of a feature relative to a data point. On the contrary, other methods such as SAGE~\cite{Covert2020} adopt a \textit{global} approach, and characterize a feature relative to the whole dataset.
    \item Explanation \textit{target}: some methods such as SHAP~\cite{Lundberg2017} aim to directly explain the value predicted by the machine learning model (e.g. a class), while others like SAGE~\cite{Covert2020} targets its performance, by explaining loss or accuracy.
\end{itemize}

In our case, we deal with a classification task, where the data point corresponds to graphs and their features to patterns. We want to assess the contribution of each pattern to the final classification performance. Therefore, we need an approach that has a global scope, and targets performance. For this reason, we select SAGE and use it to compute an approximation of the SV for each pattern. We then rank the patterns by decreasing SV: this constitutes our \textit{gold standard}, i.e. our best approximation of the true ranking of patterns depending on their discriminative power.

\subsubsection{Ranking Comparison}
\label{sec:MethRanksComp}
In order to compare a pattern ranking estimated through a quality measure and the gold standard, we need a suitable coefficient. In a similar context, \citet{LoyolaGonzalez2014} suggest using Kendall's Tau~\cite{Kendall1938}. However, it has certain limitations, so we also use the Rank-Biased Overlap (RBO) as an alternative coefficient.

\paragraph{Kendall's Tau}
This rank correlation coefficient applies to two rankings of the same set of elements. For each pair of elements that can be formed over this set, the rankings are considered as either concordant or discordant. Concordant means that they both put the two elements in the same order, and discordant means the opposite. The coefficient is the difference between the proportions of concordant and discordant pairs. 

In our case, the considered set is $\mathcal{P}_s = \{ P_1, ..., P_s \}$, and its constitutive elements are the $s$ representative patterns obtained after the selection step (Step~3 in Figure~\ref{fig:UpdateFlowchart}). Let us note $r_1(P)$ and $r_2(P)$ the ranks assigned to pattern $P$ according to the first and second rankings, respectively. Then Kendall's Tau is defined as: 
\begin{equation}
    \tau = \frac{1}{s(s-1)/2} \sum_{1 \leq i < j \leq s} \sgn\big( r_1(P_i) -r_1(P_j) \big) \sgn\big( r_2(P_i) - r_2(P_j) \big),
\end{equation}
\noindent where $\sgn$ is the signum function, which returns $-1$ if its argument is negative, and $1$ if it is positive. As a consequence, the product located inside the sum is either $-1$ (discordance) or $1$ (concordance). Kendall's Tau ranges from $-1$ (only discordant pairs, i.e. one ranking is the reverse of the other) to $+1$ (only concordant pairs, i.e. identical rankings).

We identify two limitations with Kendall's Tau. The first is that it requires both rankings to be defined over the exact same set of objects. In our experiments, there are some situations where this constraint is not respected. Second, Kendall's Tau gives the same importance to each rank. In our context, we ultimately want to select the top $s$ patterns to perform the classification. Clearly, we want to give more importance to the best-ranked patterns when comparing the rankings: discordance for the top patterns is more serious than discordance for the bottom ones. 

\paragraph{Rank-Biased Overlap}
To tackle these limitations, we propose to use the \textit{Rank-Biased Overlap} (RBO)~\cite{Webber2010} as an alternative to Kendall's Tau when comparing the rankings. Let us note $R_1(d)$ and $R_2(d)$ the $d$ first patterns according to the two considered rankings. The RBO is originally defined to compare infinite lists, as follows:
\begin{equation}
    RBO_p = (1 - p) \sum_{d=1}^{+\infty} p^{d-1} \frac{|R_1(d) \cap R_2(d)|}{d}.
\end{equation}
\noindent In our case, though, we compare rankings of finite sets, and the upper bound of the sum is $s$. Variable $d$ is used to consider an increasing number of top-ranked patterns. Inside the sum, the right-hand factor is the proportion of patterns that both rankings put in the top $d$. The left-hand factor is a weight that decreases exponentially with $d$: it allows giving more importance to the best-ranked patterns when comparing the rankings. The magnitude of this importance is controlled through parameter $p$ ($0 < p < 1$): a smaller $p$ puts more emphasis on top patterns. The $(1-p)$ term is a normalizing factor. In the end, the RBO can be seen as the weighted average of overlaps between the rankings. The coefficient ranges from 0 (rankings containing completely different elements) to 1 (exactly the same rankings).

According to its authors~\cite{Webber2010}, the advantage of the RBO over other correlation coefficients able to handle sets that do not completely overlap, such as Fagin's Intersection Metric~\cite{Fagin2003}, is that the RBO is monotonic with increasing $s$: if one ranking is a prefix of another, then increasing $s$ will not lead to a decrease of the RBO. This avoids biases in very large sets, where numerous concordances or discordances in the tails of the rankings could otherwise take too much importance.

\subsection{Classification Performance}
\label{sec:MethPerf}
To assess the performance of our classifier, we use three very standard metrics: Precision, Recall, and $F1$-Score~\cite{Rijsbergen1979}, while focusing on the positive class. Let TP, FP and FN denote the numbers of True Positives, False Positives and False Negatives, respectively. Precision and Recall are defined as:
\begin{align}
    Pre &= \frac{TP}{TP+FP} \\
    Rec &= \frac{TP}{TP +FN}.
\end{align}

In order to ease the comparison of our classification results, it is more convenient to summarize them under the form of a single value. For this purpose, we use the $F1$-Score, which is the harmonic mean of Precision and Recall: 
\begin{equation}
    F1 = 2 \frac{Pre \times Rec}{Pre + Rec}.
\end{equation}

\section{Experiments}
\label{sec:Exp}
We devise a series of experiments seeking to answer the following three major questions:
\begin{itemize}
    \item[RQ1] Is it possible to achieve more compact graph representation without compromising classification performance?
    \item[RQ2] Do quality measures behave consistently across graph datasets?
    \item[RQ3] Can we identify quality measures that tend to perform better than others? 
\end{itemize}

In the following, we first introduce the practical settings used for our experiments (Section~\ref{sec:ExpSettings}). After that, we address Q1 first, by studying how clustering affects the number of representatives, the rankings obtained with the measures, and the classification performance (Section~\ref{sec:ExpClustering}). We then tackle Q2, by assessing the correlation between the rankings produced by the considered quality measures (Section~\ref{sec:ExpPairwise}). Finally, we answer Q3, by studying how classification performance is affected by the considered quality measure (Section~\ref{sec:ExpGold}).

\subsection{Experimental Settings}
\label{sec:ExpSettings}
Let us first review the datasets as well as the pattern mining and cluster analysis tools used in our experiments. Our source code is written in Python and publicly available online\footnote{\url{https://github.com/CompNet/gpQualMeasComp}}. Each dataset is also publicly available, as indicated below.

\paragraph{Datasets}
Our selection of datasets is constrained by two aspects. First, graph datasets annotated for classification are much less common than their tabular counterparts. Second, our experimental protocol requires enumerating large numbers of patterns over the considered graph collections: this is computationally costly, especially in dense graphs~\cite{Huan2003}. For this reason, we favor attributed graphs, which ease this task, as they lead to fewer cases of subgraph isomorphisms.

We identify eight datasets fitting these constraints:
\begin{itemize}
    \item MUTAG\footnote{\url{https://www.philippe-fournier-viger.com/spmf/datasets/dang/mutag_graph.txt}}~\cite{Debnath1991}, a dataset of nitroaromatic compounds. Each compound is represented as a graph, with vertices and edges modeling atoms and chemical bonds between them, respectively. Vertices are labeled by atom type and edges by bond type.
    \item PTC\footnote{\url{https://www.philippe-fournier-viger.com/spmf/datasets/dang/ptc_graph.txt}}~\cite{Toivonen2003}, a dataset of chemical compounds used in toxicity screening. Each compound is represented as a graph, as in MUTAG. Vertices are labeled by atom type.
    \item NCI1\footnote{\url{https://www.philippe-fournier-viger.com/spmf/datasets/dang/nci1_graph.txt}}~\cite{Wale2006}, a dataset of chemical compounds used in anti-cancer screening. Each compound is represented as a graph, as in the previous datasets. Vertices are labeled by atom type and edges by bond type.
    \item D\&D\footnote{\url{https://www.philippe-fournier-viger.com/spmf/datasets/dang/dd_graph.txt}}~\cite{Dobson2003}, a dataset of chemical compounds. Each compound is represented as a graph, as before. Vertices are labeled by amino acid type.
    \item AIDS\footnote{\url{https://chrsmrrs.github.io/datasets/docs/datasets/}}~\cite{Riesen2008}, a dataset of chemical compounds tested for AIDS inhibition. Each compound is represented as a graph, as before. Vertices are labeled by atom type and edges by bond type.
    \item FOPPA\footnote{\url{https://doi.org/10.5281/zenodo.10879932}}~\cite{Potin2023d}, a dataset of public procurement contracts. Each set of contracts is represented as a graph, with vertices representing contractors and edges representing commercial relationships between them. Vertices are labeled according to contractor type, and edges are labeled by categories reflecting numbers of contracts.
    \item IMDb-BINARY\footnote{\url{https://www.philippe-fournier-viger.com/spmf/datasets/dang/IMDb_binary_graph.txt}}~\cite{Yanardag2015} (IMDb), a dataset of movie collaboration graphs. Vertices model actors, and edges represent co-appearances in movies. Each vertex has an integer label with unspecified meaning.
    \item FRANKENSTEIN\footnote{\url{https://chrsmrrs.github.io/datasets/docs/datasets/}}~\cite{Orsini2015} (FRANK), a dataset of chemical compounds. This dataset has no label on vertices nor edges.
\end{itemize}

Table~\ref{tab:Dataset} summarizes the most important characteristics of these datasets. We perform under-sampling to obtain balanced classes whenever needed. The size of the datasets ranges from hundreds to thousands of graphs, while the numbers of possible values for vertex and edge attributes range from none to several tens. They are relatively homogeneous in terms of graph size though, except D\&D whose graphs contain much more vertices and edges. The degree average and density are very heterogeneous, the latter ranging from $0.09$ to $0.52$. Similarly, the average clustering coefficient is very high in certain graphs ($0.773$ in IMDb) and very low in others ($0.003$ in NCI1). The chemical networks are sparser than the social networks, which is consistent with the literature~\cite{Melancon2006}. FOPPA contains many bipartite graphs, for this reason the clustering coefficient is not defined on this dataset. Moreover, we compute its density using the formula defined for bipartite graphs~\cite{Saryce2018}. The last row in the table shows the average number of unique patterns mined in the graphs, and also exhibits a high variability. For certain datasets (marked with a *), the pattern search is not exhaustive, because of computational limitations. In summary, our datasets exhibit a certain heterogeneity, and therefore our selection is illustrative because it covers a wide range of cases. 

\begin{table}[htbp!]
    \begin{center}
        \begin{tabular}{l rrrrrrrr}
            \hline
            \textbf{Characteristic} & \textbf{MUTAG} & \textbf{PTC} & \textbf{NCI1} & \textbf{D\&D} & \textbf{AIDS}  & \textbf{FOPPA} & \textbf{IMDb} & \textbf{FRANK}\\
            \hline
            Number of graphs & $188$ & $344$ & $4{,}110$ & $1{,}178$ & $2{,}000$ & $660$ & $1{,}000$ & $4{,}337$ \\
            Number of vertex label values & $7$ & $19$ & $37$ & $82$ & $38$ & $2$ & $65$ & -- \\
            Number of edge label values & $11$ & -- & $3$ & -- & $3$ & $3$ & -- & -- \\
            Average number of vertices & $14.58$ & $25.56$ & $29.87$ & $284.31$ & $15.69$ & $14.20$ & $19.77$ & $16.89$ \\
            Average number of edges & $19.79$ & $15.03$ & $32.30$ & $ 715.66$ & $16.20$ & $14.91$ & $96.39$ & $17.87$ \\
            Mean average degree & $2.19$ & $1.99$ & $2.16$ & $4.98$ & $2.01$ & $2.03$ & $8.88$ & $2.06$ \\
            Average density & $0.14$ & $0.21$ & $0.09$ & $0.03$ & $0.19$ & $0.40$ & $0.52$ & $0.17$ \\
            Average global clustering coefficient & $0.000$ & $0.008$ & $0.003$ & $0.458$ & $0.007$ & -- & $0.773$ & $0.010$ \\
            Average number of unique patterns & $156$ & $120*$ & $614*$ & $3{,}361*$ & $177*$ & $528$ & $106*$ & $1{,}342*$ \\
            \hline
        \end{tabular}
        \caption{Main characteristics of the eight graph datasets constituting our benchmark.} \label{tab:Dataset}
    \end{center}
\end{table}

For the sake of concision, we only focus on 6 of these 8 datasets when presenting our results in the rest of this section. The comprehensive results obtained for IMDb and FRANK are available in Appendix~\ref{sec:AppAddRes}, though. We defer them to the appendix because these results are very similar to those of AIDS and D\&D, respectively.

\paragraph{Pattern Mining}
To mine the patterns in the datasets (Step~1 of the pipeline, cf. Figure~\ref{fig:UpdateFlowchart}), we use the SPMF library~\cite{FournierViger2016}, which provides a wide range of algorithms for pattern mining. We adopt the gSpan method~\cite{Yan2002}, which is a well-known algorithm for mining frequent subgraphs. Mining patterns in graphs is a complex task, as the number of possible patterns is exponential in the number of vertices and edges. Therefore, on huge datasets, we limit the number of patterns to mine by setting a minimum support threshold. This threshold is the minimal number of graphs in which a pattern must appear to be considered as frequent.

\begin{table}[htbp!]
    \begin{center}
        \begin{tabular}{l rrrrrrrr}
            \toprule
            \textbf{Dataset} & \textbf{MUTAG} & \textbf{PTC} & \textbf{NCI1} & \textbf{D\&D} & \textbf{AIDS}  & \textbf{FOPPA} & \textbf{IMDb}  & \textbf{FRANK} \\
            \midrule
            Minimum support (\% of graphs) & $0$ & $1$ & $1$ & $25$ & $1$ & $0$ & $1$ & $20$ \\
            Minimum support (number of graphs) & $1$ & $4$ & $42$ & $277$ & $20$ & $1$ & $10$ & $837$ \\
            Number of unique patterns & $3{,}408$ & $5{,}285$ & $11{,}564$ & $10{,}000$ & $5{,}589$ & $11{,}773$ & $4{,}741$ & $5{,}000$ \\
            \bottomrule
    \end{tabular}
    \caption{Number of patterns mined for each dataset presented in Table~\ref{tab:Dataset}.}
    \label{tab:Pattern}
    \end{center}
\end{table}

In general~\cite{LoyolaGonzalez2014, AcostaMendoza2016}, this minimal threshold for frequent pattern mining is selected empirically, and expressed as a percentage of the number of graphs contained in the dataset. It is often between 5\% and 15\% of the number of graphs~\cite{Thoma2010, Rousseau2015, Wu2015}. \textcolor{highlightCol}{It is worth mentioning that our work is agnostic to the specific method used to extract patterns, as we are primarily interested in how a given set of patterns is ranked. While different pattern mining techniques may produce different sets of patterns, our approach mitigates this variability by considering the complete enumeration of patterns (whenever feasible), and applying a common threshold otherwise.}

We indicate this in Table~\ref{tab:Pattern}, along with the raw graph count corresponding to the associated support. In practice, a value of 1 means that there is no minimum support threshold. Table~\ref{tab:Pattern} also shows the number of patterns identified in each dataset. 

\paragraph{Cluster Analysis}
To perform the hierarchical clustering (Step~1a, described in Section~\ref{sec:MethClust}), we use the standard \texttt{AgglomerativeClustering} method of package sklearn\footnote{\url{https://scikit-learn.org/stable/modules/generated/sklearn.cluster.AgglomerativeClustering.html}}. In order to follow the protocol defined in Section~\ref{sec:MethClust}, we set the following parameters:
\begin{itemize}
    \item \textit{metric}: set to \texttt{precomputed}, which allows proposing a custom function to compute the distance between patterns. In our case, this function implements the Manhattan distance between footprints.
    \item \textit{linkage}: set to \texttt{complete}, i.e. use complete-linkage to assess the distance between clusters. As previously explained, this amounts to using the distance between the farthest elements of the considered clusters.
\end{itemize}

\paragraph{Gold Standard}
The elaboration of the gold standard requires computing SAGE values (cf. Section~\ref{sec:MethRanksShap}). The original implementation is publicly available online\footnote{\url{https://github.com/iancovert/sage}}, however it is very time-consuming. Instead, we use LossSHAP\footnote{\url{https://shap.readthedocs.io/en/latest/}}, an alternative implementation which is much faster. It only provides a local version of SAGE, though, which is why we compute an average over all patterns to obtain the global scores that we need.

\subsection{Clustering Assessment}
\label{sec:ExpClustering}
In this section, we present the results of the clustering step. It aims to group patterns with similar footprints, in order to reduce the number of patterns to consider without losing too much classification performance. We study the influence of clustering on the number of representatives (Section~\ref{sec:ExpClusteringReprNbr}), on the rankings produced by the quality measures (Section~\ref{sec:ExpClusteringRkComp}), and on the classification performance (Section~\ref{sec:ExpClusteringClassPerf}).

\subsubsection{Number of Representatives}
\label{sec:ExpClusteringReprNbr}
\begin{figure}[htbp!]
    \centering
    \begin{minipage}[b]{0.32\textwidth}
        \includegraphics[width=\textwidth]{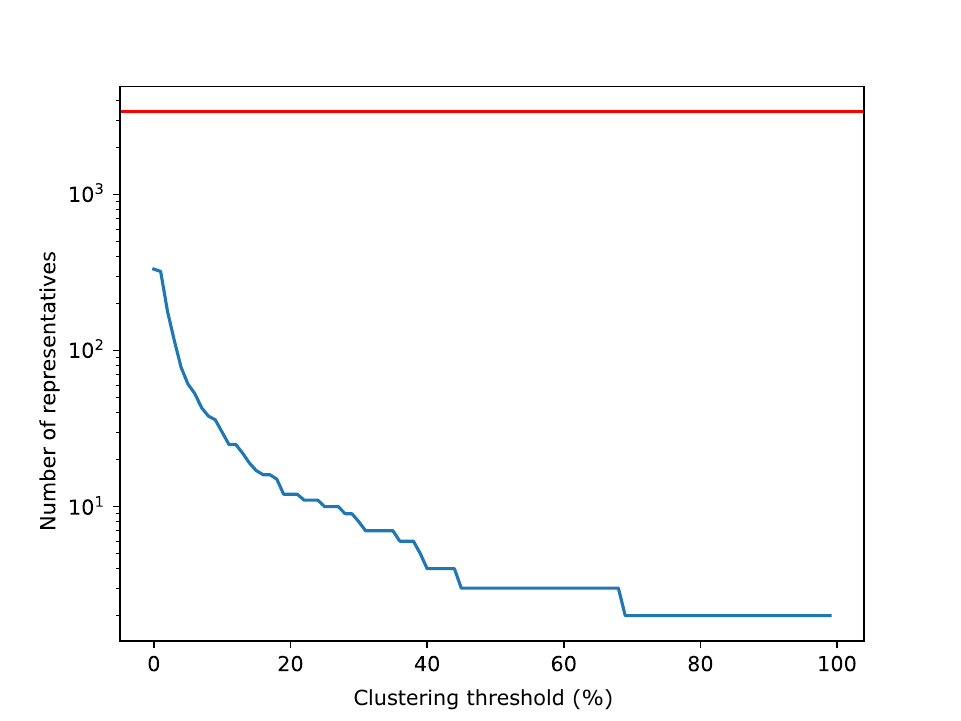}
        \subcaption{MUTAG}
        \label{fig:NbClustersMUTAG}
    \end{minipage}
    \begin{minipage}[b]{0.32\textwidth}
        \includegraphics[width=\textwidth]{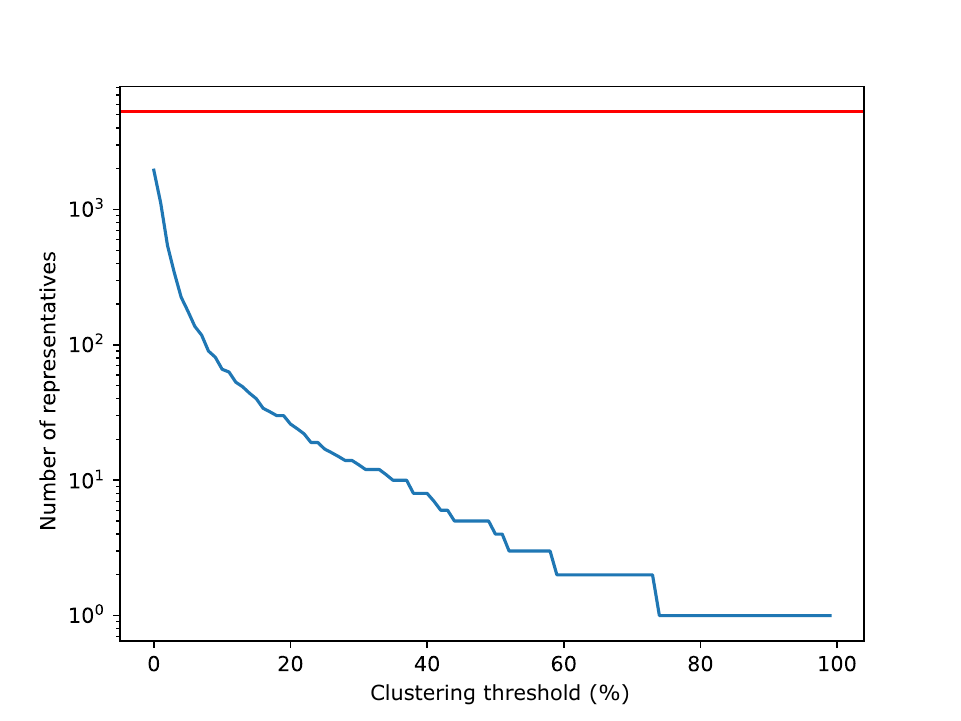}
        \subcaption{PTC}
        \label{fig:NbClustersPTC}
    \end{minipage}
    \begin{minipage}[b]{0.32\textwidth}
        \includegraphics[width=\textwidth]{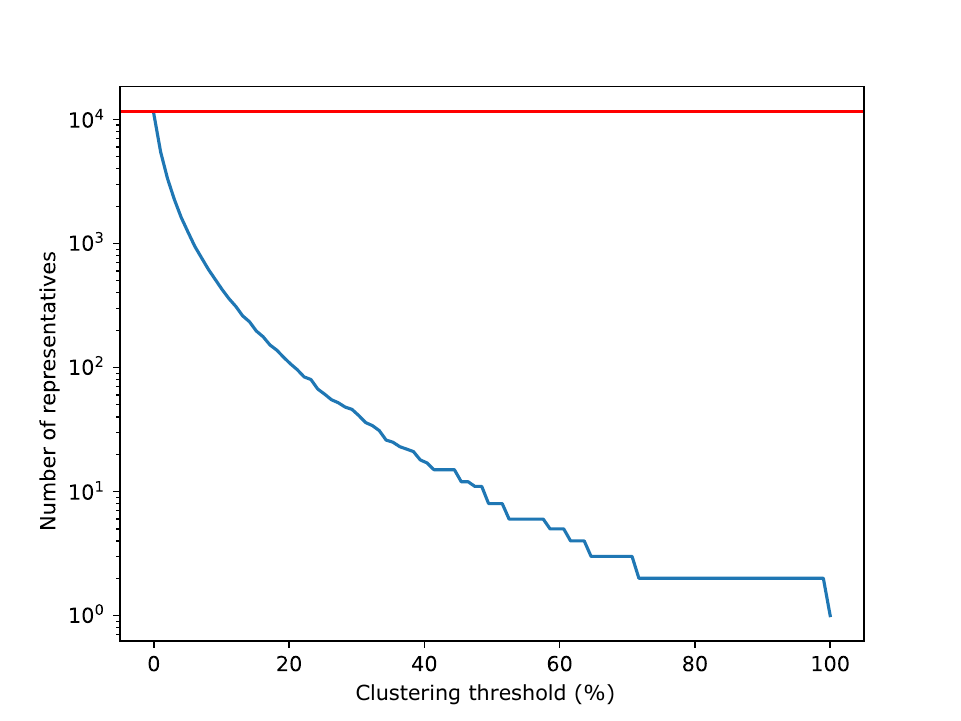}
        \subcaption{NCI1}
        \label{fig:NbClustersNCI1}
    \end{minipage}
    
    \vspace{1em}
    
    \begin{minipage}[b]{0.32\textwidth}
        \includegraphics[width=\textwidth]{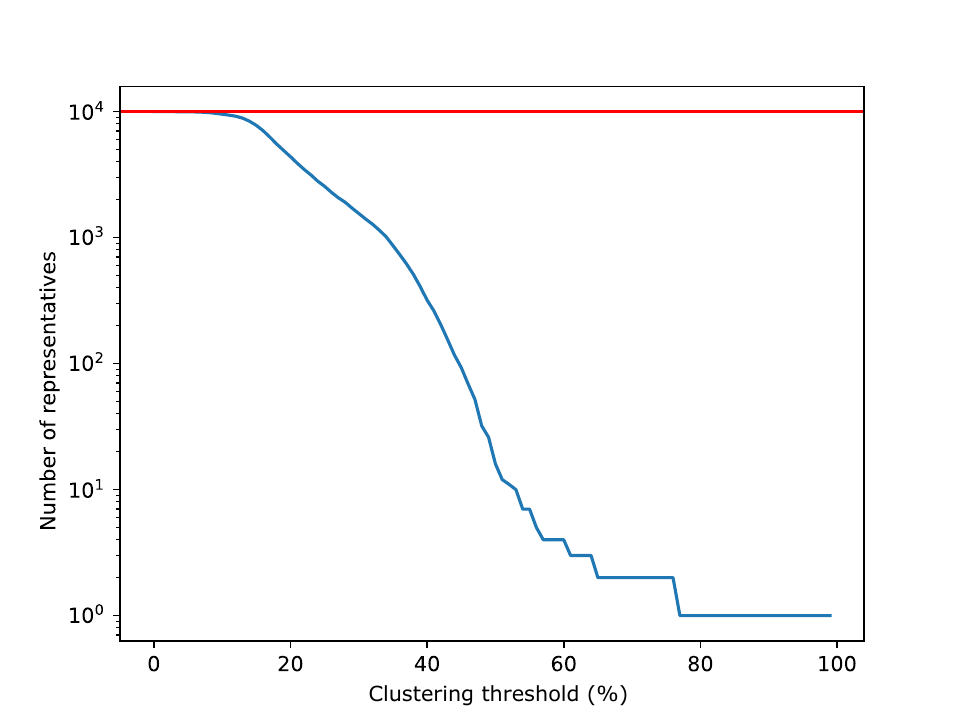}
        \subcaption{D\&D}
        \label{fig:NbClustersDD}
    \end{minipage}
    \begin{minipage}[b]{0.32\textwidth}
        \includegraphics[width=\textwidth]{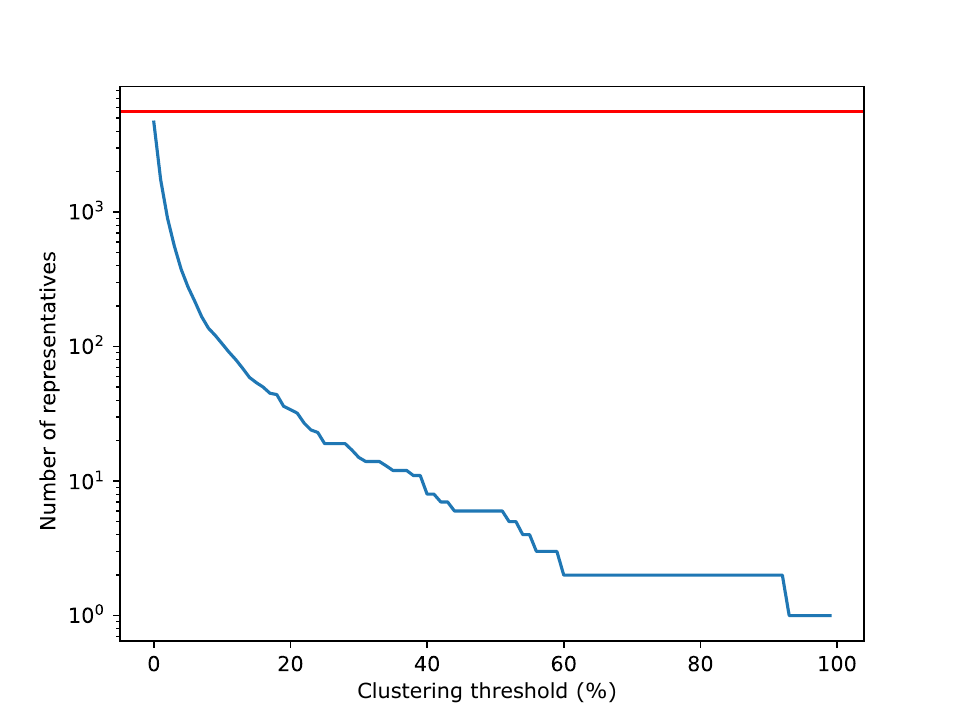}
        \subcaption{AIDS}
        \label{fig:NbClustersAIDS}
    \end{minipage}
    \begin{minipage}[b]{0.32\textwidth}
        \includegraphics[width=\textwidth]{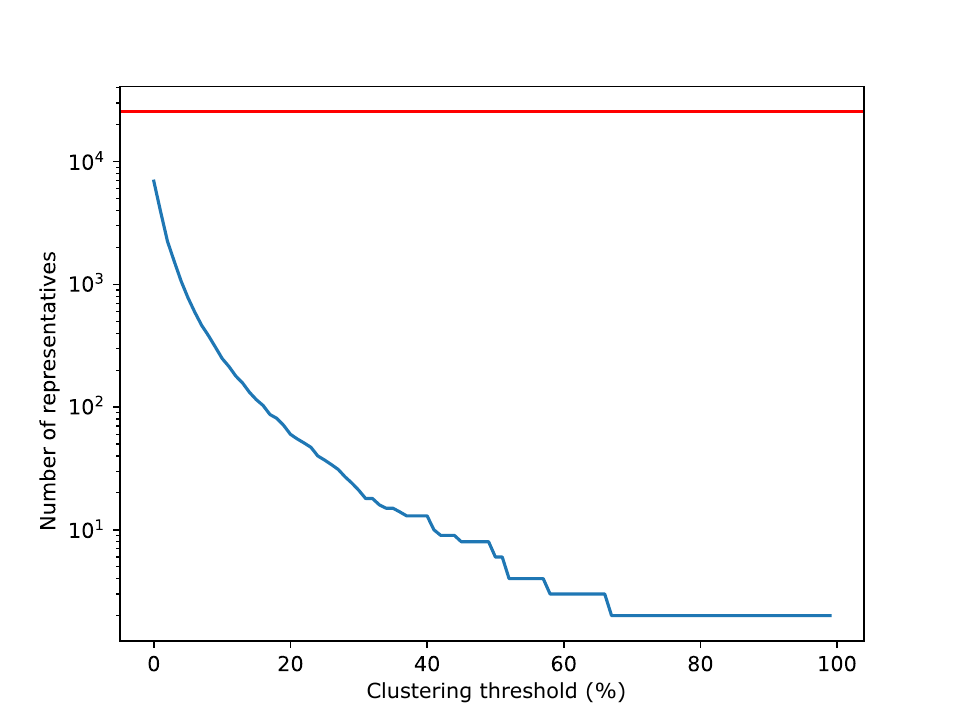}
        \subcaption{FOPPA}
        \label{fig:NbClustersFOPPA}
    \end{minipage}

    \caption{Number of representatives as a function of the clustering threshold. Note the logarithmic scale of the $y$-axis. The red horizontal lines represent the original number of patterns detected in the datasets.}
    \label{fig:NumberCluster}
    \Description{Description} 
\end{figure}

First, we discuss the number of representatives obtained by applying the clustering process with different clustering threshold values. Figure~\ref{fig:NumberCluster} shows the number of representatives identified for each dataset as a function of this threshold. The maximal Manhattan distance between two footprints corresponds to $|\mathcal{G}|$, the number of graphs in the considered collection. In order to ease the comparison between datasets, we use this value to normalize the threshold and express it as a percentage ($x$-axis).

In each plot, the red line indicates the initial number of patterns, i.e. without any clustering. Consider the strictest clustering, obtained by setting a threshold of 0\%, i.e. by grouping only patterns with identical footprints: for certain datasets, the number of representatives is significantly lower than the initial number of patterns (note the logarithmic scale of the $y$-axis). There is a reduction in the number of patterns by approximately 63\% for PTC, 92\% for MUTAG and 44\% for FOPPA. Meanwhile, NCI1, D\&D and AIDS exhibit lower reduction rates, with 4\%, 9\% and 16\% of the total number of patterns respectively. This is because pattern mining on these datasets is less comprehensive, and therefore the patterns mined are less redundant on average. In any case, the number of representatives decreases sharply as the clustering threshold increases, reaching a minimal value when there is only one cluster left.

Based on these results, it appears clearly that clustering allows for the reduction of the number of patterns to be processed later in the algorithm. We then seek to study whether it removes redundant patterns and improves the comparison of measure rankings, as intended.

\subsubsection{Ranking Comparison}
\label{sec:ExpClusteringRkComp}
As explained in Section~\ref{sec:MethClust}, the goal of our clustering step is to enable a better comparison among the considered quality measures, by avoiding considering redundant patterns (i.e. patterns exhibiting similar footprints). The quality measures are computed only over the cluster representatives, instead of all detected patterns. The clustering threshold controls how similar the grouped patterns are, and is likely to affect the rankings obtained with the quality measures, and therefore their comparison. 

\begin{figure}[htbp!]
    \centering
    \begin{minipage}[b]{0.32\textwidth}
        \includegraphics[width=\textwidth]{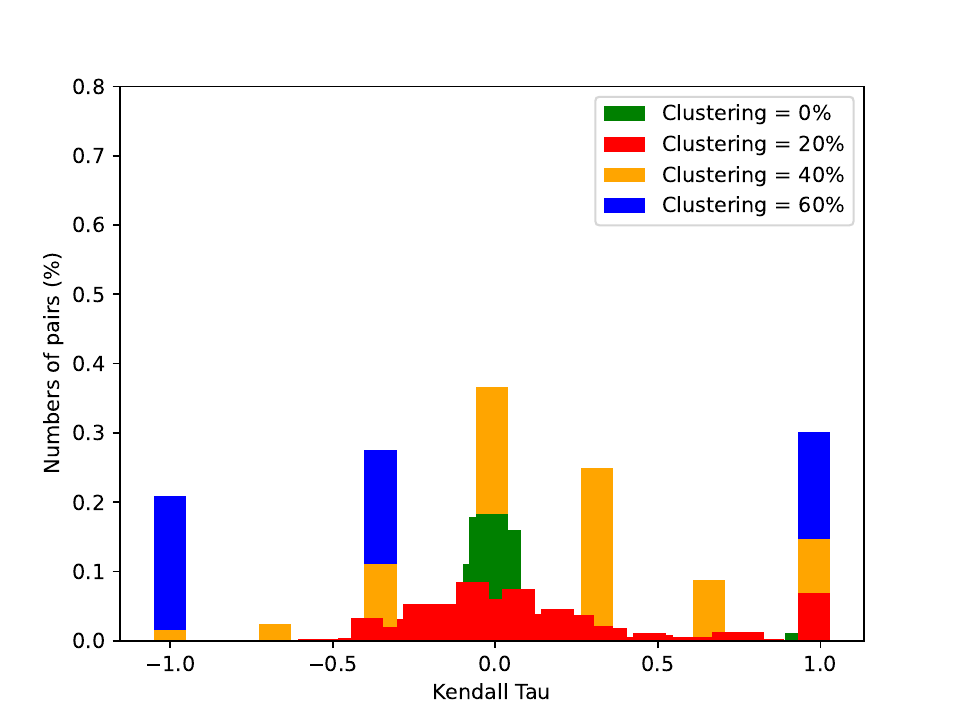}
        \subcaption{MUTAG}
        \label{fig:RBOKTMUTAG}
    \end{minipage}
    \begin{minipage}[b]{0.32\textwidth}
        \includegraphics[width=\textwidth]{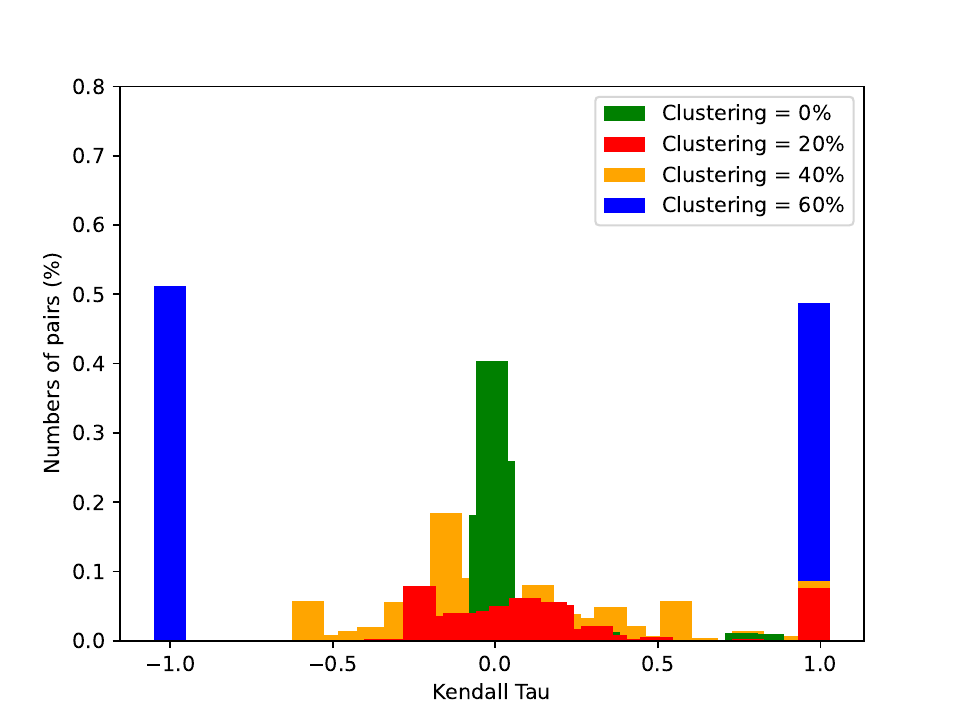}
        \subcaption{PTC}
        \label{fig:RBOKTPTC}
    \end{minipage}
    \begin{minipage}[b]{0.32\textwidth}
        \includegraphics[width=\textwidth]{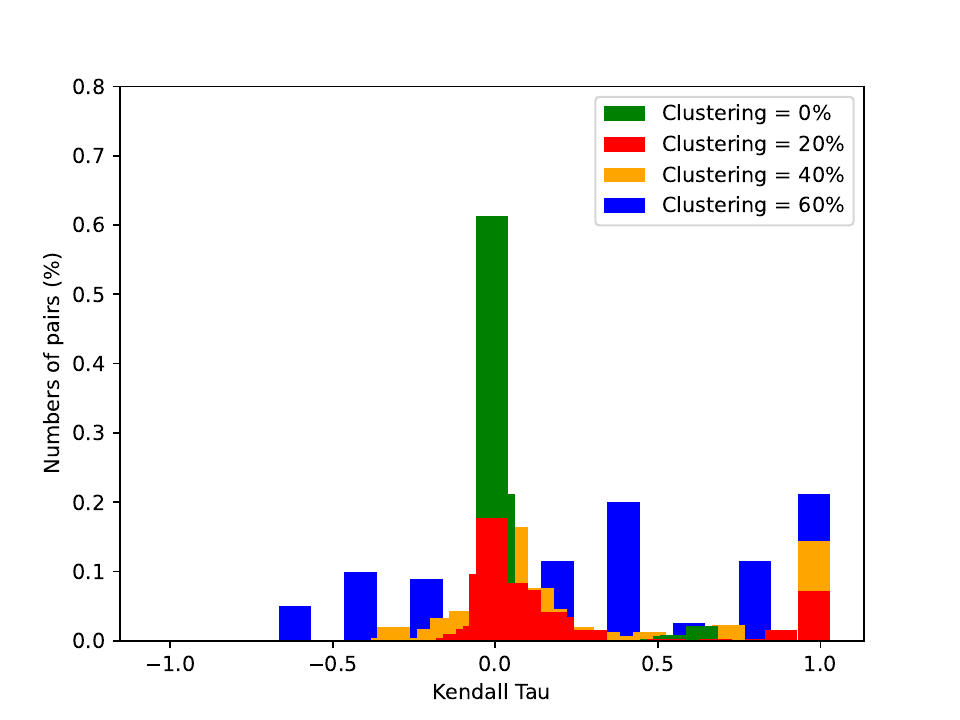}
        \subcaption{NCI1}
        \label{fig:RBOKTNCI1}
    \end{minipage}
    
    \vspace{1em}
    
    \begin{minipage}[b]{0.32\textwidth}
        \includegraphics[width=\textwidth]{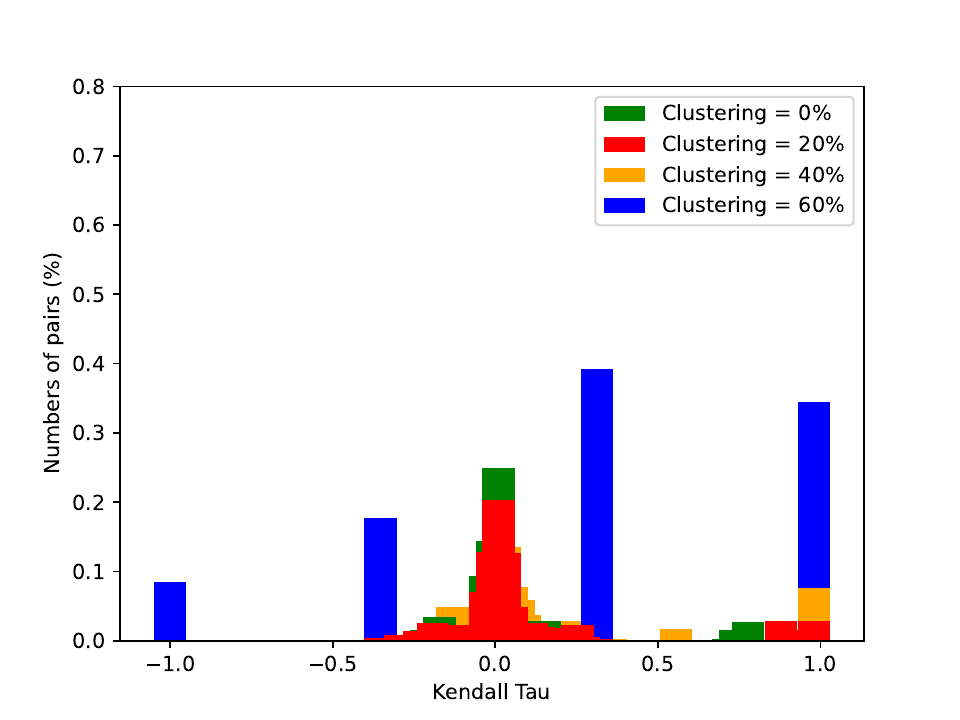}
        \subcaption{D\&D}
        \label{fig:RBOKTDD}
    \end{minipage}
    \begin{minipage}[b]{0.32\textwidth}
        \includegraphics[width=\textwidth]{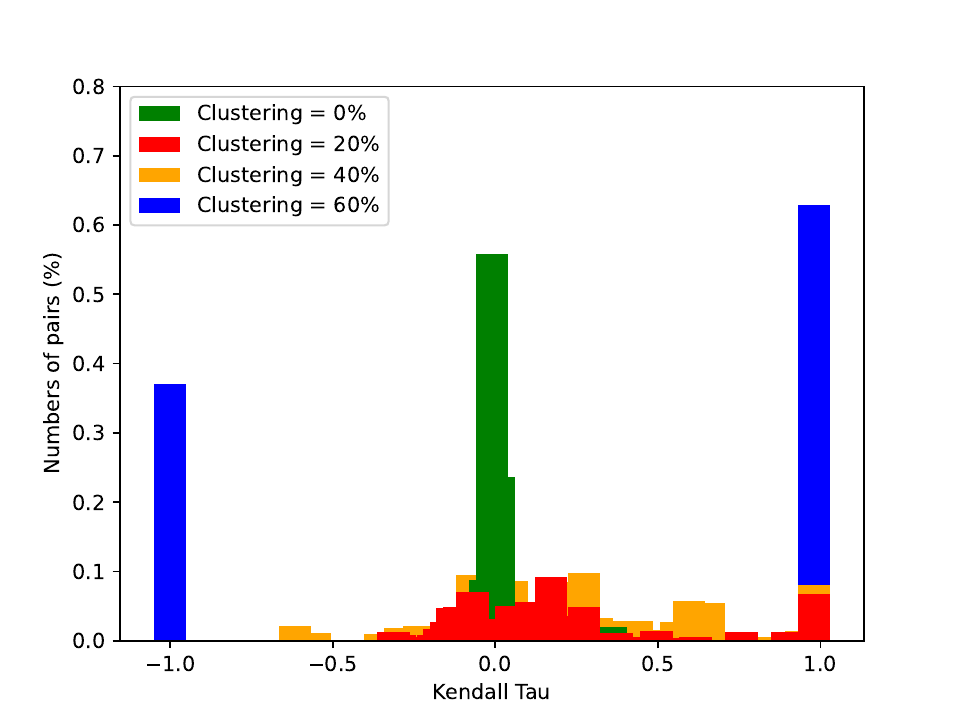}
        \subcaption{AIDS}
        \label{fig:RBOKTAIDS}
    \end{minipage}
    \begin{minipage}[b]{0.32\textwidth}
        \includegraphics[width=\textwidth]{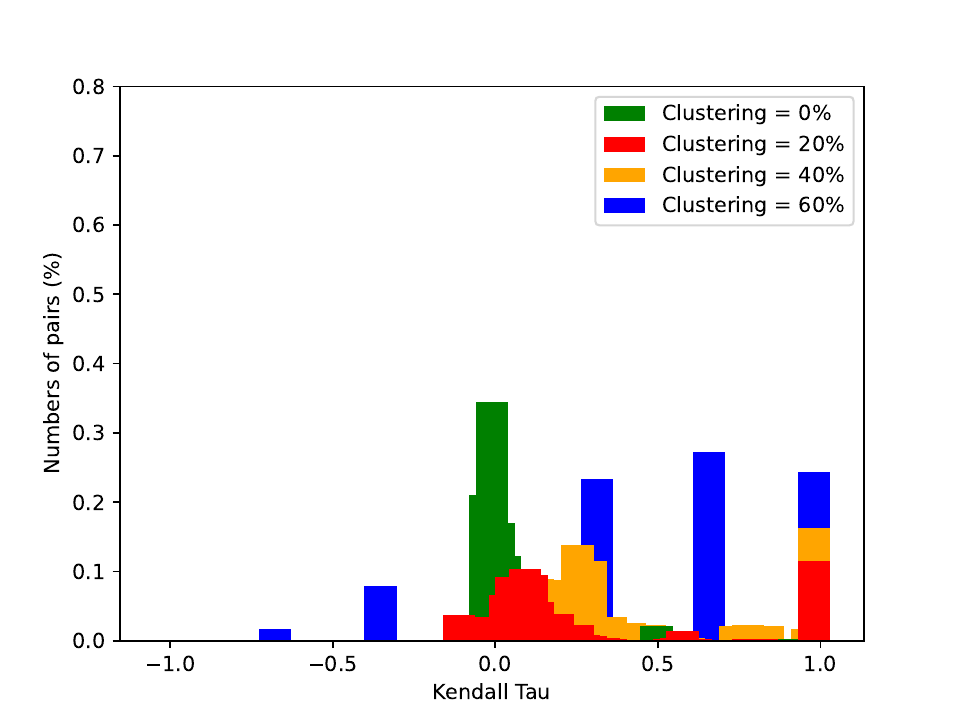}
        \subcaption{FOPPA}
        \label{fig:RBOKTFOPPA}
    \end{minipage}
    \caption{Distribution of Kendall's Tau over all pairs of quality measures, for four values of the clustering threshold (0\%, 20\%, 40\% and 60\%). See Appendix~\ref{sec:AppKTSeparate} for more plots.}
    \label{fig:CompaRBOKT}
    \Description{Description} 
\end{figure}

In order to study the impact of this parameter on the rankings, we consider four threshold values (0\%, 20\%, 40\% and 60\%) and compute Kendall's Tau between the rankings produced with all pairs of quality measures. Each plot in Figure~\ref{fig:CompaRBOKT} represents the distribution of Tau obtained for one of the considered datasets. Each such plot exhibits four distributions corresponding to the four selected threshold values, and shown as histograms of different colors. For better readability, separate plots are available in Appendix~\ref{sec:AppKTSeparate}.

We observe three different outcomes. First, when a clustering is performed with a threshold of 0 (green histograms), Tau is close to 0 for all pairs of measures. This means that there is not much similarity between their rankings, and each measure consequently provides a unique ranking. Our assumption is that this is due to the presence of many patterns that are different but have similar footprints, which makes ranking \textit{seemingly} dissimilar (see Section~\ref{sec:MethClust}). Second, when increasing the threshold (red and orange histograms), Tau is distributed more uniformly, and some pairs of measures are associated with a Tau close to 1. This means that these pairs of measures produce similar rankings from the perspective of footprints. The clustering step reveals this otherwise hidden similarity. Some pairs also exhibit low correlation values, though, which just means that some measures behave differently. Put differently, the measures form a few similarity classes (i.e. groups of measures leading to similar rankings). Third and finally, when the threshold gets very high (blue histograms), the correlation values get more extreme (close to $-1$ and $1$). The relevant differences between the rankings produced by the measures are not captured anymore, and they are considered as highly similar or dissimilar by Tau. In other terms, there are only a very few (sometimes 1 or 2) similarity classes of measures.

Our results show that our clustering step has the expected effect: by removing the redundancy among patterns, it allows making more apparent the similarity between the rankings produced by the quality measures. On the one hand, performing a strict clustering by using a very low threshold means obtaining many small clusters of patterns with exactly the same footprint, which may not differ much from using all patterns. On the other hand, performing a very relaxed clustering based on a high threshold results in a few very large clusters, likely to gather patterns exhibiting very different footprints. This would excessively reduce the number of representatives, and provide an unreliable comparison between the measure rankings. The choice of the distance threshold is therefore crucial. We next propose a method to select the most appropriate value.

\subsubsection{Classification Performance}
\label{sec:ExpClusteringClassPerf}
We turn to the evaluation of the clustering process in terms of classification performance. We perform the clustering process with different threshold values and keep all the resulting representatives. We then use them to build a vector representation of the graphs and train a classifier, before assessing its classification performance with the $F1$-Score. Our objective is to find the optimal threshold for the clustering step, meaning the one where classification performance is maximized with the smallest possible number of representatives. \textcolor{highlightCol}{To ensure a fair evaluation, we use C-Support Vector Machines~\cite{Cortes1995} as the classification model, as it is a widely used approach in this domain~\cite{Thoma2010, Rieck2019}. This choice allows us to focus on the impact of the clustering process on classification, rather than on rather than on the nature of the classifier.}

\begin{figure}[htbp!]
    \centering
    \begin{minipage}[b]{0.32\textwidth}
        \includegraphics[width=\textwidth]{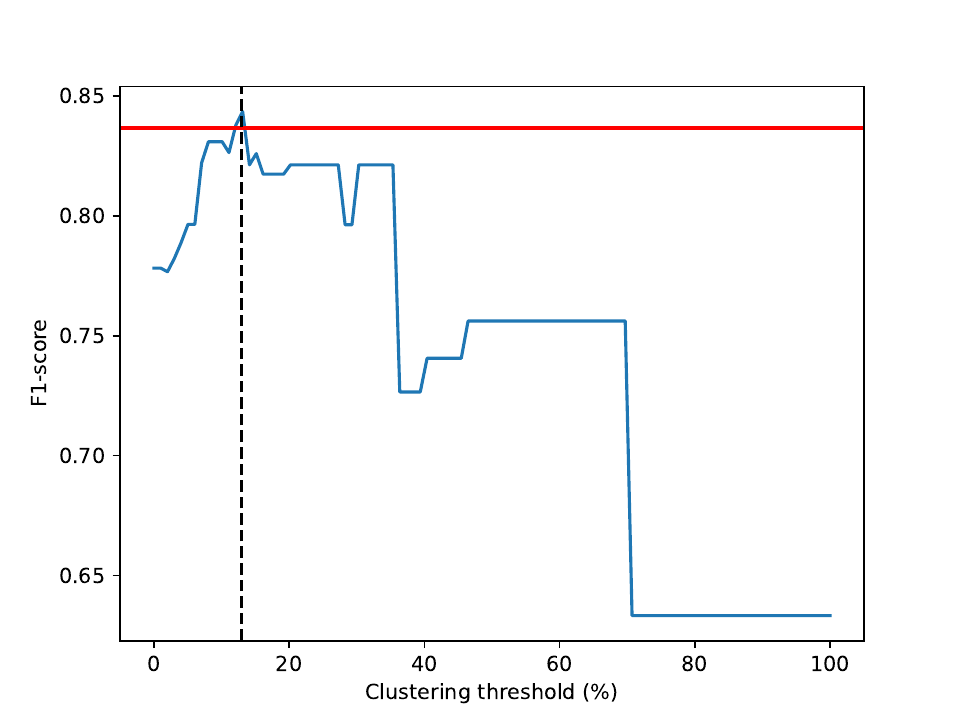}
        \subcaption{MUTAG}
        \label{fig:F1MUTAG}
    \end{minipage}
    \begin{minipage}[b]{0.32\textwidth}
        \includegraphics[width=\textwidth]{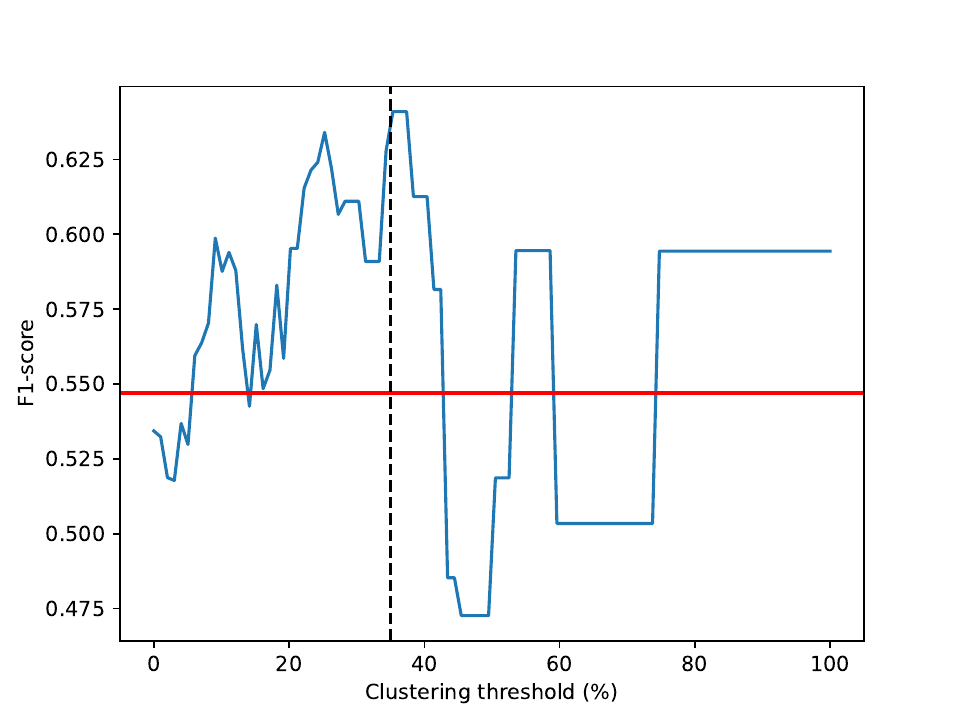}
        \subcaption{PTC}
        \label{fig:F1PTC}
    \end{minipage}
    \begin{minipage}[b]{0.32\textwidth}
        \includegraphics[width=\textwidth]{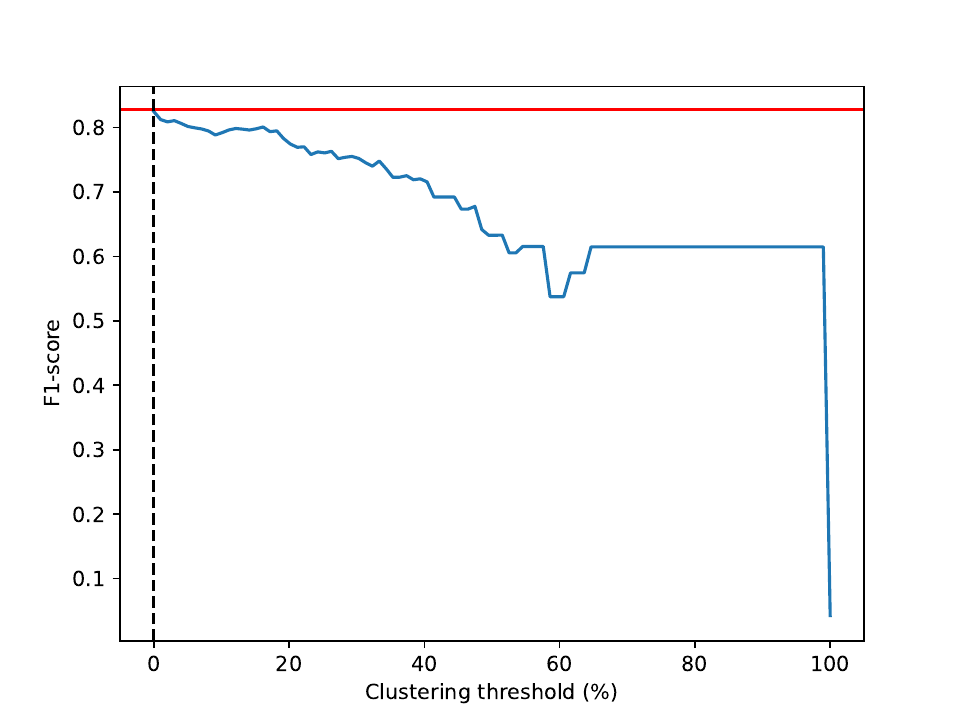}
        \subcaption{NCI1}
        \label{fig:F1NCI1}
    \end{minipage}
    
    \vspace{1em}
    
    \begin{minipage}[b]{0.32\textwidth}
        \includegraphics[width=\textwidth]{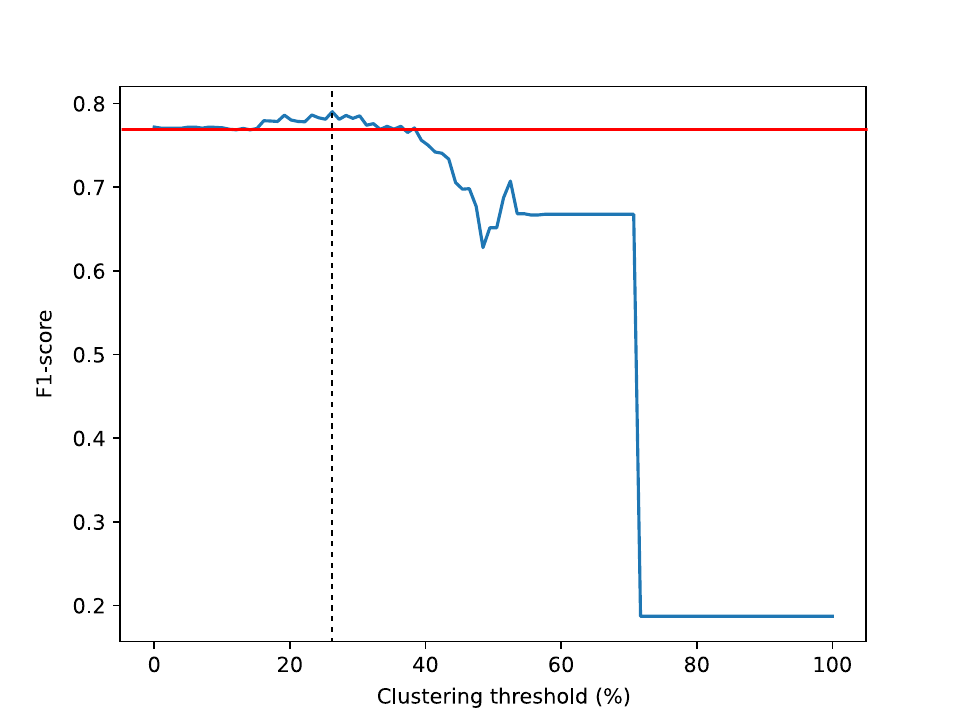}
        \subcaption{D\&D}
        \label{fig:F1DD}
    \end{minipage}
    \begin{minipage}[b]{0.32\textwidth}
        \includegraphics[width=\textwidth]{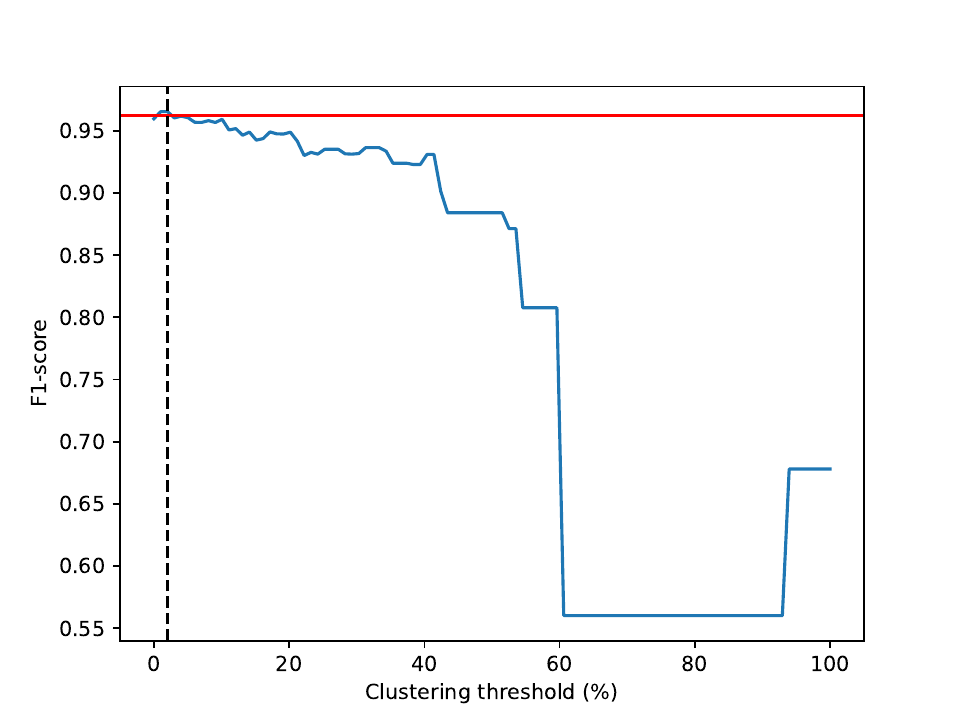}
        \subcaption{AIDS}
        \label{fig:F1AIDS}
    \end{minipage}
    \begin{minipage}[b]{0.32\textwidth}
        \includegraphics[width=\textwidth]{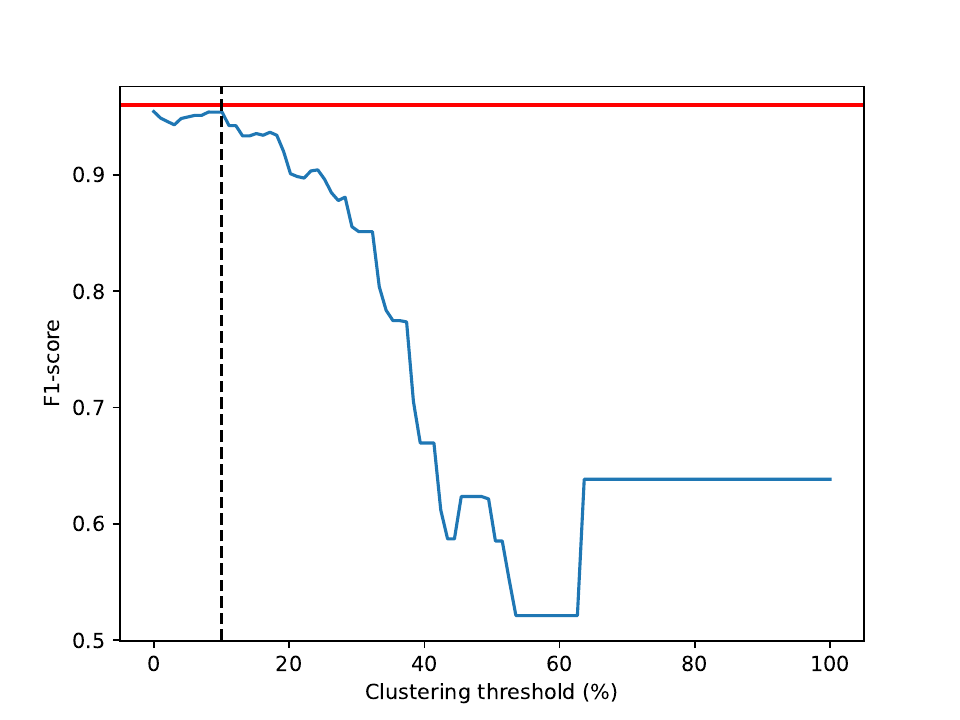}
        \subcaption{FOPPA}
        \label{fig:F1FOPPA}
    \end{minipage}

    \caption{Classification performance ($F1$-Score) as a function of the clustering threshold. The vertical dotted black lines materialize the threshold values used in the rest of our experiments.}
    \label{fig:$F1$-score}
    \Description{Description} 
\end{figure}

Each plot in Figure~\ref{fig:$F1$-score} shows the classification performance as a function of the clustering threshold, expressed as a percentage of the number of graphs in the dataset, as in Figure~\ref{fig:NumberCluster}. For reference, the horizontal red lines show the performance obtained without any clustering. The vertical dotted black lines show the threshold values that are the best trade-off between minimizing the number of representatives and maximizing the classification performance.

We observe three distinct behaviors over the considered datasets. For MUTAG and PTC, increasing the threshold leads to a better classification performance, up to a point where it starts decreasing. These sweet spots correspond to our optimal thresholds, shown as vertical black lines. These results confirm that clustering helps to reduce the number of redundant patterns, and to retain only those that are relevant for classification. When compared to the performance obtained without any clustering (red lines), we even see some substantial improvement. For D\&D and FOPPA, the classification performance initially stagnates, or only slightly improves, when increasing the threshold. After some point, it starts to decrease. In these cases, clustering improves classification performance (compared to the red lines), even if only slightly. In any case, it alleviates the computational cost by reducing the dimension of the vector representation. For these reasons, these points, shown as vertical black lines, also correspond to the best thresholds. Finally, for AIDS and NCI1, the classification performance decreases as soon as we increase the threshold. This means that clustering removes discriminative patterns essential for the classification right from the start. However, we do not notice any difference between the performance without clustering and with a threshold of 0, which implies that minimal clustering is still interesting, by reducing the number of total patterns.
For this reason, our selected thresholds values (black lines) are very low for these datasets. For all datasets, we observe a drastic decrease of the $F1$-Score when the threshold gets very high. Indeed, these values make it possible to group patterns with very different footprints in the same cluster, and therefore fail to maintain a comprehensive set of representatives for classification.

Based on our results, we can propose a rule of thumb to select the clustering threshold in case of a general dataset. Our recommended threshold amounts to approximately 20\% of the total number of graphs in the dataset. This leads either to an improvement of the classification, or in the worst case achieves a performance very close to the optimum. This threshold constitutes a balance between computational efficiency and accuracy. \textcolor{highlightCol}{However, it would be interesting to validate this heuristic on additional datasets to confirm its general applicability.}

Now that the appropriate threshold values are identified, we can focus on comparing the measures based on the pattern rankings they produce.

\subsection{Pairwise Comparison}
\label{sec:ExpPairwise}
\begin{figure}[htbp!]
    \centering
    \includegraphics[width=0.8\textwidth]{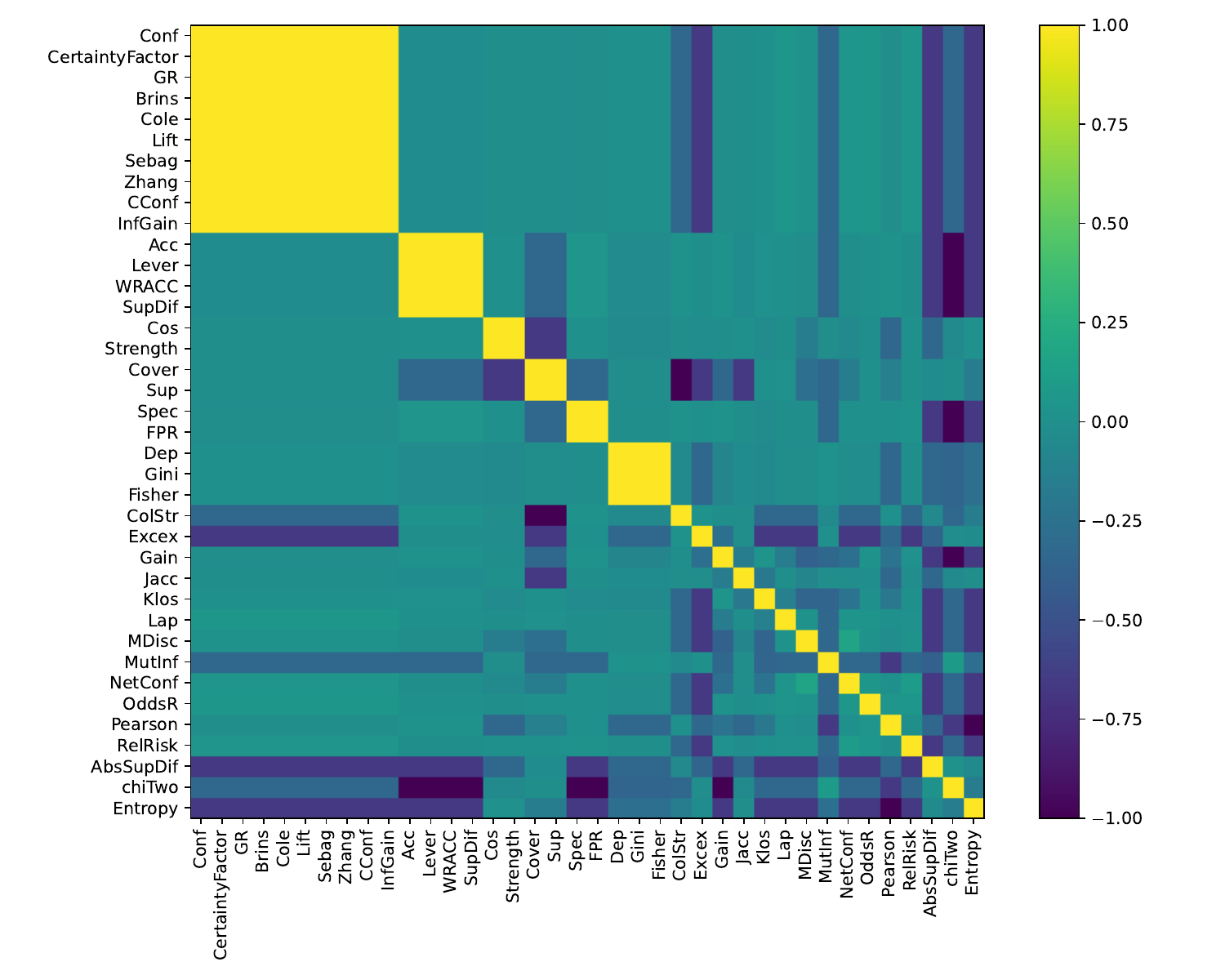}
    \caption{Minimal value of Kendall's Tau over all datasets, for each pair of quality measures.}
    \label{fig:CompaScoresMinMatri}
    \Description{Description} 
\end{figure}

We now perform a pairwise comparison of the measures described in Section~\ref{sec:QualityMeasures}. Our objective is to identify groups of measures that lead to similar pattern rankings. Following our method from Section~\ref{sec:Meth}, we first group the patterns depending on their footprint similarity, using the threshold identified in Section~\ref{sec:ExpClusteringClassPerf}, and we then rank their representatives using each quality measure. Since all measures deal with the same representatives, we compare these rankings with Kendall's Tau. There is a certain level of variability depending on the dataset: Figure~\ref{fig:CompaScoresMinMatri} presents a synthetic view of our results over all datasets. The matrix is symmetric, and each one of its rows and columns corresponds to one of the quality measures. Each element of this matrix represents the \textit{minimal} rank correlation between two measures over all datasets. A high value therefore indicates that the quality measures rank the representatives similarly regardless of the dataset. The detail of the correlations obtained for each dataset considered separately is provided in Appendix~\ref{sec:AppAllKT}. 

Yellow blocks in the figure correspond to groups of measures with a minimal correlation \textit{equal} to 1, meaning they produce \textit{identical} rankings over all datasets. We identify \textcolor{highlightCol}{six} of these blocks:
\begin{enumerate}
    \item \textsc{Conf}, \textsc{CFactor}, \textsc{GR}, \textsc{Brins}, \textsc{Cole}, \textsc{Lift}; \textsc{Sebag}, \textsc{Zhang}, \textsc{CConf} and \textsc{InfGain};
    \item \textsc{Acc}, \textsc{Lever}, \textsc{WRACC} and \textsc{SupDif};
    \item \textsc{Cos} and \textsc{Strength};
    \item \textsc{Cover} and \textsc{Sup};
    \item \textsc{Spec} and \textsc{FPR}.
    \item \textcolor{highlightCol}{\textsc{Dep}, \textsc{Gini} and \textsc{Fisher}}
\end{enumerate}
The rest of the measures do not exhibit such similarity, except punctually, for some datasets (e.g. PTC, see Appendix~\ref{sec:AppAllKT}). The measures belonging to the same block rank the patterns in the exact same way, therefore they also lead to identical classification performances. For this reason, in the rest of our experiments, we select only one measure to represent each block, in order to simplify the presentation and discussion of our results.

The first block contains measures that do not possess the Jumpiness property (see Definition~\ref{def:Jumpiness}), and are based on $p(\mathcal{G}^+ \mid P)$. They favor patterns that do not appear in the negative class, regardless of their frequency in the positive class. This block is represented in the rest of our results by \textsc{GR}.

The measures that form the second block also favor patterns that appear more frequently in positive than negative graphs, but they possess the Jumpiness property. As a result, they are able to order jumping emergent patterns among themselves, in contrast to Block~1. Block~2 is represented in the rest of our results by \textsc{Acc}.

The measures of the third block prioritize patterns that are more present in the positive class than in the negative class, with a high frequency of appearance overall. In contrast to Blocks~1 and~2, patterns that are only present in the positive class, but very infrequent, will not necessarily be ranked before more frequent patterns. This block is represented in the rest of our results by \textsc{Cos}.

The fourth block is constituted of measures that do not respect the Contrastivity property (see Definition~\ref{def:Contrastivity}), and are only based on $p(P,\mathcal{G}^+)$. The presence of the pattern in the negative class no longer matters, compared to the other blocks. This block is represented in the rest of our results by \textsc{Sup}.

The measures of the fifth block prioritize patterns that are often absent from the negative class. Similarly to the first block, they favor patterns that never appear in the negative class, regardless of their frequency in the positive class. However, they also rank well patterns that appear only a few times in the negative class, even if they are infrequent in the positive class. This block is represented in the rest of our results by \textsc{Spec}. 

\textcolor{highlightCol}{The measures composing the sixth block focus on patterns that exhibit a strong contrast between the positive and negative classes. However, unlike the other blocks, they also assign high scores to patterns where this contrast favors the negative class. This block is represented in the rest of our results by \textsc{Dep}.}

\citet{LoyolaGonzalez2014} carried out a similar experiment, but worked on tabular data rather than graphs. They mine the most frequent patterns from 61 datasets, keeping only those that are more present in the positive than in the negative class. The patterns obtained are then ranked with each quality measure, and the different rankings are compared using Kendall's Tau. Overall, their results are similar to ours, but diverge in the following aspects:
\begin{itemize}
    \item \textsc{Excex} and \textsc{Dep} are not in the same block as \textsc{GR}.
    \item \textsc{Pearson} and $\chi^2$ are not correlated.
    \item \textsc{OddsRatio} and \textsc{MDisc} are not correlated.
\end{itemize}
In addition to these differences, \textsc{Lever} is not associated with the \textsc{Acc} block, but it would appear that this is due to a mistake in the definition of the formula in~\cite{LoyolaGonzalez2014}, compared with that used in the literature~\cite{Webb2005}. 

All these differences can be explained by the fact that in our case, some patterns are more present in the negative than in the positive class, unlike the study conducted by \citet{LoyolaGonzalez2014}, which only focuses on patterns more frequent in the positive class. As a result, some patterns that are often present in the negative class are well ranked by some scores (\textsc{Dep}, \textsc{Excex}, \textsc{MDisc}, $\chi^2$) and not so much by others (\textsc{GR}, \textsc{OddsRatio}, \textsc{Pearson}). 

At this stage, we have determined that some measures are so correlated that it is not worth examining all of them, and that we can focus only on a subset constituted of \textcolor{highlightCol}{21} measures instead of \textcolor{highlightCol}{38}, in the rest of our experiments. We next compare the rankings obtained with this subset, to our gold standard ranking.

\subsection{Gold Standard Comparison}
\label{sec:ExpGold}
We now compare the measures selected at the previous section with our gold standard, on two aspects. First, in terms of pattern ranking, by assessing the similarity between the ranking obtained with each quality measure and the ranking produced based on the gold standard (Section~\ref{sec:ExpGoldRanks}). Second, in terms of classification performance, by comparing the $F1$-Score obtained when using the top representatives according to each quality measure to represent the graph, against the top representatives according to the gold standard (Section~\ref{sec:ExpGoldPerfs}). In each case, we examine the impact of the parameter $s$, which indicates the number of representatives used to represent the graphs and perform the classification.

\subsubsection{Ranking Comparison}
\label{sec:ExpGoldRanks}
Let us first compare the rankings estimated using the quality measures to the gold standard one. Using each measure, we rank the representatives identified at the clustering step, and focus on the top $s$ patterns. As a result, this list of $s$ patterns may differ from one measure to the other, and also from the gold standard. For this reason, we cannot use Kendall's Tau, and turn to the RBO instead (see Section~\ref{sec:MethRanksComp}). An RBO close to 1 indicates that the rankings share simultaneously the same elements and the same order.

\begin{figure}[htbp!]
    \centering
    \begin{minipage}[b]{0.32\textwidth}
        \includegraphics[width=\textwidth]{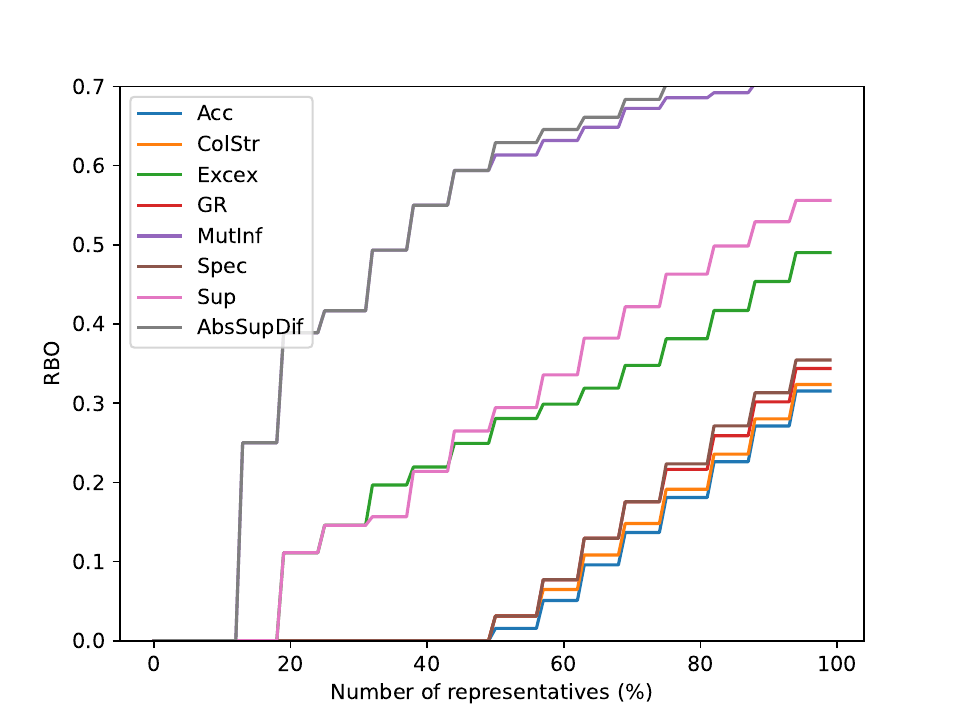}
        \subcaption{MUTAG}
        \label{fig:Shapley_RBO_MUTAG_main}
    \end{minipage}
    \begin{minipage}[b]{0.32\textwidth}
        \includegraphics[width=\textwidth]{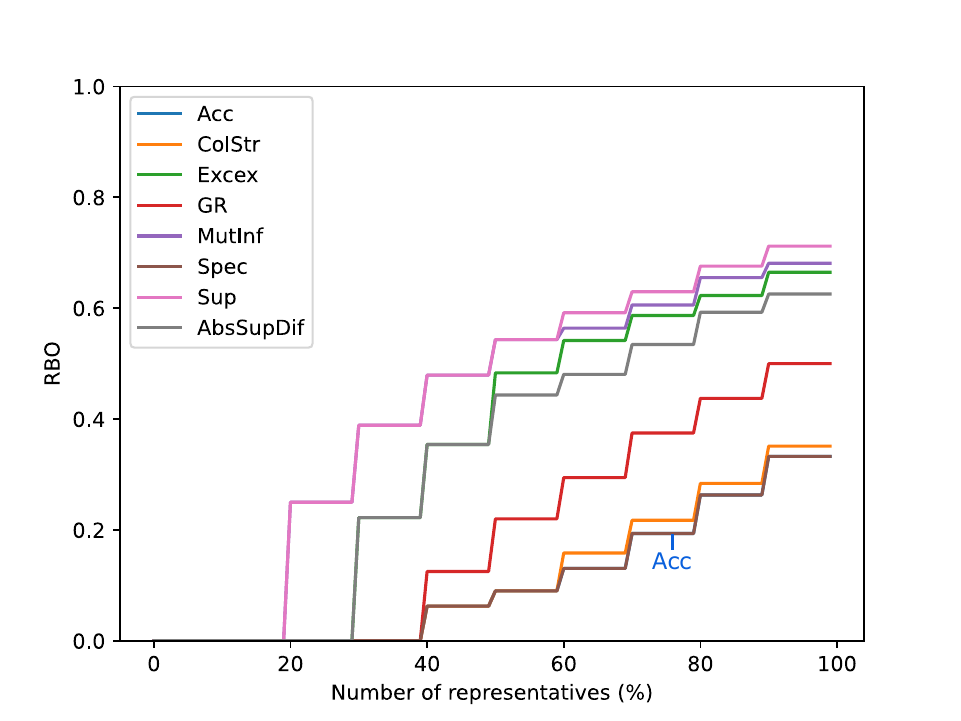}
        \subcaption{PTC}
        \label{fig:Shapley_RBO_PTC_main}
    \end{minipage}
    \begin{minipage}[b]{0.32\textwidth}
        \includegraphics[width=\textwidth]{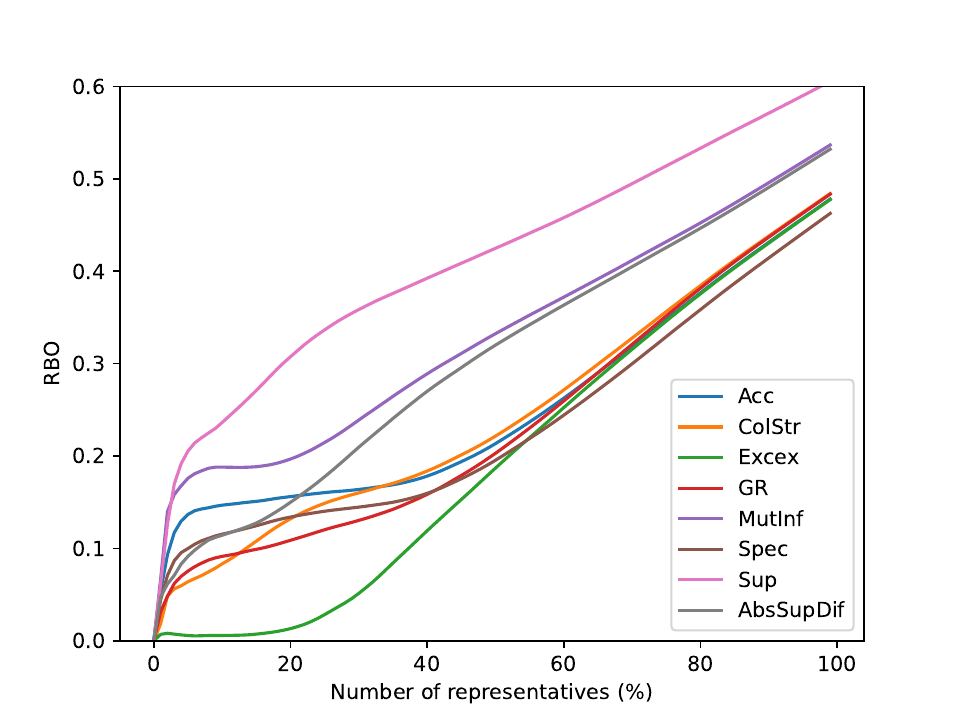}
        \subcaption{NCI1}
        \label{fig:Shapley_RBO_NCI1_main}
    \end{minipage}
    
    \vspace{1em}
    
    \begin{minipage}[b]{0.32\textwidth}
        \includegraphics[width=\textwidth]{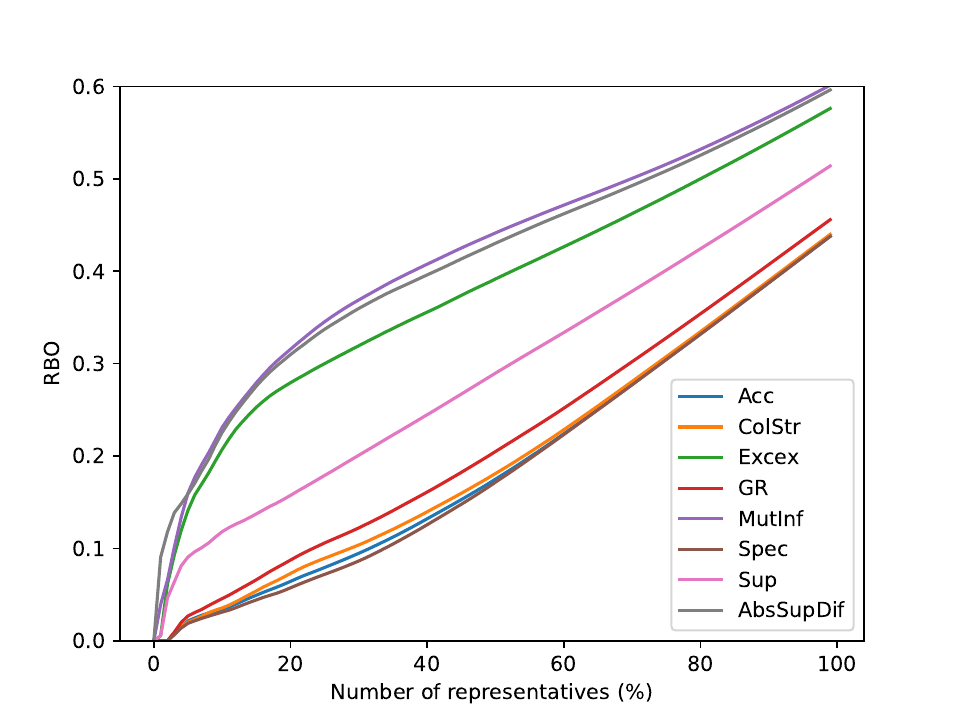}
        \subcaption{D\&D}
        \label{fig:Shapley_RBO_DD_main}
    \end{minipage}
    \begin{minipage}[b]{0.32\textwidth}
        \includegraphics[width=\textwidth]{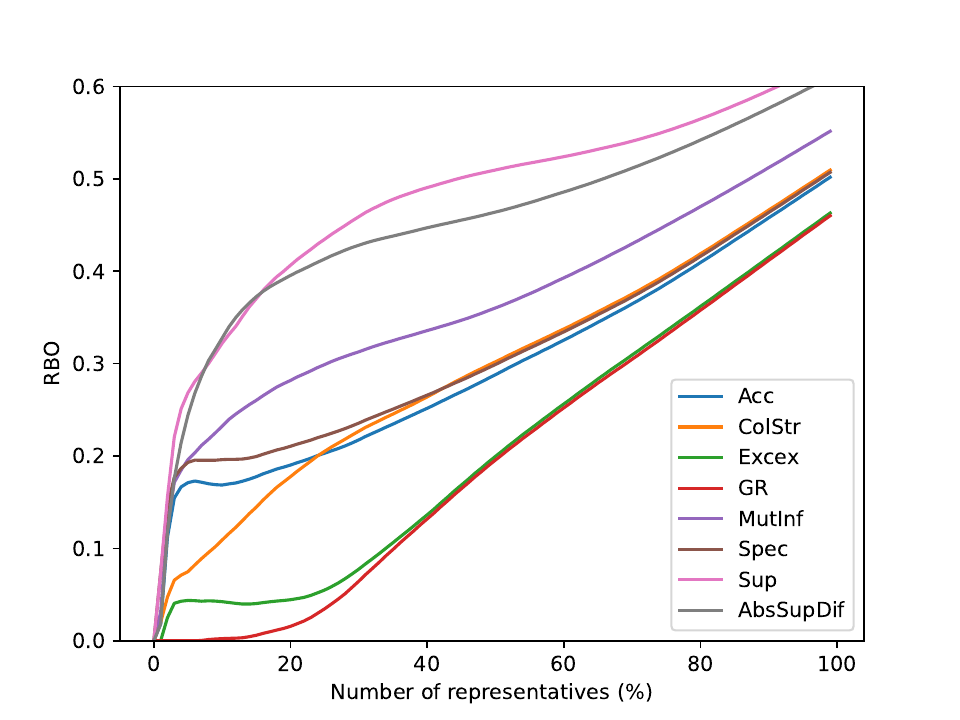}
        \subcaption{AIDS}
        \label{fig:Shapley_RBO_AIDS_main}
    \end{minipage}
    \begin{minipage}[b]{0.32\textwidth}
        \includegraphics[width=\textwidth]{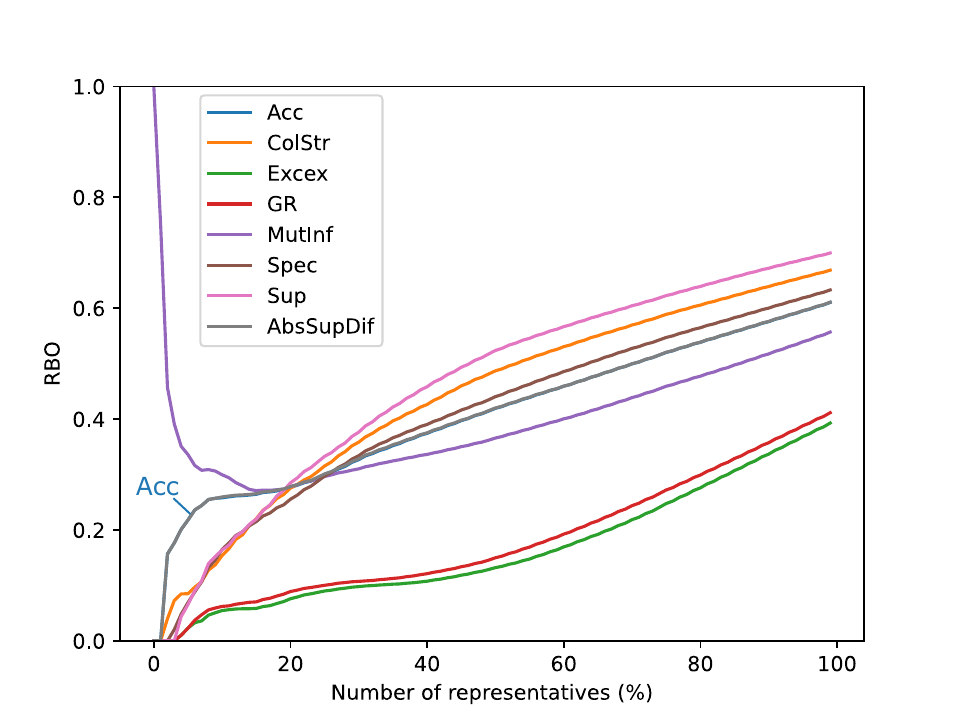}
        \subcaption{FOPPA}
        \label{fig:Shapley_RBO_FOPPA_main}
    \end{minipage}
    \caption{RBO between the gold standard and the rankings obtained for the eight quality measure of interest, as a function of $s$, the number of top representatives considered. For all quality measures, see Appendix~\ref{sec:AppFullRBO}.}
    \label{fig:ShapleyRBO_main}
    \Description{Description} 
\end{figure}

Figure~\ref{fig:ShapleyRBO_main} shows the RBO obtained between quality measures and our gold standard, as a function of $s$, the number of representatives considered. To ease the comparison between the datasets, this quantity is expressed as a percentage of the total number of representatives. To improve readability, we only display the results for a selection of 8 quality measures of interest: \textsc{Acc}, \textsc{Cos}, \textsc{Excex}, \textsc{GR}, \textsc{MutInf}, \textsc{Spec}, \textsc{Sup} and \textsc{AbsSupDif}. We decide to choose these eight measures because they provide an overall view of all observed behaviors. Full results are available in Appendix~\ref{sec:AppFullRBO}. The staircase effect observed for some of the datasets (MUTAG, PTC) is due to the small number of representatives remaining after Step~3 of our process.

There is an overall trend: at some point, for all the measures, the RBO starts increasing with $s$, and keeps doing so until reaching the maximal $s$ value considered. There seems to be some kind of convergence, for large $s$ values. This can be explained by the fact that a greater $s$ means more patterns, and therefore more chance to overlap with the gold standard. This results in a larger RBO, even if the patterns are not placed in the same order. Besides this similarity, the measures exhibit differences for smaller values of $s$. One can distinguish three types of behavior. 
The first one is the most common. It corresponds to measures that start with a zero RBO, before undergoing a regular increase with $s$ (e.g. all the measures in Figure~\ref{fig:Shapley_RBO_MUTAG_main}). This means that the top patterns according to these measures are not those of the gold standard, but that they appear farther in the estimated rankings, when considering more patterns. From an operational perspective, these experimental results can help to select an appropriate value for parameter $s$, depending on the measure of interest. Indeed, the RBO of some measures starts increasing much later than others. For instance, in Figure~\ref{fig:Shapley_RBO_MUTAG_main}, \textsc{Sup} starts around 20\%, when \textsc{GR} has a zero RBO until almost 50\% of the patterns are considered. 

The second type of behavior concerns only a few measures. They start strong, then see their RBO decrease, before reaching the increase discussed before (e.g. \textsc{MutInf} in Figure~\ref{fig:Shapley_RBO_FOPPA_main}). Unlike the first type, these measures are good at identifying the top patterns, however they disagree with the gold standard regarding the patterns that come after. 
The third type of behavior is also followed by a few measures. They start from zero, have a strong increase at first, before undergoing a decrease, followed by the usual increase (e.g. \textsc{AbsSupDif} in Figure~\ref{fig:Shapley_RBO_FOPPA_main}). These are not good at detecting the top patterns, but manage to match the gold standard on the intermediary ones.

Many measures exhibit the same behavior across all datasets (\textsc{Acc}, \textsc{ColStr}, \textsc{GR}, \textsc{Sup}...) but this is not truth for all of them. For example, \textsc{MutInf} follows the first type of behavior on D\&D (continuous increase), whereas it follows the second type (increase followed by stagnation) on NCI1. 
One can also distinguish measures in terms of how quick they approach the gold standard ranking. From this perspective, \textsc{AbsSupDif} is particularly efficient on all datasets, and \textsc{Sup} on datasets all but D\&D. Alternatively, \textsc{GR} and \textsc{Acc}, the two measures representing the largest block of correlated measures found in Section~\ref{sec:ExpPairwise}, are not particularly close to the ranking obtained by the gold standard, regardless of the dataset.

\subsubsection{Classification Comparison}
\label{sec:ExpGoldPerfs}
\begin{figure}[htbp!]
    \centering
    \begin{minipage}[b]{0.32\textwidth}
        \includegraphics[width=\textwidth]{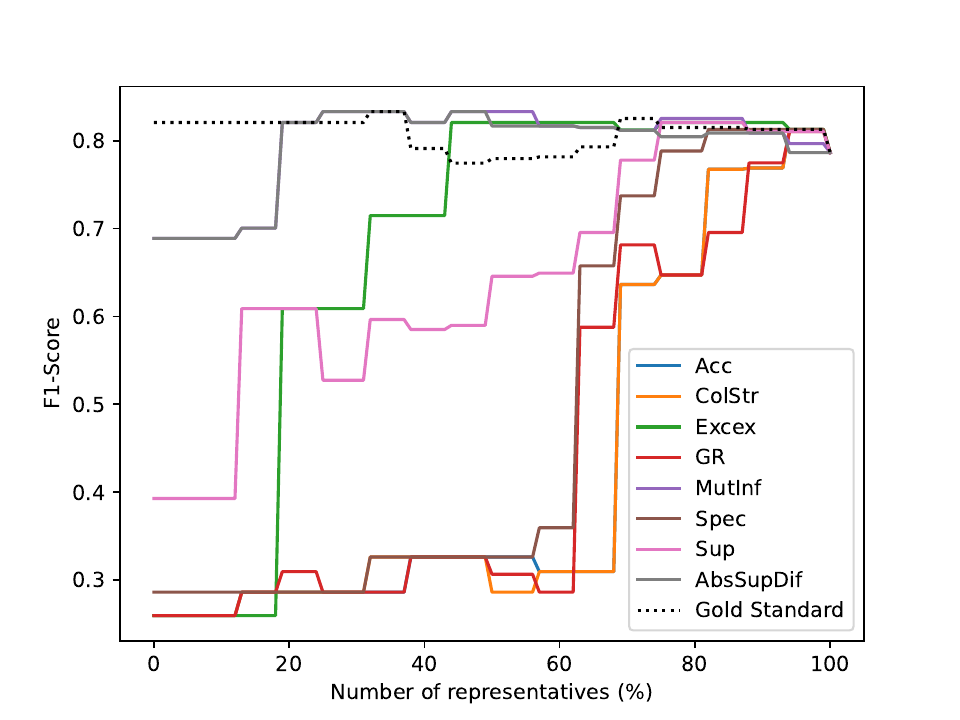}
        \subcaption{MUTAG}
        \label{fig:Shapley_F1_MUTAG_main}
    \end{minipage}
    \begin{minipage}[b]{0.32\textwidth}
        \includegraphics[width=\textwidth]{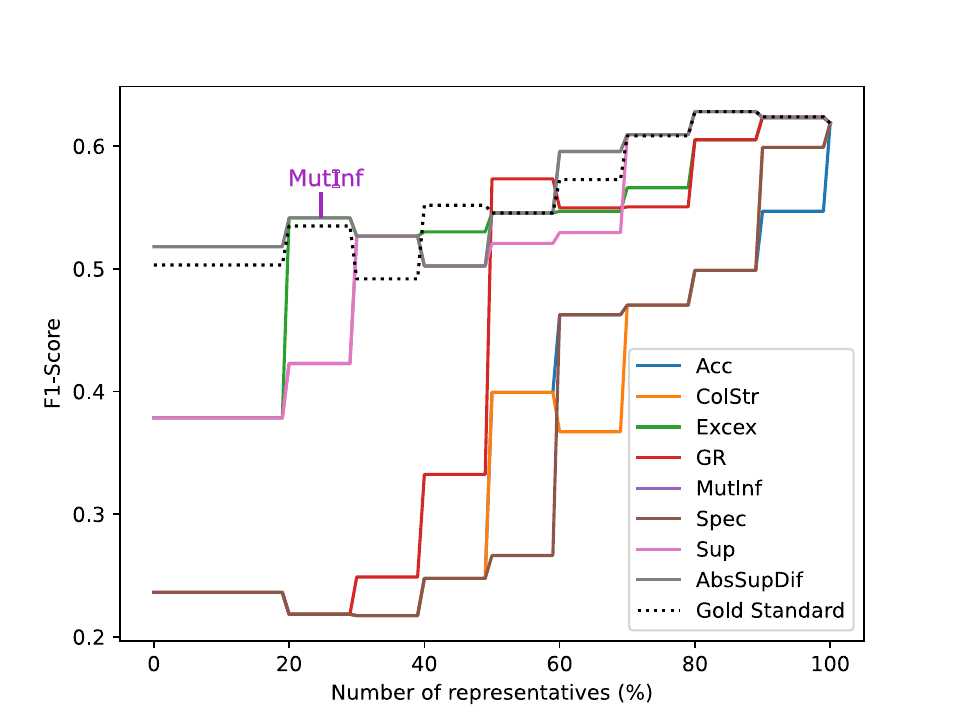}
        \subcaption{PTC}
        \label{fig:Shapley_F1_PTC_main}
    \end{minipage}
    \begin{minipage}[b]{0.32\textwidth}
        \includegraphics[width=\textwidth]{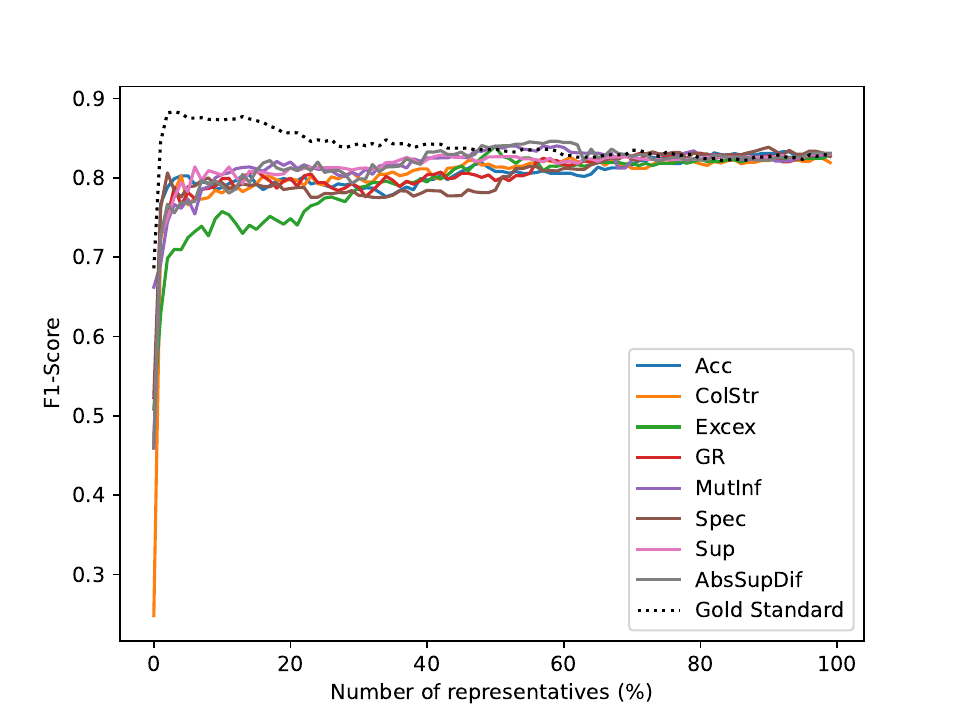}
        \subcaption{NCI1}
        \label{fig:Shapley_F1_NCI1_main}
    \end{minipage}
    
    \vspace{1em}
    
    \begin{minipage}[b]{0.32\textwidth}
        \includegraphics[width=\textwidth]{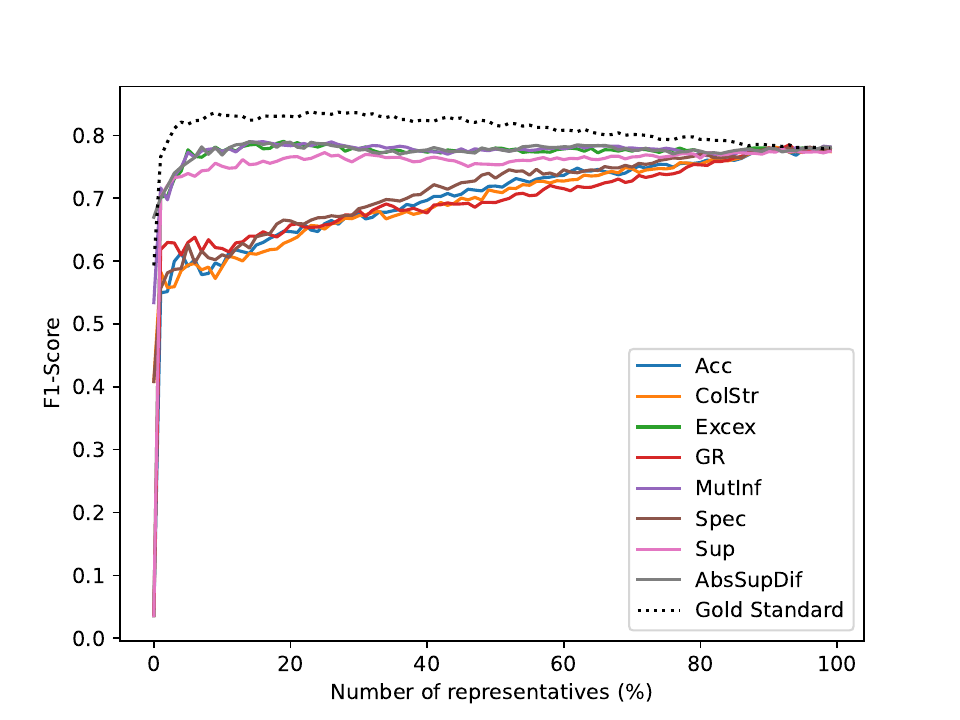}
        \subcaption{D\&D}
        \label{fig:Shapley_F1_DD_main}
    \end{minipage}
    \begin{minipage}[b]{0.32\textwidth}
        \includegraphics[width=\textwidth]{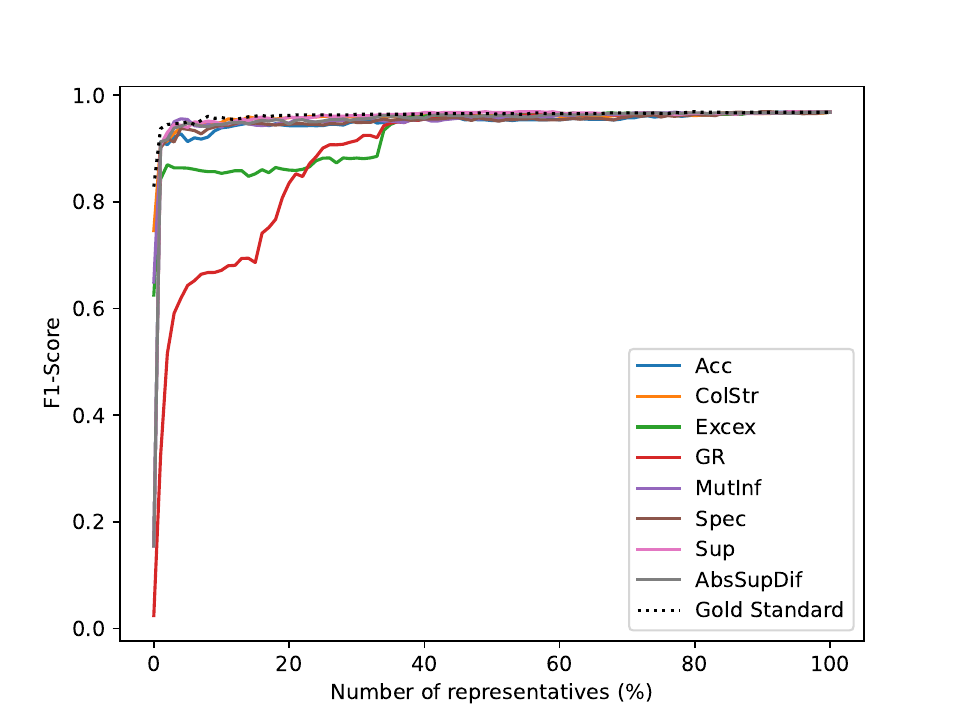}
        \subcaption{AIDS}
        \label{fig:Shapley_F1_AIDS_main}
    \end{minipage}
    \begin{minipage}[b]{0.32\textwidth}
        \includegraphics[width=\textwidth]{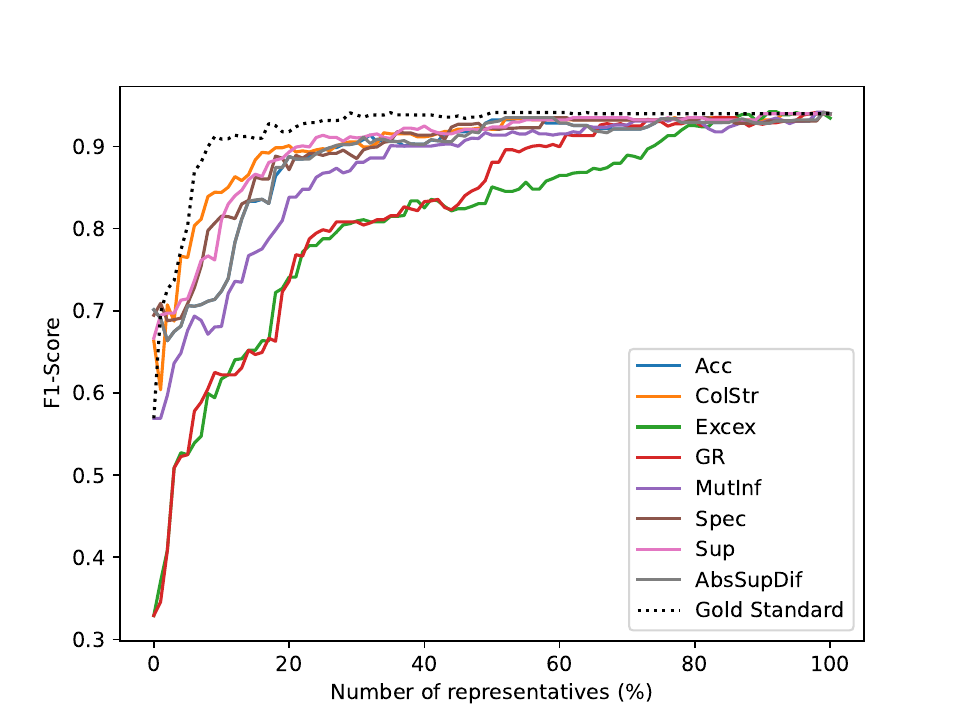}
        \subcaption{FOPPA}
        \label{fig:Shapley_F1_FOPPA_main}
    \end{minipage}
    \caption{$F1$-Score as a function of the proportion of representatives selected, for each quality measure of interest, as well as the gold standard (dotted line). To visualize all quality measures, see Appendix~\ref{sec:AppFullF1}.}
    \label{fig:ShapleyF1_main}
    \Description{Description} 
\end{figure}

We now compare the measures to the gold standard in terms of classification performance. As before, we select the top $s$ representatives identified by each quality measure, but this time we use them to build a representation of the graphs, train the classifier and compute the classification performance. The reference performance is obtained by proceeding similarly with the gold standard. Figure~\ref{fig:ShapleyF1_main} shows the classification performance for each quality measure, expressed in terms of $F1$-Score, as a function of $s$, the number of representatives selected, expressed as a percentage as before. The dotted line represents the performance obtained using the gold standard ranking. We focus on the same eight measures of interest as before, while the full results are provided in Appendix~\ref{sec:AppFullF1}.

As expected, selecting the best representatives according to the gold standard leads to the fastest increase in $F1$-Score, for all datasets. This supports our decision to consider SAGE values as a good proxy for a ground truth, regarding pattern ranking. As one would expect, increasing the number of patterns used for classification generally leads to better performance, but it is not always the case. For instance, the $F1$-Score obtained for NCI1 and D\&D quickly starts decreasing when using more than 4\% and 13\% of the representatives, respectively. And for AIDS and FOPPA, the $F1$-Score reaches a plateau after using only a fraction of the representatives: increasing $s$ further leads to a higher computational cost without any improvement in terms of classification.

We observe two main behaviors among the measures. Some of them undergo a brutal increase in classification performance from the start, which then stays relatively stable when increasing $s$ further. A good example is \textsc{Sup} in dataset D\&D (Figure~\ref{fig:Shapley_F1_DD_main}). These measures are able to rank the most discriminative patterns first. 
The second behavior corresponds to a much more progressive increase with $s$. This is illustrated by \textsc{GR} in the same dataset. These measures require considering more patterns in order to capture the same level of discriminative power. Generally speaking, a larger $s$ means using more patterns, which results in a higher classification performance for most measures. Indeed, the specific ranking produced by a measure only affects which patterns are selected for the classification step. When many or all patterns are selected, this ranking is irrelevant. This explains why all measures end up with the same $F1$-Score for maximal $s$, despite possibly very low RBO scores (Figure~\ref{fig:ShapleyRBO_main}). 

The measures that produce rankings correlated to the gold standard, in terms of RBO, also perform well in terms of classification. For example, in the case of dataset D\&D, (Figure~\ref{fig:Shapley_F1_DD_main}), \textsc{AbsSupDif}, \textsc{Sup} and \textsc{Excex} are the best measures both in terms of classification performance and RBO score. However, certain measures also get a good classification performance despite exhibiting only moderate RBO correlation with the gold standard. For instance, for dataset AIDS, (Figure~\ref{fig:Shapley_F1_AIDS_main}) all the measures except \textsc{Excex} and \textsc{GR} reach a comparably good $F1$-score, despite low RBO scores (Figure~\ref{fig:Shapley_RBO_AIDS_main}). 

Measures \textsc{Sup} and \textsc{AbsSupDif} are associated with good performances in all datasets and can be considered as safe choices, when no prior knowledge is available. \textcolor{highlightCol}{This is due to the fact that \textsc{AbsSupDif} possesses the Class Symmetry property. Unlike measures that prioritize one class over the other, it evaluates patterns based on their frequency in both the positive and negative classes.} In contrast, the results for \textsc{GR} and \textsc{Acc} are not among the best quality measures. This can be explained by the fact that the measures belonging to the block containing \textsc{GR} correspond to measures that do not respect the \textit{Jumpiness} property. As a result, the patterns selected first only represent a few graphs, which implies that the measure follows the second behavior, synonymous with a slow and gradual increase and therefore a reduced performance. 
Measures belonging to the \textsc{Acc} block favor patterns that are more frequent in the positive than in the negative class, but neglect patterns with a higher frequency in the negative class. \textsc{AbsSupDif}, which considers the most frequent patterns in each of the two classes, obtains better results.

\textcolor{highlightCol}{Among all measures, \textsc{Excex} produces the least effective rankings. While it possesses the \textit{Contrastivity} property, it fails to satisfy \textit{Jumpiness} and therefore suffers from the same issues as \textsc{GR}: it does not properly differentiate patterns that are exclusively present in the positive class. However, \textsc{Excex} presents an additional weakness: it is not directly expressed as a function of $p(\mathcal{G}^+ \mid P)$. This formulation leads to unstable rankings, particularly when patterns have low support. This combination of factors explains why it systematically gives the lowest results in our benchmark.}

\section{Conclusion}
\label{sec:Conclusion}
In this work, we deal with the problem of pattern-based graph classification. We provide a comprehensive review of \textcolor{highlightCol}{38} quality measures proposed in the literature to assess the discriminative power of such patterns. We characterize these measures through four properties that are relevant to our task. We constitute a benchmark of graph datasets and elaborate gold standard rankings of their patterns by leveraging the Shapley value. We use these resources to empirically assess and compare the measures depending on two aspects: the way they rank the patterns, and their effect on classification performance. It turns out \textsc{AbsSupDif} and \textsc{Sup} give good results overall, when some measures commonly used in the literature, such as \textsc{InfGain}~\cite{Thoma2009}, are considerably less relevant. 
In addition, we propose a preprocessing step based on cluster analysis to decrease the number of patterns used during classification. These clusters are obtained by grouping patterns exhibiting similar footprints, i.e. that are present and absent from the same graphs. Not only does this step allow reducing the graph representation dimension and lowering the computational cost, but it also tends to improve the classification performance. These clusters are also interesting from the perspective of interpretation, as they correspond to groups of patterns that are possibly very different, but are characteristic of the same class. Finally, we also show empirically that restricting pattern mining to specific types of patterns, such as induced or closed ones, also results in a smaller selection of patterns for equal performance.

Our work opens several perspectives. It is limited to the case of balanced classes, so the first extension is straightforward: consider unbalanced classes, which requires handling an extra parameter, the level of imbalance, in order to study its effect. Similarly, our experiments focus on two-class datasets and measures: a second extension is to turn to the multiclass case. This means considering one-vs-all approaches to apply the same measures as in this survey, identifying other measures able to directly handle multiple classes, and finding multiclass datasets. Here too, there is an additional parameter, the number of classes, whose effect must be studied. 
A third and more indirect extension of our work is to compare the effectiveness of quality measures with methods that directly mine subsets of discriminative patterns, such as CORK~\cite{Thoma2009}. In particular, it would be interesting to study how the patterns identified by such methods are distributed over our gold standard ranking. 
Fourth and finally, another research lead could be to work directly on the pattern mining method itself. Existing approaches are agnostic, in the sense they are independent of the final task (in our case, classification). As a consequence, they typically work by starting with small patterns and iteratively extending them. It could be interesting to develop a method tailored for classification, that would work on the pattern \textit{footprints} rather than the patterns themselves.

\bibliographystyle{ACM-Reference-Format}
\bibliography{Potin2024_biblio.bib}

\appendix

\section{Additional Information About Quality Measures}
This appendix contains additional information related to the quality measures and their properties.

\subsection{Excluded Quality Measures}
\label{sec:AppQM}
As explained in Section~\ref{sec:QualityMeasures}, we discarded three quality measures of the literature~\cite{Chen2022} from our experiments. The first one, \textsc{GenQuotient}, requires the user to set a specific parameter value. The second, \textsc{SupMaxK}, is designed for itemsets, and cannot handle graphs natively. It could be possible to adapt it to this use case, but this is out of the scope of this paper. The third measure, \textsc{PValue}, is unsuitable to large datasets due to its computational cost. 

\begin{table}[htbp!]
    \centering
    \begin{tabular}{l l}
        \toprule
        Quality Measure & Definition \\
        \midrule
        \textsc{Gen Quotient} & $\frac{p(P \mid \mathcal{G}^+)}{p(\overline{P} \mid \mathcal{G}^+) + g}$ \\
        [+8pt]
        \textsc{SupMaxK} & $p(P \mid \mathcal{G}^+) - \max_{P_i \subset P} p(P_i \mid \mathcal{G}^-) $ \\
        [+8pt]
        \textsc{pValue} & $\sum_{i=0}^{\max(n_{12},n_{21})}  \frac{t_{1}!t_{2}!|\mathcal{G^+}|!|\mathcal{G^-}|!}{|\mathcal{G}|!(n_{11}+i)!(n_{12}-i)!(n_{21}-i)!(n_{22}+i)!} $ \\
        \bottomrule
    \end{tabular}
    \caption{Name and formula of the three quality measures that appear in the literature~\cite{Chen2022}, but that we discarded from our experiments.}
    \label{tab:QMNotConsidered}
\end{table}

Table~\ref{tab:QMNotConsidered} provides their definitions, for the sake of completeness. In order to enhance the readability of the \textsc{pValue} formula, we note $n_{11}$ the support of $P$ in $\mathcal{G^+}$, $n_{12}$ the support of $P$ in $\mathcal{G^-}$, $n_{21}$ the support of $\overline{P}$ in $\mathcal{G^+}$ and $n_{22}$ the support of $\overline{P}$ in $\mathcal{G^-}$

\subsection{Balanced Classes and Independence / Equilibrium}
\label{sec:Proofs}
As mentioned in Section~\ref{sec:QualityMeasuresProp}, here is the proof that the properties of \textit{Independence} and \textit{Equilibrium} defined by \citet{LoyolaGonzalez2014} are equivalent under the assumption that the considered classes are balanced.

\begin{proof}
    As explained in Section~\ref{sec:QualityMeasures}, \textit{Independence} is defined as $p(P,\mathcal{G}^+) = p(P) p(\mathcal{G}^+)$, whereas \textit{Equilibrium} is defined as $p(\mathcal{G}^+ \mid P) = p(\mathcal{G}^- \mid P)$. We want to show that, under the assumption that both classes have the same size, i.e. $|\mathcal{G}^+| = |\mathcal{G}^-|$, both properties are equivalent. 
    
    We first focus on proving \textit{Independence} $\Rightarrow$ \textit{Equilibrium}. Starting with the definition of conditional probability applied to the positive class, we have
    \begin{align}
        p(\mathcal{G}^+ \mid P) &= \frac{p(P,\mathcal{G}^+)}{p(P)}. \\
    \intertext{Assuming independance yields}
        p(\mathcal{G}^+ \mid P) &= \frac{p(P) p(\mathcal{G}^+)}{p(P)} \\
                                &= p(\mathcal{G}^+).
    \end{align}
    
    We use the definition of conditional probability on the negative class
    \begin{align}
        p(\mathcal{G}^- \mid P) &= \frac{p(P,\mathcal{G}^-)}{p(P)} \\
    \intertext{Using the law of total probability, we have $p(P,\mathcal{G}^+)+p(P,\mathcal{G}^-)=p(P)$, and therefore}
        p(\mathcal{G}^- \mid P) &= \frac{p(P) - p(P,\mathcal{G}^+)}{p(P)}\\
        p(\mathcal{G}^- \mid P) &= \frac{p(P) - p(P) p(\mathcal{G}^+)}{p(P)}\\
                                &= 1 - p(\mathcal{G}^+).
    \end{align}
    
    Now, if $|\mathcal{G}^+| = |\mathcal{G}^-|$, then $p(\mathcal{G}^+) = 0.5$. Consequently,
    \begin{align}
        p(\mathcal{G}^+ \mid P) &= p(\mathcal{G}^- \mid P) = 0.5,
    \end{align}
    and the \textit{Equilibrium} property is verified.
    
    We now turn to proving \textit{Equilibrium} $\Rightarrow$ \textit{Independence}. On the one hand, the \textit{Equilibrium} property states
    \begin{align}
        \label{eq:bal1}
        p(\mathcal{G}^+ \mid P) &= p(\mathcal{G}^- \mid P).
    \end{align}
    On the other hand, we have
    \begin{align}
        \label{eq:bal2}
        p(\mathcal{G}^+ \mid P) + p(\mathcal{G}^- \mid P) &= 1.
    \end{align}
    Combining (\ref{eq:bal1}) and (\ref{eq:bal2}) yields $p(\mathcal{G}^+ \mid P)=p(\mathcal{G}^- \mid P) = 0.5$. In addition, as shown before, $p(\mathcal{G}^+) = 0.5$, thus
    \begin{align}
        p(P,\mathcal{G}^+)  &= p(\mathcal{G}^+ \mid P) p(P) \\
                            &= 0.5 \cdot p(P) \\
                            &= p(\mathcal{G}^+) p(P),
    \end{align}
    and the \textit{Independence} property is verified.
\end{proof}

\subsection{Additional Properties}
\label{sec:AddProps}
\citet{Ventura2016} list seven properties defined to characterize quality measures in the context of \textit{association rule mining}. The rules have the form $X \rightarrow Y$, where $X$ and $Y$ are itemsets. If we assume instead that $X$ denotes the presence of a graph pattern $P$, and that $Y$ denotes our positive class $\mathcal{G}^+$, then these rules can be considered as classification rules, and the properties can be applied to characterize the quality measures listed in Section~\ref{sec:QualityMeasuresDef}. In the following, we first consider the three properties by \citet{PiatetskyShapiro1991a} (Appendix~\ref{sec:AddPropsPSF}), then the four properties by \citet{Tan2004} (Appendix~\ref{sec:AddPropsT}).

\subsubsection{Properties of Piatetsky-Shapiro}
\label{sec:AddPropsPSF}
We note PS1 the first property of~\citet{PiatetskyShapiro1991a}. It is defined as follows:
\begin{definition}[PS1]
    Let $X$ and $Y$ be two itemsets with no item in common. Quality measure $q$ possesses Property 1 of~\citet{PiatetskyShapiro1991a} iff
    \center
    $q(X \rightarrow Y) = 0 \text{ when } p(X,Y) = p(X)p(Y)$.
\end{definition}
In other words, the measure is zero when $X$ and $Y$ are independent. If we translate in terms of patterns and classes, we get $p(P, \mathcal{G}^+) = p(P)p(\mathcal{G}^+)$. This is equivalent to the \textit{Independence} property of \citet{LoyolaGonzalez2014}, as described in Section~\ref{sec:QualityMeasuresProp}.

The second property of \citet{PiatetskyShapiro1991a}, which we note PS2, is defined as:
\begin{definition}[PS2]
    Let $X$ and $Y$ be two itemsets with no item in common. Quality measure $q$ possesses Property 2 of~\citet{PiatetskyShapiro1991a} iff
    \center
    $q(X \rightarrow Y) \text{ monotonically increases with } p(X,Y) \text{ when } p(X) \text{ and } p(Y) \text{ remain the same}$.
\end{definition}
In our case, for a given dataset, the classes are fixed and only the patterns can exhibit different distributions over the graphs. Therefore, $p(G^+)$ (the counterpart of $p(Y)$) is constant, and only $p(P)$ (the counterpart of $p(X)$) can vary. If we translate the property in terms of patterns and classes, under a form comparable to our own properties from Section~\ref{sec:QualityMeasuresProp}, then we get
\begin{equation}
    \forall P_i,P_j,\; \big[ p(P_i, \mathcal{G}^+) > p(P_j, \mathcal{G}^+) \text{ and } p(P_i) = p(P_j) \big] \Rightarrow \big[ q(P_i,\mathcal{G}^+, \mathcal{G}^-) > q(P_j,\mathcal{G}^+, \mathcal{G}^-)  \big].
    \label{eq:PS2}
\end{equation}
This property is very similar to our \textit{Contrastivity} property from Section~\ref{sec:QualityMeasuresProp}, with the difference that here we assume $p(P_i, \mathcal{G}^+) > p(P_j, \mathcal{G}^+)$ instead of $p(P_i, \mathcal{G}^+) = p(P_j, \mathcal{G}^+)$. Intuitively, the \textit{Contrastivity} states that the quality measure must increase when one \textit{deletes} a pattern occurrence from the negative class, whereas SP2 states that it must increase when one \textit{switches} a pattern occurrence from the negative to the positive class.

Importantly, PS2 is mutually exclusive with \textit{Class Symmetry} (cf. Section~\ref{sec:QualityMeasuresProp}), i.e. a measure cannot simultaneously possess both properties. We provide a proof in Appendix~\ref{sec:ProofsProperty2}). In addition, a quality measure which verifies neither PS2 nor \textit{Class Symmetry} is not particularly effective: it is not able to identify patterns that are very frequent in either the positive or the negative class. As a result, it is not necessary to consider both measures when describing quality measures. In this work, we focus on the \textit{Class symmetry}.

The third property of \citet{PiatetskyShapiro1991a}, which we note PS3, is defined as:
\begin{definition}[PS3]
    Let $X$ and $Y$ be two itemsets with no item in common. Quality measure $q$ possesses Property 3 of~\citet{PiatetskyShapiro1991a} iff
    \center
    $q(X \rightarrow Y) \text{ monotonically decreases with } p(X) \text{ or with } p(Y) \text{ when } p(X,Y) \text{ and } p(Y) \text{ or } p(X) \text{ remain the same}$.
\end{definition}
Same as before, in our case $p(Y)$ cannot change. If we translate this property in terms of patterns and classes, we get
\begin{equation}
    \forall P_i,P_j,\; \big[ p(P_i) > p(P_j) \text{ and } p(P_i, \mathcal{G}^+) = p(P_j, \mathcal{G}^+) \big] \Rightarrow \big[ q(P_i,\mathcal{G}^+, \mathcal{G}^-) < q(P_j,\mathcal{G}^+, \mathcal{G}^-)  \big].
\end{equation}
If $P_i$ is more frequent than $P_j$ while they are equally frequent in $\mathcal{G}^+$, then $P_i$ is more frequent than $P_j$ in $G^-$. Consequently, the property can be rewritten as
\begin{equation}
    \forall P_i,P_j,\; \big[ p(P_i, \mathcal{G}^-) > p(P_j, \mathcal{G}^-) \text{ and } p(P_i, \mathcal{G}^+) = p(P_j, \mathcal{G}^+) \big] \Rightarrow \big[ q(P_i,\mathcal{G}^+, \mathcal{G}^-) < q(P_j,\mathcal{G}^+, \mathcal{G}^-)  \big].
\end{equation}
In the end, this property is equivalent to our \textit{Contrastivity} property from Section~\ref{sec:QualityMeasuresProp}.

\subsubsection{Properties of Tan \textit{et al}.}
\label{sec:AddPropsT}
The first property of~\citet{Tan2004} is related to the symmetry under variable permutation. We note it T1, and it is defined as 
\begin{definition}[T1]
     Let $X$ and $Y$ be two itemsets with no item in common. Quality measure $q$ is symmetric under variable permutation iff
    \center
     $q(X \rightarrow Y) = q(Y \rightarrow X)$.
\end{definition}
In our situation, this property does not apply since we focus only on classification rules, i.e. rules of the form $X \rightarrow Y$ (where $X$ corresponds to a pattern and $Y$ to a class). Thus, variable permutation is irrelevant. 

The second property of~\citet{Tan2004} is related to the notion of antisymmetry in the following matrix, called \textit{Table of relative frequencies} in~\cite{Ventura2016}:
\begin{equation}
    \begin{bmatrix}
        p(X,Y) & p(X,\overline{Y}) \\
        p(\overline{X},Y) & p(\overline{X},\overline{Y})
    \end{bmatrix}.
\end{equation}
This antisymmetry property has two variants. The first focuses on the \textit{rows} of this matrix:
\begin{definition}[T2a]
     Let $X$ and $Y$ be two itemsets with no item in common. Quality measure $q$ is antisymmetric under row permutation iff
    \center
     $q(\overline{X} \rightarrow Y) = -q(X \rightarrow Y)$.
\end{definition}
If we translate to patterns and classes, we get 
\begin{equation}
    q(\overline{P}, \mathcal{G}^+, \mathcal{G}^-) = -q(P, \mathcal{G}^+, \mathcal{G}^-).
\end{equation}
This property is similar to our \textit{Pattern Symmetry} property from Section~\ref{sec:QualityMeasuresProp}, with the difference of the minus sign in the right-hand term.

The second variant of this second property focuses on the \textit{columns} of the matrix:
\begin{definition}[T2b]
     Let $X$ and $Y$ be two itemsets with no item in common. Quality measure $q$ is antisymmetric under column permutation iff
    \center
     $q(X \rightarrow \overline{Y}) = -q(X \rightarrow Y)$.
\end{definition}
If we translate to patterns and classes, we get 
\begin{equation}
    q(P, \mathcal{G}^-, \mathcal{G}^+) = -q(P, \mathcal{G}^+, \mathcal{G}^-).
\end{equation}
This property is similar to our \textit{Class Symmetry} property from Section~\ref{sec:QualityMeasuresProp}, but like before, it differs in the minus sign present in the right-hand term.

The third property of~\citet{Tan2004} considers both types of permutations:
\begin{definition}[T3]
     Let $X$ and $Y$ be two itemsets with no item in common. Quality measure $q$ is symmetric under simultaneous row and column permutations iff
    \center
     $q(\overline{X} \rightarrow \overline{Y}) = q(X \rightarrow Y)$.
\end{definition}
If we translate to patterns and classes, we get 
\begin{equation}
    q(\overline{P}, \mathcal{G}^-, \mathcal{G}^+) = q(P, \mathcal{G}^+, \mathcal{G}^-).
\end{equation}
Note that this is not equivalent to possessing both T2a and T2b. For instance, in our selected measures, \textsc{Acc} does not respect T2a nor T2b, but it possesses T3. 

Finally, the fourth property of~\citet{Tan2004}, which we note T4, is called \textit{Null-Invariance}. It concerns measures that do not vary when considering a new dataset with more records not containing $X$ and $Y$. This property requires breaking class balance, so it is not relevant to our case.

\subsection{Mutual Exclusivity Between PS2 and Class Symmetry}
\label{sec:ProofsProperty2}
As mentioned in Appendix~\ref{sec:AddPropsPSF}, here is the proof that property PS2 of~\citet{PiatetskyShapiro1991a} (cf. Appendix~\ref{sec:AddPropsPSF}) and \textit{Class Symmetry} (cf. Section~\ref{sec:QualityMeasuresProp}) are mutually exclusive.

\begin{proof}
    As mentioned in Appendix~\ref{sec:AddPropsPSF} (\ref{eq:PS2}), property PS2 is defined as
    \begin{equation}
        \label{eq:P2}
        \forall P_i,P_j,\; \big[ p(P_i, \mathcal{G}^+) > p(P_j, \mathcal{G}^+) \text{ and } p(P_i) = p(P_j) \big] \Rightarrow \big[ q(P_i,\mathcal{G}^+, \mathcal{G}^-) > q(P_j,\mathcal{G}^+, \mathcal{G}^-)  \big].
    \end{equation}
    Moreover, \textit{Class Symmetry} is defined as: $\forall P,\; q(P,\mathcal{G}^+,\mathcal{G}^-) = q(P,\mathcal{G}^-,\mathcal{G}^+)$ (Definition~\ref{def:ClassSym}). We reformulate this property under a more convenient form by using two distinct but class-symmetrical patterns $P_i$ and $P_j$: 
    \begin{equation}
        \label{eq:Sym2Patterns}
        \forall P_i,P_j,\; \big[ p(P_i, \mathcal{G}^+) = p(P_j, \mathcal{G}^-) \text{ and } \big[ p(P_i, \mathcal{G}^-) = p(P_j, \mathcal{G}^+) \big] \Rightarrow \big[ q(P_i,\mathcal{G}^+, \mathcal{G}^-) = q(P_j,\mathcal{G}^+, \mathcal{G}^-)  \big].
    \end{equation}
    
    In the rest of our proof, we use two such class-symmetric patterns $P_1$ and $P_2$ defined as follows
    \begin{align}
        \label{eq:PiPj1}
        support(P_1,\mathcal{G}^+) &= support(P_2,\mathcal{G}^-) = x \\
        \label{eq:PiPj2}
        support(P_1,\mathcal{G}^-) &= support(P_2,\mathcal{G}^+) = y,
    \end{align}
    where $x > y$.
    
    First, we show that if $q$ verifies PS2, then it is not class-symmetric, i.e. PS2 $\Rightarrow \neg$ \textit{Class Symmetry}. Let us assume that $q$ is a quality measure satisfying P2. Using (\ref{eq:PiPj1}) and (\ref{eq:PiPj2}), we have
        \begin{align}
    p(P_1,\mathcal{G^+}) &>p(P_2,\mathcal{G^+}) \\
        p(P_1) &= P(P_2) = (x + y)/|\mathcal{G}|.
    \end{align}
    Consequently, according to PS2,
    \begin{equation}
        \label{eq:P2_PattSym}
        q(P_i,\mathcal{G}^+,\mathcal{G}^-) >
        q(P_j,\mathcal{G}^+,\mathcal{G}^-).
    \end{equation}
    The antecedent of (\ref{eq:Sym2Patterns}) is true for $P_1$ and $P_2$, but not its consequent. As a result, measure $q$ is not class-symmetric.
    
    Second, we turn to showing \textit{Class Symmetry} $\Rightarrow$ $\neg$P2. Let us assume that $q$ is a quality measure that satisfies the \textit{Class Symmetry} property. Then, given (\ref{eq:PiPj1}) and (\ref{eq:PiPj2}), we get 
    \begin{equation}
        \label{eq:Defsym}
        q(P_1,\mathcal{G^+},\mathcal{G^-}) = q(P_2,\mathcal{G^+},\mathcal{G^-}).
    \end{equation}
    The antecedent of P2 is true for $P_1$ and $P_2$, but not its consequent. Therefore, $q$ does not possess the PS2 property.
\end{proof}

\section{Additional Results Regarding Measure Comparison}
This appendix contains additional results related to the comparison of measures through their rankings, using Kendall's Tau.

\subsection{Separated Distribution Plots}
\label{sec:AppKTSeparate}
\begin{figure}[htbp!]
    \centering
    \begin{minipage}[b]{0.32\textwidth}
        \includegraphics[width=\textwidth]{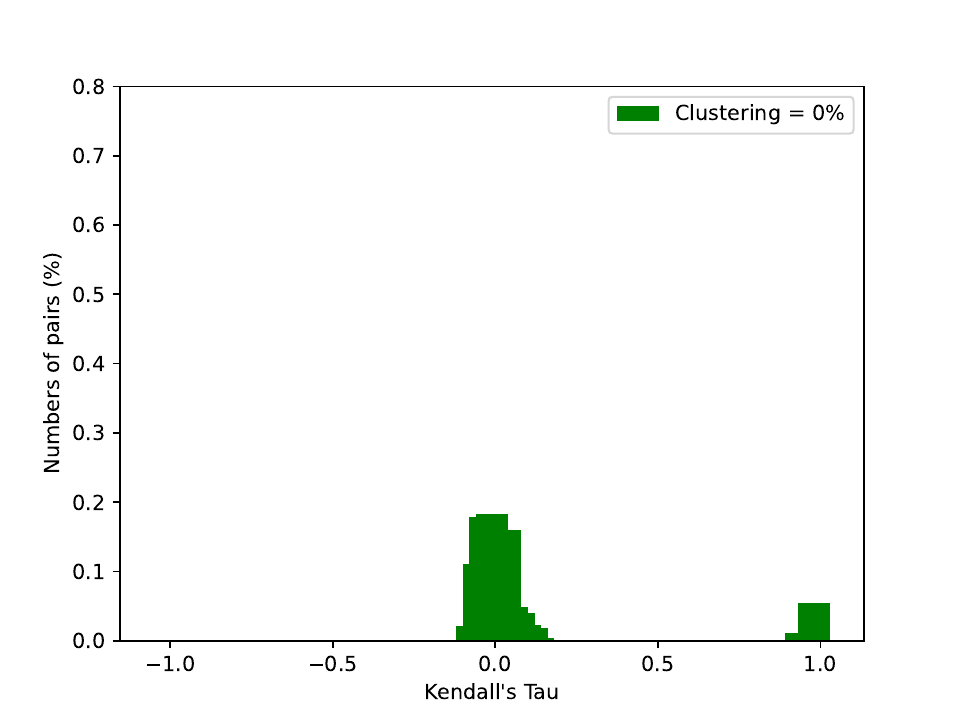}
        \subcaption{MUTAG}
        \label{fig:AppKTMUTAG_0}
    \end{minipage}
    \begin{minipage}[b]{0.32\textwidth}
        \includegraphics[width=\textwidth]{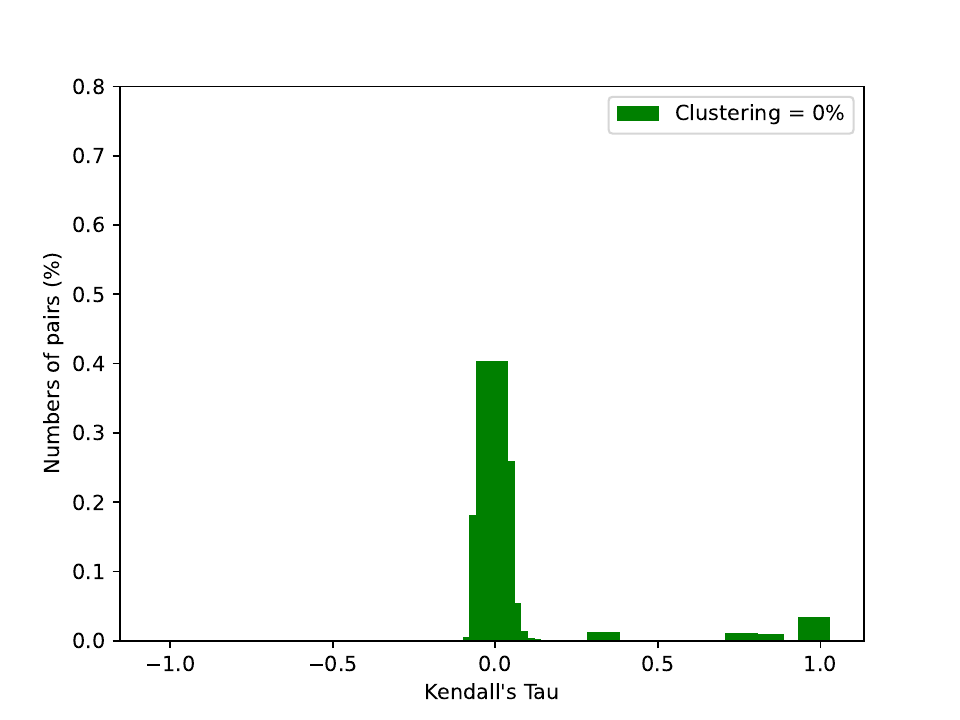}
        \subcaption{PTC}
        \label{fig:AppKTPTC_0}
    \end{minipage}
    \begin{minipage}[b]{0.32\textwidth}
        \includegraphics[width=\textwidth]{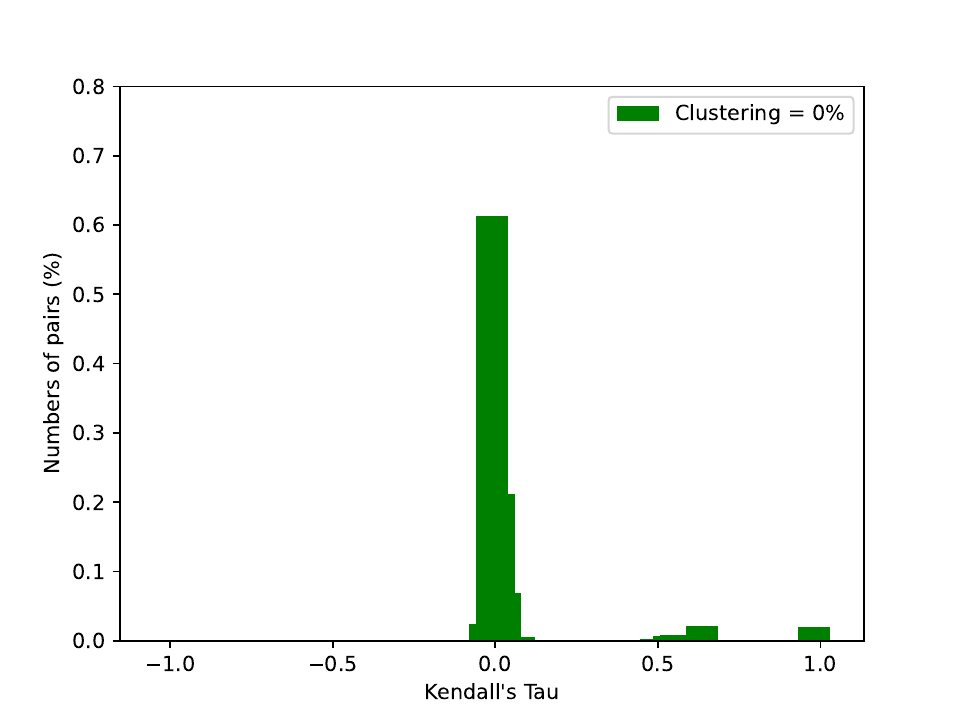}
        \subcaption{NCI1}
        \label{fig:AppKTNCI1}
    \end{minipage}
    
    \vspace{1em}
    
    \begin{minipage}[b]{0.32\textwidth}
        \includegraphics[width=\textwidth]{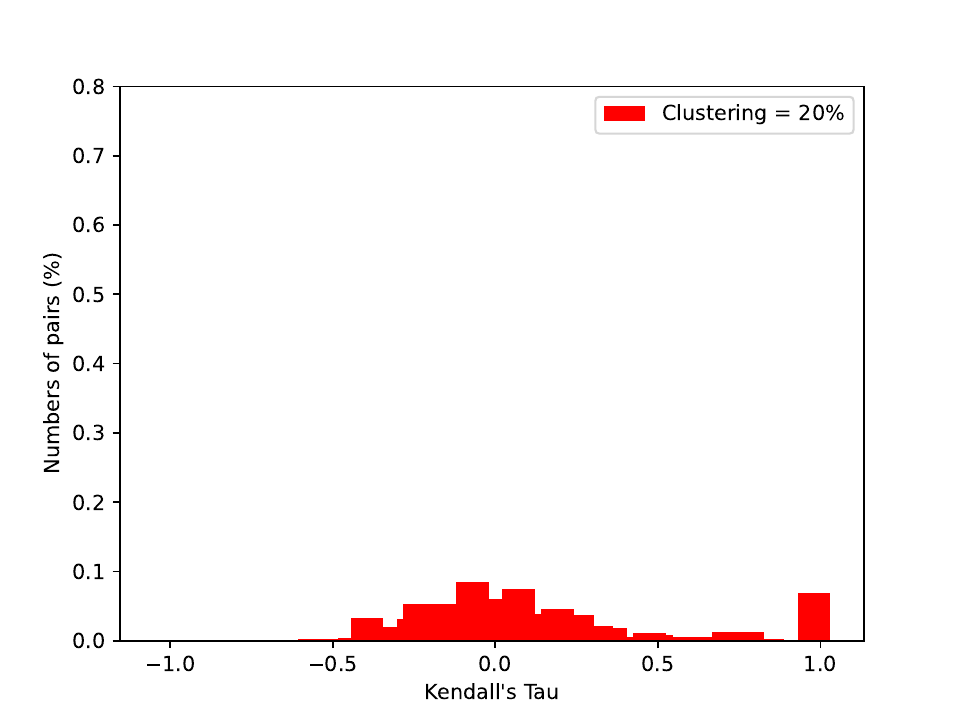}
        \subcaption{MUTAG}
        \label{fig:AppKTMUTAG_20}
    \end{minipage}
    \begin{minipage}[b]{0.32\textwidth}
        \includegraphics[width=\textwidth]{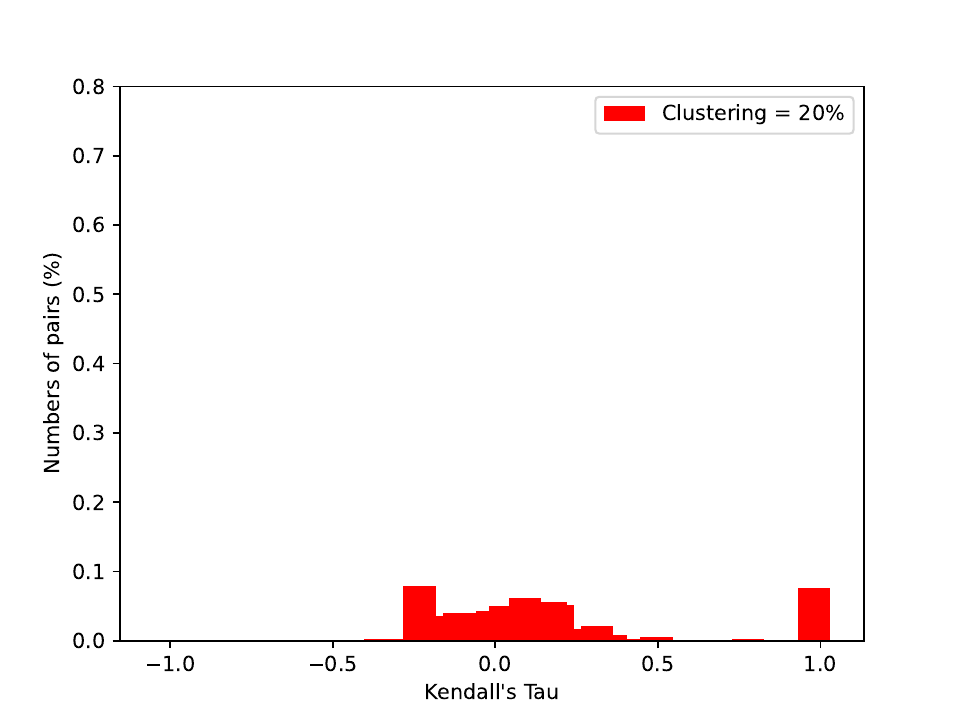}
        \subcaption{PTC}
        \label{fig:AppKTPTC_20}
    \end{minipage}
    \begin{minipage}[b]{0.32\textwidth}
        \includegraphics[width=\textwidth]{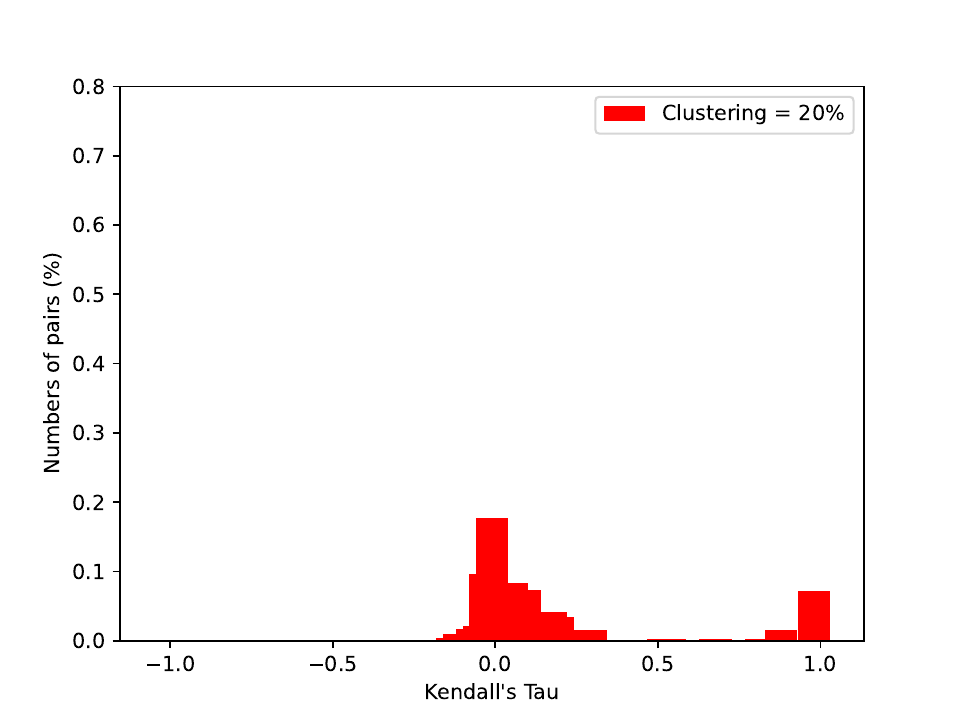}
        \subcaption{NCI1}
        \label{fig:AppKTNCI1_20}
    \end{minipage}

    \vspace{1em}
    
    \begin{minipage}[b]{0.32\textwidth}
        \includegraphics[width=\textwidth]{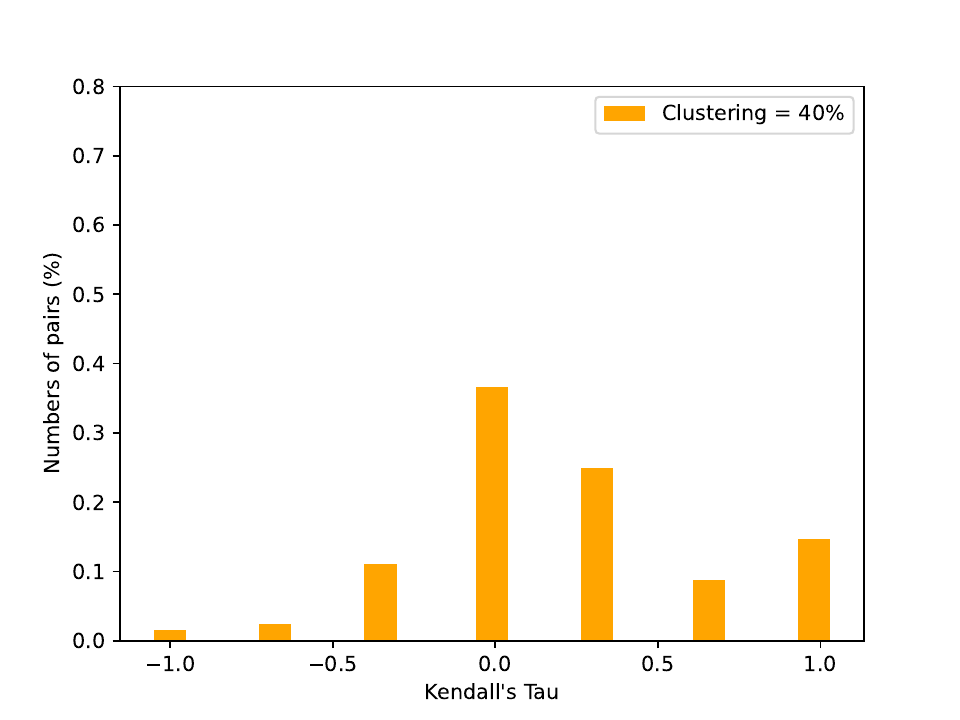}
        \subcaption{MUTAG}
        \label{fig:AppKTMUTAG_40}
    \end{minipage}
    \begin{minipage}[b]{0.32\textwidth}
        \includegraphics[width=\textwidth]{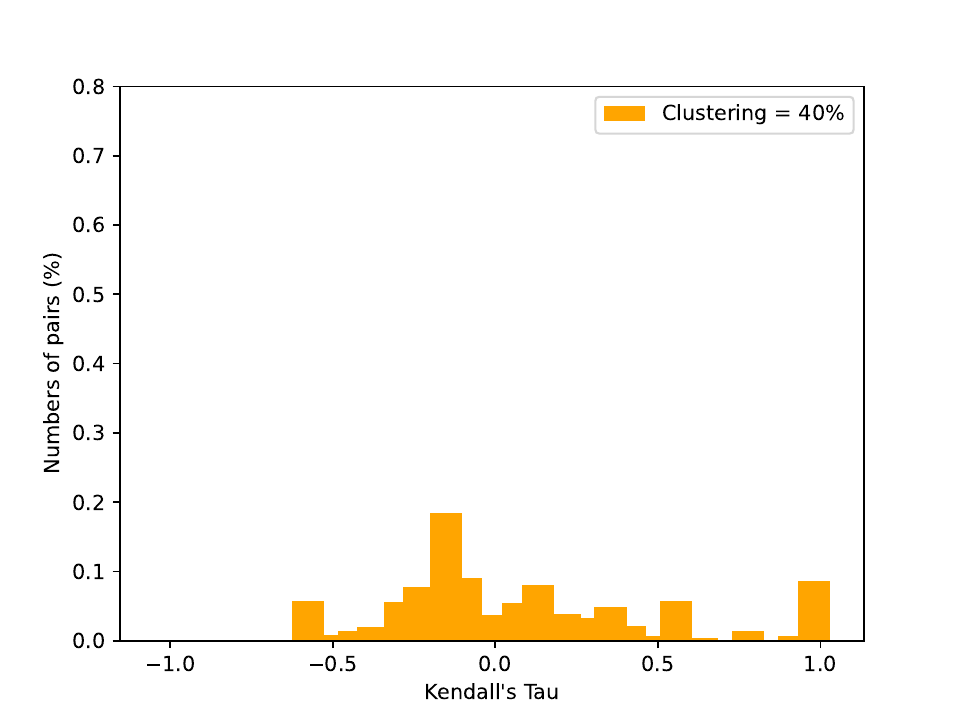}
        \subcaption{PTC}
        \label{fig:AppKTPTC_40}
    \end{minipage}
    \begin{minipage}[b]{0.32\textwidth}
        \includegraphics[width=\textwidth]{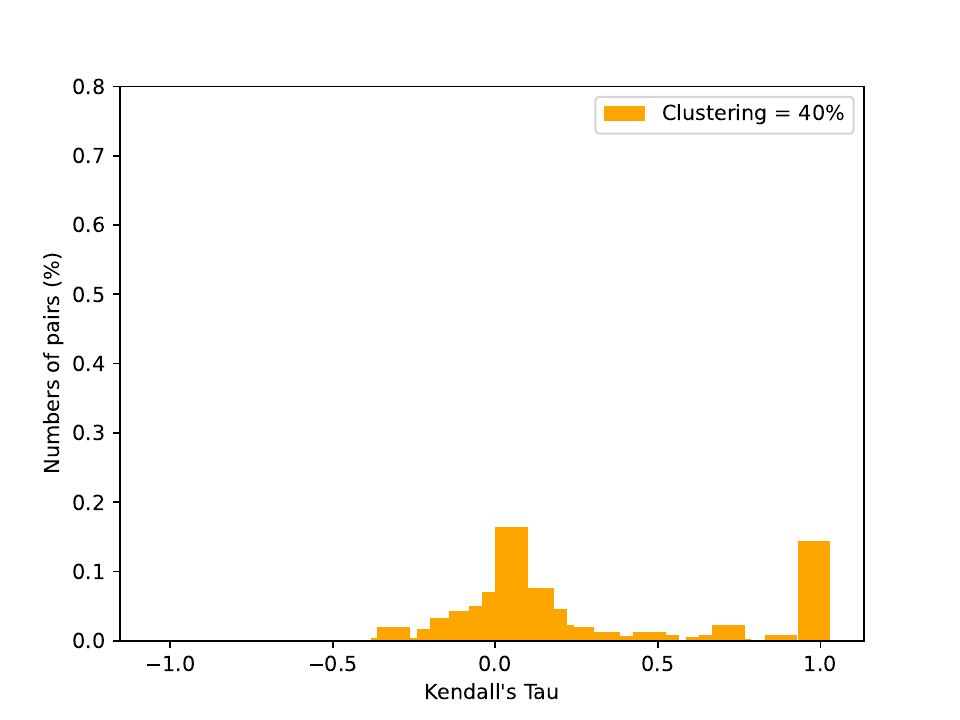}
        \subcaption{NCI1}
        \label{fig:AppKTNCI1_40}
    \end{minipage}

    \vspace{1em}
    
    \begin{minipage}[b]{0.32\textwidth}
        \includegraphics[width=\textwidth]{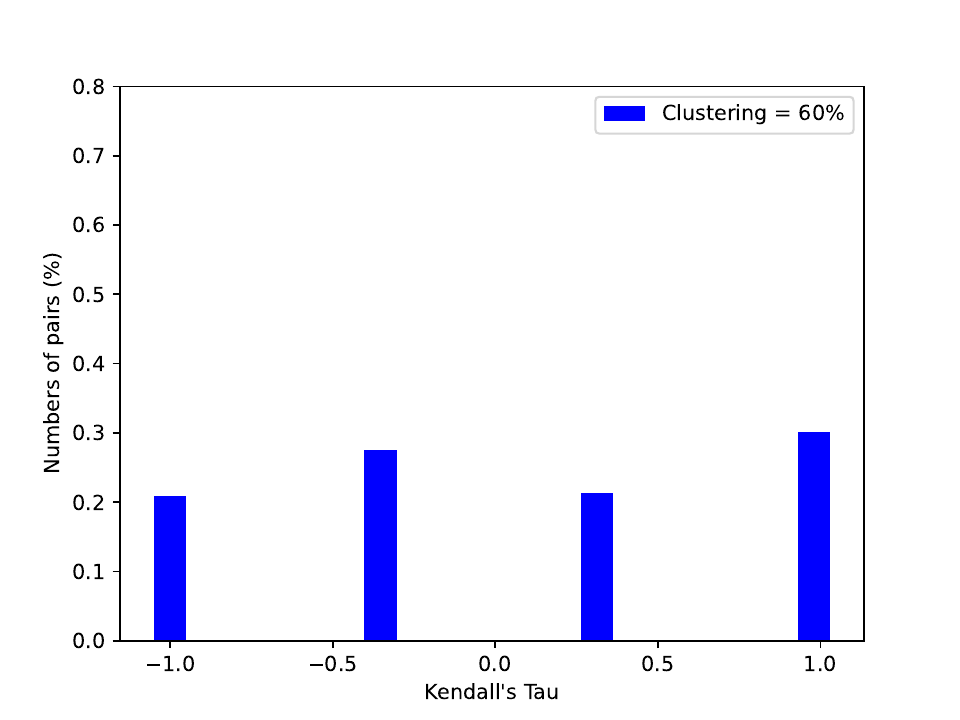}
        \subcaption{MUTAG}
        \label{fig:AppKTMUTAG_60}
    \end{minipage}
    \begin{minipage}[b]{0.32\textwidth}
        \includegraphics[width=\textwidth]{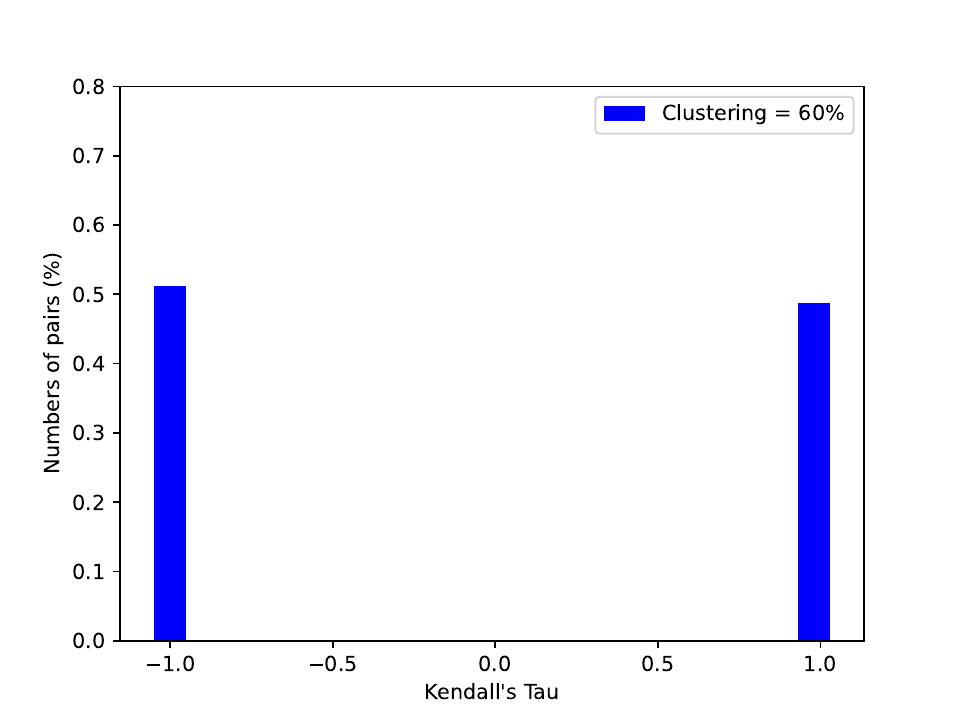}
        \subcaption{PTC}
        \label{fig:AppKTPTC_60}
    \end{minipage}
    \begin{minipage}[b]{0.32\textwidth}
        \includegraphics[width=\textwidth]{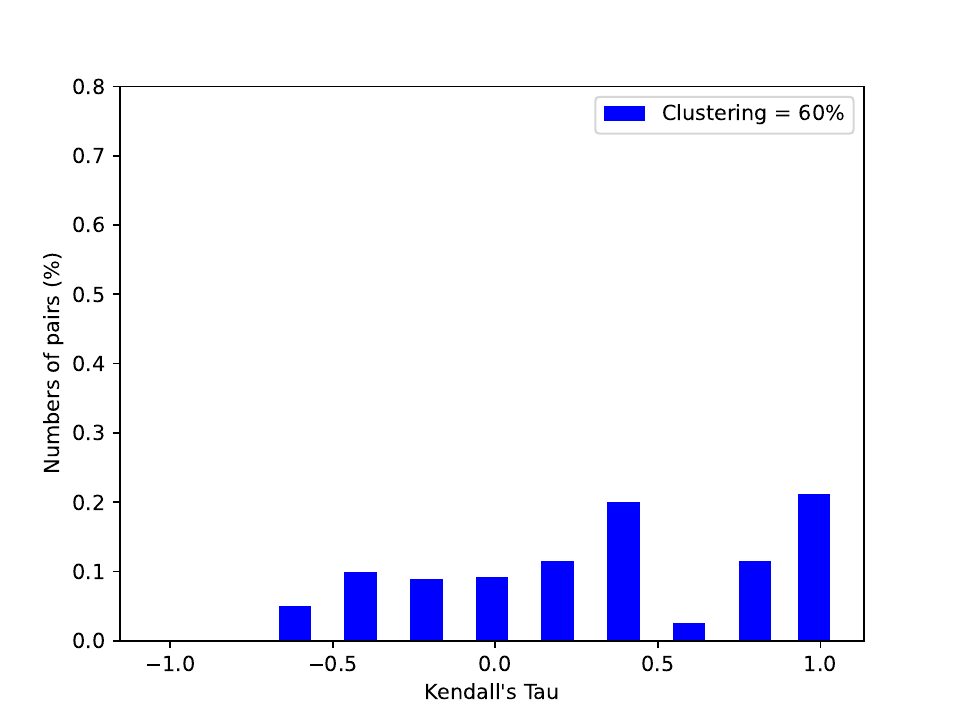}
        \subcaption{NCI1}
        \label{fig:AppKTNCI1_60}
    \end{minipage}

    \caption{Distribution of Kendall's Tau coefficient computed over all pairs of quality measure, for datasets MUTAG, PTC, and NCI1. Each row correspond to a different clustering threshold. The rest of the datasets are shown in Figure~\ref{fig:KtHisto2}. The top row of Figure~\ref{fig:CompaRBOKT} from the main paper shows a column-wise collapsed version of these plots.}
    \label{fig:KtHisto1}
    \Description{Description} 
\end{figure}

\begin{figure}[htbp!]
    \centering
    \begin{minipage}[b]{0.32\textwidth}
        \includegraphics[width=\textwidth]{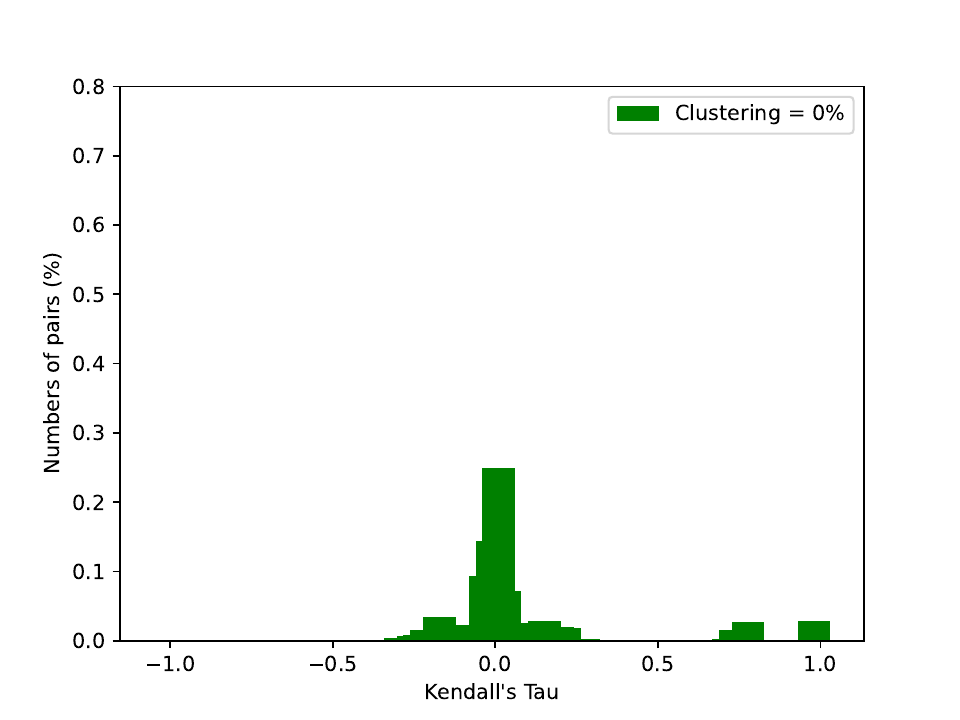}
        \subcaption{D\&D}
        \label{fig:AppKTDD_0}
    \end{minipage}
    \begin{minipage}[b]{0.32\textwidth}
        \includegraphics[width=\textwidth]{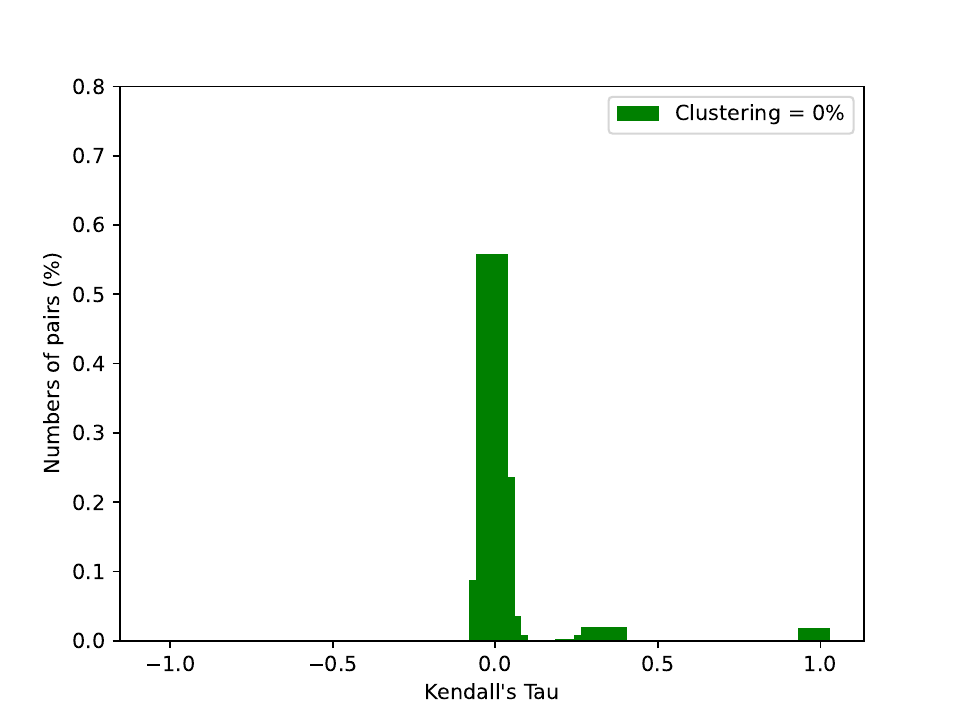}
        \subcaption{AIDS}
        \label{fig:AppKTAIDS_0}
    \end{minipage}
    \begin{minipage}[b]{0.32\textwidth}
        \includegraphics[width=\textwidth]{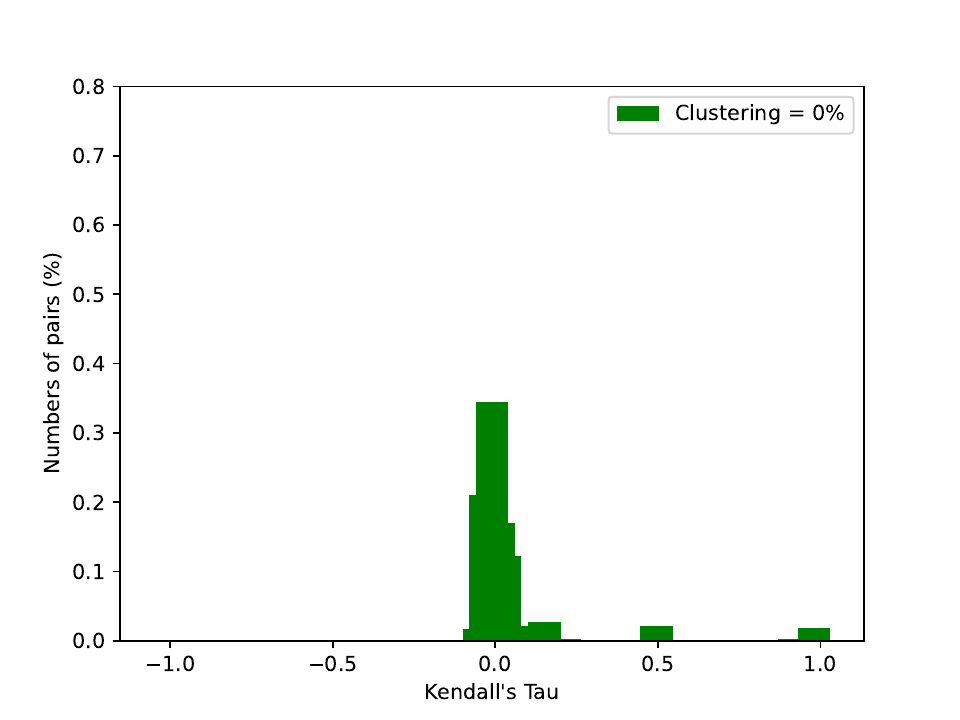}
        \subcaption{FOPPA}
        \label{fig:AppKTFOPPA_0}
    \end{minipage}
    
    \vspace{1em}
    
    \begin{minipage}[b]{0.32\textwidth}
        \includegraphics[width=\textwidth]{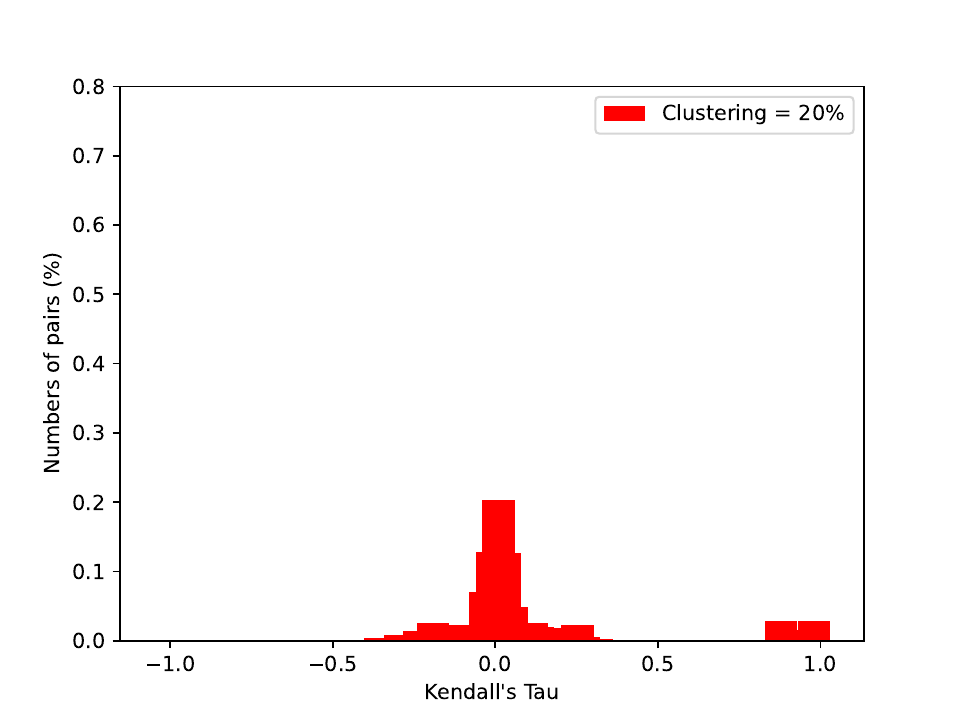}
        \subcaption{D\&D}
        \label{fig:AppKTDD_20}
    \end{minipage}
    \begin{minipage}[b]{0.32\textwidth}
        \includegraphics[width=\textwidth]{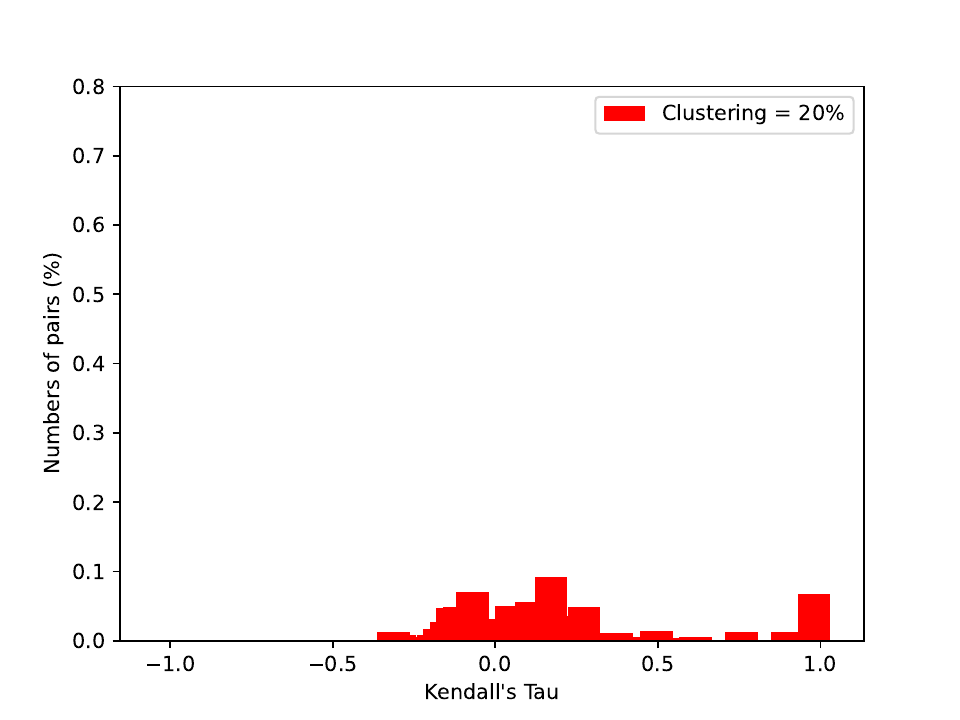}
        \subcaption{AIDS}
        \label{fig:AppKTAIDS_20}
    \end{minipage}
    \begin{minipage}[b]{0.32\textwidth}
        \includegraphics[width=\textwidth]{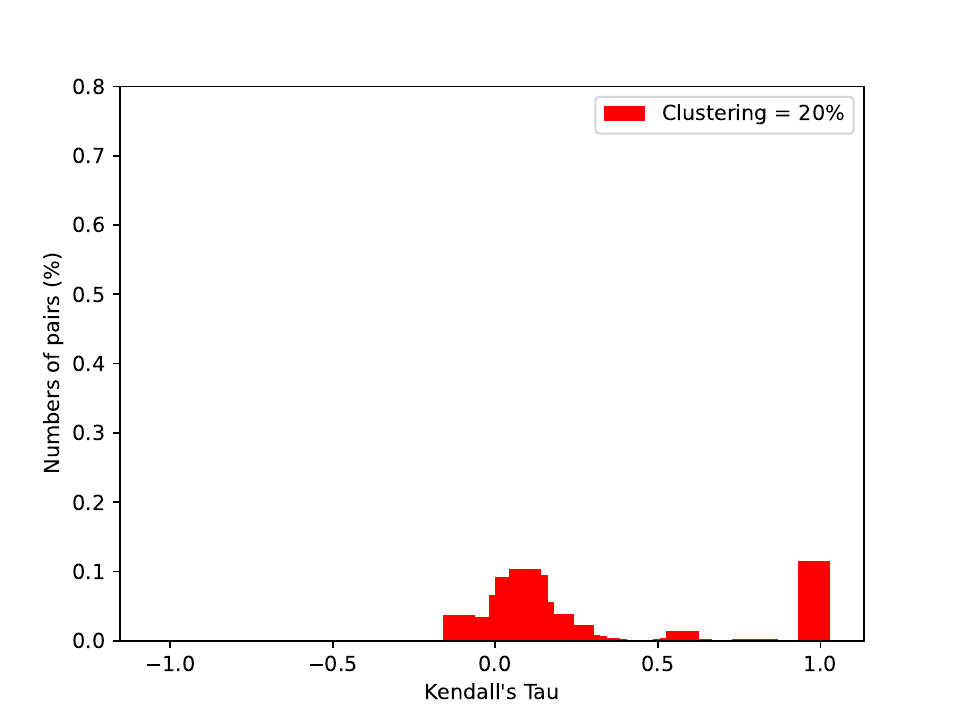}
        \subcaption{FOPPA}
        \label{fig:AppKTFOPPA_20}
    \end{minipage}

    \vspace{1em}
    
    \begin{minipage}[b]{0.32\textwidth}
        \includegraphics[width=\textwidth]{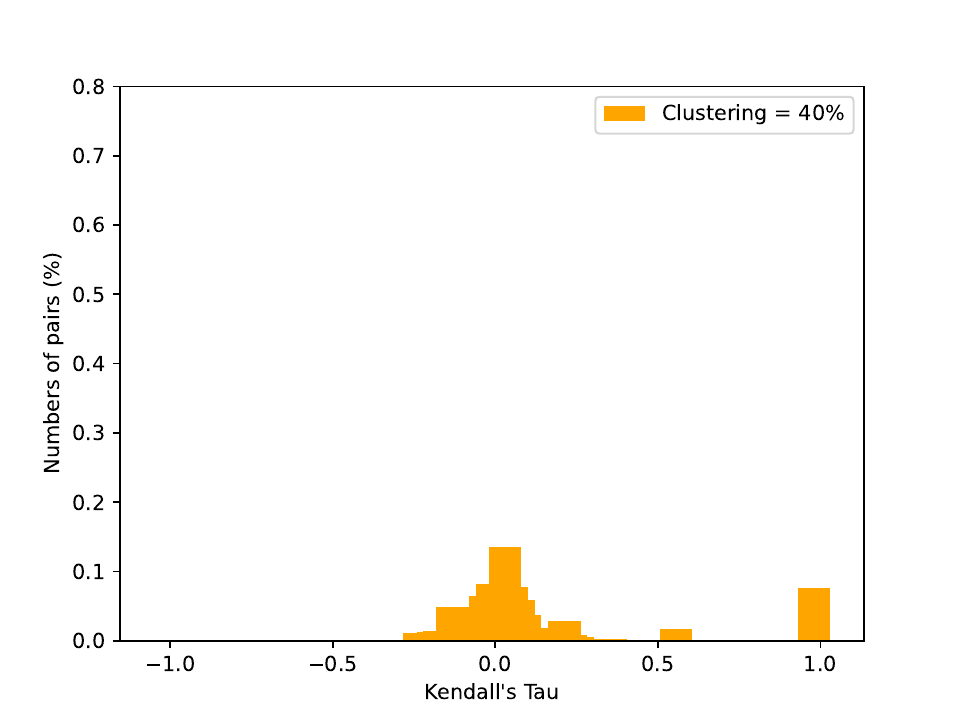}
        \subcaption{D\&D}
        \label{fig:AppKTDD_40}
    \end{minipage}
    \begin{minipage}[b]{0.32\textwidth}
        \includegraphics[width=\textwidth]{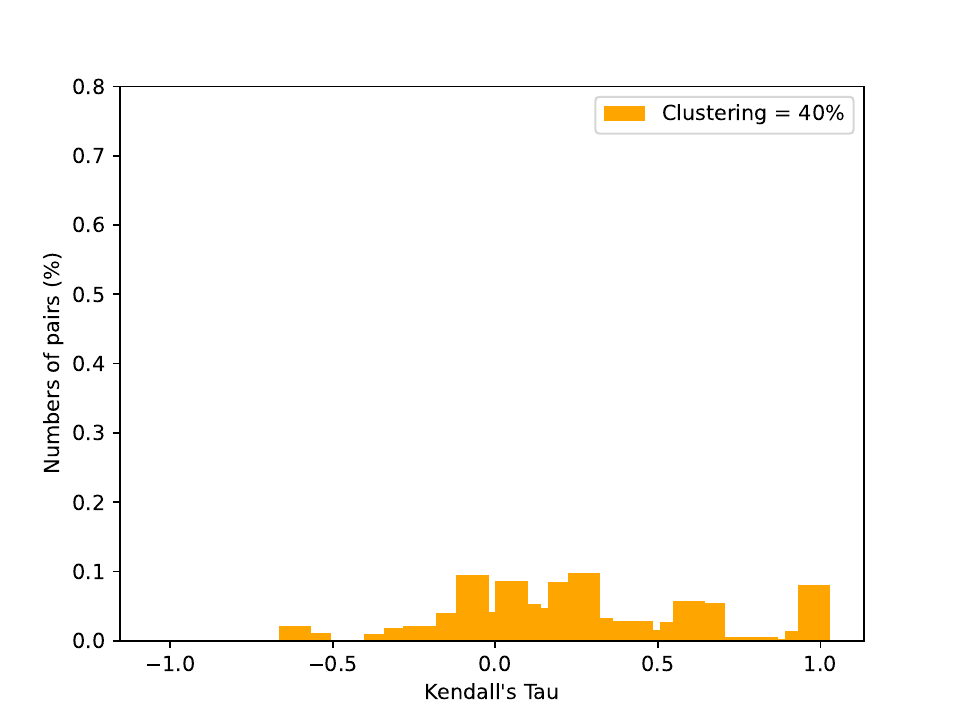}
        \subcaption{AIDS}
        \label{fig:AppKTAIDS_40}
    \end{minipage}
    \begin{minipage}[b]{0.32\textwidth}
        \includegraphics[width=\textwidth]{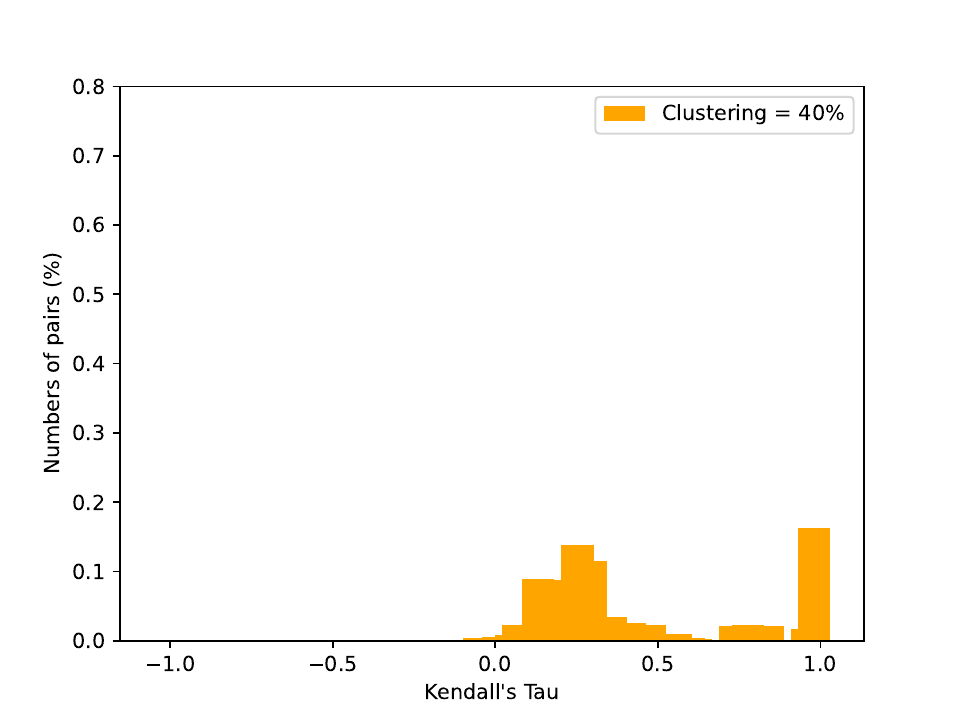}
        \subcaption{FOPPA}
        \label{fig:AppKTFOPPA_40}
    \end{minipage}

    \vspace{1em}
    
    \begin{minipage}[b]{0.32\textwidth}
        \includegraphics[width=\textwidth]{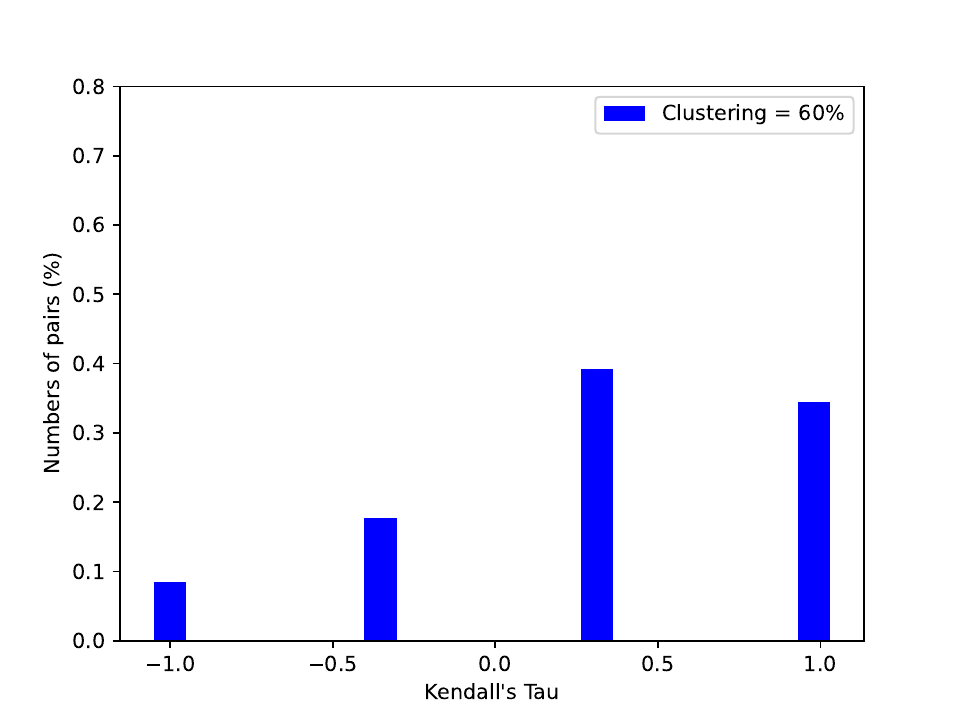}
        \subcaption{D\&D}
        \label{fig:AppKTDD_60}
    \end{minipage}
    \begin{minipage}[b]{0.32\textwidth}
        \includegraphics[width=\textwidth]{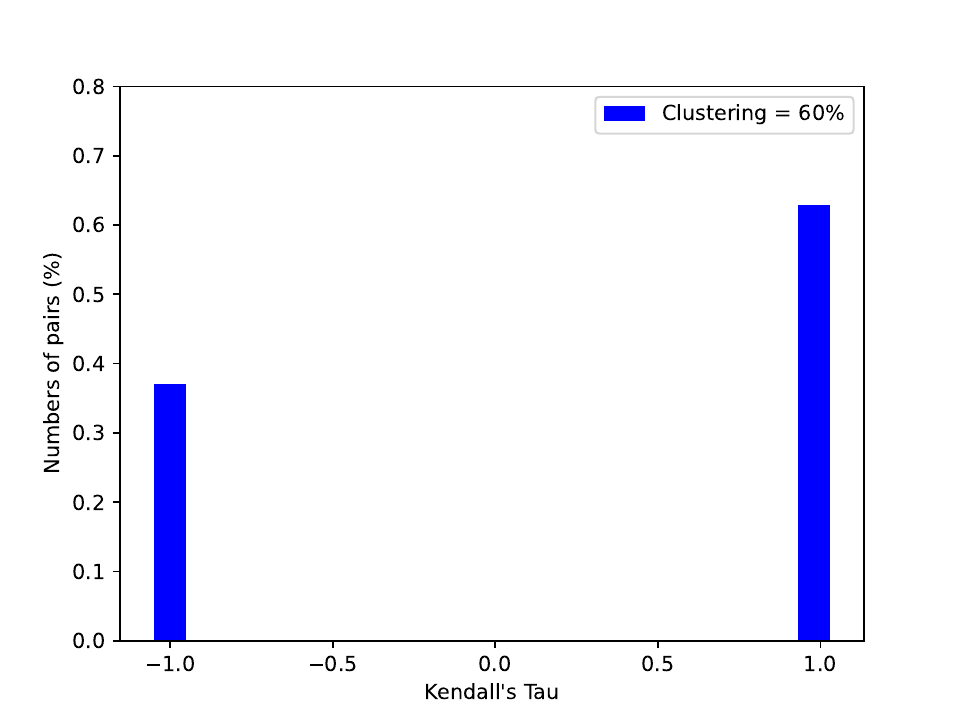}
        \subcaption{AIDS}
        \label{fig:AppKTAIDS_60}
    \end{minipage}
    \begin{minipage}[b]{0.32\textwidth}
        \includegraphics[width=\textwidth]{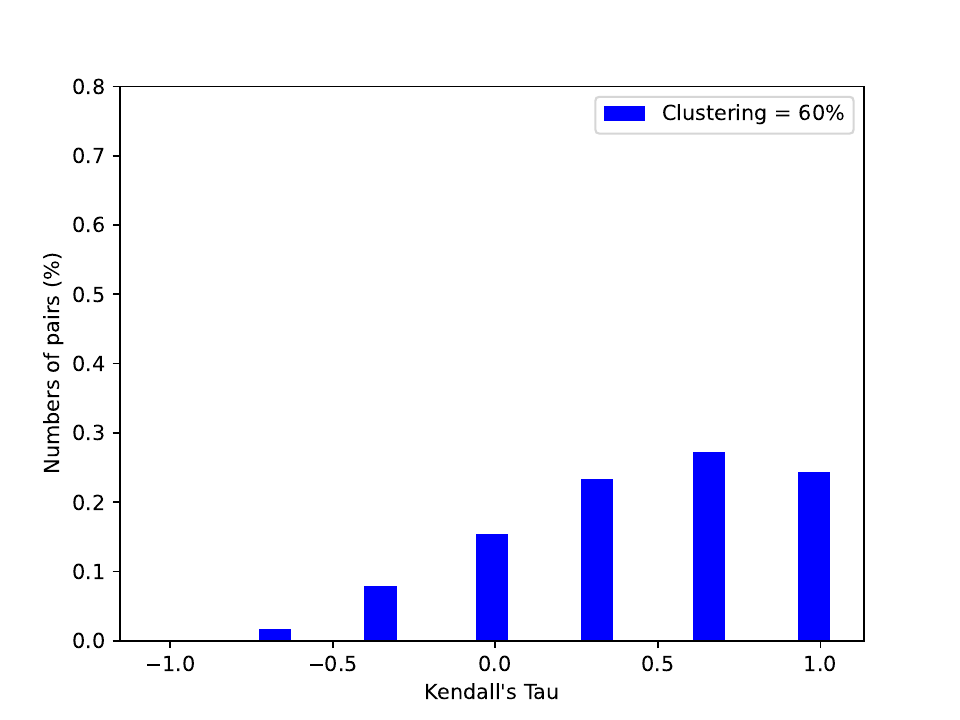}
        \subcaption{FOPPA}
        \label{fig:AppKTFOPPA_60}
    \end{minipage}

    \caption{Distribution of Kendall's Tau coefficient computed over all pairs of quality measure, for datasets D\&D, AIDS, and FOPPA. Each row correspond to a different clustering threshold. The rest of the datasets are shown in Figure~\ref{fig:KtHisto1}. The bottom row of Figure~\ref{fig:CompaRBOKT} from the main paper shows a column-wise collapsed version of these plots.}
    \label{fig:KtHisto2}
    \Description{Description} 
\end{figure}

Figures~\ref{fig:KtHisto1} (datasets  MUTAG, PTC, and NCI1) and~\ref{fig:KtHisto2} (datasets D\&D, AIDS, and FOPPA) represent the distribution of Kendall's Tau obtained for each dataset when comparing all \textcolor{highlightCol}{38} pairs of quality measures. They display the same information as Figure~\ref{fig:CompaRBOKT} from Section~\ref{sec:ExpClusteringRkComp}, except that each considered value of the clustering threshold (0, 20, 40 and 60 \%) is shown as a separate plot, instead of putting them all in the same plot.

\subsection{Dataset Correlation Matrices}
\label{sec:AppAllKT}
\begin{figure}[htbp!]
    \centering
    \begin{minipage}[b]{0.32\textwidth}
        \includegraphics[width=\textwidth]{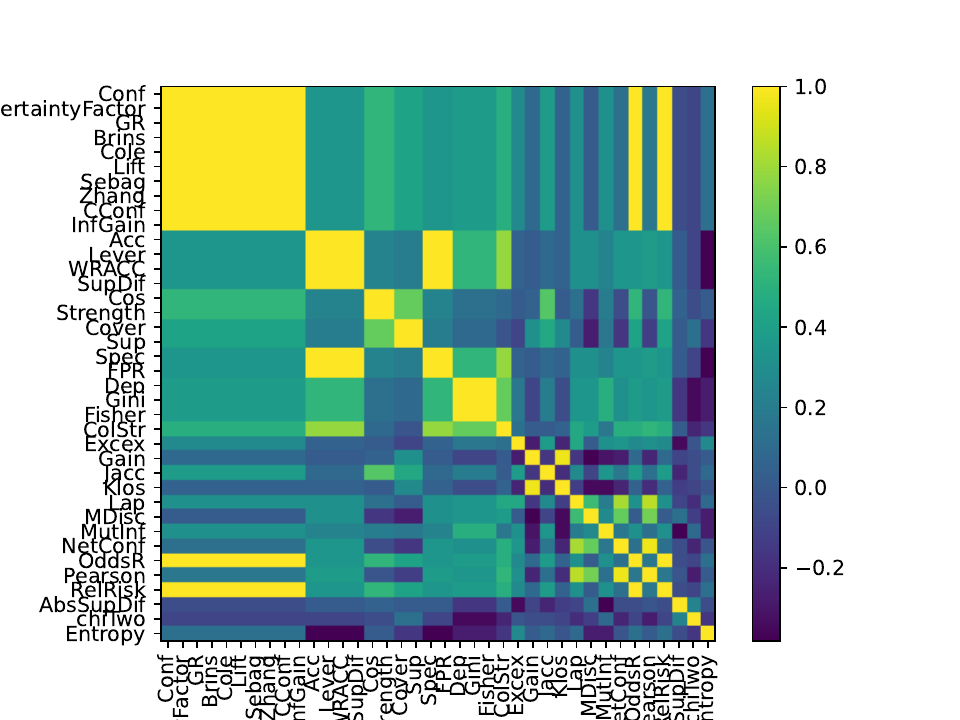}
        \subcaption{MUTAG}
        \label{fig:KTMUTAG}
    \end{minipage}
    \begin{minipage}[b]{0.32\textwidth}
        \includegraphics[width=\textwidth]{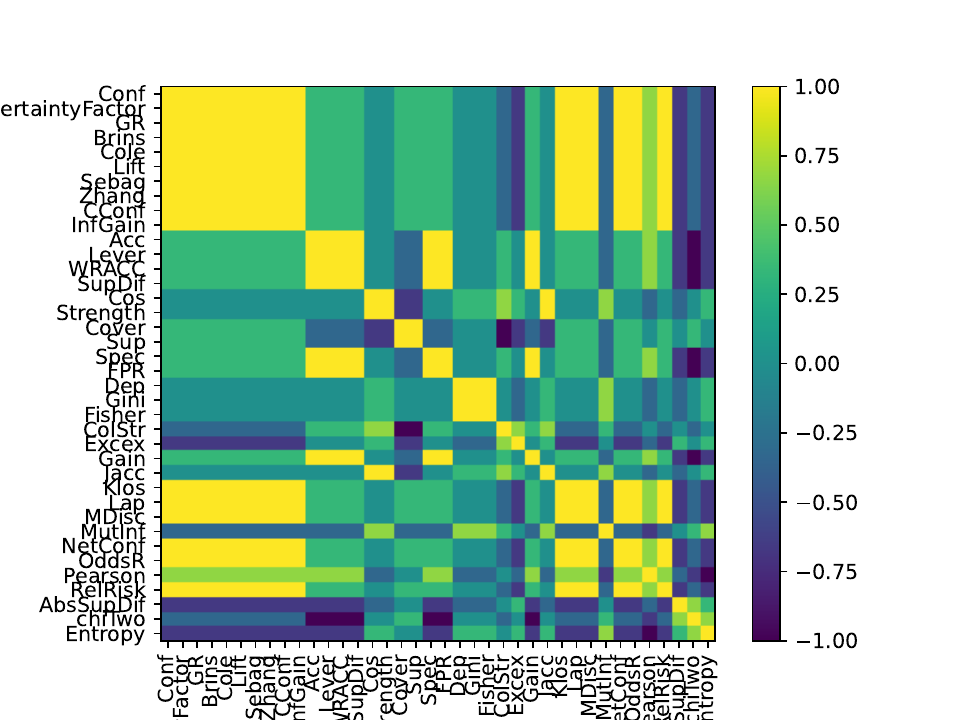}
        \subcaption{PTC}
        \label{fig:KTPTC}
    \end{minipage}
    \begin{minipage}[b]{0.32\textwidth}
        \includegraphics[width=\textwidth]{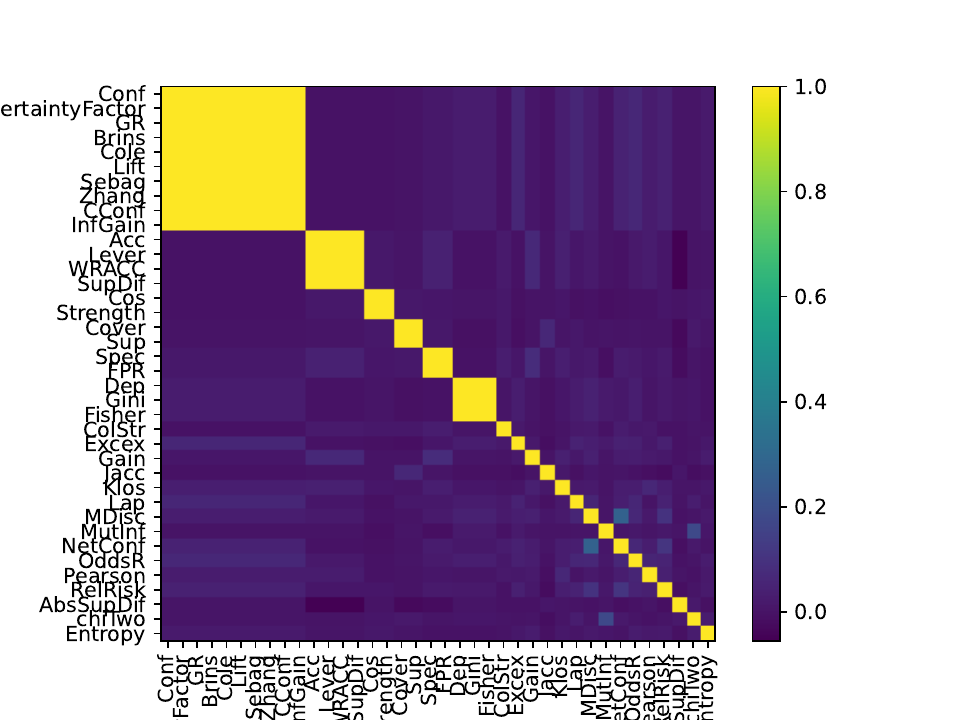}
        \subcaption{NCI1}
        \label{fig:KTNCI1}
    \end{minipage}
    
    \vspace{1em}
    
    \begin{minipage}[b]{0.32\textwidth}
        \includegraphics[width=\textwidth]{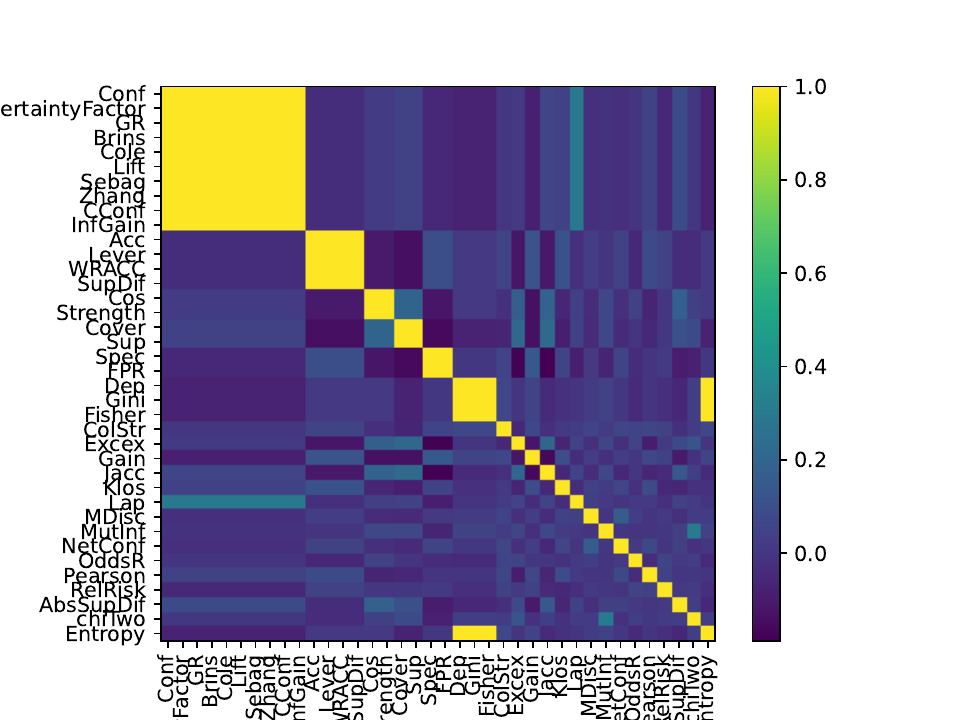}
        \subcaption{D\&D}
        \label{fig:KTDD}
    \end{minipage}
    \begin{minipage}[b]{0.32\textwidth}
        \includegraphics[width=\textwidth]{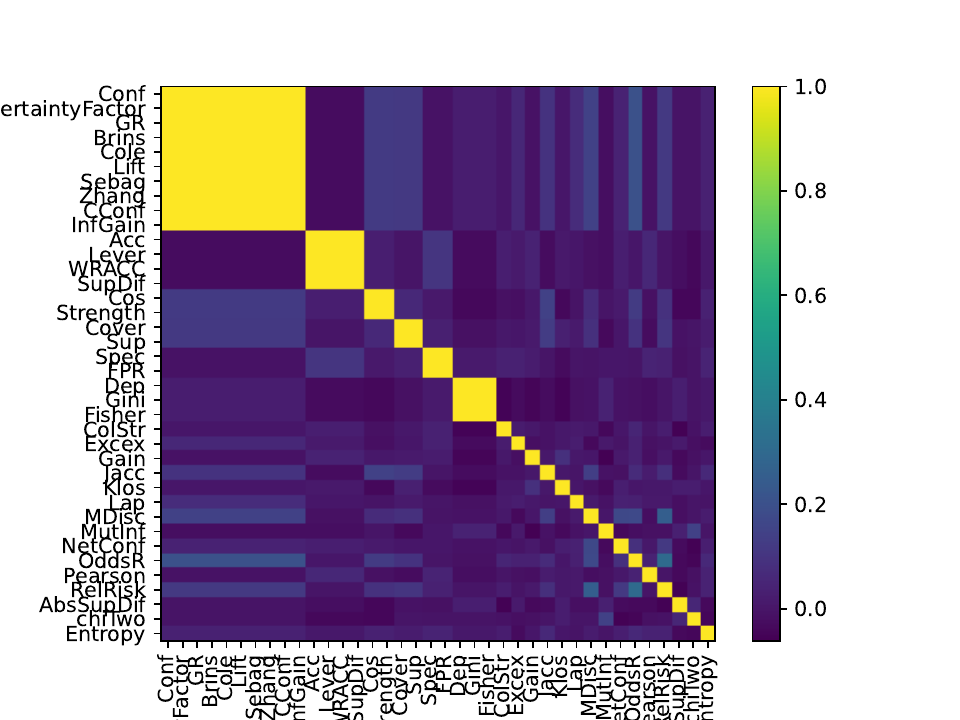}
        \subcaption{AIDS}
        \label{fig:KTAIDS}
    \end{minipage}
    \begin{minipage}[b]{0.32\textwidth}
        \includegraphics[width=\textwidth]{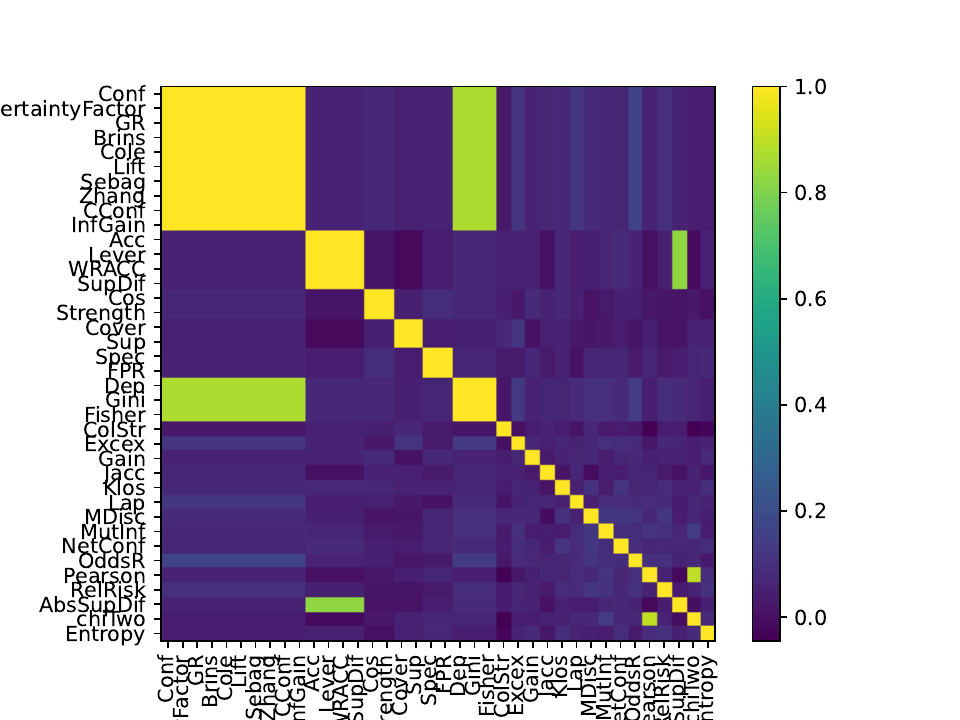}
        \subcaption{FOPPA}
        \label{fig:KTFOPPA}
    \end{minipage}
    
    \caption{Kendall's Tau for each pair of quality measures, shown separately for each dataset. Note that the color scale is not fixed over the plots, to improve contrast.}
    \label{fig:KTALLDataset}
    \Description{Description} 
\end{figure}

Figure~\ref{fig:KTALLDataset} represents Kendall's Tau computed for each pair of the \textcolor{highlightCol}{38} quality measures. The differences with Figure~\ref{sec:ExpPairwise} from Section~\ref{fig:CompaScoresMinMatri} is that instead of showing the minimal correlation value over all datasets, this figure contains a distinct plot for each dataset.

\section{Additional Results Related to Gold Standard Comparison}
This appendix provides plots comparing all measures to the gold standard, in terms of ranking correlation (Appendix~\ref{sec:AppFullRBO}) and classification performance (Appendix~\ref{sec:AppFullF1}). By comparison, the plots provided in the main article only focus on eight of these measures, for the sake of concision.

\subsection{Ranking Comparison}
\label{sec:AppFullRBO}
\begin{figure}[htbp!]
    \centering
    \begin{minipage}[b]{0.32\textwidth}
        \includegraphics[width=\textwidth]{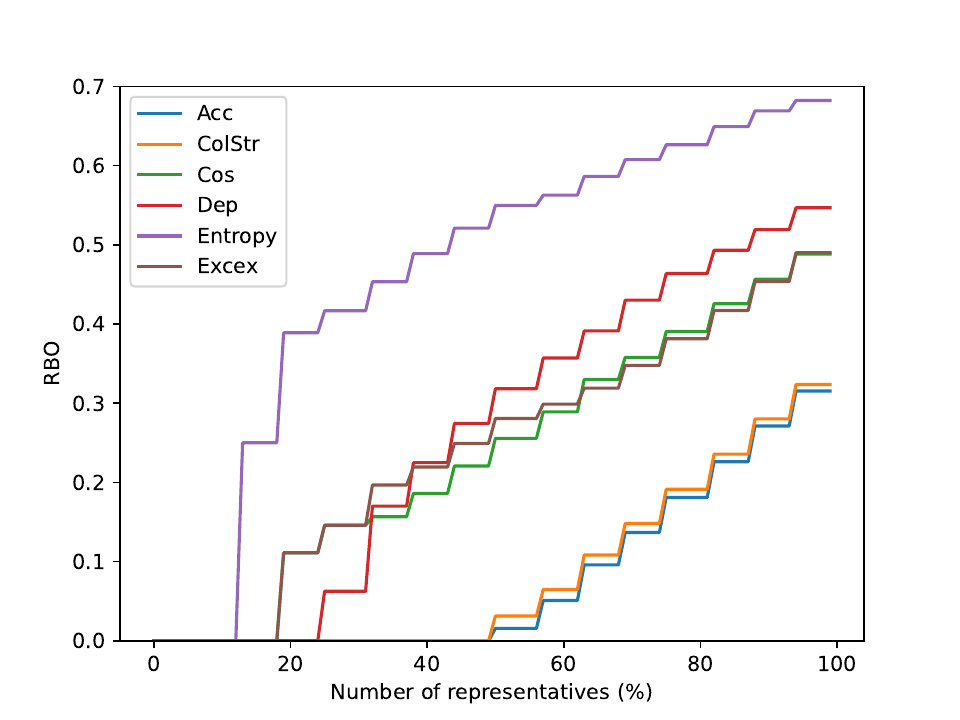}
        \subcaption{MUTAG}
        \label{fig:AppRBOMUTAG_1}
    \end{minipage}
    \begin{minipage}[b]{0.32\textwidth}
        \includegraphics[width=\textwidth]{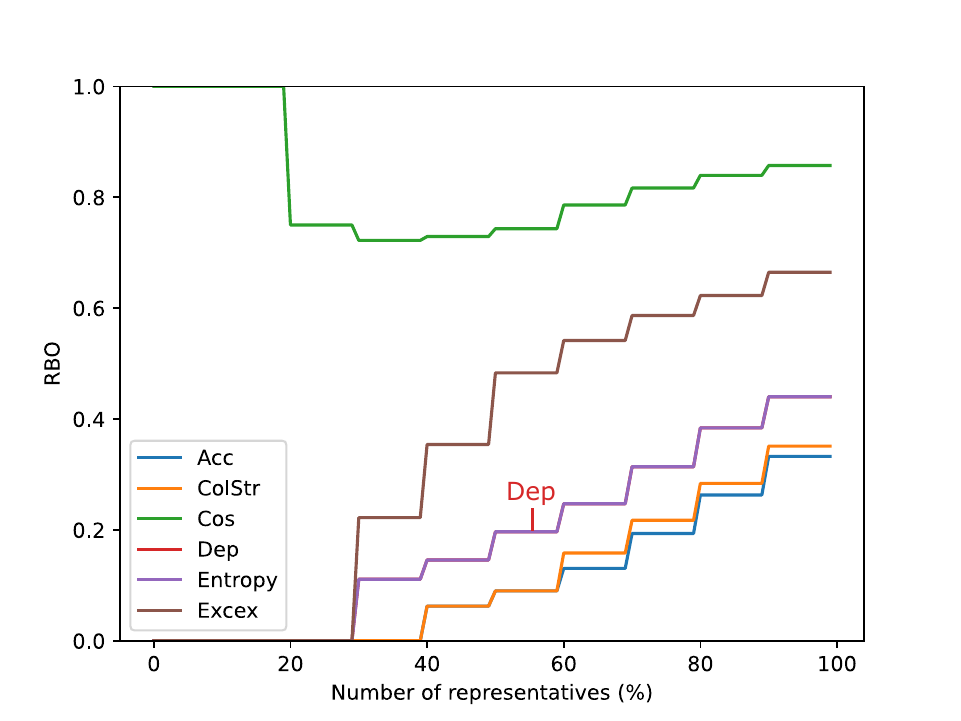}
        \subcaption{PTC}
        \label{fig:AppRBOPTC_1}
    \end{minipage}
    \begin{minipage}[b]{0.32\textwidth}
        \includegraphics[width=\textwidth]{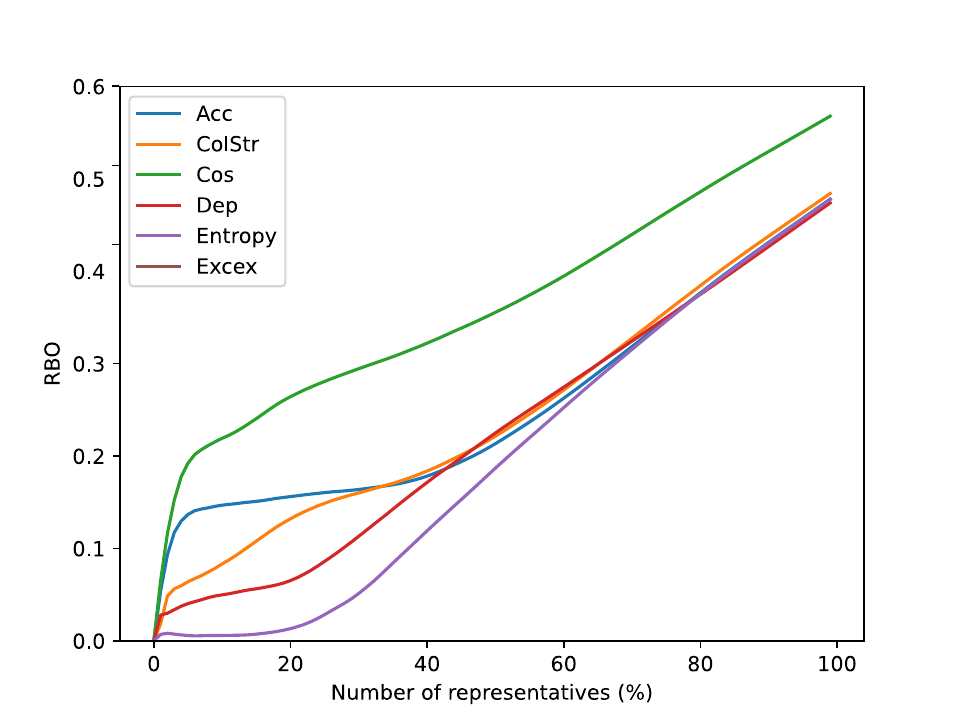}
        \subcaption{NCI1}
        \label{fig:AppRBONCI1_1}
    \end{minipage}
    
    \vspace{1em}
    
    \begin{minipage}[b]{0.32\textwidth}
        \includegraphics[width=\textwidth]{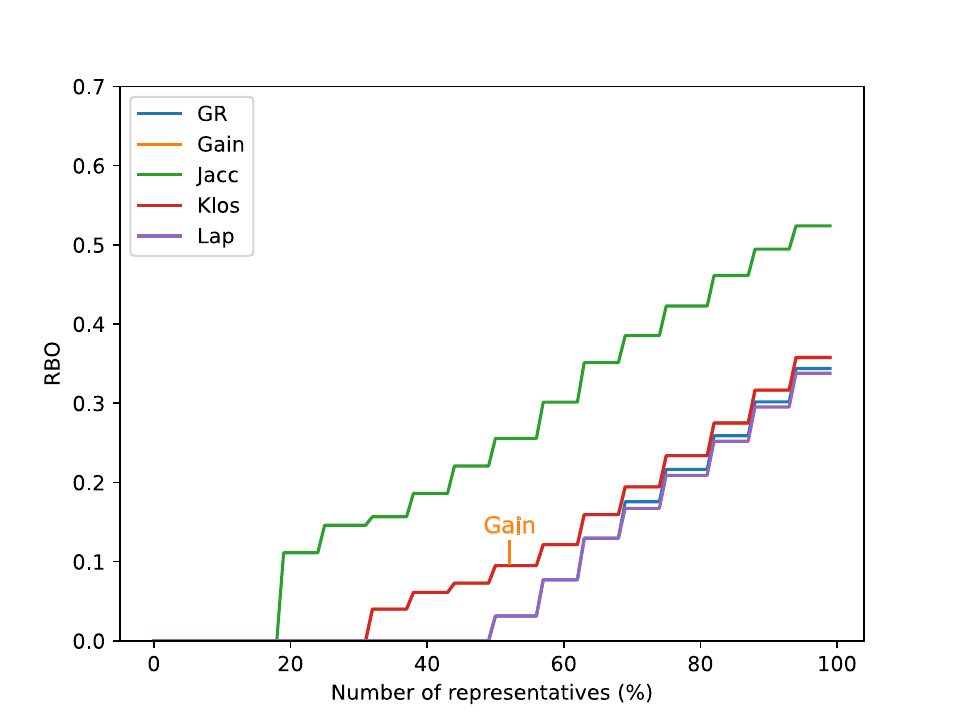}
        \subcaption{MUTAG}
        \label{fig:AppRBOMUTAG_2}
    \end{minipage}
    \begin{minipage}[b]{0.32\textwidth}
        \includegraphics[width=\textwidth]{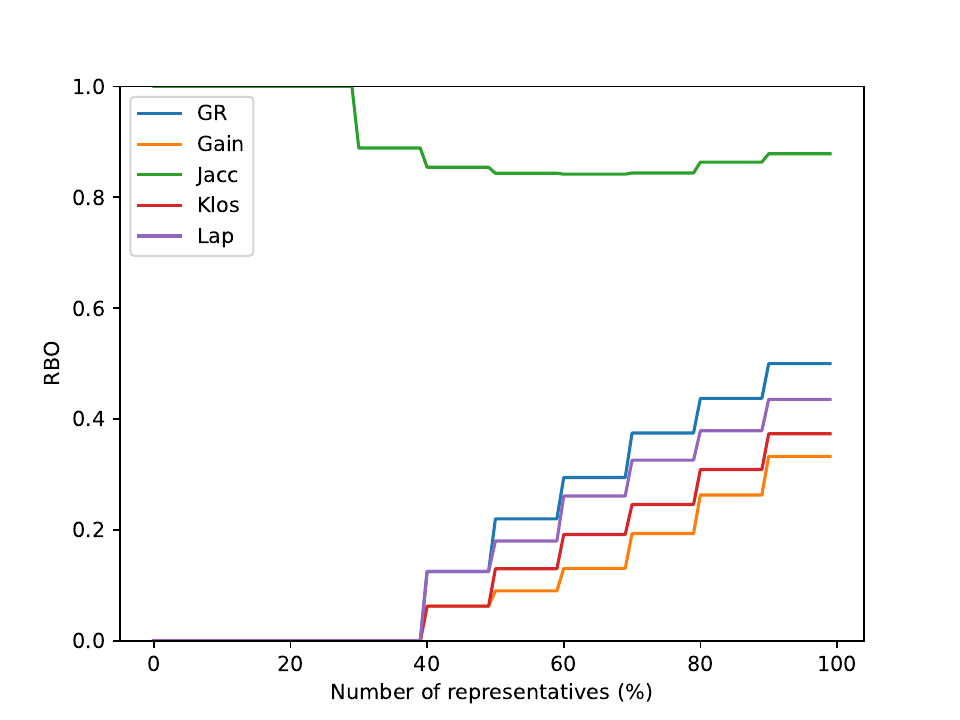}
        \subcaption{PTC}
        \label{fig:AppRBOPTC_2}
    \end{minipage}
    \begin{minipage}[b]{0.32\textwidth}
        \includegraphics[width=\textwidth]{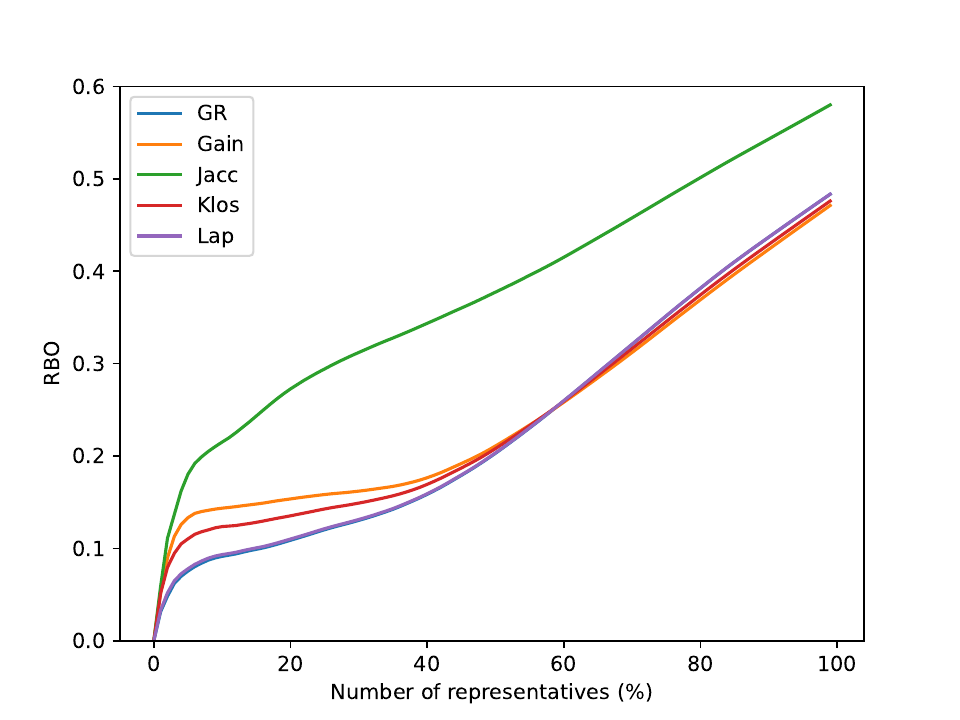}
        \subcaption{NCI1}
        \label{fig:AppRBONCI1_2}
    \end{minipage}

    \vspace{1em}
    
    \begin{minipage}[b]{0.32\textwidth}
        \includegraphics[width=\textwidth]{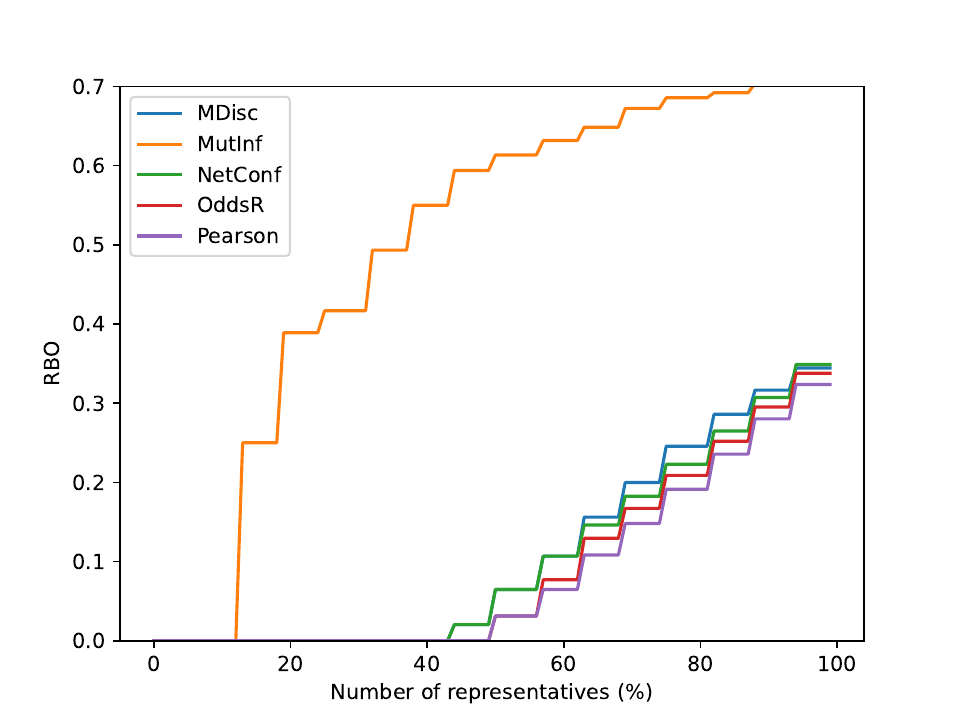}
        \subcaption{MUTAG}
        \label{fig:AppRBOMUTAG_3}
    \end{minipage}
    \begin{minipage}[b]{0.32\textwidth}
        \includegraphics[width=\textwidth]{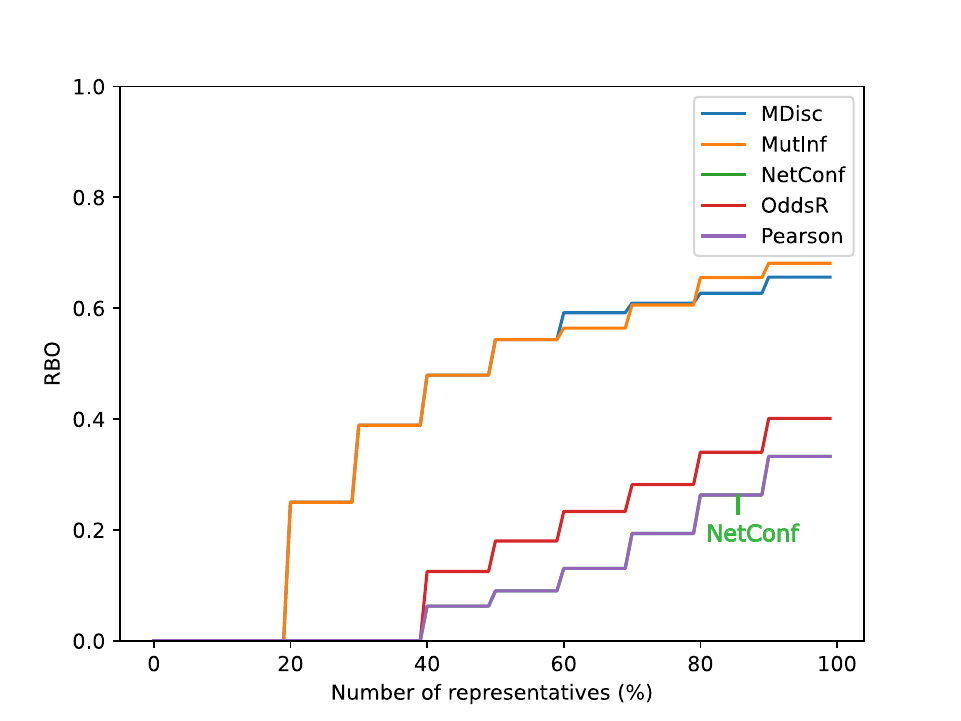}
        \subcaption{PTC}
        \label{fig:AppRBOPTC_3}
    \end{minipage}
    \begin{minipage}[b]{0.32\textwidth}
        \includegraphics[width=\textwidth]{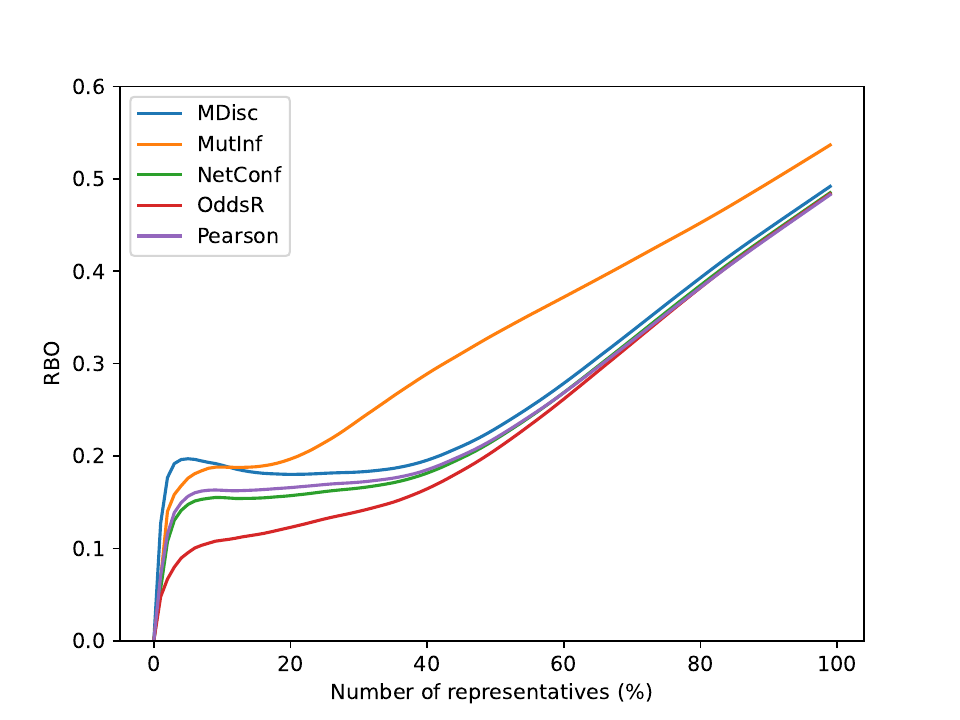}
        \subcaption{NCI1}
        \label{fig:AppRBONCI1_3}
    \end{minipage}

    \vspace{1em}
    
    \begin{minipage}[b]{0.32\textwidth}
        \includegraphics[width=\textwidth]{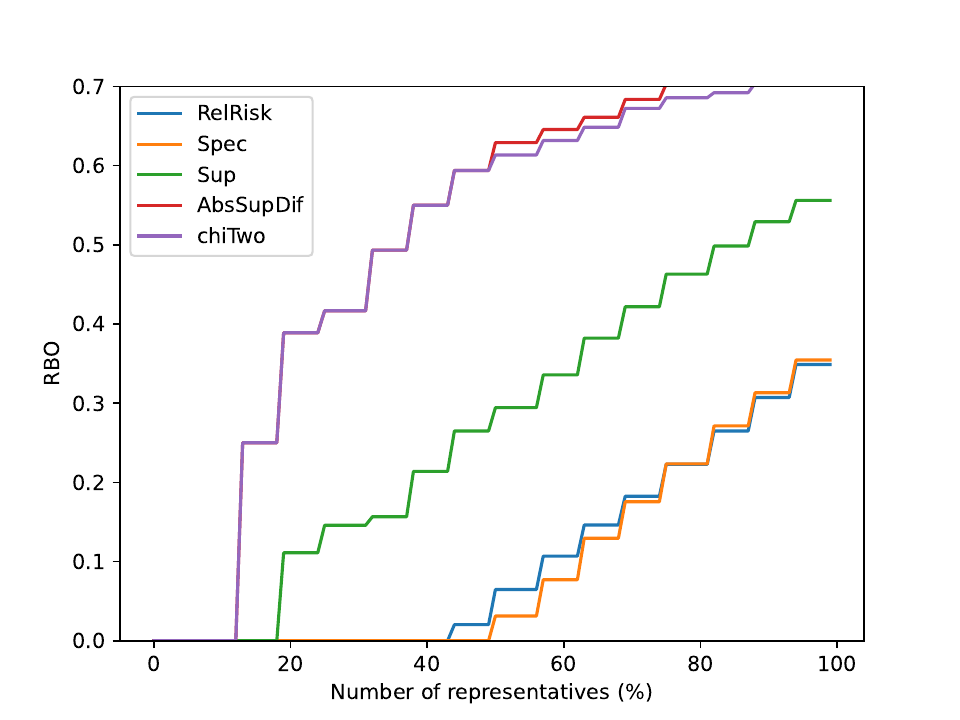}
        \subcaption{MUTAG}
        \label{fig:AppRBOMUTAG_4}
    \end{minipage}
    \begin{minipage}[b]{0.32\textwidth}
        \includegraphics[width=\textwidth]{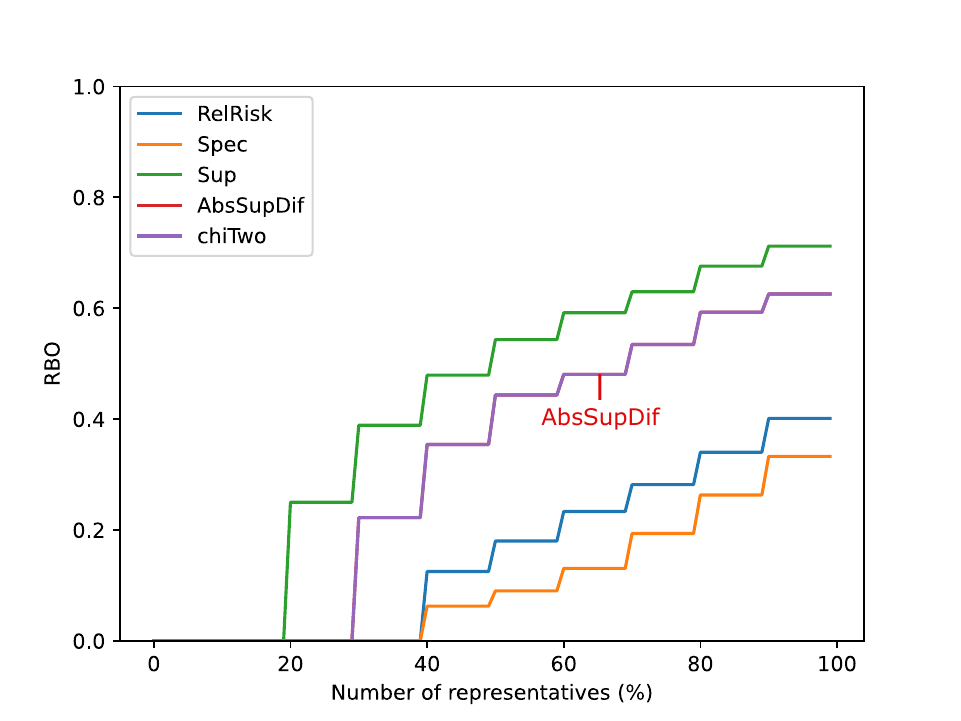}
        \subcaption{PTC}
        \label{fig:AppRBOPTC_4}
    \end{minipage}
    \begin{minipage}[b]{0.32\textwidth}
        \includegraphics[width=\textwidth]{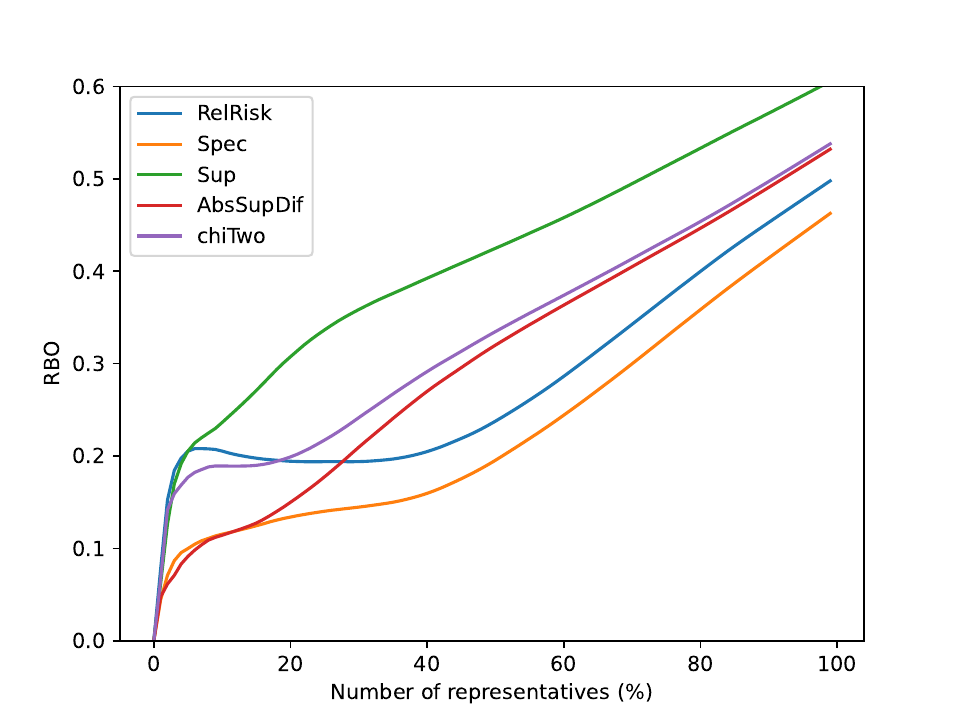}
        \subcaption{NCI1}
        \label{fig:AppRBONCI1_4}
    \end{minipage}
    \caption{RBO between the rankings obtained for each selected quality measure and the gold standard, as a function of $s$, the number of top representatives considered for datasets MUTAG, PTC and NCI1. The rest of the datasets are shown in Figure~\ref{fig:ShapleyRBO_2}.}
    \label{fig:ShapleyRBO_1}
    \Description{Description} 
\end{figure}

\begin{figure}[htbp!]
    \centering
    \begin{minipage}[b]{0.32\textwidth}
        \includegraphics[width=\textwidth]{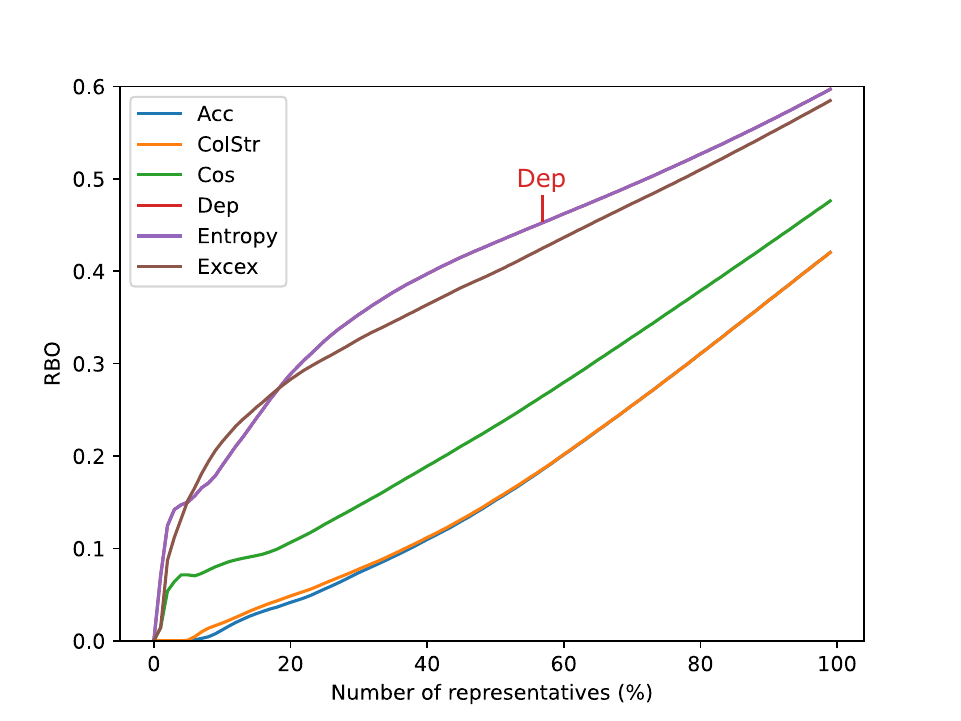}
        \subcaption{D\&D}
        \label{fig:AppRBODD_1}
    \end{minipage}
    \begin{minipage}[b]{0.32\textwidth}
        \includegraphics[width=\textwidth]{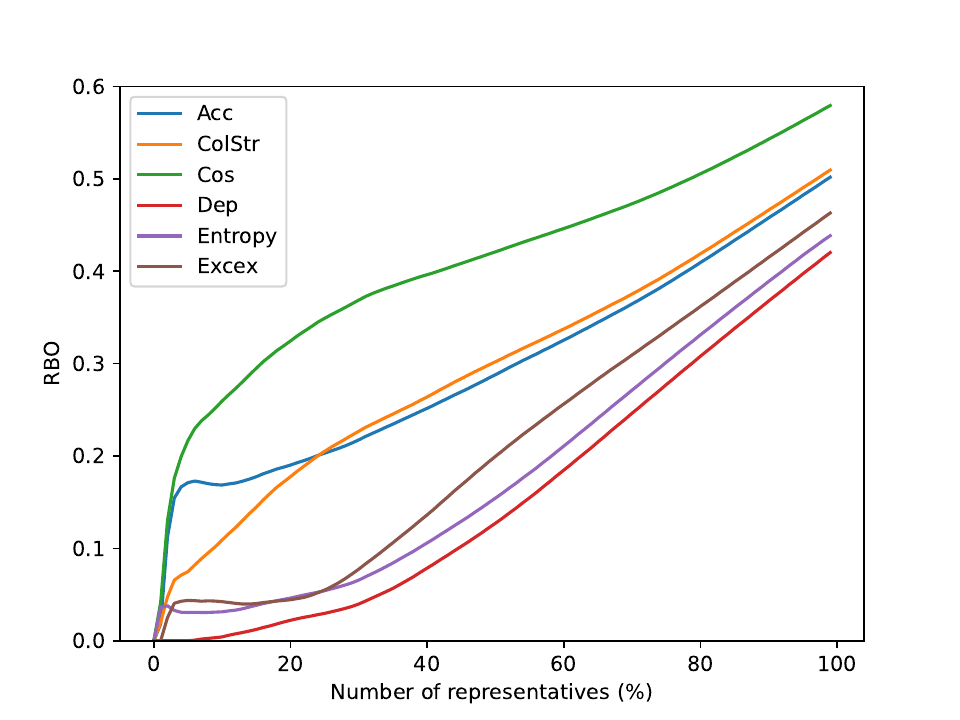}
        \subcaption{AIDS}
        \label{fig:AppRBOAIDS_1}
    \end{minipage}
    \begin{minipage}[b]{0.32\textwidth}
        \includegraphics[width=\textwidth]{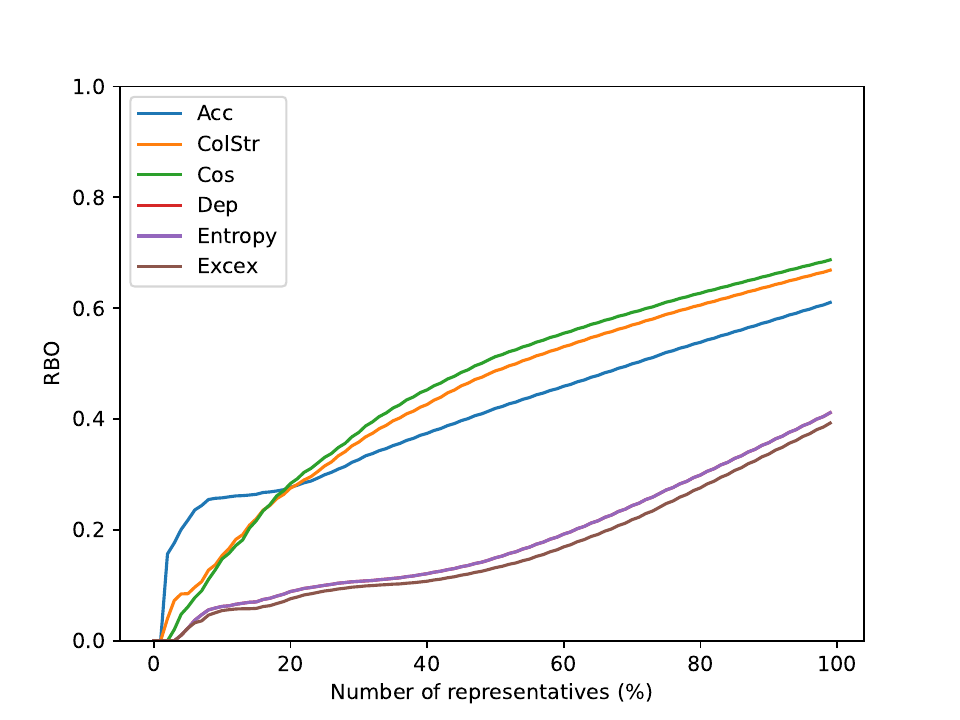}
        \subcaption{FOPPA}
        \label{fig:AppRBOFOPPA_1}
    \end{minipage}
    
    \vspace{1em}
    
    \begin{minipage}[b]{0.32\textwidth}
        \includegraphics[width=\textwidth]{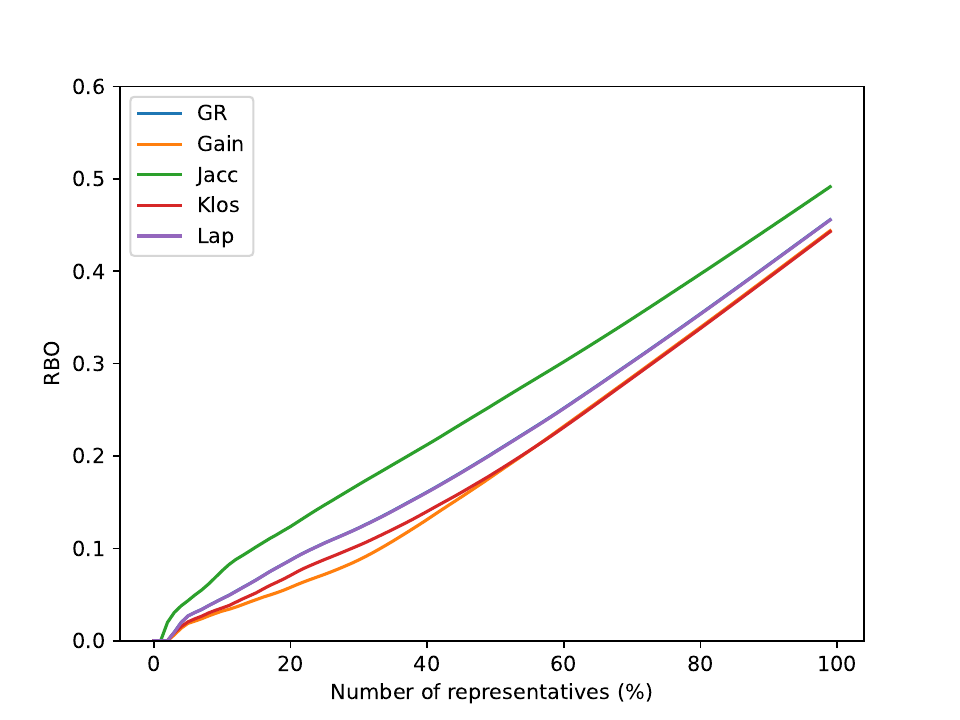}
        \subcaption{D\&D}
        \label{fig:AppRBODD_2}
    \end{minipage}
    \begin{minipage}[b]{0.32\textwidth}
        \includegraphics[width=\textwidth]{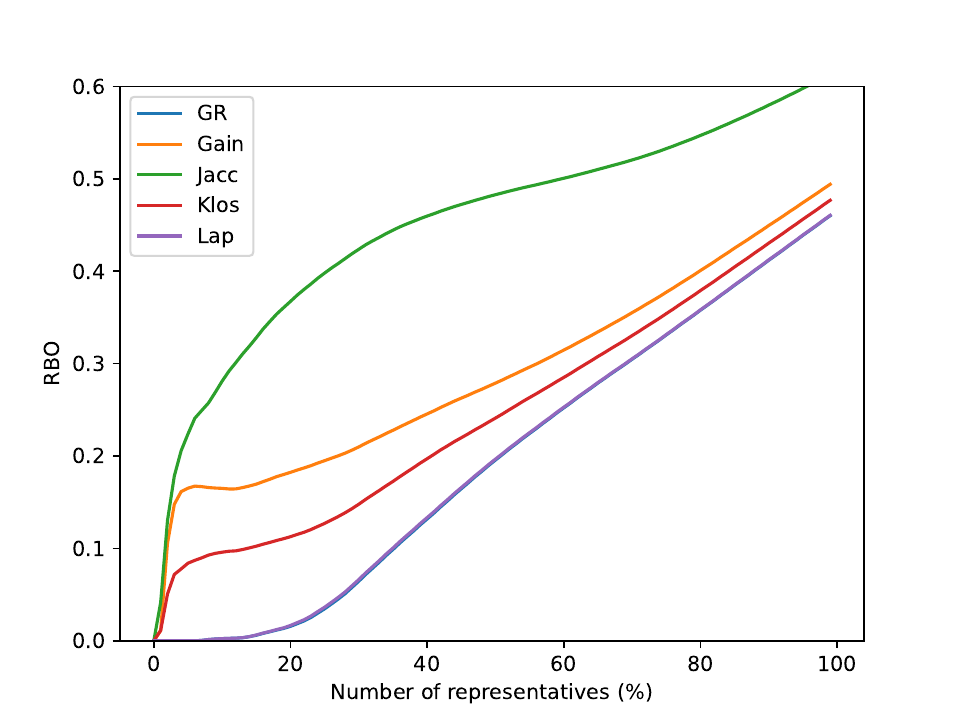}
        \subcaption{AIDS}
        \label{fig:AppRBOAIDS_2}
    \end{minipage}
    \begin{minipage}[b]{0.32\textwidth}
        \includegraphics[width=\textwidth]{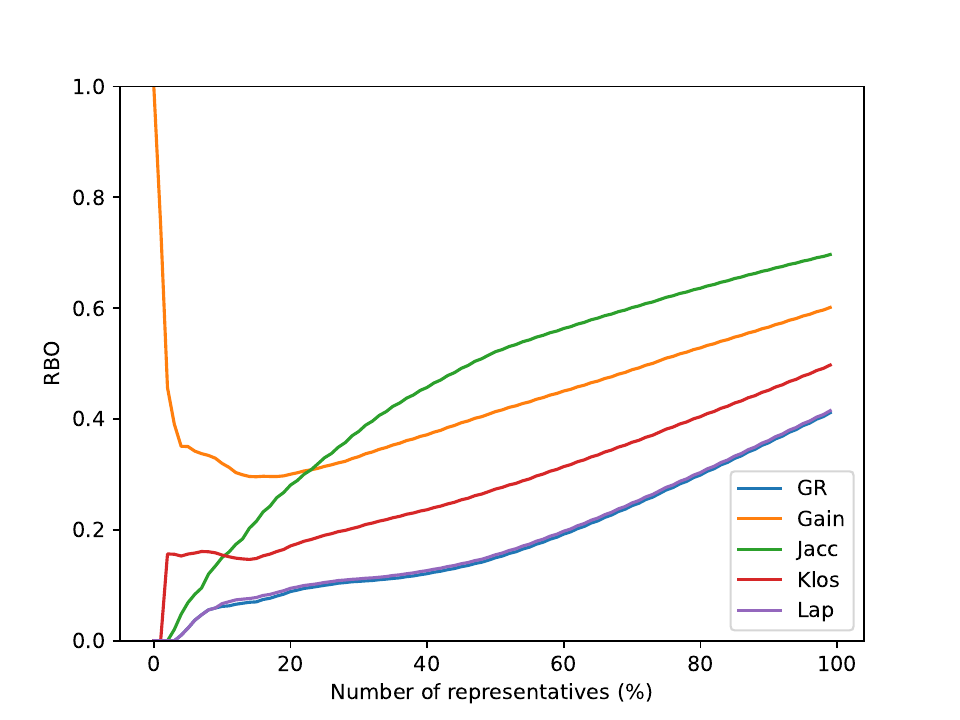}
        \subcaption{FOPPA}
        \label{fig:AppRBOFOPPA_2}
    \end{minipage}

    \vspace{1em}
    
    \begin{minipage}[b]{0.32\textwidth}
        \includegraphics[width=\textwidth]{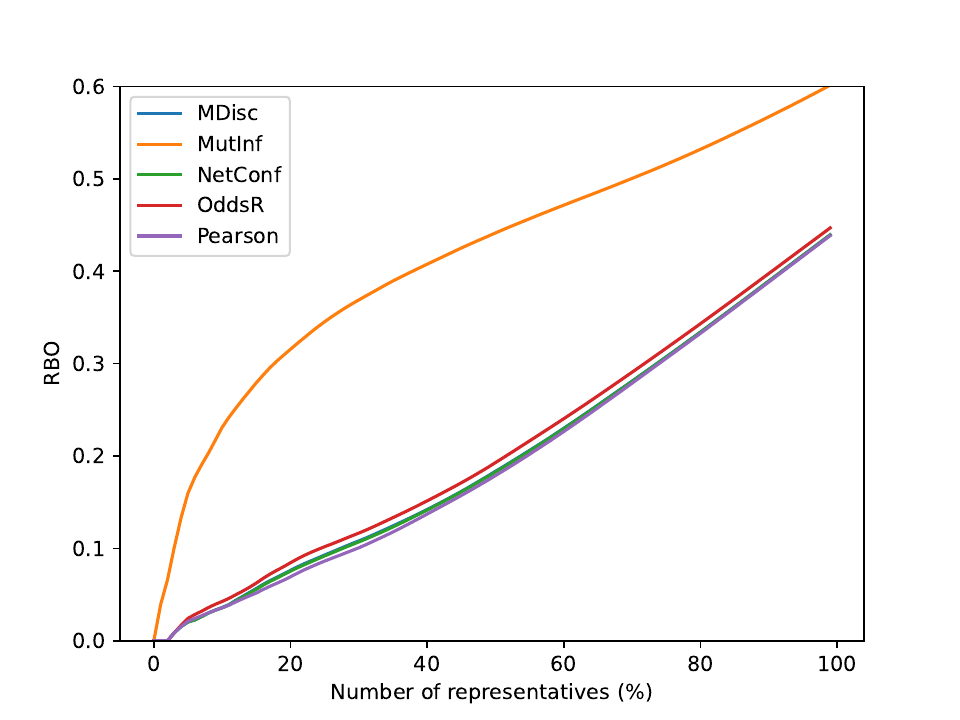}
        \subcaption{D\&D}
        \label{fig:AppRBODD_3}
    \end{minipage}
    \begin{minipage}[b]{0.32\textwidth}
        \includegraphics[width=\textwidth]{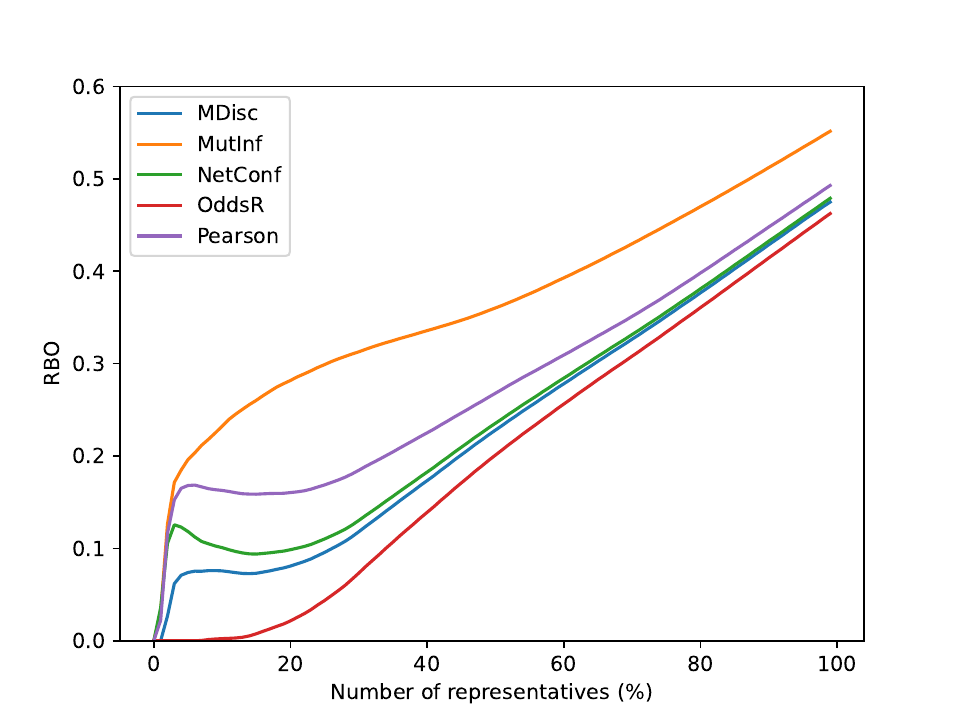}
        \subcaption{AIDS}
        \label{fig:AppRBOAIDS_3}
    \end{minipage}
    \begin{minipage}[b]{0.32\textwidth}
        \includegraphics[width=\textwidth]{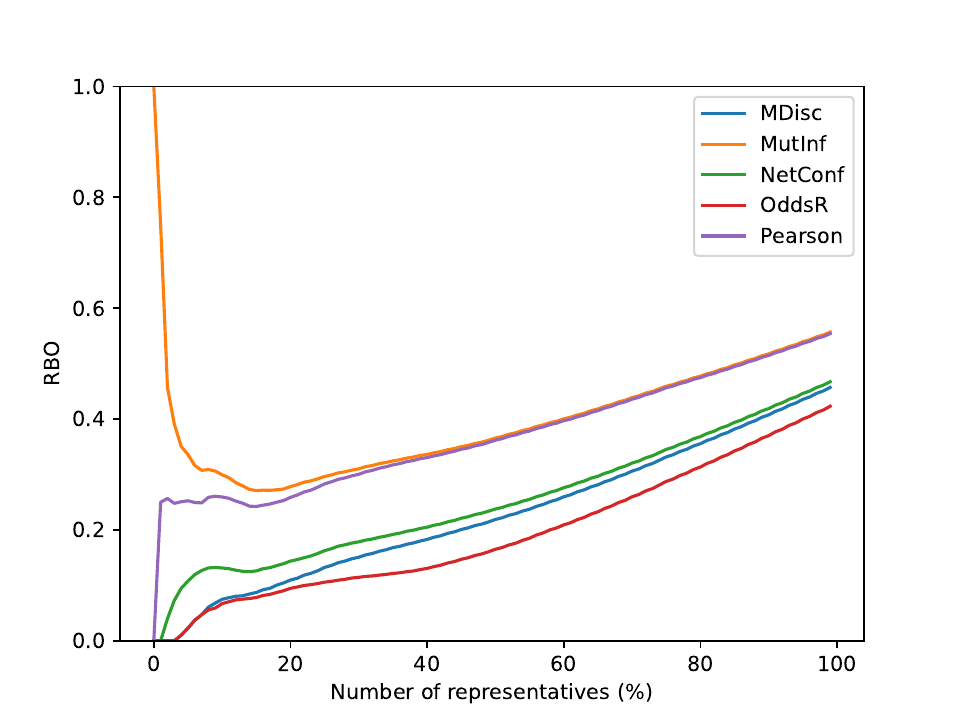}
        \subcaption{FOPPA}
        \label{fig:AppRBOFOPPA_3}
    \end{minipage}

    \vspace{1em}
    
    \begin{minipage}[b]{0.32\textwidth}
        \includegraphics[width=\textwidth]{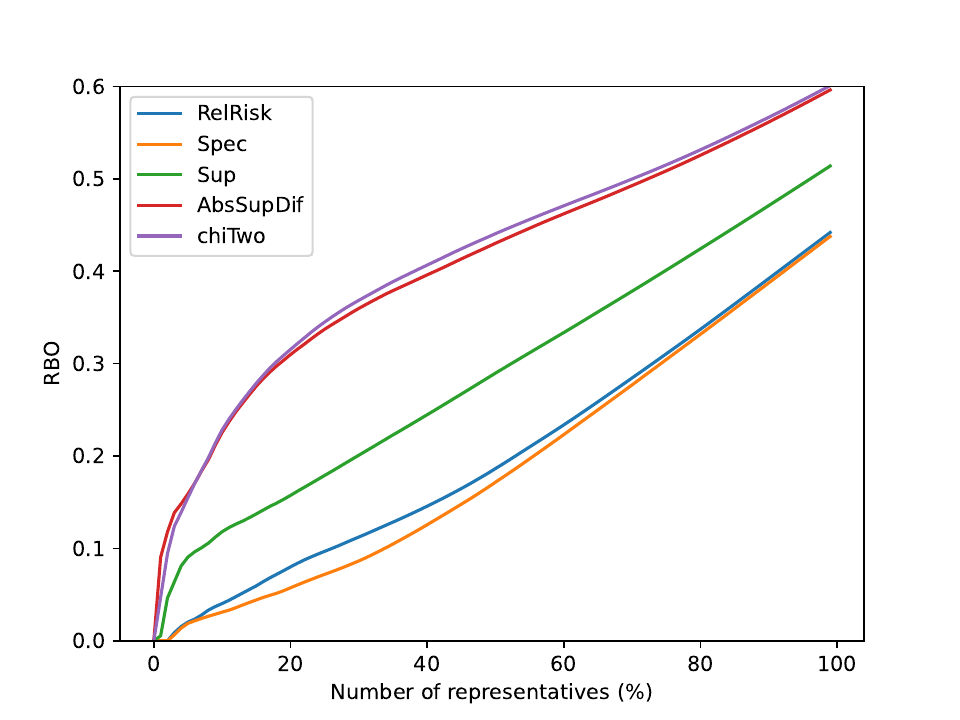}
        \subcaption{D\&D}
        \label{fig:AppRBODD_4}
    \end{minipage}
    \begin{minipage}[b]{0.32\textwidth}
        \includegraphics[width=\textwidth]{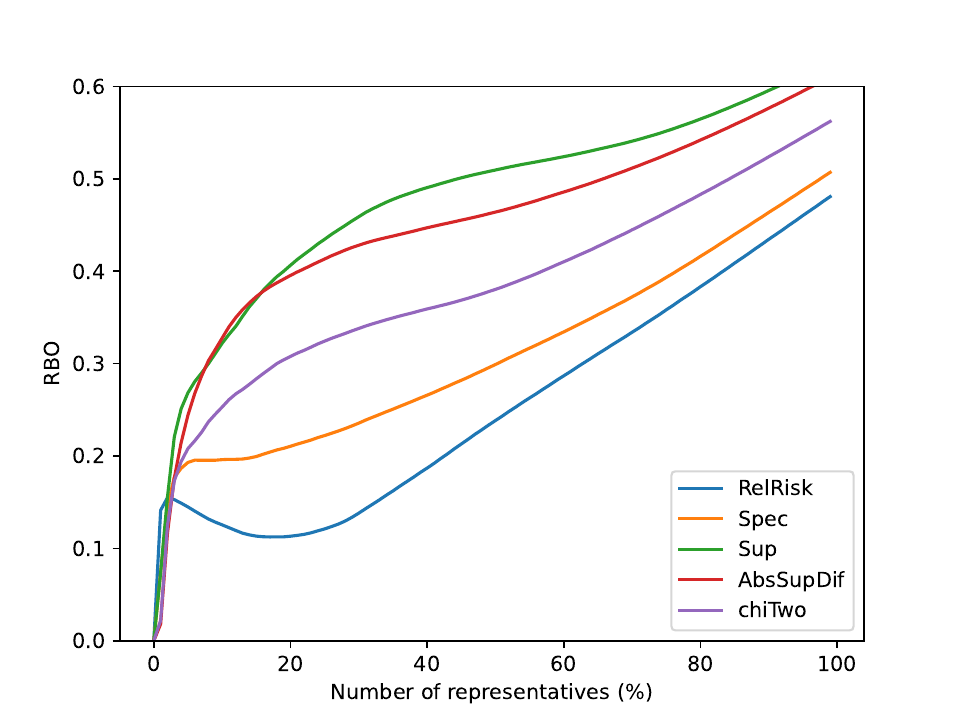}
        \subcaption{AIDS}
        \label{fig:AppRBOAIDS_4}
    \end{minipage}
    \begin{minipage}[b]{0.32\textwidth}
        \includegraphics[width=\textwidth]{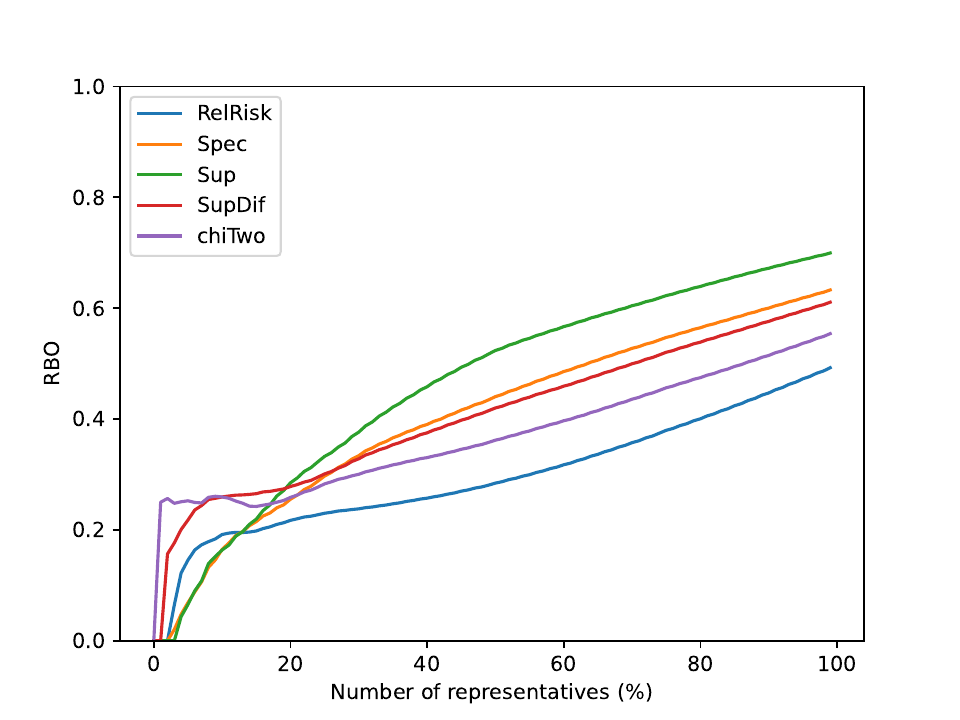}
        \subcaption{FOPPA}
        \label{fig:AppRBOFOPPA_4}
    \end{minipage}
    \caption{RBO between the rankings obtained for each selected quality measure and the gold standard, as a function of $s$, the number of top representatives considered for datasets D\&D, AIDS and FOPPA. The rest of the datasets are shown in Figure~\ref{fig:ShapleyRBO_1}.}
    \label{fig:ShapleyRBO_2}
    \Description{Description} 
\end{figure}

Figures~\ref{fig:ShapleyRBO_1} and \ref{fig:ShapleyRBO_2} show the RBO obtained between each measure and our gold standard, as a function of $s$, the number of representatives considered. These figures are extensions of Figure~\ref{fig:ShapleyRBO_main}, this time displaying the full set of \textcolor{highlightCol}{38} quality measures. 

\subsection{Classification Comparison}
\label{sec:AppFullF1}
Figures~\ref{fig:ShapleyRBO_1} and \ref{fig:ShapleyRBO_2} show the $F1$-Score obtained for each measure as well as our gold standard, as a function of $s$, the number of representatives considered. These figures are extensions of Figure~\ref{fig:ShapleyF1_main}, this time displaying all \textcolor{highlightCol}{38} quality measures. 

\begin{figure}[htbp!]
    \centering
    \begin{minipage}[b]{0.32\textwidth}
        \includegraphics[width=\textwidth]{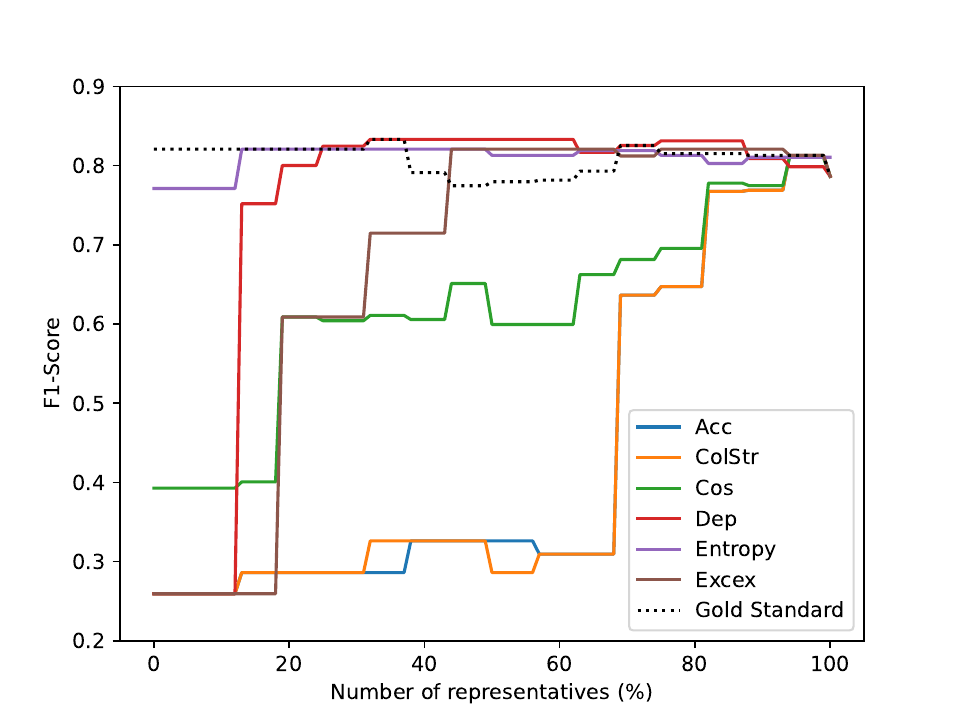}
        \subcaption{MUTAG}
        \label{fig:AppF1MUTAG_1}
    \end{minipage}
    \begin{minipage}[b]{0.32\textwidth}
        \includegraphics[width=\textwidth]{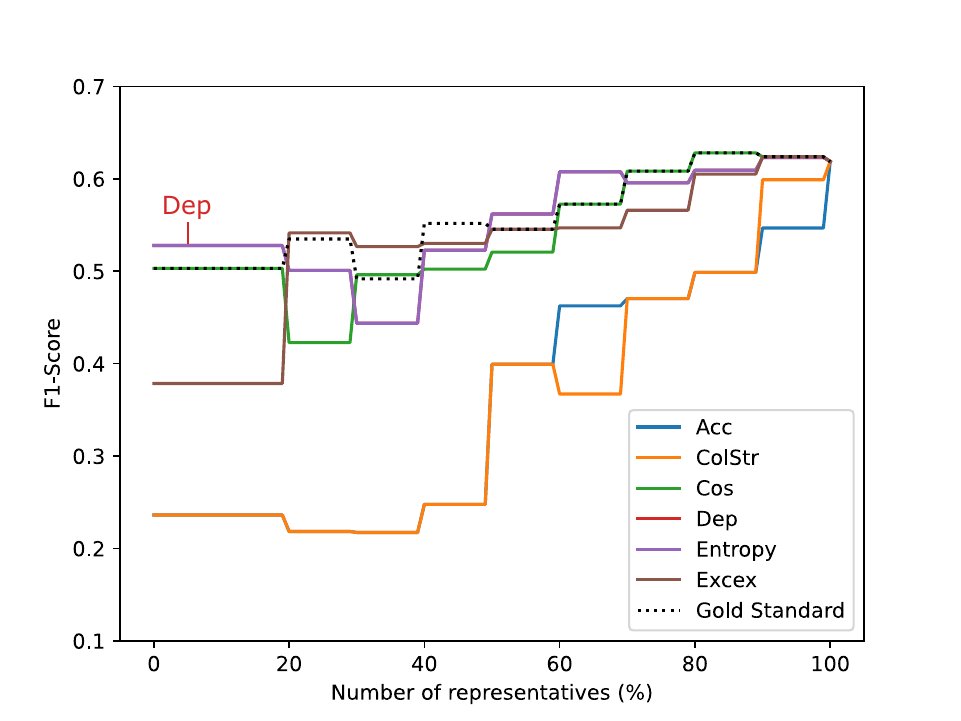}
        \subcaption{PTC}
        \label{fig:AppF1PTC_1}
    \end{minipage}
    \begin{minipage}[b]{0.32\textwidth}
        \includegraphics[width=\textwidth]{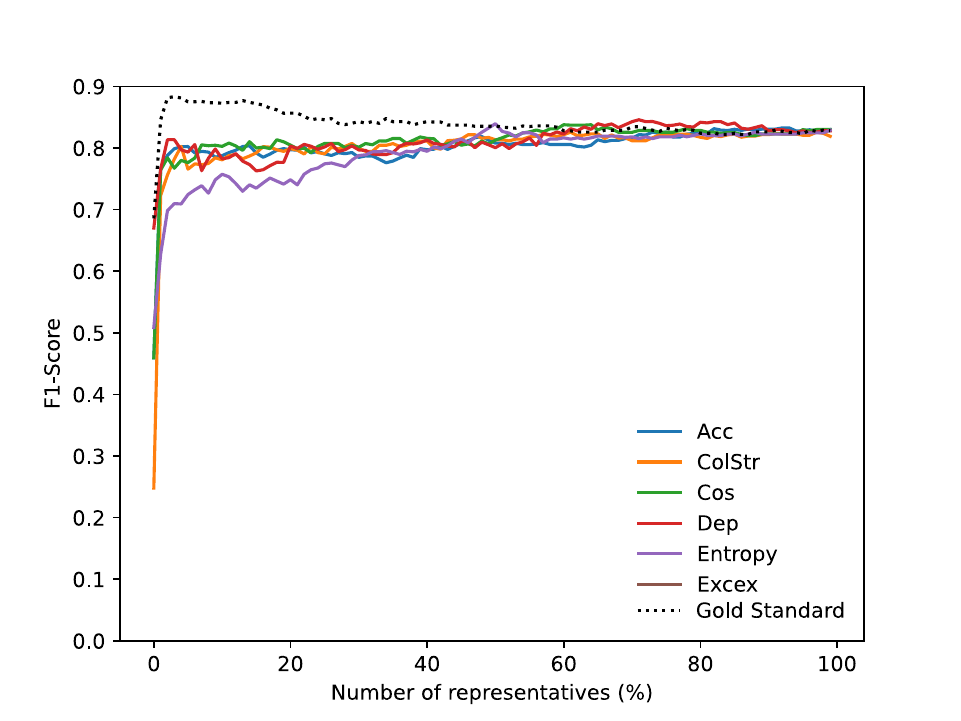}
        \subcaption{NCI1}
        \label{fig:AppF1NCI1_1}
    \end{minipage}
    
    \vspace{1em}
    
    \begin{minipage}[b]{0.32\textwidth}
        \includegraphics[width=\textwidth]{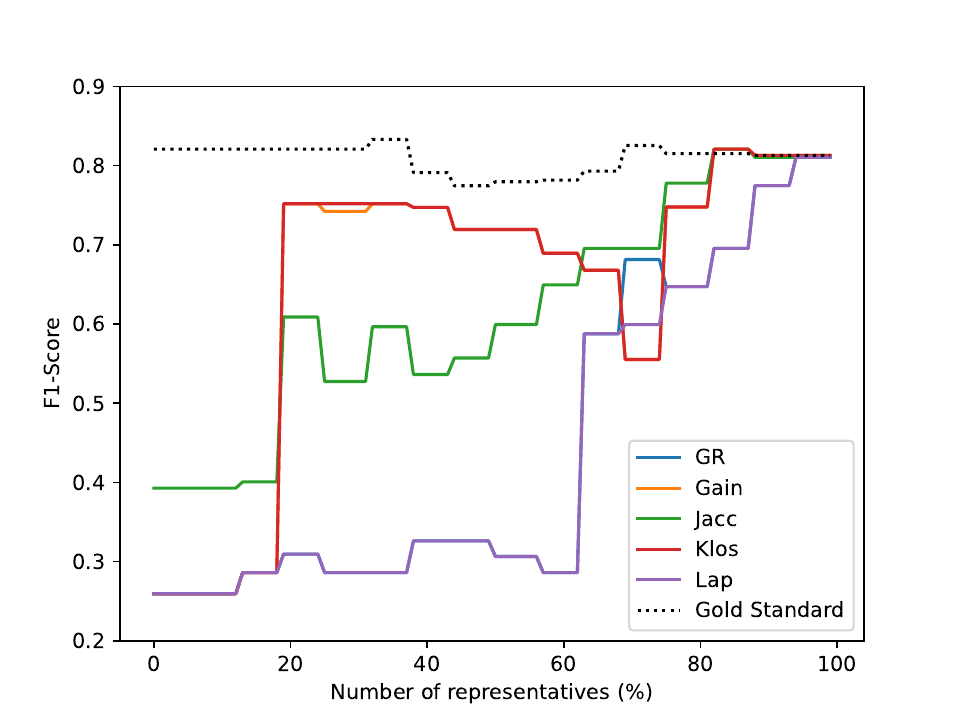}
        \subcaption{MUTAG}
        \label{fig:AppF1MUTAG_2}
    \end{minipage}
    \begin{minipage}[b]{0.32\textwidth}
        \includegraphics[width=\textwidth]{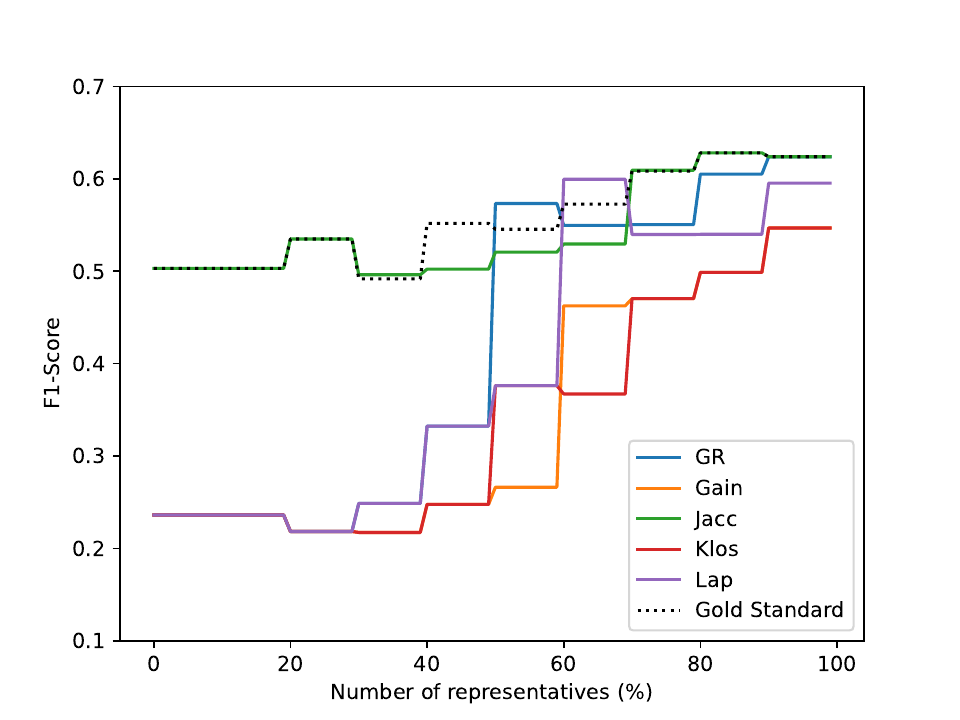}
        \subcaption{PTC}
        \label{fig:AppF1PTC_2}
    \end{minipage}
    \begin{minipage}[b]{0.32\textwidth}
        \includegraphics[width=\textwidth]{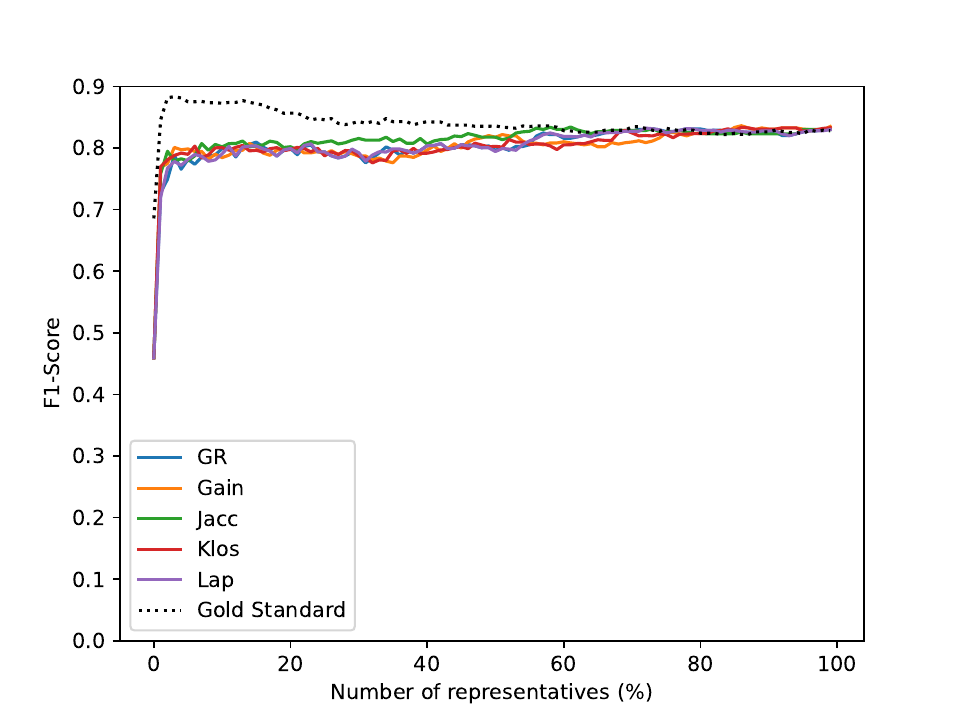}
        \subcaption{NCI1}
        \label{fig:AppF1NCI1_2}
    \end{minipage}

    \vspace{1em}
    
   \begin{minipage}[b]{0.32\textwidth}
        \includegraphics[width=\textwidth]{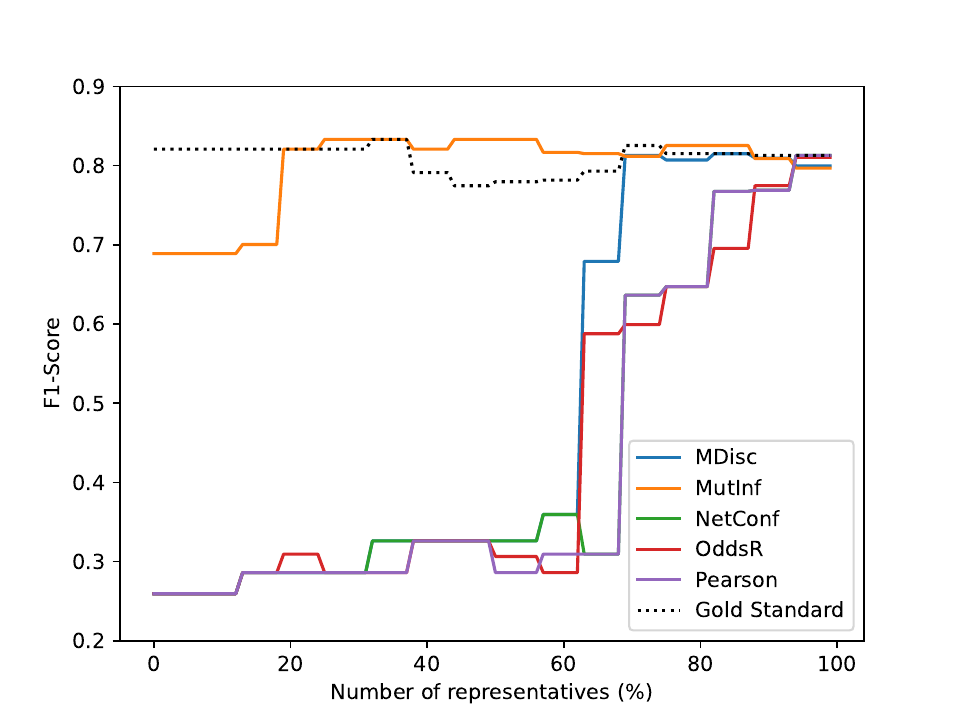}
        \subcaption{MUTAG}
        \label{fig:AppF1MUTAG_3}
    \end{minipage}
    \begin{minipage}[b]{0.32\textwidth}
        \includegraphics[width=\textwidth]{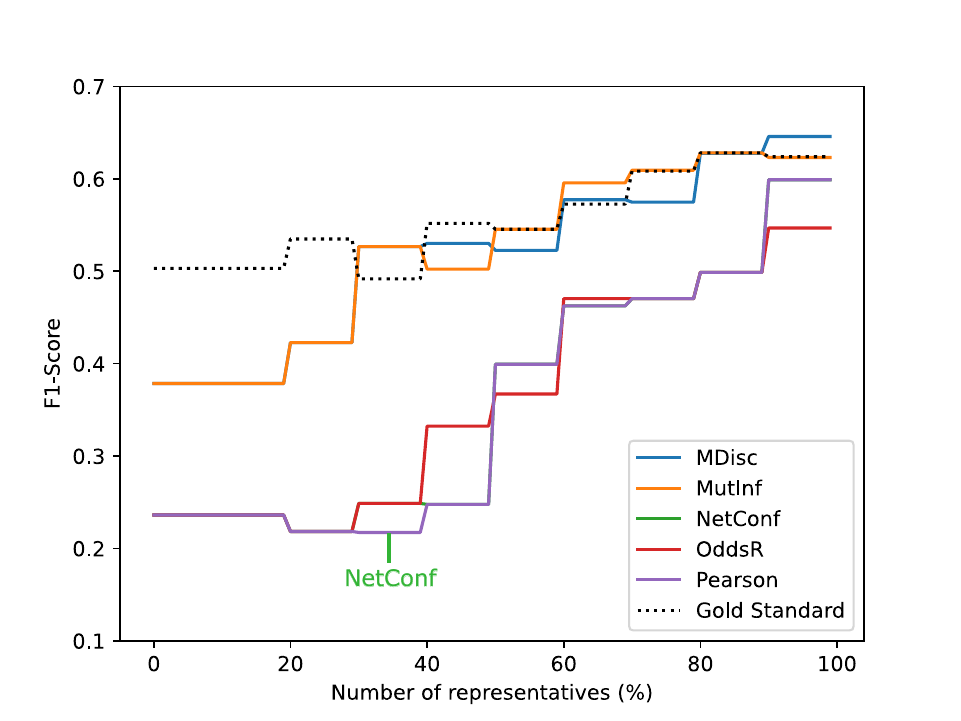}
        \subcaption{PTC}
        \label{fig:AppF1PTC_3}
    \end{minipage}
    \begin{minipage}[b]{0.32\textwidth}
        \includegraphics[width=\textwidth]{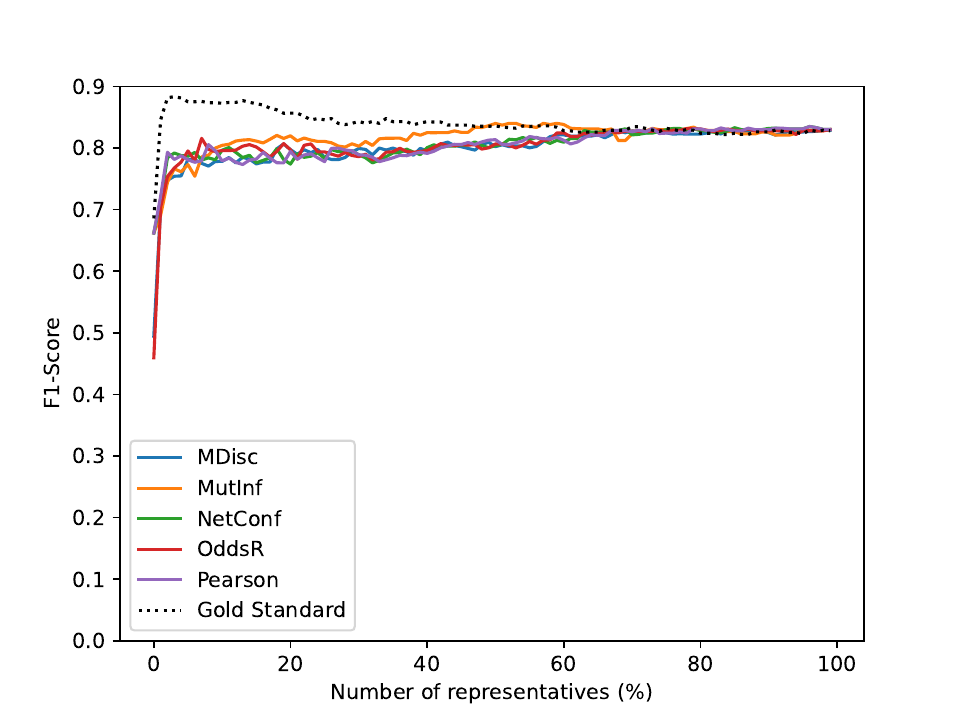}
        \subcaption{NCI1}
        \label{fig:AppF1NCI1_3}
    \end{minipage}

    \vspace{1em}
    
    \begin{minipage}[b]{0.32\textwidth}
        \includegraphics[width=\textwidth]{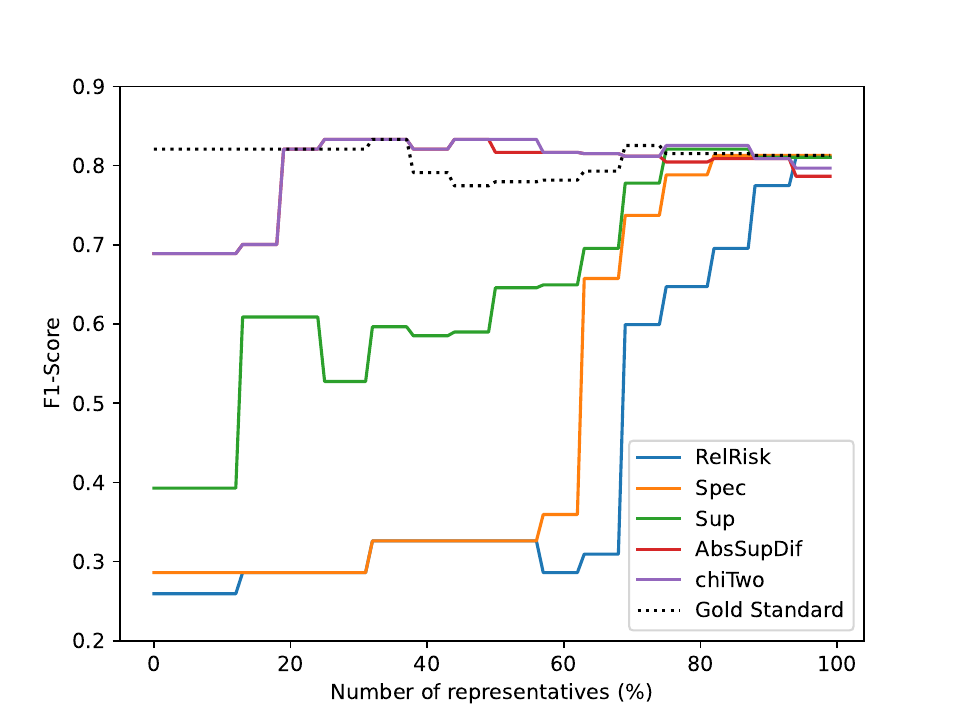}
        \subcaption{MUTAG}
        \label{fig:AppF1MUTAG_4}
    \end{minipage}
    \begin{minipage}[b]{0.32\textwidth}
        \includegraphics[width=\textwidth]{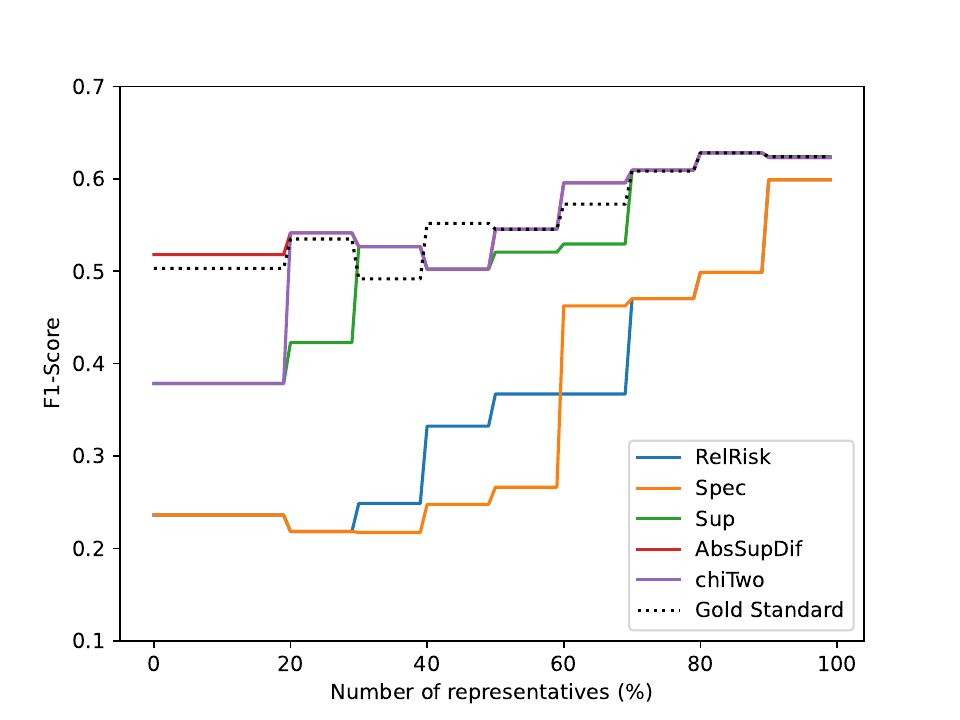}
        \subcaption{PTC}
        \label{fig:AppF1PTC_4}
    \end{minipage}
    \begin{minipage}[b]{0.32\textwidth}
        \includegraphics[width=\textwidth]{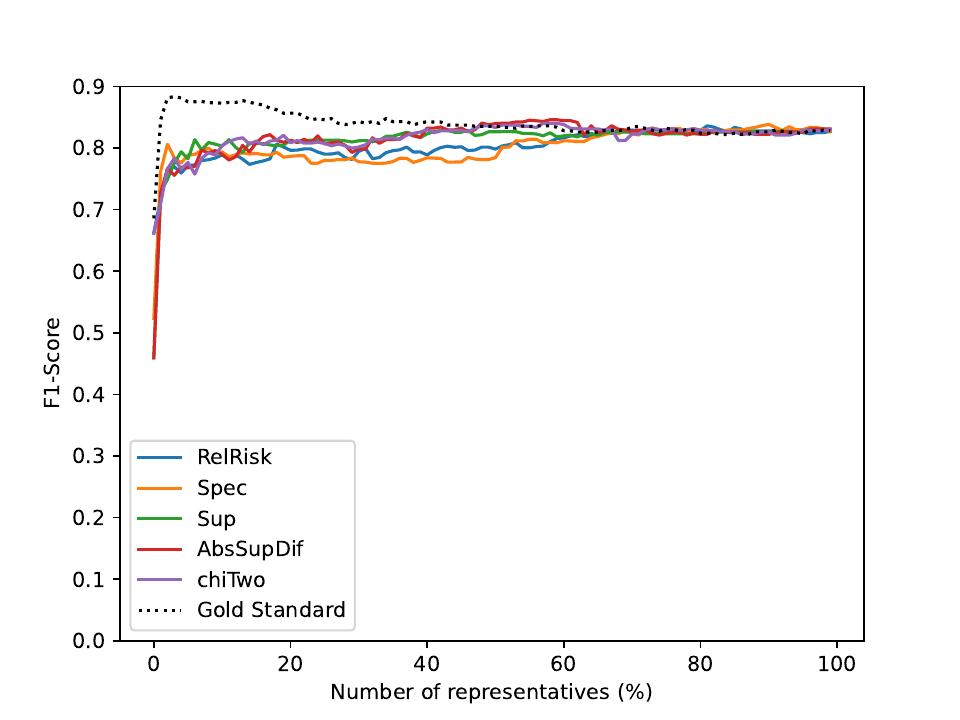}
        \subcaption{NCI1}
        \label{fig:AppF1NCI1_4}
    \end{minipage}
    \caption{$F1$-Score as a function of the proportion of representatives selected for each quality measure and gold standard for datasets MUTAG, PTC and NCI1. The rest of the datasets are shown in Figure~\ref{fig:ShapleyF1_2})}
    \label{fig:ShapleyF1_1}
    \Description{Description} 
\end{figure}


\begin{figure}[htbp!]
    \centering
    \begin{minipage}[b]{0.32\textwidth}
        \includegraphics[width=\textwidth]{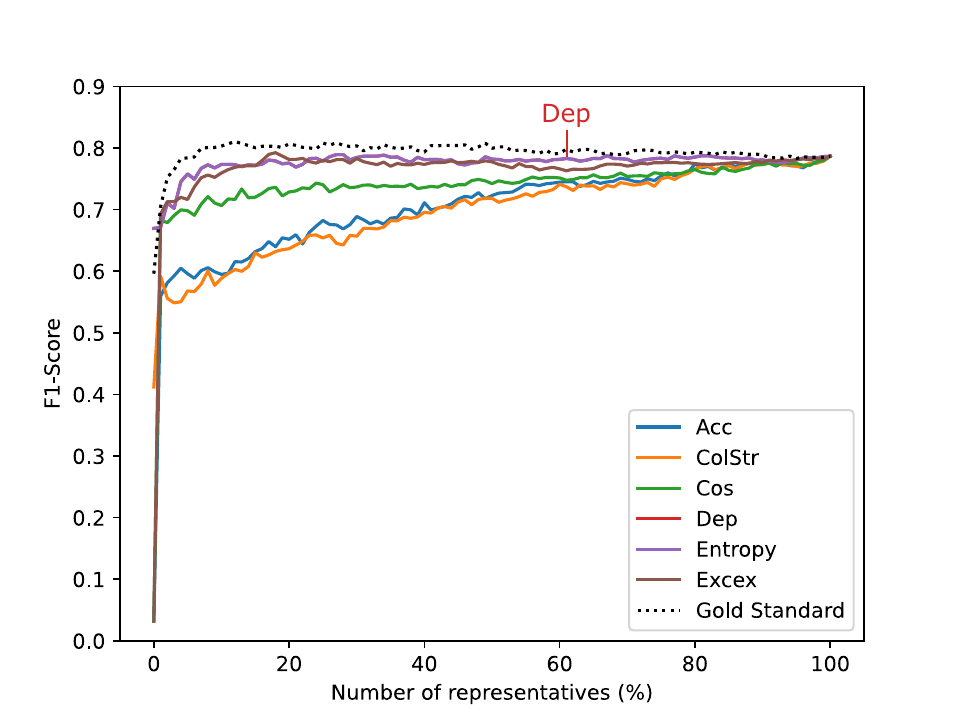}
        \subcaption{D\&D}
        \label{fig:AppF1DD_1}
    \end{minipage}
    \begin{minipage}[b]{0.32\textwidth}
        \includegraphics[width=\textwidth]{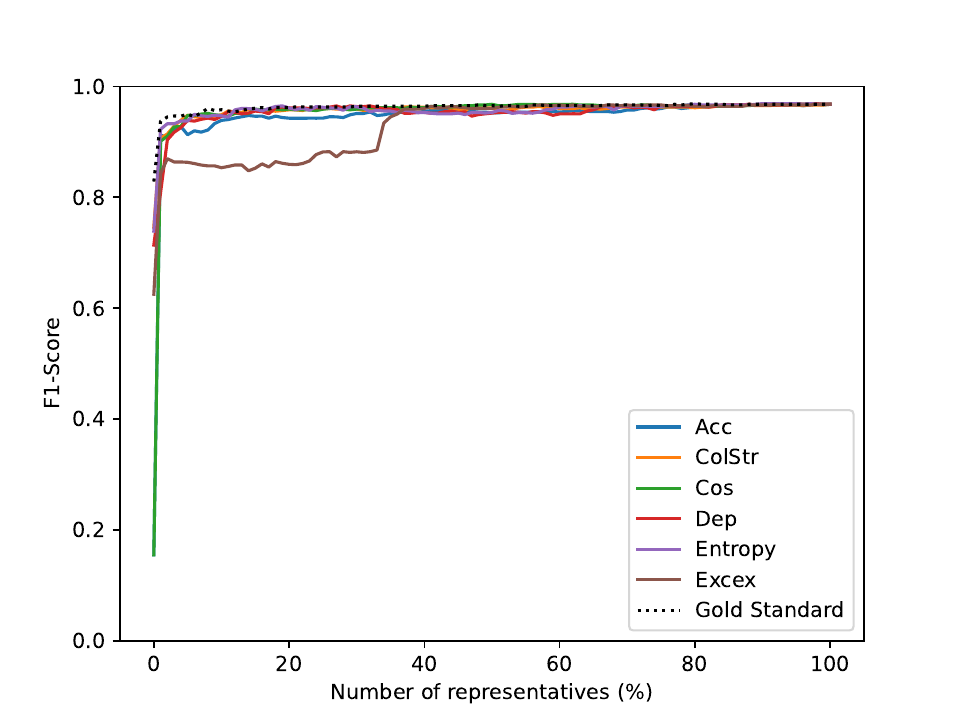}
        \subcaption{AIDS}
        \label{fig:AppF1AIDS_1}
    \end{minipage}
    \begin{minipage}[b]{0.32\textwidth}
        \includegraphics[width=\textwidth]{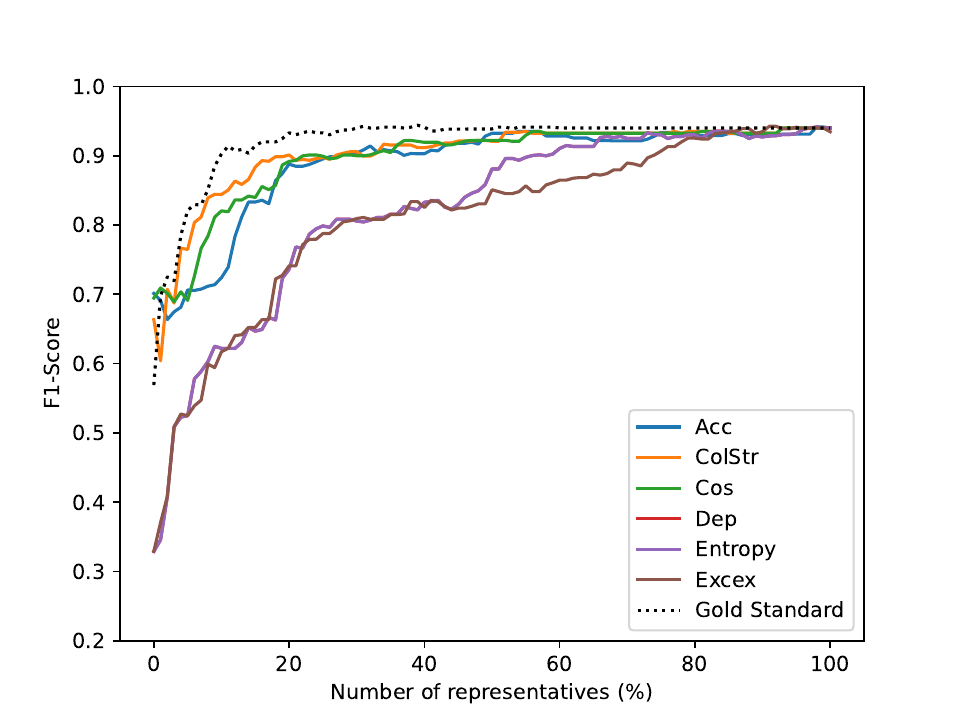}
        \subcaption{FOPPA}
        \label{fig:AppF1FOPPA_1}
    \end{minipage}
    
    \vspace{1em}
    
    \begin{minipage}[b]{0.32\textwidth}
        \includegraphics[width=\textwidth]{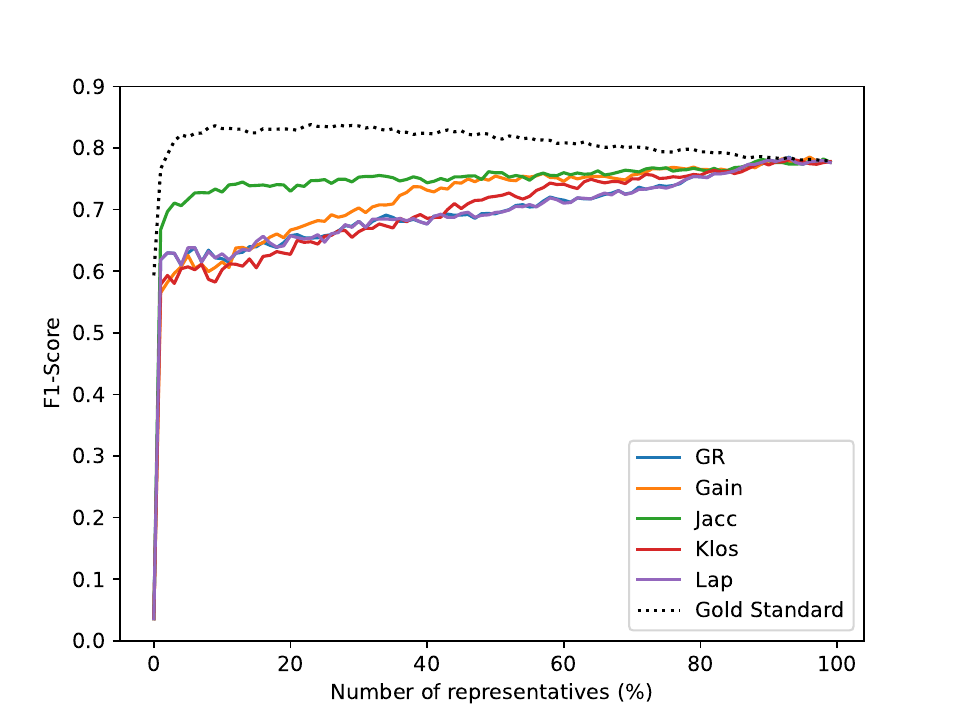}
        \subcaption{D\&D}
        \label{fig:AppF1DD_2}
    \end{minipage}
    \begin{minipage}[b]{0.32\textwidth}
        \includegraphics[width=\textwidth]{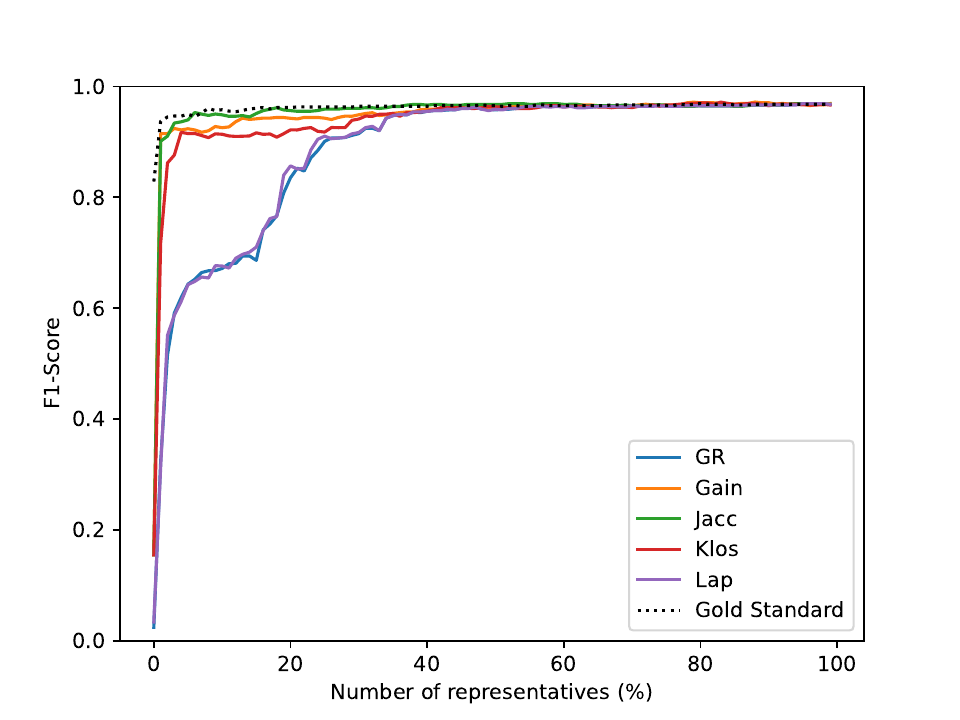}
        \subcaption{AIDS}
        \label{fig:AppF1AIDS_2}
    \end{minipage}
    \begin{minipage}[b]{0.32\textwidth}
        \includegraphics[width=\textwidth]{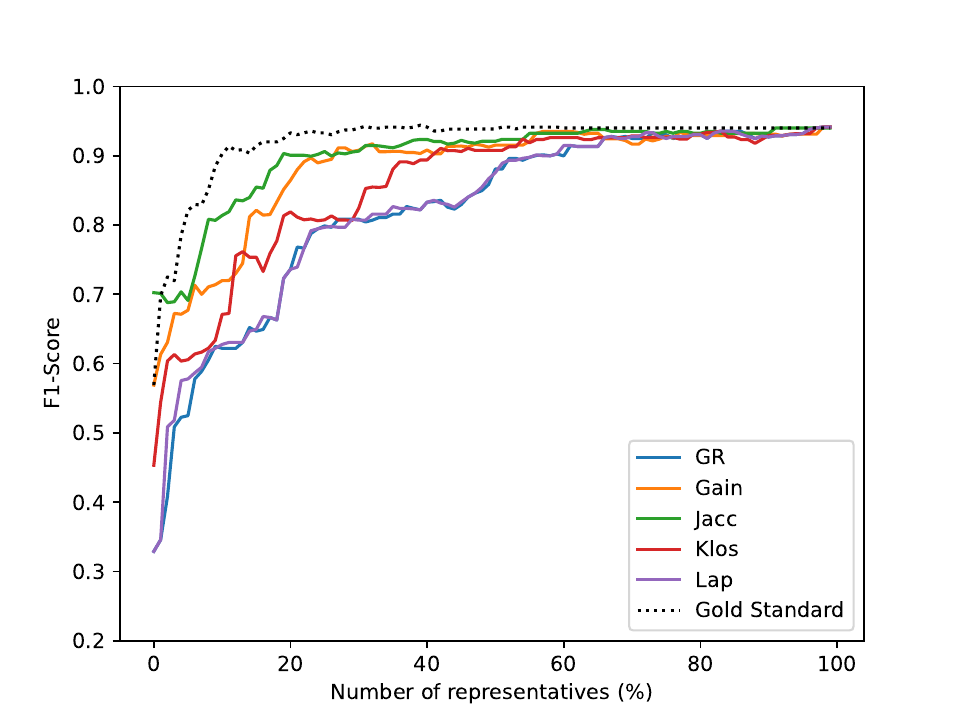}
        \subcaption{FOPPA}
        \label{fig:AppF1FOPPA_2}
    \end{minipage}

    \vspace{1em}
    
    \begin{minipage}[b]{0.32\textwidth}
        \includegraphics[width=\textwidth]{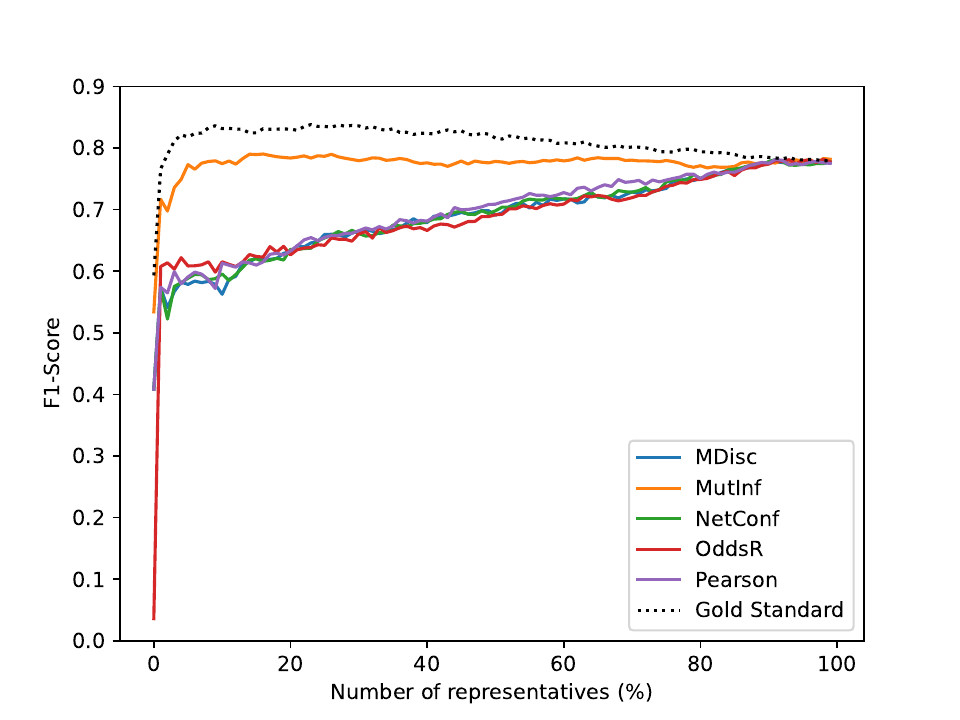}
        \subcaption{D\&D}
        \label{fig:AppF1DD_3}
    \end{minipage}
    \begin{minipage}[b]{0.32\textwidth}
        \includegraphics[width=\textwidth]{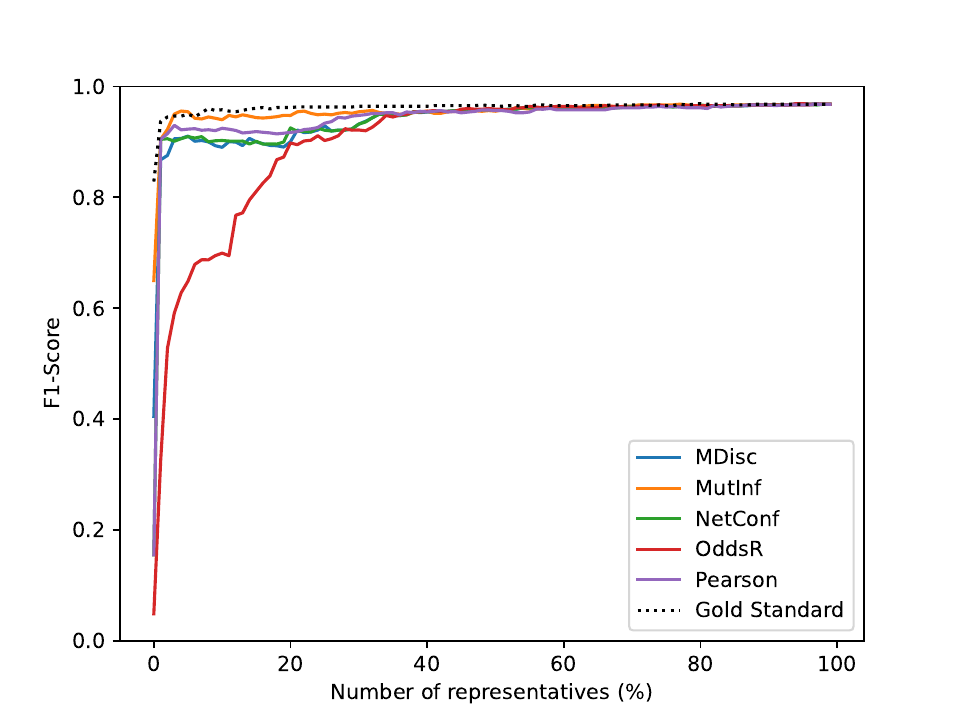}
        \subcaption{AIDS}
        \label{fig:AppF1AIDS_3}
    \end{minipage}
    \begin{minipage}[b]{0.32\textwidth}
        \includegraphics[width=\textwidth]{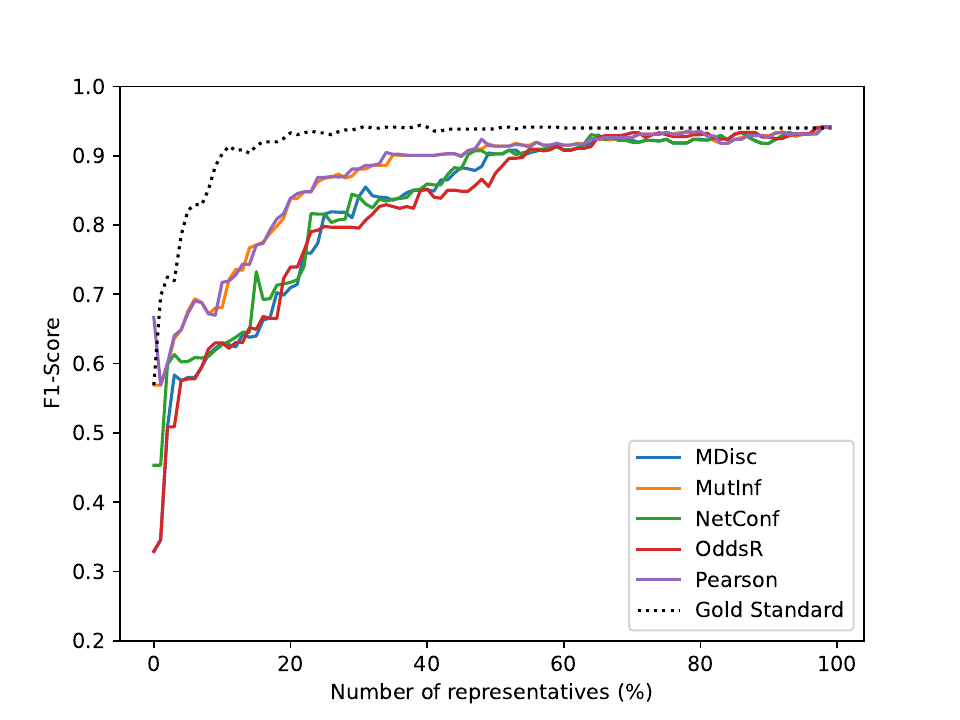}
        \subcaption{FOPPA}
        \label{fig:AppF1FOPPA_3}
    \end{minipage}

    \vspace{1em}
    
    \begin{minipage}[b]{0.32\textwidth}
        \includegraphics[width=\textwidth]{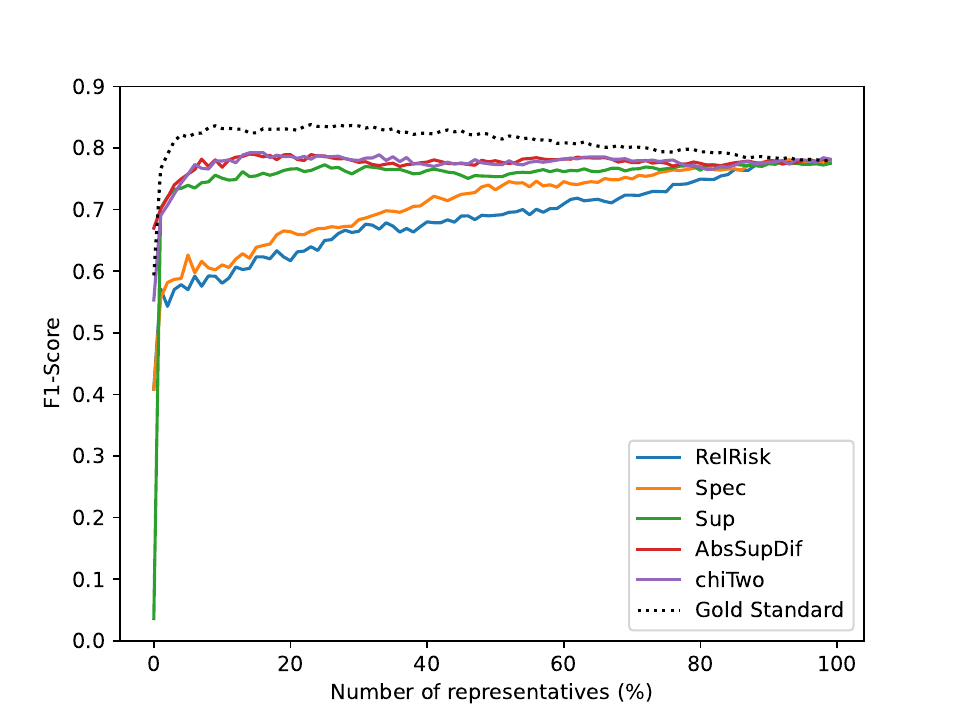}
        \subcaption{D\&D}
        \label{fig:AppF1DD_4}
    \end{minipage}
    \begin{minipage}[b]{0.32\textwidth}
        \includegraphics[width=\textwidth]{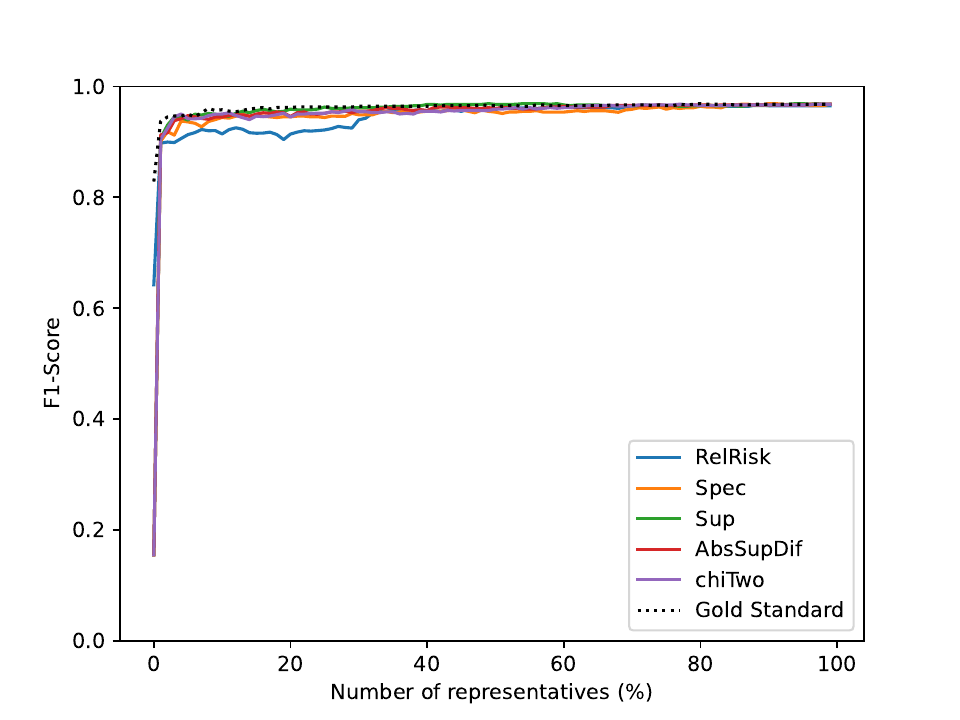}
        \subcaption{AIDS}
        \label{fig:AppF1AIDS_4}
    \end{minipage}
    \begin{minipage}[b]{0.32\textwidth}
        \includegraphics[width=\textwidth]{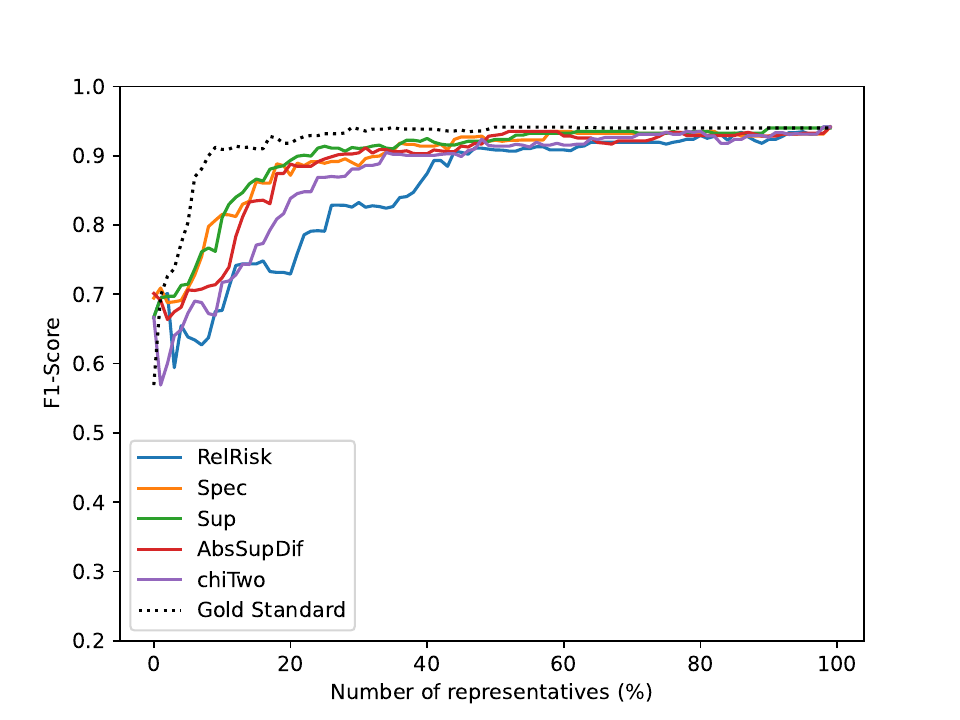}
        \subcaption{FOPPA}
        \label{fig:AppF1FOPPA_4}
    \end{minipage}
    \caption{$F1$-Score as a function of the proportion of representatives selected for each quality measure and gold standard for datasets D\&D, AIDS and FOPPA. The rest of the datasets are shown in Figure~\ref{fig:ShapleyF1_1})}
    \label{fig:ShapleyF1_2}
    \Description{Description} 
\end{figure}

\section{Additional Results for Two Datasets}
\label{sec:AppAddRes}
This appendix shows results obtained for two datasets that are not shown in the main paper, for the sake of concision: FRANK (Appendix~\ref{sec:AppAddResFRANK}) and IMDb (Appendix~\ref{sec:AppAddResIMDb}).

\subsection{FRANK Results}
\label{sec:AppAddResFRANK}
\begin{figure}[htbp!]
    \centering
    \begin{minipage}[b]{0.32\textwidth}
        \includegraphics[width=\textwidth]{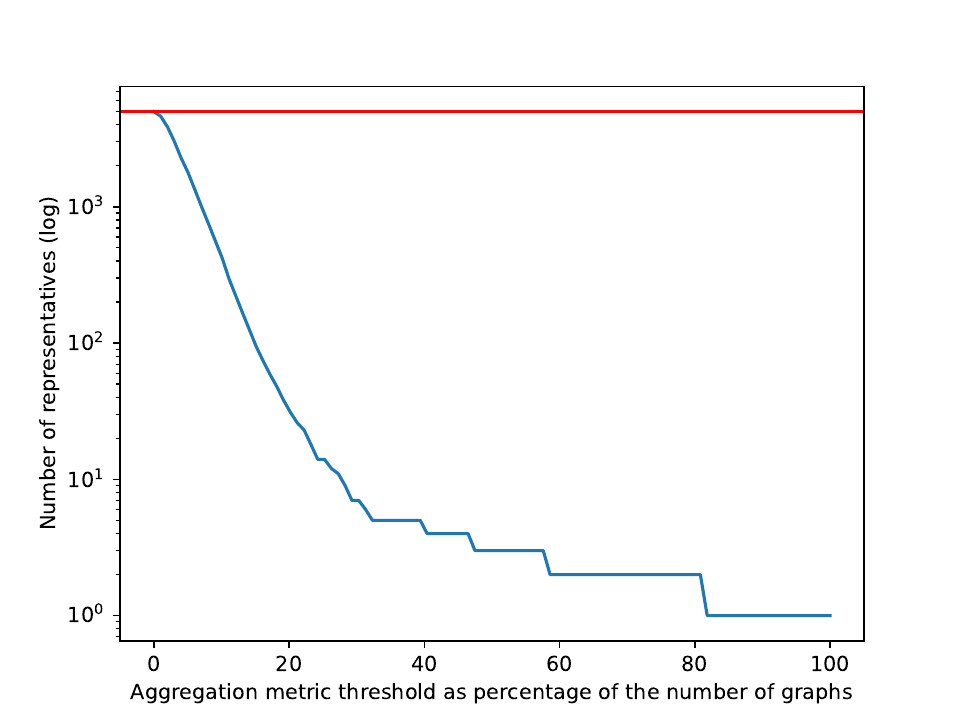}
        \subcaption{Number of representatives}
        \label{fig:Frank_NbR}
    \end{minipage}
    \begin{minipage}[b]{0.32\textwidth}
        \includegraphics[width=\textwidth]{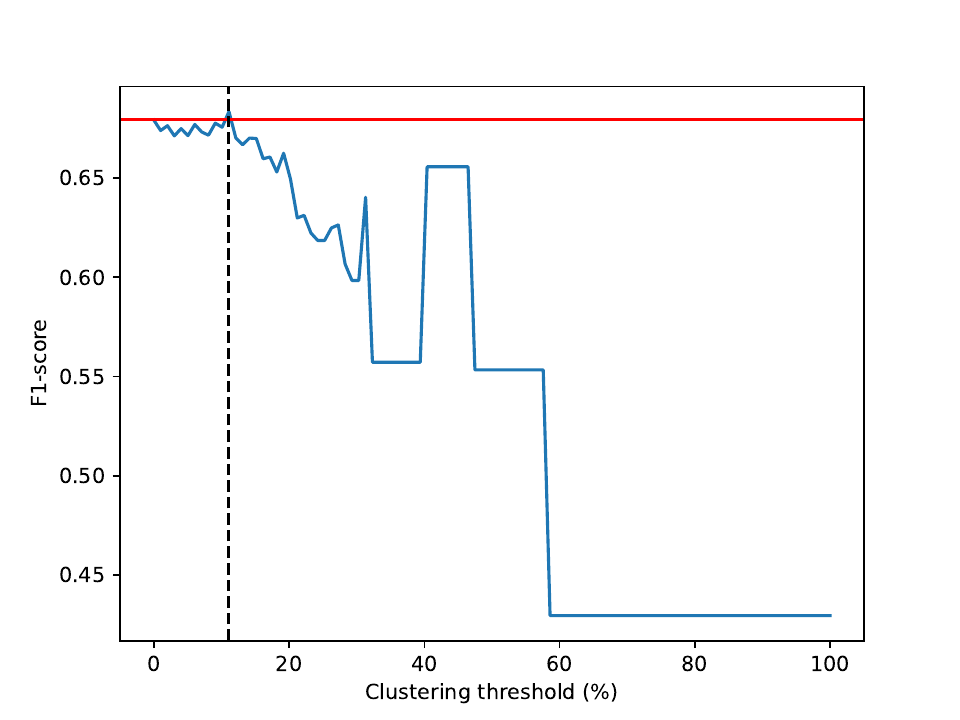}
        \subcaption{$F1$-Score (clustering threshold)}
        \label{fig:Frank_F1_cluster}
    \end{minipage}
    \begin{minipage}[b]{0.32\textwidth}
        \includegraphics[width=\textwidth]{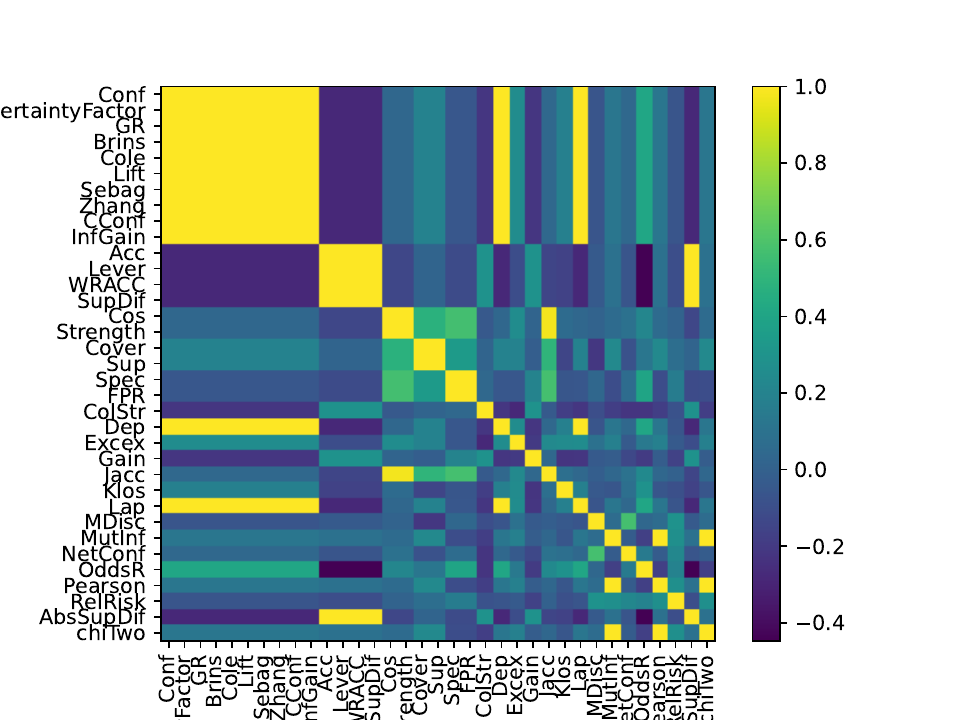}
        \subcaption{RBO matrix correlation}
        \label{fig:Frank_Matrix}
    \end{minipage}
    
    \vspace{1em}
    
    \begin{minipage}[b]{0.32\textwidth}
        \includegraphics[width=\textwidth]{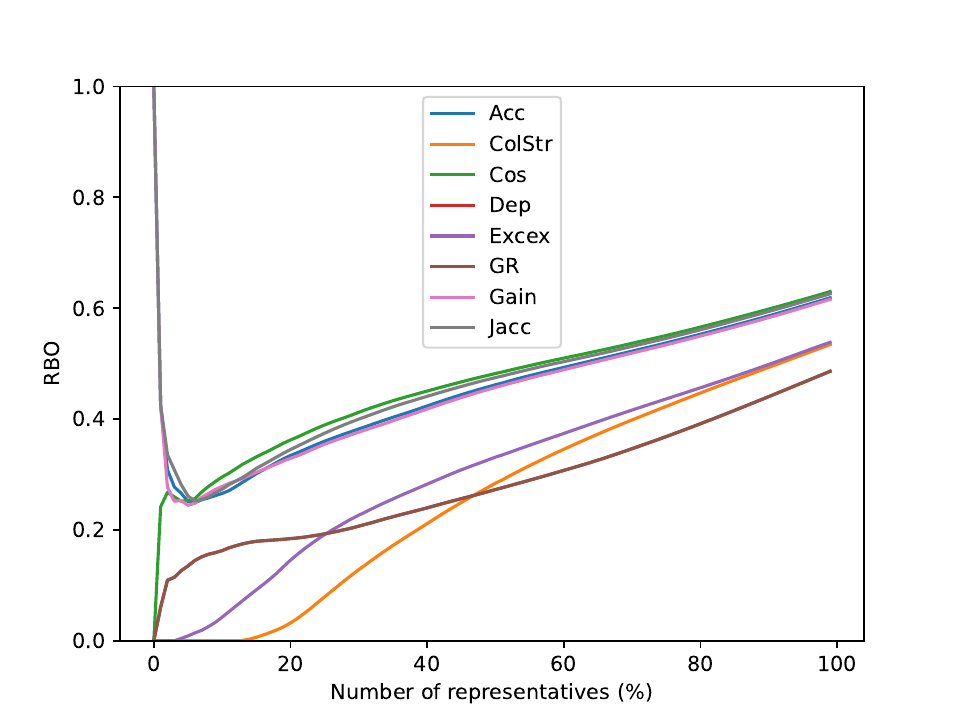}
        \subcaption{RBO (QM 1-8)}
        \label{fig:Frank_RBO_1}
    \end{minipage}
    \begin{minipage}[b]{0.32\textwidth}
        \includegraphics[width=\textwidth]{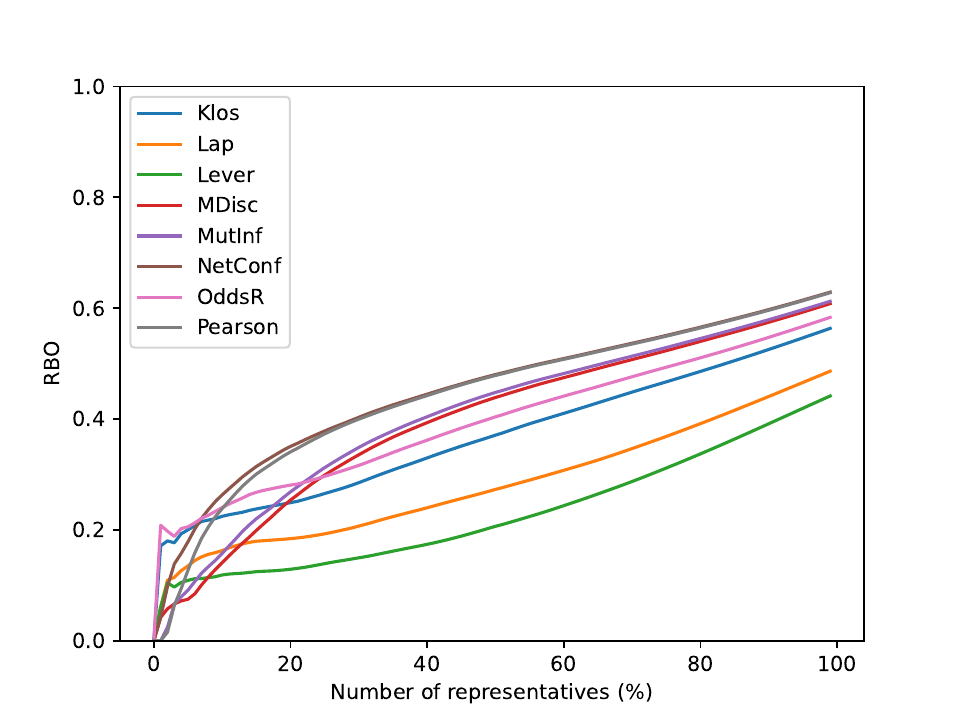}
        \subcaption{RBO (QM 9-16)}
        \label{fig:Frank_RBO_2}
    \end{minipage}
    \begin{minipage}[b]{0.32\textwidth}
        \includegraphics[width=\textwidth]{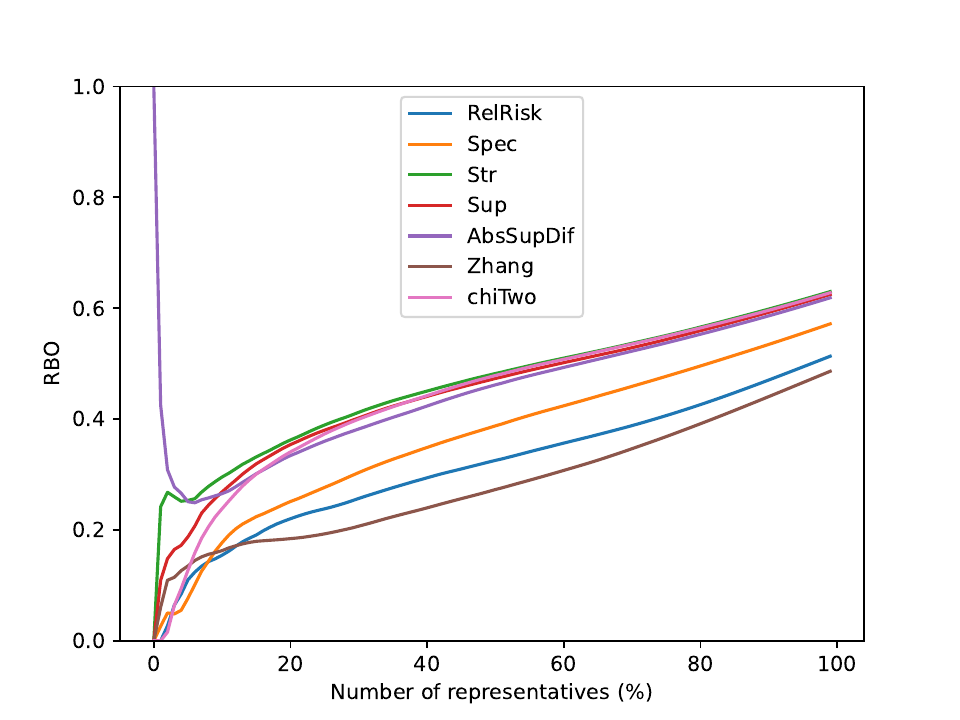}
        \subcaption{RBO (QM 17-23)}
        \label{fig:Frank_RBO_3}
    \end{minipage}

    \vspace{1em}
    
    \begin{minipage}[b]{0.32\textwidth}
        \includegraphics[width=\textwidth]{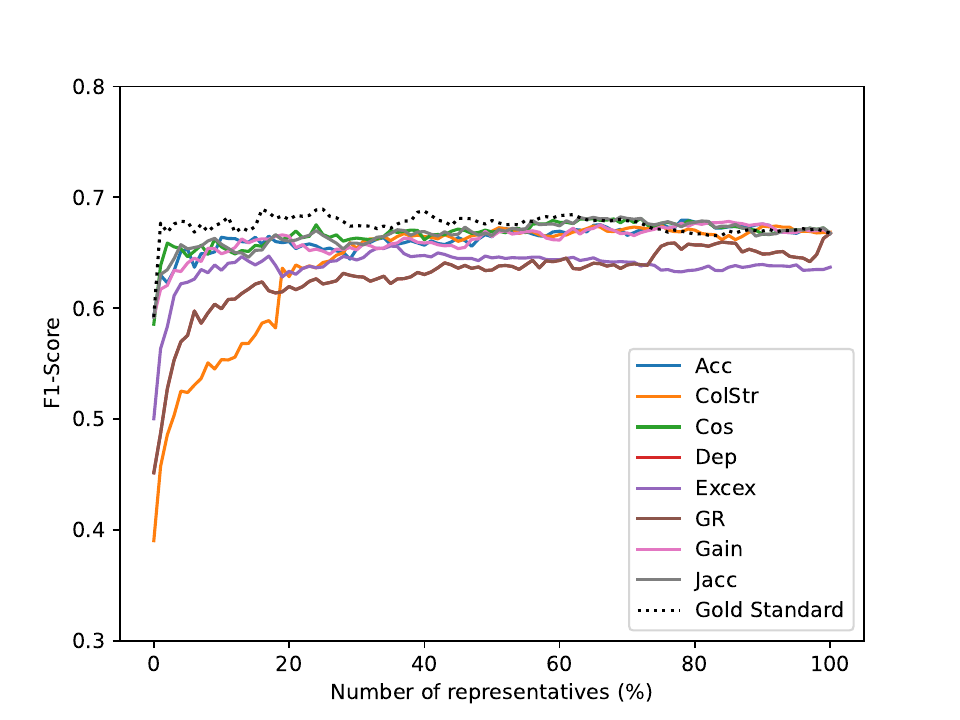}
        \subcaption{$F1$-Score (QM 1-8)}
        \label{fig:Frank_F1}
    \end{minipage}
    \begin{minipage}[b]{0.32\textwidth}
        \includegraphics[width=\textwidth]{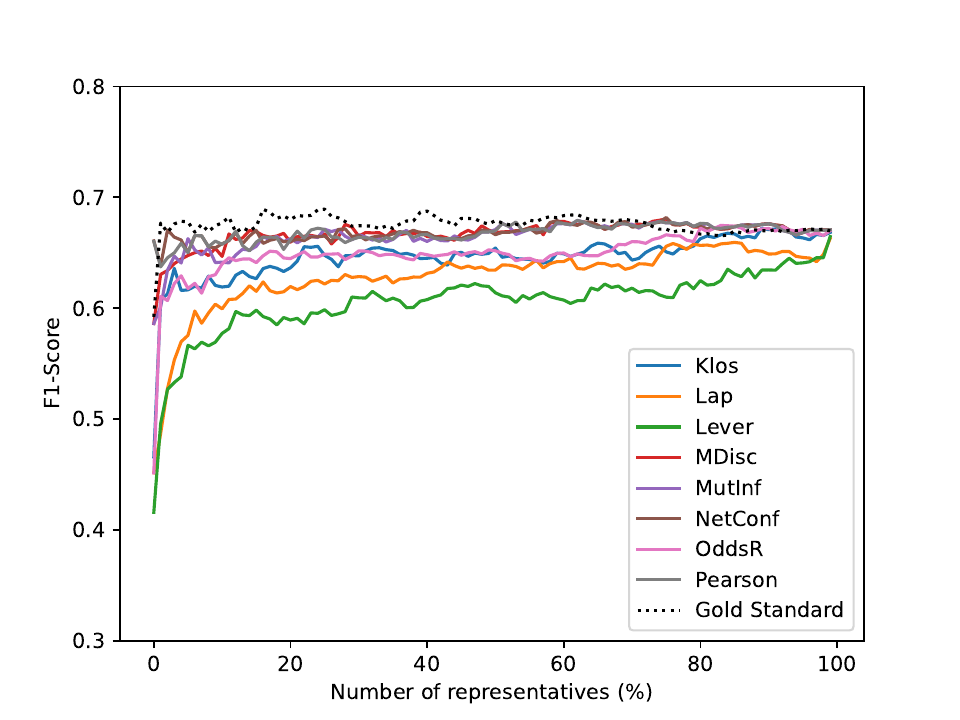}
        \subcaption{$F1$-Score (QM 9-16)}
        \label{fig:Frank_F1_2}
    \end{minipage}
    \begin{minipage}[b]{0.32\textwidth}
        \includegraphics[width=\textwidth]{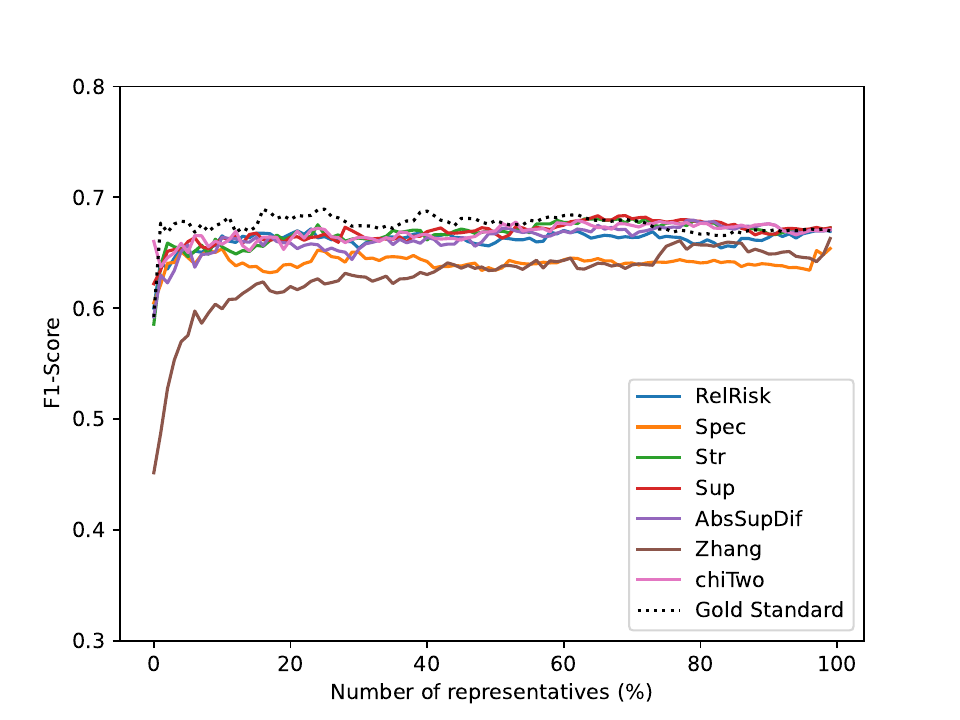}
        \subcaption{$F1$-Score (QM 17-23)}
        \label{fig:Frank_F1_3}
    \end{minipage}
    \caption{Experiments for the FRANK dataset.}
    \label{fig:FRANK_ALL}
    \Description{Description} 
\end{figure}

The results obtained for the FRANK dataset are not presented in the main article due to their similarity with D\&D. Figure~\ref{fig:Frank_NbR} and~\ref{fig:Frank_F1_cluster} correspond to the experiments from Section~\ref{sec:ExpClustering}. Figure~\ref{fig:Frank_Matrix} shows Kendall's Tau correlation matrix, as in Section~\ref{sec:ExpPairwise}. Figures~\ref{fig:Frank_RBO_1}, \ref{fig:Frank_RBO_2}, and~\ref{fig:Frank_RBO_3} show comparisons with the gold standard in terms of RBO, similarly to what we do in Section~\ref{sec:ExpGoldRanks}. Figures~\ref{fig:Frank_F1}, \ref{fig:Frank_F1_2}, and \ref{fig:Frank_F1_3} show the classification performance in terms of $F1$-Score, as in Section~\ref{sec:ExpGoldPerfs}.

The main difference between this dataset and the others, in terms of results, is in the blocks of correlation quality measures identified using Kendall's Tau (Figure~\ref{fig:Frank_Matrix}), which are not exactly the same as for the other datasets:
\begin{itemize}
    \item \textsc{Dep} and \textsc{Lap} are included in the \textsc{GR} block.
    \item \textsc{SupDif} is included in the \textsc{Acc} block.
    \item  \textsc{Jacc} is included in the \textsc{Cos} block.
    \item \textsc{MutInf}, \textsc{Pearson} and $\chi^{2}$ share a block together.
\end{itemize}

The overall classification performance is lower than for the other datasets, which can be explained by the absence of labels, resulting in a large number of generic patterns, present in many graphs regardless of class.

\subsection{IMDb Results}
\label{sec:AppAddResIMDb}
\begin{figure}[htbp!]
    \centering
    \begin{minipage}[b]{0.32\textwidth}
        \includegraphics[width=\textwidth]{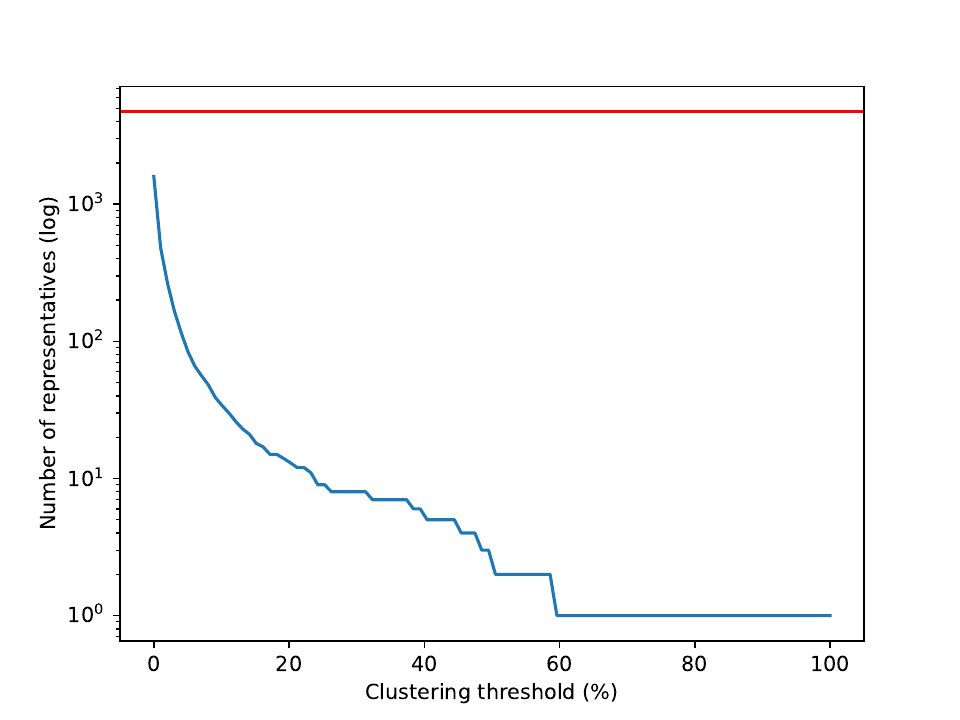}
        \subcaption{Number of representatives}
        \label{fig:IMDB_NbR}
    \end{minipage}
    \begin{minipage}[b]{0.32\textwidth}
        \includegraphics[width=\textwidth]{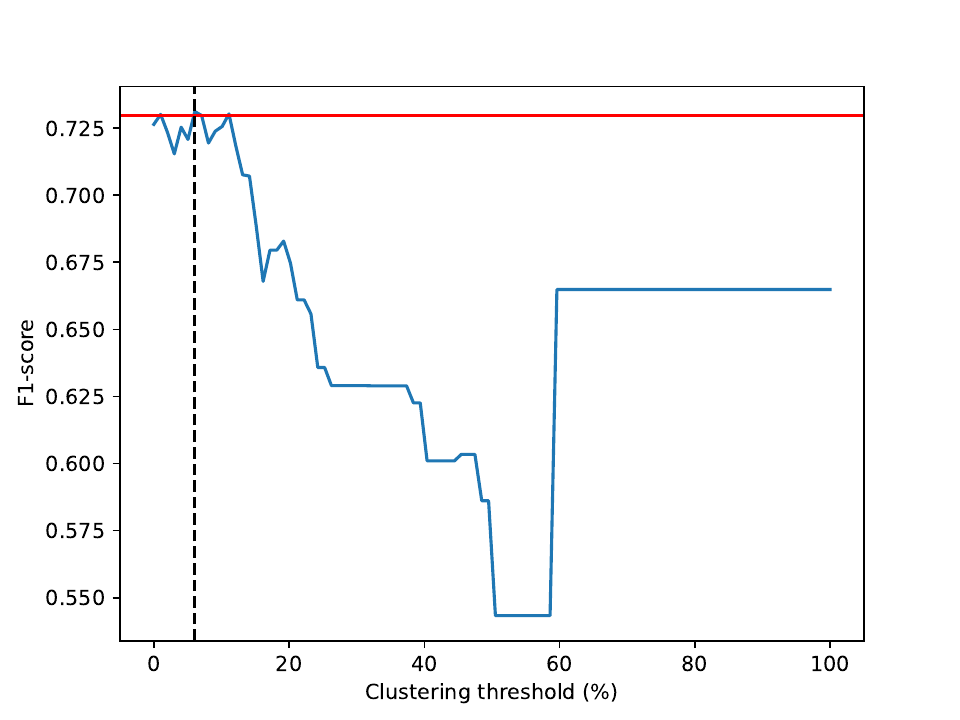}
        \subcaption{$F1$-Score (clustering threshold)}
        \label{fig:IMDB_F1_cluster}
    \end{minipage}
    \begin{minipage}[b]{0.32\textwidth}
        \includegraphics[width=\textwidth]{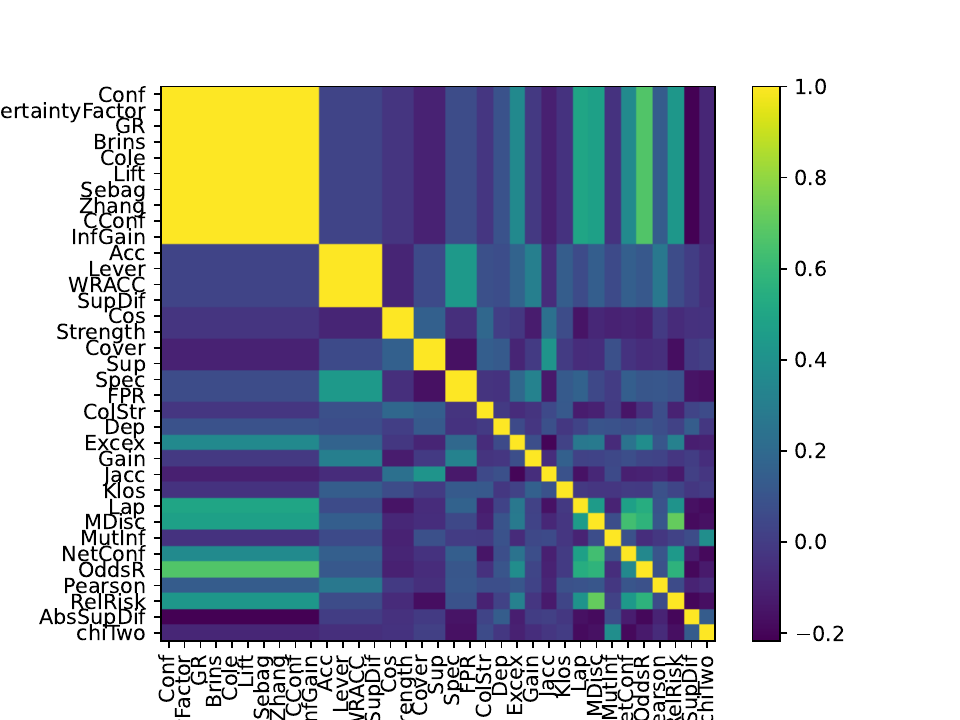}
        \subcaption{RBO matrix correlation}
        \label{fig:IMDB_Matrix}
    \end{minipage}
    
    \vspace{1em}
    
    \begin{minipage}[b]{0.32\textwidth}
        \includegraphics[width=\textwidth]{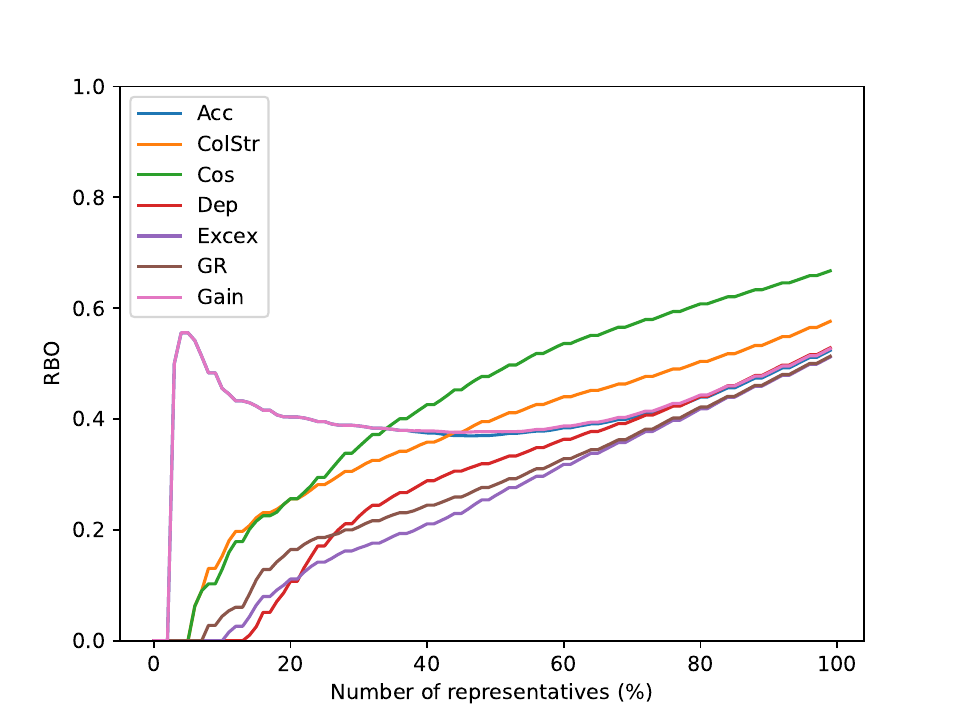}
        \subcaption{RBO (QM 1-8)}
        \label{fig:IMDB_RBO_1}
    \end{minipage}
    \begin{minipage}[b]{0.32\textwidth}
        \includegraphics[width=\textwidth]{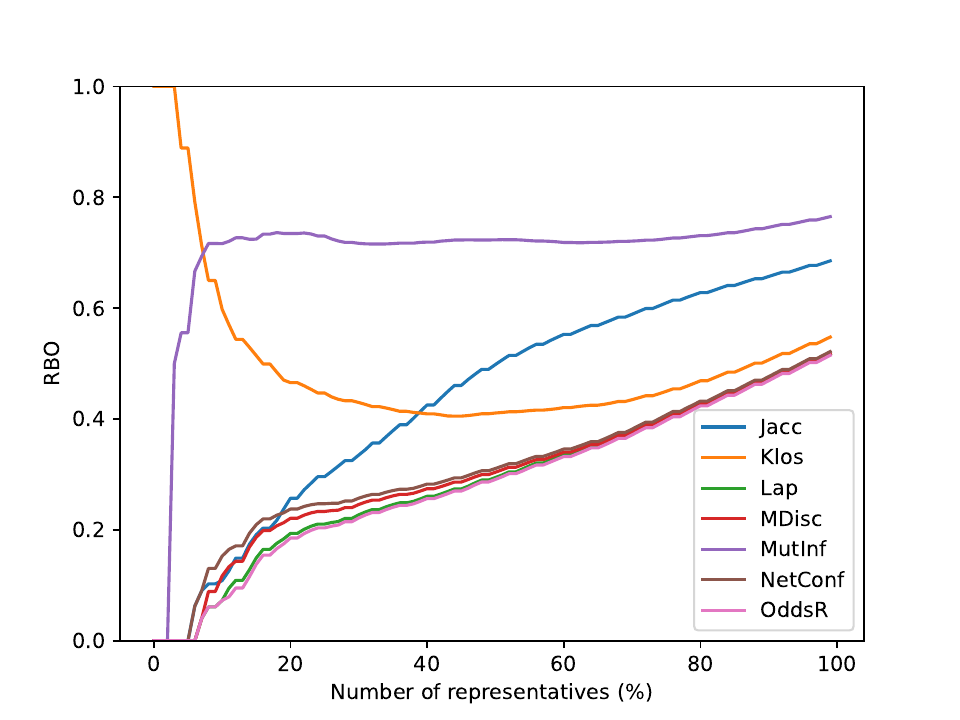}
        \subcaption{RBO (QM 9-16)}
        \label{fig:IMDB_RBO_2}
    \end{minipage}
    \begin{minipage}[b]{0.32\textwidth}
        \includegraphics[width=\textwidth]{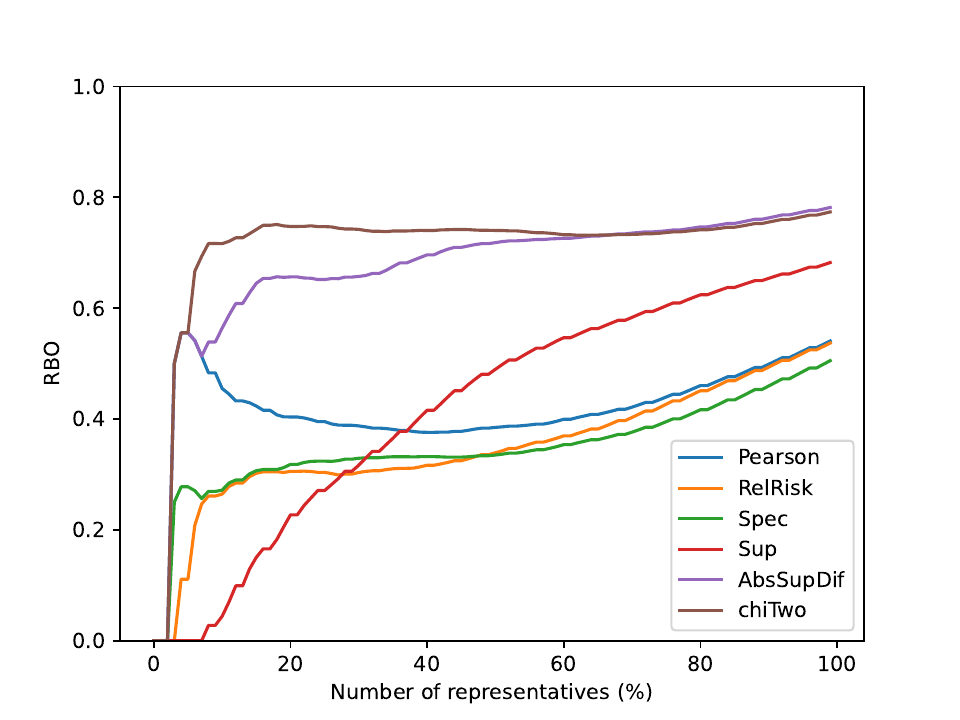}
        \subcaption{RBO (QM 17-23)}
        \label{fig:IMDB_RBO_3}
    \end{minipage}

    \vspace{1em}
    
    \begin{minipage}[b]{0.32\textwidth}
        \includegraphics[width=\textwidth]{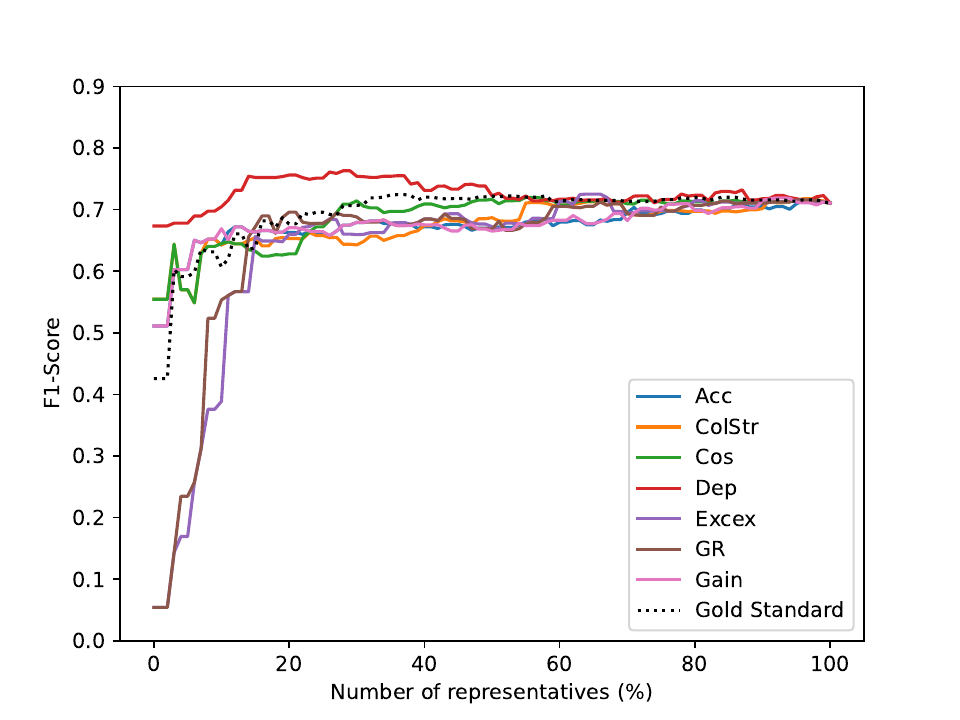}
        \subcaption{$F1$-Score (QM 1-7)}
        \label{fig:IMDB_F1}
    \end{minipage}
    \begin{minipage}[b]{0.32\textwidth}
        \includegraphics[width=\textwidth]{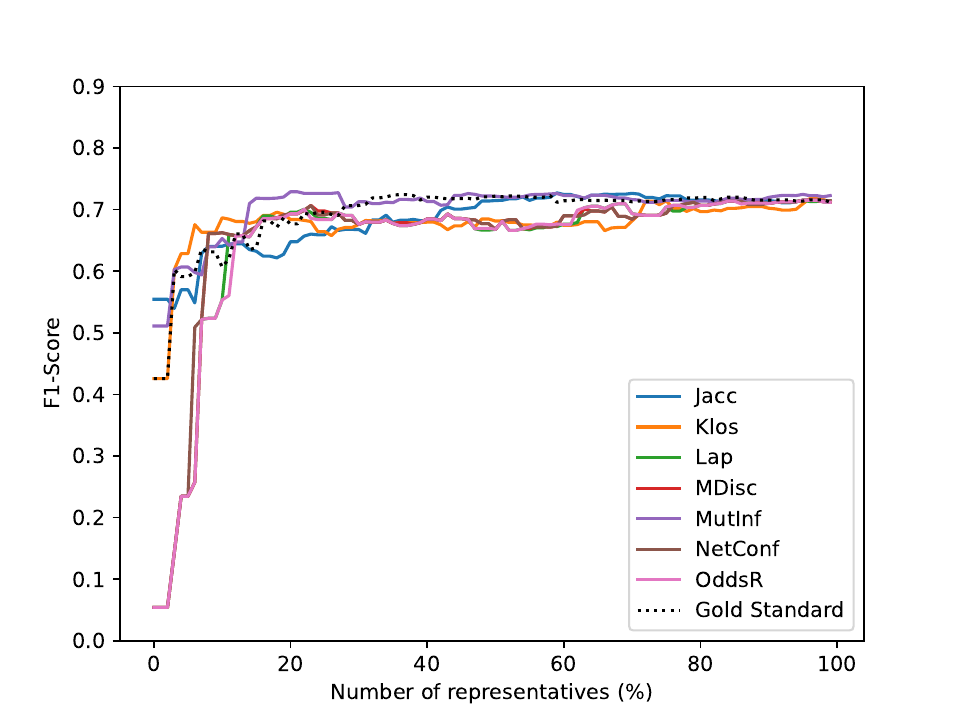}
        \subcaption{$F1$-Score (QM 8-14)}
        \label{fig:IMDB_F1_2}
    \end{minipage}
    \begin{minipage}[b]{0.32\textwidth}
        \includegraphics[width=\textwidth]{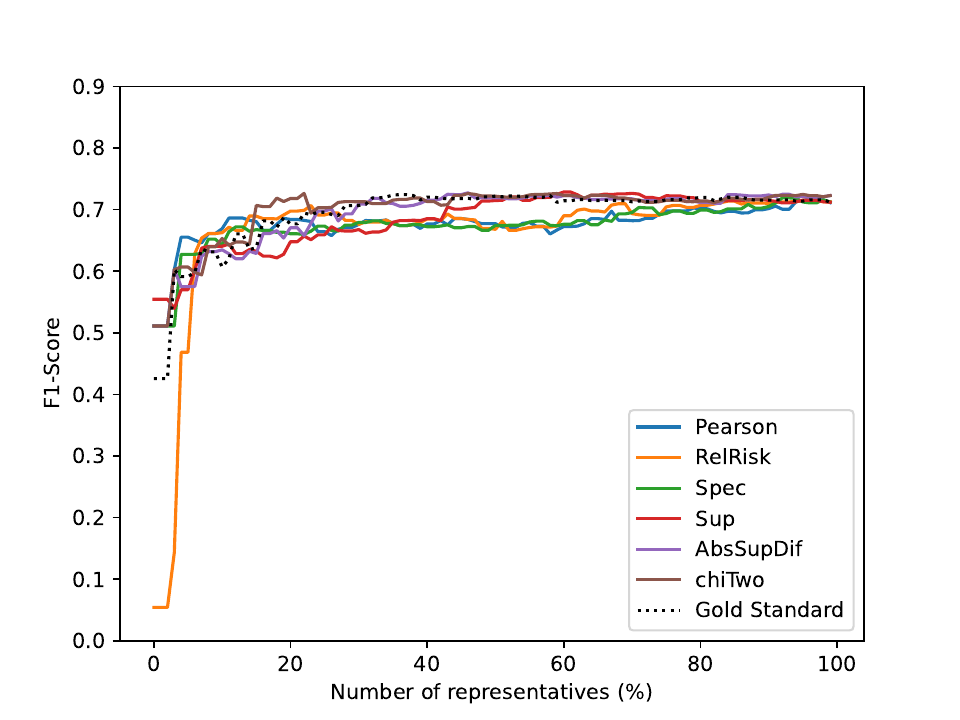}
        \subcaption{$F1$-Score (QM 15-20)}
        \label{fig:IMDB_F1_3}
    \end{minipage}
    \caption{Experiments for the IMDb dataset.}
    \label{fig:IMDB_ALL}
    \Description{Description} 
\end{figure}

The results obtained for the IMDb dataset are not presented in the main article due to their similarity with those of AIDS. Figure~\ref{fig:IMDB_NbR} and~\ref{fig:IMDB_F1_cluster} correspond to the experiments from Section~\ref{sec:ExpClustering}. Figure~\ref{fig:IMDB_Matrix} shows Kendall's Tau correlation matrix, as in Section~\ref{sec:ExpPairwise}. Figures~\ref{fig:IMDB_RBO_1}, \ref{fig:IMDB_RBO_2}, and~\ref{fig:IMDB_RBO_3} show comparisons with the gold standard in terms of RBO, similarly to what we do in Section~\ref{sec:ExpGoldRanks}. Figures~\ref{fig:IMDB_F1}, \ref{fig:IMDB_F1_2}, and \ref{fig:IMDB_F1_3} show the classification performance in terms of $F1$-Score, as in Section~\ref{sec:ExpGoldPerfs}.

The blocks of measures are identical to the general case. However, a difference can be observed regarding classification performance. Measure \textsc{Dep} achieves a better $F1$-score than our gold standard, despite a low RBO between the two. This is because our gold standard is only an approximation of the ground truth ranking, as it is based on an approximation of the Shapley Value (cf. Section~\ref{sec:MethRanksShap}). 

\end{document}